\documentclass[12pt]{cmuthesis}
\pdfoutput=1
\usepackage{floatrow} 
\usepackage{color}
\usepackage{fullpage}
\usepackage[numbers,sort&compress]{natbib}
\usepackage{moreverb}
\usepackage{float}
\usepackage[shortlabels]{enumitem}
\usepackage{multicol}
\usepackage{subcaption}
\usepackage[font=footnotesize,labelfont=bf]{caption}
\usepackage{graphicx}
\usepackage{booktabs} 
\usepackage{transparent}
\graphicspath{./fig/}

\usepackage{mwe} 

\usepackage{amsmath}
\usepackage{amssymb}
\usepackage{amsfonts}
\usepackage{mathtools,bbm}
\usepackage{bm}
\usepackage{mathrsfs}
\usepackage[binary-units=true]{siunitx}
\usepackage{pgffor}
\usepackage{dsfont}

\usepackage{physics}
\usepackage[
backref,
hidelinks,
breaklinks,
colorlinks,
linkcolor=black,
citecolor=black,
filecolor=red,
urlcolor=blue
]{hyperref}
\usepackage[export]{adjustbox}
\usepackage{url}
\usepackage{makeidx}
\usepackage{csquotes}
\usepackage{xinttools}
\usepackage{gensymb}
\usepackage[space]{grffile}
\usepackage{soul}
\usepackage[letterpaper, twoside, vscale=0.8, hscale=0.75, nomarginpar,
hmarginratio=1:1]{geometry}
\usepackage{xspace}
\usepackage[multiple]{footmisc}

\usepackage{xparse} 
\usepackage{environ}

\usepackage{amsthm}
\usepackage{algorithm}
\usepackage[noend]{algpseudocode}
\usepackage{csvsimple}

\usepackage{titlesec} 
\usepackage{pbox} 

\usepackage{tabularx} 

\usepackage{pgfplotstable} 
\usepackage{pgfplots}
\usepackage{tikz}

\usepgfplotslibrary{external}
\pgfplotsset{compat=newest} 
\pgfplotsset{plot coordinates/math parser=false}

\DeclareGraphicsExtensions{.pdf,.png,.jpg}

\newlength\figureheight
\newlength\figurewidth
\newlength\subfiguresheight
\newlength\phantomheight
\newlength\compositewidth

\newlength\notationsymbol
\newlength\notationdescription

\newcommand{\hyph}{-\penalty0\hskip0pt\relax}

\newif\ifshowdevelopment

\showdevelopmentfalse

\newcommand{\todo}[1]{%
  \ifshowdevelopment%
    \textcolor{blue}{\emph{\bf TODO:\ #1}}
  \else\fi%
}

\NewEnviron{development}{\color{blue}\ifshowdevelopment\BODY\fi}

\newcommand{\datapath}{{./fig}}
\newcommand{\setdatapath}[1]{%
  \renewcommand{\datapath}{#1}%
  \graphicspath{{#1/}}%
}

\newcommand{\belowtoprule}{\midrule[\heavyrulewidth]}

\newtheorem{theorem}{Theorem}
\newtheorem{definition}{Definition}
\newtheorem{algorithmdef}[algorithm]{Algorithm}
\newtheorem{problem}{Problem}
\newtheorem{corollary}{Corollary}[theorem]

\newtheorem{lemma}[theorem]{Lemma}
\newtheorem*{remark}{Remark}
\newtheorem{assumption}{Assumption}

\algnewcommand{\CommentLine}[1]{\State \(\triangleright\) #1}

\DeclareMathOperator*{\argmin}{arg\,min}
\DeclareMathOperator*{\argmax}{arg\,max}

\newcommand{\spacefont}[1]{{\mathbb{#1}}}

\renewcommand{\real}{{\spacefont{R}}}
\newcommand{\integer}{{\spacefont{Z}}}

\newcommand{\SOthree}{{\spacefont{SO}(3)}}
\newcommand{\SEthree}{{\spacefont{SE}(3)}}

\renewcommand{\vec}[1]{\mathbf{#1}}

\newcommand{\I}{{\mathbf{I}}}

\newcommand{\probabilityfont}[1]{{\mathds{#1}}}
\newcommand{\E}{{\probabilityfont{E}}}
\renewcommand{\P}{{\probabilityfont{P}}}
\renewcommand{\H}{{\probabilityfont{H}}} 
\newcommand{\h}{{\probabilityfont{h}}} 
\newcommand{\p}{{p}} 
\newcommand{\MI}{{\probabilityfont{I}}}
\newcommand{\SMI}{{\MI_{\mathrm{S}}}}
\newcommand{\CSQMI}{{\MI_{\mathrm{CS}}}}
\newcommand{\D}{{\probabilityfont{D}}}
\newcommand{\DKL}{{\D_{\mathrm{KL}}}}
\newcommand{\DCS}{{\D_{\mathrm{CS}}}}

\newcommand{\setsystemfont}[1]{{\mathscr{#1}}}
\newcommand{\ground}{\setsystemfont{U}}
\newcommand{\independence}{\setsystemfont{I}}
\newcommand{\coverset}{\setsystemfont{C}}
\newcommand{\block}{\setsystemfont{B}}
\newcommand{\matroidrank}{\mbox{rank}}

\newcommand{\setfun}{g} 

\newcommand{\graphfont}[1]{{\mathcal{#1}}}
\newcommand{\graph}{{\graphfont{G}}}
\newcommand{\edges}{{\graphfont{E}}}
\newcommand{\deletededges}{{\graphfont{\tilde E}}}
\newcommand{\weights}{{\graphfont{W}}}
\newcommand{\approxweights}{{\graphfont{\widehat W}}}

\newcommand{\opt}{{\star}}

\newcommand{\state}{\vec{x}}
\newcommand{\states}{\vec{X}}
\newcommand{\observation}{\vec{y}}
\newcommand{\observations}{\vec{Y}}
\newcommand{\environment}{E}
\newcommand{\environments}{\mathcal{E}}
\newcommand{\policy}{\pi}
\newcommand{\sensingquality}{J}
\newcommand{\quota}{B}
\newcommand{\control}{u}
\newcommand{\dynamics}{f}
\newcommand{\observationfunction}{h}

\newcommand{\camera}{{F^\mathrm{cam}}}
\newcommand{\occupancy}{{F^\mathrm{occ}}}

\newcommand{\initialstates}{\mathcal{X}_{0}}
\newcommand{\safespace}{\mathcal{X}_{\text{safe}}}

\newcommand{\goalspace}{\mathcal{X}_{\text{goal}}} 
\newcommand{\viewthreshold}{\epsilon_{\text{view}}}
\newcommand{\goaldistance}{d_{\text{goal}}}
\newcommand{\distancefactor}{\alpha}
\newcommand{\environmentsguess}{\mathcal{E}_{\text{guess}}}
\newcommand{\cell}{c}
\newcommand{\covercells}{C_{\text{cov}}}
\newcommand{\cellweight}{w_{\text{cell}}}
\newcommand{\newcellweight}{w_{\text{new}}} 

\newcommand{\neighbor}{\mathcal{N}}
\newcommand{\rangeneighbor}{\mathcal{B}}
\newcommand{\events}{\setsystemfont{E}}
\newcommand{\agents}{\mathcal{A}}
\newcommand{\budget}{\gamma}
\newcommand{\agentdensity}{\rho}
\newcommand{\maxredundancy}{\hat r}
\newcommand{\redundancygraph}{\graph}
\newcommand{\numevents}{{n_\mathrm{e}}}

\newcommand{\agentrange}{{r_\mathrm{a}}}
\newcommand{\sensorrange}{{r_\mathrm{s}}}
\newcommand{\communicationrange}{{r_\mathrm{c}}}

\definecolor{cmu_red}{rgb}{0.6, 0.0, 0.0} 
\definecolor{endcolor}{RGB}{196,18,48}
\definecolor{differencecolor}{RGB}{0,45,114}

\newcommand{\robots}{{\mathcal{R}}}
\newcommand{\cells}{{\mathcal{C}}}
\newcommand{\numrobot}{{n_\mathrm{r}}}
\newcommand{\numagent}{{n_\mathrm{a}}}
\newcommand{\numcells}{{n_\mathrm{m}}}
\newcommand{\numrounds}{{n_\mathrm{d}}}
\newcommand{\numneighbors}{{n_\mathrm{n}}}
\newcommand{\controls}{\mathcal{U}}

\newcommand{\neighbordistance}{d_\mathrm{n}}
\newcommand{\networkdiameter}{d_\mathrm{m}}

\newcommand{\optionalsubscript}[1][]{%
  \ifthenelse{\equal{#1}{}}{%
  }{%
    $_{#1}$%
  }%
}

\NewDocumentCommand{\distalgnamed}{O{} O{} m}{%
  \ifthenelse{\equal{#2}{}}{%
    {\text{\textnormal{#3}\optionalsubscript[#1]{}}}%
  }{%
    {\text{#2{#3}\optionalsubscript[#1]{}}}%
  }%
}

\newcommand{\greedy}[1][]{{\distalgnamed[][#1]{SGA}}}
\NewDocumentCommand{\dgreedy}{O{} O{}}
{%
  \distalgnamed[#1][#2]{DSGA}%
}

\NewDocumentCommand{\rsp}{O{} O{}}
{%
  {\distalgnamed[#1][#2]{RSP}}%
}
\NewDocumentCommand{\rrsp}{O{} O{}}
{%
  {\distalgnamed[#1][#2]{R-lRSP}}%
}

\newcommand{\inputfigure}[2][]
{%
  \ifthenelse{\equal{#1}{}}%
  {}{\tikzsetnextfilename{#1}}%
  \input{#2}%
}

\newcommand{\setfunsim}{\setfun^\mathrm{sim}} 
\newcommand{\setfuns}{\mathscr{G}} 

\newcommand{\ignore}{{\mathcal{\hat N}}}
\newcommand{\targets}{\mathcal{T}}
\newcommand{\numtargets}{{n_\mathrm{t}}}
\newcommand{\numrobots}{{\numrobot}}

\newcommand{\method}[1]{{\textsc{#1}}}
\newcommand{\inneighbor}{\neighbor^\mathrm{in}}
\newcommand{\outneighbor}{{\neighbor^\mathrm{out}}}
\newcommand{\data}{\theta}
\newcommand{\approxsetfun}{\widetilde\setfun}

\newcommand{\cost}{\gamma}
\newcommand{\objectivecost}{\cost^\mathrm{obj}}
\newcommand{\plannercost}{\cost^\mathrm{plan}}
\newcommand{\distcost}{\cost^\mathrm{dist}}
\newcommand{\robotstates}{\vec{X}^\mathrm{r}}
\newcommand{\targetstates}{\vec{X}^\mathrm{t}}
\newcommand{\robotposition}{\vec{p}^\mathrm{r}}
\newcommand{\targetposition}{\vec{p}^\mathrm{t}}

\newcommand{\robotstate}{\vec{x}^\mathrm{r}}
\newcommand{\targetstate}{\vec{x}^\mathrm{t}}

\newcommand{\robotspace}{\real^{d^\mathrm{r}}}
\newcommand{\targetspace}{\real^{d^\mathrm{t}}}

\newcommand{\robotdynamics}{f^\mathrm{r}}
\newcommand{\targetdynamics}{f^\mathrm{t}}

\newcommand{\controlspace}{\mathcal{U}}

\newcommand{\targetnoise}{\epsilon^\mathrm{t}}
\newcommand{\observationnoise}{\epsilon^\mathrm{y}}

\newcommand{\capacity}{C}

\newcommand{\viewvalue}{\setfun_{\text{view}}}
\newcommand{\distancevalue}{\setfun_{\text{dist}}}
\newcommand{\expectedcoverage}{\setfun_{\text{cov}}}
\newcommand{\actionstostates}{\Phi}

\newcommand{\distributedplanningduration}{\Delta^\mathrm{d}}
\newcommand{\planningduration}{\Delta^\mathrm{p}}

\newcommand{\epoch}{n_\mathrm{e}}
\newcommand{\communicationneighbors}{\neighbor^\mathrm{c}}

\begin{document}

\frontmatter

\pagestyle{empty}
\title{
  {\bf\LARGE Sensor Planning for Large Numbers of Robots}
}
\author{Micah Corah}
\date{September 21, 2020}
\Year{2020}
\trnumber{CMU-RI-TR-20-53}

\committee{
  Nathan Michael, \emph{Chair} \\
  Anupam Gupta \\
  Katia Sycara \\
  Mac Schwager, Stanford
}

\support{}


\maketitle

\begin{dedication}
  \begin{minipage}{0.48\linewidth}
    For my parents, David and Barbara, who supported and encouraged my
    interest in robotics at every step of this journey.
  \end{minipage}
\end{dedication}

\pagestyle{plain}

\begin{abstract}
  In the wake of a natural disaster, locating and extracting victims quickly
is critical because mortality rises rapidly after the first forty-eight hours.
In order to assist search and rescue teams and improve response times, teams of
aerial robots equipped with sensors and cameras can engage in sensing tasks
such as mapping buildings, assessing structural integrity, and locating victims.
We seek to enable large numbers of robots to cooperate to complete such sensing
tasks more quickly and thereby to improve response times for first responders.

When formalized, such sensing tasks encapsulate numerous computationally
difficult (at least NP-Hard) problems related to routing, sensor coverage, and
decision processes.
One way to simplify planning for these tasks is to focus on maximizing
sensing performance over a short time horizon.
Specifically, consider the problem of how to select motions for a team of robots
to maximize a notion of sensing quality (the sensing objective) over the near
future, say by maximizing the amount of unknown space in a map that robots will
observe over the next several seconds.
By repeating this process regularly (about once a second), the robots can react
quickly to new observations as they work to complete the sensing task.
In technical terms, this planning and control process forms an example of
receding-horizon control.
Fortunately, common sensing objectives for these problems benefit from well-known
monotonicity properties (e.g. submodularity), and greedy algorithms can exploit
these monotonicity properties to solve the receding-horizon optimization
problems that we study near-optimally.

However, greedy algorithms typically force robots to make decisions
sequentially.
Thus, planning time grows with the number of robots and eventually exceeds time
constraints to replan in real time.
This is particularly important in distributed settings as the accumulation of
communication latencies between robots in sequence can be significant on its
own.
Further, recent works that have begun to investigate sequential greedy planning,
have demonstrated that reducing the number of sequential steps while retaining
suboptimality guarantees can be hard or impossible.

This thesis demonstrates that halting such growth in planning time is possible
for many sensing problems.
To do so, we introduce new greedy methods, Randomized Sequential Partitions
(\rsp{})
and Range-limited \rsp{}.
These methods enable planning with a fixed number of sequential steps that does
not grow with the number of robots.
Additionally, we prove that our algorithms approach the initial near-optimality
guarantees for sequential planning for many sensing problems.
In doing so, we develop new methods for quantifying redundancy between potential
future observations and highlight the importance of a relatively unknown
monotonicity property of some sensing objectives.

\pagebreak[2]
We apply our algorithms to autonomous mapping (known as exploration) and target
tracking problems which serve as proxies for the variety of tasks and
combinations of tasks that may arise in search and rescue scenarios.
Simulation results demonstrate that our greedy planners often approach the
performance of sequential planning  (in terms of target position uncertainty)
given only a few planning steps (2-4), even for very large numbers of robots
(96).
This amounts to a $24\times$ reduction in the number of sequential steps
and an equivalent or greater reduction in the duration of  distributed planning.

With exploration, we apply our methods in the
context of a complex system where robots equipped with depth cameras
map unknown, three-dimensional environments such as an office space or a cave.
Moreover, the analysis we present when applying our planning methods
to mapping objectives also provides valuable insights into the design of such
exploration systems.
While consistently improving completion times via greedy methods proves
challenging,
we demonstrate that sequential planning and \rsp{}
increase coverage rates early in simulation trials and
reliably improve solution quality.
Finally, we present a distributed implementation of \rsp{} and simulation
results for exploration of an office environment in real time, along with a 5\%
improvement in task completion time in this setting.

\end{abstract}

\begin{acknowledgments}
  I would like to begin by thanking my family for their ongoing support.
In particular, my parents have encouraged my interests in robotics for quite a
long time and put up with quite a nerdy youngster.
They deserve more than just my thanks.
I should also thank my partner, Srujana.
She kept me alive, happy, and well fed during the final and most harrowing
months of my doctorate.

I also have many to thank in the Resilient Intelligent Systems Lab.
My officemates, Ellen and Derek were ever-present companions.
Ellen, I believe our fish tank set new standards for office decor as well as
undergraduate achievement.
Derek, aside from being my office compatriot, you were also an early mentor
when I was still an intern.
Thank you for that time.
And, Vibhav, you were my replacement, but I preferred being students together.
Kshitij and Cormac, we collaborated on some of the best work that is not in this
thesis.
Thank you Curt for keeping my robots alive before I learned to avoid them.
And, thanks Wennie for doing the hard work to develop systems for exploration
and mapping.
Vishnu, thank you for demonstrating the first feasible solution to the
graduation problem.
Your leadership was invaluable.
Shaurya, thank you for being someone who I can talk to about research and
everything else.
Also, John, the pandemic has taken our pizza, and I want it back.
Thanks to everyone else in the lab who has made my time here brighter, including
Alex, Xuning, Aditya, Arjav, Mosam, Moses, Tim, Matt, Mike, Erik, and anyone
else who I may have forgotten.
Also, Karen Widmaier, you put in a good deal of hard work for our lab, and you
make our hearts warmer.

Many thanks to my friends in the Robotics Institute.
Achal and Radhika, Dhruv and Cara, and now I am compelled to mention Xuning
again, thank you for the adventures that brought so much joy to this time.
Shushman, you are a caring friend.
I did not introduce you to my family; I introduced my family to you.
Thank you Reuben, you speak softly, but your words carry great weight.
Nick, you have done great things for the RI, but you also do great things for
frogs.

Thanks to everyone else who contributed to my time here.
Rachel Burcin, I am only one of a vast number of people who have been
significantly impacted by your tireless enthusiasm administrating RISS.
I could not keep up with you.
Reid, you welcomed me to CMU twice, and I hope to someday become more like you.
Thank you Michael Erdmann, you allowed me to babble about submodular functions
in front of 16-811.
Thanks Sankalp, your words in front of same class introduced me to same topic.
My thanks also to Roie for validating my interest in 3-increasing functions
and for our too infrequent exchanges on the topic.

Thank you Nate for guiding my journey in robotics at CMU.
You accepted my stubbornness and pushed me to excellence.
Thanks also to the rest of my committee, Anupam, Katia, and Mac for direction
and advice.
I hope these first few pages are to your liking;
for the next hundred, well I am not Asimov.

\end{acknowledgments}

\tableofcontents
\listoffigures
\listoftables
\listofalgorithms

\mainmatter

\chapter*{Notation}
\label{chapter:notation}

\setlength{\notationsymbol}{3cm}
\setlength{\notationdescription}{10.0cm}

\textbf{General mathematical notation}\\
\begin{tabular}{p{\notationsymbol}p{\notationdescription}}
  $\real$, $\real_{>0}$  & The set of real numbers and real numbers greater than
  zero respectively\\%
  $\integer$ & The set of integers\\%
  $\SOthree$ &
  The special orthogonal group in three dimensions (rotations)\\
  $\SEthree$ &
  The special Euclidean group for three dimensions (rigid-body motions)\\
  $\vec p$, $\vec v$ & Examples of vectors\\
  $A_{1:n}$, $A_{X}$ &
  Examples of indexing ``sets.'' Elements of sets will generally be
  implicitly associated with time or agent indices which will be accessed
  using ranges and sets of indices\\
  $X^\opt$, $X^\text{g}$, $X^\text{d}$ & Superscripts designating (elements of)
  an optimal solution to an optimization problem and examples of output from
  greedy and distributed algorithms as will be evident from context%
  \vspace{0.2cm}
\end{tabular}\\
\textbf{Information theory and probability}\\
\begin{tabular}{p{\notationsymbol}p{\notationdescription}}
  $\P(X\!=\!x|Y\!=\!y)$ & Probability that a random variable $X$ equals $x$ conditional
  on $Y\!=\!y$ \\
  $\E[X]$ & The expected value of $X$ \\
  $\H(X|Y)$ & Entropy of $X$ conditional on $Y$ \\
  $\MI(X;Y|Z)$ & Mutual information between $X$ and $Y$ conditional on $Z$\\
  \vspace{0.2cm}
\end{tabular}\\

\pagebreak[4]
\noindent\textbf{Combinatorial optimization}\\
\begin{tabular}{p{\notationsymbol}p{\notationdescription}}
  $\ground$ & The ground set \\
  $\independence$ & An independence system \\
  $\block$ & A block of a partition (matroid)\\
  $\setfun(A,B)$, $\setfun(x)$ &
  Evaluation of a set function $\setfun$ at $A \cup B$ where $A,B\subseteq \ground$
  and implicit evaluation at $\{x\}$ where $x\in\ground$\\
  $\setfun(A|B)$, $\setfun(A;B|C)$ &
  The former shows evaluation at $A$ ``conditional'' on $B$
  ($\setfun(A,B) - \setfun(A)$) using similar notation as for mutual information.
  More generally, these two are first and second discrete derivatives of
  $\setfun$
  where
  $\setfun(A;B|C) = \setfun(B;A|C) = \setfun(A,B,C) - \setfun(A,C)
                                     - \setfun(B,C) + \setfun(C)$
  (with $A$, $B$, $C$ disjoint).
  \vspace{0.2cm}
\end{tabular}\\
\textbf{Graphs}\\
\begin{tabular}{p{\notationsymbol}p{\notationdescription}}
  $\graph$ & A (directed or undirected) graph\\
  $\edges$ & The set of edges of a graph\\
  $\deletededges$ & Edges removed from a complete graph,
  $\robots\times\robots\setminus\edges$ if robots are vertices\\
  $\weights$ & Edge weights
  \vspace{0.2cm}
\end{tabular}\\

As hinted above, we will forgo use of a symbol for vertices in favor of the
symbol for the set of objects---generally robots or agents---in question.
\vspace{0.2cm}\\
\textbf{Robotics and control}\\
\begin{tabular}{p{\notationsymbol}p{\notationdescription}}
  $\robots$ & A set of (mobile) robots \\
  $\agents$ & A set of (immobile) agents \\
  $\controls$ & A set of controls (or sometimes actions)
  \vspace{0.1cm}
  \\
  $\dynamics(\state, \control)$ &
  System dynamics
  given state $\state$ and control $\control$
  (generally for an individual robot) \\
  $\observationfunction(\state)$ & Sensing or observation function
  (also for individual robots)
  \vspace{0.2cm}
\end{tabular}\\
\textbf{Algorithm acronyms}\\
\begin{tabular}{p{\notationsymbol}p{\notationdescription}}
  $\greedy$ & Sequential greedy assignment for submodular maximization on a
  partition matroid
  (acronym used in Chapter~\ref{chapter:distributed_multi-robot_exploration})
  \\
  $\dgreedy$, $\dgreedy[\numrounds]$ & Distributed sequential greedy assignment
  in general and in $\numrounds$ rounds, an early version of the methods this
  thesis proposes
  \\
  $\rsp$, $\rrsp$ & Randomized sequential partitions and the range-limited
  variant, the main contributions of this thesis
  \vspace{0.2cm}
\end{tabular}\\

\chapter{Introduction}
\setdatapath{./fig/introduction}
\label{chapter:introduction}

Robots are often characterized as being able to \emph{sense} their environment
and \emph{act} upon it~\citep{brooks1991intelligence}.
While the actions that a robot can perform upon an environment---say folding
laundry~\citep{miller2012ijrr} or preparing a meal---are
important~\citep{harmo2005},
sensing is becoming increasingly prominent.
This shift has been driven by changes in the markets for both robot platforms
and the sensors they carry:
\begin{itemize}
  \item Depth sensors and cameras have become increasingly prevalent, cheaper,
    and lighter, and
  \item Aerial robots have become both popular and
    pervasive~\citep[p.~41--51]{faa2019}.
\end{itemize}
Moreover, aerial robots, when equipped with appropriate sensors, are able to
navigate complex~\citep{deits2015icra} 
and unknown~\cite{liu2016icra}
environments to accomplish sensing tasks such as
videography~\citep{bonatti2018},
monitoring crops~\citep{berni2009tro},
and mapping~\citep{charrow2015rss}.

Sensing problems arising from urban disaster and emergency response and in
defense settings are time-sensitive and involve operation in unknown and
cluttered environments.
Sometimes, the time-sensitive nature of a task is immediate.
One such case is a fire wherein robots may assist firefighters to outpace the
effects of a conflagration.
Here, a team of robots might engage in sensing tasks immediately preceding
or in parallel with firefighters entering a building.\footnote{%
  While human-robot interaction is not a subject of this thesis, humans can be
  treated as additional members of a team that may provide additional sensor
  data, dictate sensing objectives, and must not be collided with.
}
Disaster response has similar features, but the time-sensitivity is driven by
scale.
In the case of a widespread disaster, teams may have ample time to inspect
individual buildings, but when viewed as a whole, the number of locations that
must be inspected and the paucity of responders motivate rapid action.
In particular, mortality typically increases rapidly forty-eight hours after the
start of a disaster while search and rescue teams often operate on a
first-come-first-serve basis~\citep{murphy2016}.

\begin{figure}
  \begin{center}
    \includegraphics[width=0.7\linewidth]{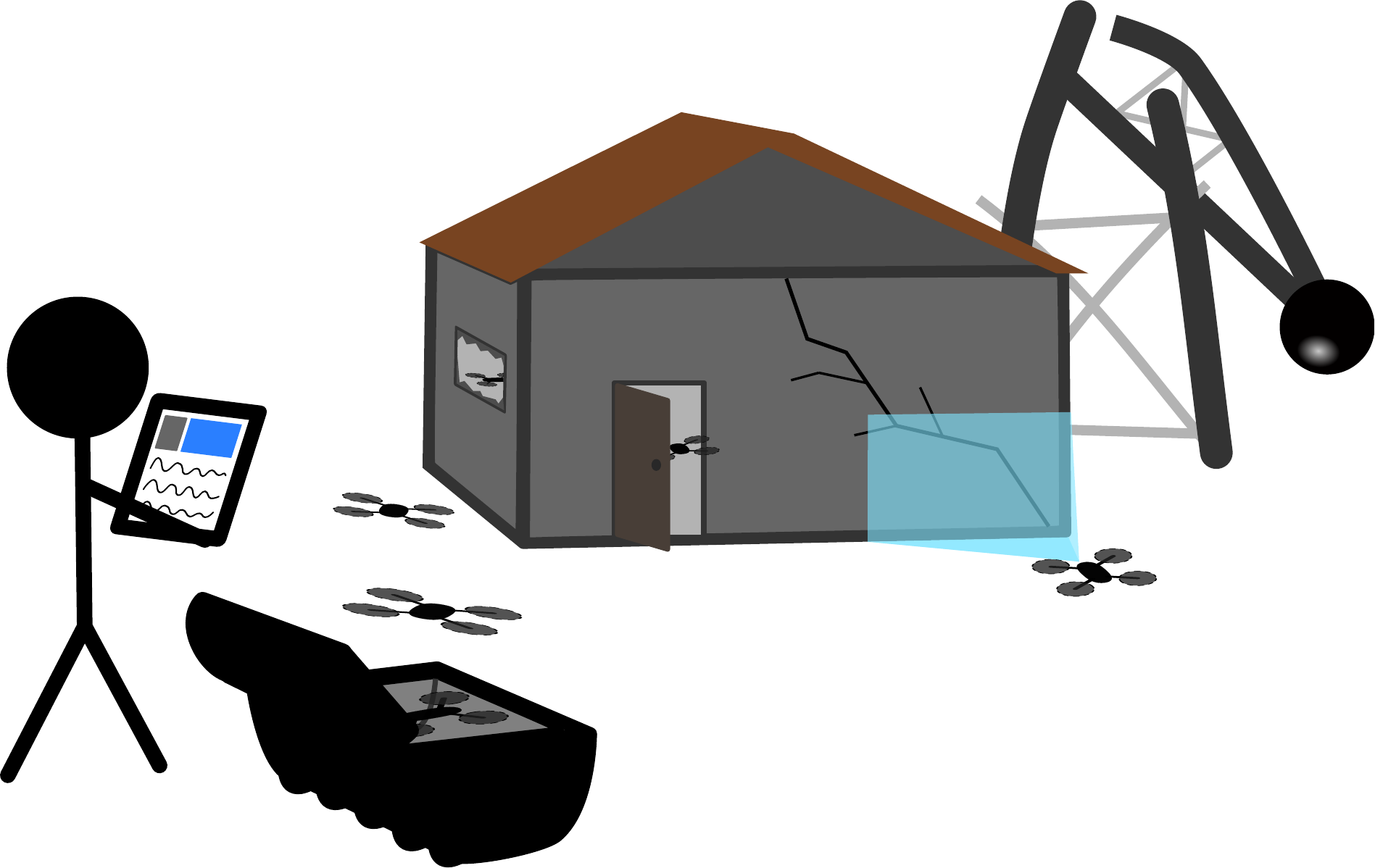}
  \end{center}
  \caption[Illustration of a disaster response scenario]{%
    This figure illustrates a typical disaster response scenario.
    Here, first responders wish to inspect a disaster site, such as after an
    earthquake, in order to ascertain structural integrity and to locate any
    survivors.
    To do so, they deploy a team of robots from a mobile base station that they
    transport along with the rest of their equipment, and the operator uses a
    tablet interface to instruct the robots to inspect the disaster site.
    Because communication infrastructure may have been destroyed in the
    disaster, the team has access to little prior information on the
    environment, and robots take advantage of distributed, onboard computation
    to mitigate effects of unreliable communication.
  }\label{fig:disaster_response_illustration}
\end{figure}

Figure~\ref{fig:disaster_response_illustration} illustrates an example of how
a team of robots may contribute to sensing tasks that arise while searching a
building in a disaster response scenario.
For the purpose of this thesis, we will focus on applications and challenges
related to disaster and emergency response scenarios such as this, although many
of the contributions are relevant to other sensing tasks and optimization
problems.

When seeking to improve response times, a designer may work to improve the
sensing platform itself along with planning and control algorithms for
individual robots~\citep{goel2019tr,goel2019fsr}.
However, platform constraints will ultimately limit the performance of
individual robots.
Additionally, teams of robots can cooperate in sensing tasks to cover more space
at once or to complete tasks more rapidly.
We focus on this latter dimension of the problem of improving
response times and ask:
\emph{how can a designer compose multiple engineered sensing platforms to
improve completion time in sensing tasks?}
Addressing this question for large teams of robots will involve solving
increasingly large planning problems, addressing communication constraints, and
accounting for complications such as inter-robot collision constraints.

As sensing tasks progress, the robots' individual and collective understanding
of the world (commonly referred to as belief) can also change rapidly.
The process of updating control actions in response to new sensor data is
commonly referred to as
adaptation~\citep{golovin2011jair,asadpour2015,hollinger2013ijrr}.
Adaptation is particularly important for robots operating in cluttered and urban
environments.
In these scenarios, as robots obtain new sensor data, they can become aware of
both new regions of interest and new unoccupied and traversable space.
Often, as for a robot that is moving toward unobserved
space~\citep{goel2019tr,goel2019fsr}, this data is immediately pertinent to the
selection of control actions.
By extension, teams of robots that are able complete sensing tasks rapidly while
operating in unknown environments must also adapt frequently.

\section{\emergencystretch 3em
  Challenges for distributed sensing and time-sensitive sensing domains}

The challenges that arise in this thesis can be divided into two categories:
technical challenges related to the design of sensing systems and domain
challenges for urban search and rescue.
While the domain challenges will motivate and shape the methods we propose,
the following technical challenges apply more generally to aerial sensing
systems:

\begin{itemize}
  \item \textbf{Size, Weight, and Power (SWaP) constraints:}
    Robots, especially aerial robots, have limitations in size and weight based
    on the application domain.
    These constraints are pervasive and can lead to further constraints
    on thrust, flight time, onboard computation, and sensor payloads
  \item \textbf{Safety:}
    Robots may operate in close proximity, replan frequently, and have
    non-trivial dynamics.
    Maintaining safety involves avoiding collisions between
    robots and other objects despite uncertainties in the states of the
    robots and the environment.
    Moreover, systems should be robust to planners that may sometimes fail to
    provide results, and
    safety may also include more complex conditions such as the ability to
    return to an initial state or avoiding hazards
  \item \textbf{Sensing:}
    The sensors that aerial robots carry are not trivial.
    Camera views are sensitive to orientation, and observations are complicated
    by the geometry of occlusions.
    Robots may also carry a variety of sensors such as thermal cameras,
    range sensors, or gas detectors, and autonomous sensing systems must account
    for sensor models, fuse data, and reason about the contributions of sensing
    actions
  \item \textbf{Communication:}
    Communication bandwidth between robots and operators may also be limited,
    and communication links may be unreliable.
    Communication constraints also affect dissemination of sensor data,
    operation of distributed algorithms, and interaction with human operators
  \item \textbf{Computation:}
    Algorithms related to planning and control may also scale poorly
    (e.g. being NP-Hard) and have to be solved suboptimally or approximately to
    be tractable.
    Computational units situated on the robots themselves or
    elsewhere must also have access to relevant information such as the robots'
    locations and sensor data
\end{itemize}

\noindent
Then, considering the application of methods for time-sensitive sensing for
disaster and emergency response in urban environments narrows these challenges
and produces more concrete details:

\begin{itemize}
  \item \textbf{Size:}
    Systems---referring to the entire multi-robot system and related
    equipment---that are usable by teams of first responders are likely limited
    in size to what one or two people can carry or what is portable with a small
    vehicle~\citep{murphy2016}
  \item \textbf{Hazardous environments:}
    Environments may contain water, dust, gasses, and other materials that are
    hazardous to people and robots~\citep{murphy2016}.
    Similarly, the structural integrity of a building may be compromised, and
    rubble may be unstable.
    Not only should robots be resilient to these hazards, but these hazards may
    themselves be the focus of sensing tasks, such as to provide situational
    awareness to responders who wish to avoid hazards as well
  \item \textbf{Complex geometry:}
    Rubble creates complex formations and small and irregular voids.
    Because casualties are often buried deep within the
    rubble~\citep{murphy2016}, sensing robots should be able navigate such voids
    or else be able to recognize voids and mark them for further inspection
  \item \textbf{Limited communication:}
    Disasters frequently destroy communication infrastructure, and wireless
    communication within a disaster site may be unreliable such as due to
    rubble~\citep{murphy2016}.
    Teams of robots may also employ ad-hoc networks~\citep{george2010}, and the
    communication graphs linking robots may be incomplete or disconnected
  \item \textbf{Lack of information:}
    Emergency and disaster response activities typically occur within hours of
    an event~\citep{murphy2016}, and teams often work independently and with
    limited access to information~\citep{murphy2008}.
    By extension and considering other constraints on communication,
    robot teams will also have limited prior information on the environments
    that they operate in
\end{itemize}

\noindent
Disaster and emergency response scenarios do not always exhibit all of these
challenges.
However, first responders and search and rescue teams are individually
responsible for a wide variety of disasters and emergencies.
For this reason,
any aerial sensing systems that are useful in practice will likely be those that
are readily accessible to responders and which those responders have been
trained to use.
As such, in order to be useful, robots may have to address many challenges
related to different emergencies and disaster
scenarios~\citep{murphy2016,murphy2013ssrr}.

\section{Approach and scope}

This thesis focuses on the process of jointly planning sensing actions over
short time horizons (e.g. seconds) and across
large teams of robots, with a focus on sensing tasks arising from urban
search and rescue.
The proposed approach draws from sequential greedy algorithms for maximization
of submodular functions~\citep{nemhauser1978,fisher1978}
(discussed in detail in Chapter~\ref{chapter:background})
which are becoming increasingly popular for multi-robot sensor planning
problems~\citep{%
singh2009,atanasov2015icra,regev2016iros,zhou2019ral,jorgensen2017iros,
schlotfeldt2018iros,hollinger2009ijrr%
}.

Lack of reliable communication between robots and damage to infrastructure
also motivate our focus on distributed planning.
Specifically, robots may not be able to communicate reliably with centralized or
remote computational resources (e.g. a mobile base station or remote server)
so robots should be able to plan for themselves.
However, we note that some of the techniques we develop in this thesis could be
adapted to produce efficient centralized and parallel implementations.

While some works that apply submodular maximization to related sensor planning
problems allude to distributed
implementation~\citep{regev2016iros,atanasov2015icra}, planning time for greedy
techniques still scales at least linearly with the number of robots.
Challenges related to distributed variations of these algorithms have also
only begun to be addressed~\citep{%
  gharesifard2017,grimsman2018tcns,grimsman2018cdc%
}, and yet existing approaches still scale poorly in planning time or
suboptimality as the number of robots increases.
Conversely, this thesis proposes techniques for planning via submodular
maximization that scale to any number of robots,
and we prove that the techniques we propose maintain constant-factor
suboptimality for a variety of sensing tasks.

\begin{figure}
  \centering
  \def\svgwidth{\linewidth}
  \input{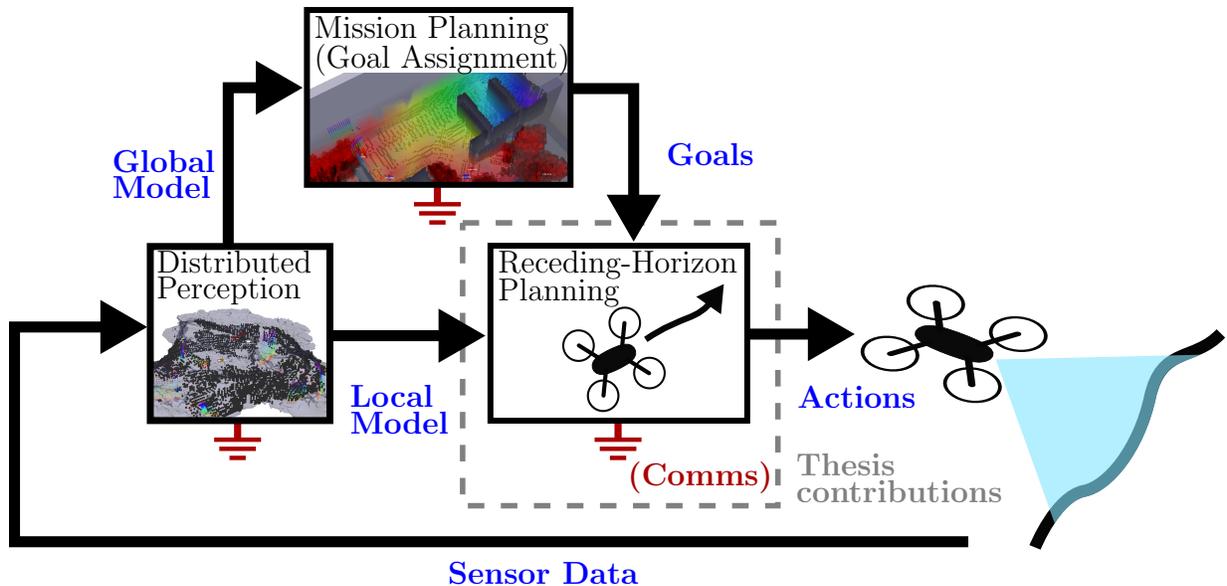}
  \caption[Diagram of a typical sensing system]{%
    Diagram of a typical sensing system.
    This thesis focuses on planning sensing actions for teams of robots
    over a receding-horizon
    (e.g. several seconds).
    The inputs to the receding-horizon planner are long-term goals for the
    sensing process and models of the environment that are locally consistent in
    space and time.
    The outputs of the receding-horizon planner are control actions,
    typically a smooth trajectory, which the robot will execute to collect
    sensor data.
  }%
  \label{fig:system_diagram}
\end{figure}

Specifically, we develop a receding-horizon approach
to sensor planning---illustrated in Fig.~\ref{fig:system_diagram}---whereby
robots collectively optimize sensing actions over a short time horizon,
typically several seconds.
In general, such planners take high-level goals and locally consistent
models of the environment as inputs and output control actions in the form of
smooth trajectories.
This thesis focuses primarily on the distributed aspect of these
receding-horizon optimization problems.
While we address some challenges related to receding-horizon
planning for \emph{individual robots}, that topic can also be addressed
separately~\citep{goel2019tr,goel2019fsr}.
We will also provide little detail on methods for high-level
mission planning (or goal assignment) and distributed perception which are each
necessary components of complete sensing systems and, themselves, important
topics of study.
Maintaining distributed representations of the environment can also incur
significant communication costs.
While there are ways to mitigate such communication costs either by
intelligently communicating or compressing sensor data%
~\citep{ganesh2017tr,corah2019ral}
or by distributed robots in ways that reduce the need for
communication~\citep{luo2019aamas}, the methods in this thesis will not address
this challenge.
Likewise, problems related to estimating robot position and states, particularly
simultaneous localization and mapping (SLAM), are out of scope both for passive
estimation and as an active control problem~\citep{indelman2014icra}.
Additionally, since our analysis will apply primarily to receding-horizon
sub-problems, translating improvements in solution quality on these sub-problems
into improvements in task performance---such as for robots to map an environment
more quickly---will itself pose an important challenge in this thesis.

Further, this thesis will apply the methods for receding-horizon planning which
we develop to two application areas: \emph{exploration} (autonomous mapping)
and \emph{target tracking}.
Here, exploration serves as a simple proxy for more complex tasks related to
search and rescue.
In particular, planning for exploration addresses the challenge of selecting
sensing actions while operating in an environment with unknown geometry.
On the other hand, the target tracking problems which we study are not
immediately relevant to the search and rescue domain.
Instead, these target tracking problems provide an opportunity to study a
special class of factored objectives; for target tracking this arises in an
objective which can be written as a sum over terms associated with each of the
targets.
More generally, this class of objectives can capture the multi-faceted sensing
tasks that arise in search and rescue.
In this sense, our analysis for such factored objectives could enable designers
to extend the systems we develop for exploration to simultaneously address a
variety of nuanced sensing tasks such as for locating survivors, assessing
structural integrity, or identifying hazards.

\section{Assumptions}

The formulations of the exploration and target tracking tasks will also make a
few simplifying assumptions.
These assumptions are not necessarily realistic in practice but can often be
relaxed with minor modifications to the approach.
The following assumptions are important to various parts of this thesis:
\begin{itemize}
  \item \textbf{Homogeneity:} We assume teams of robots are \emph{homogeneous}
    to simplify exposition, but our methods could be applied to heterogeneous
    teams with only minor modifications
  \item \textbf{Spatial locality:}
    A stricter assumption is that robots'
    \emph{incremental motions and sensor ranges are bounded}.
    However, we make this assumption primarily for the purpose of analysis of
    large teams of robots, and this assumption need only apply loosely to actual
    teams of robots.
    Additionally, spatial locality will arise in different forms for different
    tasks and will be treated on a case-by-base basis
  \item \textbf{Static environments:}
    For exploration, we assume that robots map \emph{static environments}.
    This assumption that environments are largely static while being searched is
    likely realistic---searchers would generally avoid disturbing objects and
    rubble to avoid causing collapses and endangering themselves or any
    survivors.
    However, robots may also have to avoid collisions with responders who are
    moving about the environment or may have to recognize minor changes in the
    map such as after responders clear some rubble or open a door.
    Thus, while the environment may remain predominantly static for the purpose
    of the sensing task, addressing extraneous dynamics in and around the
    environment will be left to future work
\end{itemize}


\section{Contributions and outline}
This thesis provides two kinds of contributions: development of scalable,
distributed algorithms with performance guarantees for multi-robot sensing and
development of sensing applications for exploration (autonomous mapping) and
target tracking.
Moreover, the central contribution of this thesis is a planning algorithm that
guarantees constant factor sensing performance on average for a variety of
sensing problems and requires only a fixed number of sequential planning steps
for any number of robots.
The rest of the thesis will involve implementing such algorithms, analysis
of relevant sensing objectives, and developing methods for operating in and
exploring unknown environments.
These contributions are outlined below:

\begin{itemize}
  \item \textbf{Planning for single- and multi- robot exploration:}
    Chapter~\ref{chapter:distributed_multi-robot_exploration}
    proposes a first attempt toward a distributed sensing algorithm for sensor
    planning via submodular maximization---by
    modifying to existing greedy methods~\citep{fisher1978}.
    The approach provides post-hoc guarantees and benefits from parallel
    computation when single-robot planning is expensive.
    We develop this approach, along with methods for single-robot planning based
    on Monte-Carlo tree search, to develop a receding-horizon planner for
    exploration with teams of aerial robots.
    And, \emph{the rest of this thesis applies similar approaches to
    single-robot planning, throughout}.
    Additionally, we guarantee that robots do not collide with each other or the
    surrounding environment and describe some weak conditions that are
    sufficient to guarantee liveness.
    Notably, this is also the only chapter to include experiments with
    physical robots
  \item \textbf{Scalable multi-agent coverage:}
    Chapter~\ref{chapter:scalable_multi-agent_coverage} presents a greedy
    algorithm (Randomized Sequential Partitions or \rsp{}) for coverage and
    any other submodular maximization problems which exhibit a certain
    higher-order monotonicity condition.
    Our algorithm guarantees constant factor suboptimality in the sensing
    objective (e.g. coverage) in expectation with fixed numbers of planning and
    communication rounds
    (versus one per robot for sequential greedy methods).
    This guarantee extends to any number of robots given certain conditions on
    spatial locality.
    \emph{To our knowledge ours is the first algorithm to provide such
    guarantees for a non-trivial class of submodular functions.}
    As such, the remainder of the contributions continue to develop applications
    and implementations of \rsp{} planning
  \item \textbf{Target tracking and quantifying inter-robot redundancy with
      factored objectives:}
    Chapter~\ref{chapter:target_tracking} introduces target tracking problems
    and applies similar methods for planning as the preceding chapters.
    Although we found in Chapter~\ref{chapter:scalable_multi-agent_coverage}
    that some sensing objectives satisfy a higher-order monotonicity condition
    that enables development of scalable planners with suboptimality guarantees,
    these results do not apply in general to some popular objectives (e.g.
    mutual information).
    However, we prove that some special cases of mutual information (those that
    can be factored as sums) satisfy similar suboptimality guarantees.
    We also provide detailed analysis for applying this result to target
    tracking problems.
    Further, this sum decomposition is relevant to the kinds of multi-objective
    problems which may arise when robots engage in more complex sensing tasks.
    This chapter also provides new analysis for approximate receding-horizon
    planning in real time that is relevant to sensing problems throughout this
    thesis.
    Last, simulation studies establish performance improvements for \rsp{}
    planning in terms of target uncertainty and demonstrate that the
    suboptimality guarantees remain well-behaved for large numbers of robots
    (with results for up to 96 robots, providing a $24\times$ reduction in the
    number of planning steps compared to planning sequentially)
  \item \textbf{Time-sensitive sensing in unknown environments:}
    Chapter~\ref{chapter:time_sensitive_sensing} revisits the problem of
    multi-robot exploration.
    We provide more in-depth analysis of objective functions and present one
    case of a mutual information objective that satisfies the monotonicity
    conditions we study in this thesis.
    This chapter also incorporates some changes in planner design that
    significantly improve on what
    Chapter~\ref{chapter:distributed_multi-robot_exploration} presents.
    Regarding simulation results, we find that improving completion times
    via sequential or \rsp{} planning
    (versus having no explicit coordination in planning)
    can be challenging.
    Still, we demonstrate that planning with \rsp{} improves coverage rates
    early in simulation trials and reliably improves solution quality
    (suboptimality)
  \item \textbf{Design and implementation of a distributed sensor planner:}
    While the previous chapters \emph{described} distributed algorithms,
    Chapter~\ref{chapter:distributed_communication} is the first to provide a
    distributed implementation of an \rsp{} planner.
    Toward this end, we provide detailed discussion of design decisions and
    results that investigate communication costs.
    Simulation results demonstrate this distributed, synchronous, anytime
    planner in exploration of an office-like environment, verify that the system
    behaves as expected, and identify a 5\% improvement in task completion time
    for coordination via \rsp{}, albeit for a small set of experiments
\end{itemize}

\subsection{Code release}

Some of the source code for these contributions is available via the BSD
license.
The numerical experiments for coverage in
Chapter~\ref{chapter:scalable_multi-agent_coverage},
the target tracking simulations in Chapter~\ref{chapter:target_tracking},
and the communication study in Chapter~\ref{chapter:distributed_communication}
rely on a common code-base,\footnote{%
  \url{https://github.com/mcorah/MultiAgentSensing}
}
written in Julia.
We have also released the distributed implementation of \rsp{}\footnote{%
  \url{https://github.com/mcorah/distributed_randomized_sequential_partitions}
}
that Chapter~\ref{chapter:distributed_communication} describes which has
been implemented in C++ via ROS.
This release encompasses the distributed scheduling and communication
aspects of \rsp{} and excludes methods for exploration.
We hope that the design of this package can enable incorporation of \rsp{}
methods alongside various single-robot receding-horizon sensor planners with
minimal effort.

\chapter{Sensing Problems}
\setdatapath{./fig/problem}
\label{chapter:problem}

Before developing the contributions of this thesis, let us pause to discuss the
nature of the problems that we study.
So far, the introduction (Chapter~\ref{chapter:introduction}) has discussed
sensing problems at a high level and from the perspective of applications,
particularly search and rescue.
This chapter will focus more closely on problem form and evaluation
for time-sensitive sensing and for target tracking.
Later, Chapter~\ref{chapter:background} will provide technical background
methods and detailed discussion of related works.

We provide one caveat: the discussion in this chapter is primarily for the
purpose of empirical evaluation of the systems we propose and to broadly
characterize how features of sensing problems relate to the requirements of
potential solution methods.
Where some sensing processes admit algorithms with formal
guarantees~\citep{golovin2011jair}, features such as robot motion confound
existing methods.
Likewise, the receding-horizon techniques and other methods that we develop
involve approximations and heuristics, and we would not expect the
resulting systems to satisfy rigorous suboptimality guarantees.


\section{Time-sensitive sensing}
\label{sec:time_sensitive_sensing}

The \emph{time-sensitive sensing} moniker intends to provide some formalism and
structure to the problems we study in this thesis.
Chapter~\ref{chapter:introduction} provided motivation for why completing
sensing tasks quickly can be important.
However, designers might consider other criteria.
For example, even early methods for exploration based on
frontiers~\citep{yamauchi1997} could be described as \emph{complete} in the
sense that the system will eventually explore the entire environment under
reasonable assumptions.

In defining the time-sensitive sensing problem, we take a cue from
\citet{golovin2011jair} and describe the controller using notation for policies.
However, in line with the rest of this thesis and to provide a direct
description of the system, we also draw on notation from control theory.

Consider an \emph{unknown} static environment
$\environment\!\sim\!\environments$
drawn from some \emph{known} distribution over possible environments
$\environments$.
One or more robots
(this problem definition can be interpreted as describing arbitrary systems,
consisting of any number of robots)
navigate the environment and
obtain observations according to the known dynamics and observation model
\begin{align}
  \state_t &= \dynamics(\state_{t-1}, \control_{t-1})
           &
  \observation_t &= \observationfunction(\state_t, \environment)
  \label{eq:problem_dynamics}
\end{align}
where $\state_t$, $\observation_t$, and $\control_t$
are the state, observation, and control input at the (discrete) time $t$.
Like the environment, the initial state $\state_0\!\sim\!\initialstates$
is random and drawn from a known distribution.
Given that the robot(s) learn about the environment through the sequence of
observations,
the controller is a policy:\footnote{%
  For now, we write the policy and system model as if deterministic.
  However, the reasoning we present also applies to stochastic systems and
  policies, as we study elsewhere in this thesis.
}
\begin{align}
  \control_t = \policy(\states_{0:t-1}, \observations_{0:t-1})
  \label{eq:problem_policy}
\end{align}
where the capital letters represent vectors of states and
observations.
This policy states that the robot makes decisions at each time $t$ given
the available information, the states, and observations.
Further, as the robot navigates the environment, it must also avoid collisions
or otherwise unsafe states.
In general, the robot must remain in some set of safe states
$\safespace(\environment)$ at all times.
We seek to minimize the expected completion time\footnote{
  Similar theoretic works~\citep{chekuri2005,singh2009,golovin2011jair}
  typically use a more abstract monotonic or modular cost which could represent
  quantities like time or energy.
  Although planning with energy costs is an important topic of
  study~\citep{tabib2016iros,sudhakar}, such general cost models do not arise in
  this thesis.
}
(given the distribution over environments and starting positions)
for a given sensing task---say mapping a building.
Progress in the task is measured by the sensing quality function
$\sensingquality(\states_{0:t},\observations_{0:t})$
which depends on states and observations while completion is modeled by a quota
$\quota(\environment)$ which depends on the environment.
For example, this sensing quality may represent how much of a building that a
robot has observed or mapped by a given time, and the quota may require the
robot to map a certain fraction of the building.
Putting this all together, the time-sensitive sensing problem is as follows:
\begin{align}
  \min_{\policy}~& \E_{\environment,\state_0}[T]
  \nonumber \\
  \mathrm{s.t.}~
             & \sensingquality(\states_{0:T},\observations_{0:T})
             \geq \quota(\environment)
  \nonumber \\
              & \state_t \in \safespace(\environment)
  \nonumber \\
              & \control_t = \policy(\states_{0:t-1}, \observations_{0:t-1})
  \nonumber \\
              & \state_t = \dynamics(\state_{t-1}, \control_{t-1})
  \nonumber \\
              & \observation_t = \observationfunction(\state_{t}, \environment)
  \nonumber \\
              &\mbox{and the above for all } t \mbox{ in } \{1\ldots T\}
              \label{eq:time_sensitive_sensing}
\end{align}

Solving this problem, even approximately, is challenging.
This problem is imbued with challenges from several different fields such as
control theory, artificial intelligence, and combinatorial optimization as
we hint by mixing notation from these fields.

\subsection{Characteristics of time-sensitive sensing}
The following paragraphs discuss some of the challenges and
properties of time-sensitive sensing problems.

\subsubsection{Inference and uncertainty}

The robot gains information about the environment by collecting observations
from a variety of states.
In particular, the set of possible environments that are consistent with a
collection of states $\states_{0:t}$ and $\observations_{0:t}$ is\footnote{
  We could write this more generally for noisy observations using Bayes' rule.
  However, some relevant solution methods~\citep{javdani2013icra} only apply to
  deterministic models.
  We emphasize the deterministic case to avoid unduly restricting applicable
  solutions and because the sensing noise for common depth sensors in
  mapping~\citep{henderson2020icra}
  is frequently small compared to range (meters) and the
  environment discretization (typically $\SI{10}{\centi\metre}$).
}
\begin{align}
  \{ \environment : \observations_{0:t} = \observationfunction(\states_{0:t},
    \environment)
  \mbox{ for all possible environments } \environment \}.
  \label{eq:possible_environments}
\end{align}
Referring back to~\eqref{eq:time_sensitive_sensing}, several features of the
problem formulation depend on the environment:
the observations $\observationfunction(\state, \environment)$,
the safe set $\safespace(\environment)$, and the quota $\quota(\environment)$.
Effects of uncertainty in observations will naturally arise frequently as
this thesis focuses on sensing problems.
On the other hand, uncertainty in the quota could be interpreted as complicating
certain solution strategies, but the quota will only appear again in
experimental evaluation where its value will be known.

Although, we will discuss the safe set less frequently, the uncertainty in which
actions are safe has strong implications on system performance such as by
bounding maximum speeds~\citep{goel2019fsr}.
Likewise, this uncertainty implies that we cannot plan complete paths
(such as to map a building) a priori because those paths may not be safe.
Thus, methods for path planning cannot solve~\eqref{eq:time_sensitive_sensing}
on their own.
We will continue this discussion later in this section in terms of \emph{safety}
and \emph{feasibility}.


\subsubsection{Form of the sensing quality}
The sensing quality $\sensingquality(\states_{0:t},\observations_{0:t})$ is
immediately in terms of known quantities: states and observations.
However, the sensing quality is also a function of the states and environment
\begin{align}
  \sensingquality(\states_{0:t}, \observations_{0:t}) &=
  \sensingquality(\states_{0:t},
  \observationfunction(\states_{0:t}, \environment)),
  \label{eq:sensing_quality}
\end{align}
wherein we abuse notation by applying the observation function to a vector of
states.
Given this latter form, we can make inferences about future observations.
In this sense, future sensing progress is uncertain to the extent that the
environment remains uncertain but can also be optimized by selecting robot
motions.

Numerous recent works seek to optimize common measures of future sensing
progress.
However, common sensing functions (such as those we discuss in this thesis)
have sometimes surprising differences in important functional properties.
For this reason, we do not yet specify the functional form of
the sensing quality $\sensingquality$.
For example, \citet{golovin2011jair} demonstrate that certain sensor
coverage functions have a monotonicity property called
\emph{adaptive submodularity}
even though important measures of information gain do
not~\citep{golovin2010nips,chen2015aaai}.
We will frequently revisit similar functions for coverage and information gain.
As these often have similar properties, distinctions such as this are important.
Here, adaptive submodularity is one factor that determines whether greedy
methods can solve~\eqref{eq:time_sensitive_sensing} to
near-optimality~\citep{golovin2011jair}.
However, we will also find that adaptive submodularity does not apply to many of
the problems in this thesis.

\subsubsection{Robot dynamics and available observations}

The principal effect of the dynamics model is that it changes which information
gathering actions are available at a given time.
Here, the information available at a given state consists of state,
observation pairs $(\state,\observation)$.
Robots gain information about the environment by visiting states and inspecting
such pairs as $\observation=\observationfunction(\state,\environment)$.
The information gathering actions that the robot has available at a state
$\state$ consists of the set of states the robot can visit next
\begin{align}
  \{ \state' : \state'=\dynamics(\state, \control)
  \mbox{ for all control inputs } \control \}
  \label{eq:information_actions}
\end{align}
As such, the set of available observations at any given time depend on the prior
selections.
This excludes some methods for adaptive
sensing~\citep{golovin2011jair,choudhury2017rss} which require the set of
information gathering actions not to change over time.

Another important property of the system dynamics $\dynamics$ is whether it produces
a directed or undirected relationship between states.
The system dynamics \emph{induce an undirected relationship} on the system
states if, for any possible state $\state_1$ and control $\control_1$ pair
\begin{align}
  \state_2 &= \dynamics(\state_1, \control_1) \nonumber
  \intertext{there exists some $\control_2$ such that}
  \state_1 &= \dynamics(\state_2, \control_2).
  \label{eq:undirected_dynamics}
\end{align}
Alternatively, the dynamics produce a \emph{directed} relationship if the above
does not hold.
Which of these properties holds can determine which kinds of algorithms and
analysis~\citep{chekuri2005,chekuri2012talg} apply to the planning problems we
discuss in this thesis.

Many of the systems in this thesis will involve trivialized kinematic models
with undirected dynamics.
However, Chapters~\ref{chapter:distributed_multi-robot_exploration}
and~\ref{chapter:distributed_communication}
each provide results for systems with more general, directed dynamics
models.
Likewise, the methods for distributed coordination which we develop are
independent of the robots' dynamics.

\subsubsection{Stopping and inevitability of collision}
\label{sec:safety_and_stopping}
Let us consider how dynamics affect safety.
Often, we will consider problems where robots can always stop
immediately.
In terms of control theory, this means that for all possible states $\state_t$
there exists some control input $\control_t$ that stops the robot so that
\begin{align}
  \state_{t+1} = \state_{t} = \dynamics(\state_t, \control_t) \quad
  \mbox{(for some $\control_t$)},
  \label{eq:invariant_state}
\end{align}
ensuring that the robot remains in the same state.
Individual states where \eqref{eq:invariant_state} holds are called
\emph{invariant states}\footnote{%
  Similar properties also apply for invariant \emph{sets} where each state in
  the set has some control action which can keep the robot in the invariant set.
}~\citep{rawlings1993tac}.
Not all states are invariant states in practice---consider a robot with non-zero
velocity and non-trivial dynamics.
Now, consider a robot flying rapidly toward a brick wall;
that robot will eventually reach a point of no return where it cannot stop
without hitting
the wall---such is the heightening intensity of two children playing ``chicken''
who must decide when to act to avoid collision.
States past such points of no return,
where no sequence of control inputs can prevent the robot from leaving
the safe set $\safespace(\environment)$, are \emph{inevitable collision states}.
Controllers that ensure safety by preventing robots from entering inevitable
collision states are a frequent
topic of study
in contemporary control
theory in the form of control barrier
functions~\citep{prajna2004,ames2016tac,notomista2018ral}
and also
for applications to mobile
robots~\citep{watterson2015,liu2016icra,janson2018rss,goel2019tr,luo2019}.
This will arise in Chapter~\ref{chapter:distributed_multi-robot_exploration}
where we will maintain safe stopping actions for all robots at all times.
Further, some works have begun to investigate the relationship between safety
and inference in mapping~\citep{janson2018rss}.
For example, a robot can take a turn widely while accelerating around a corner
to obtain advance warning of what lies beyond.
Now, building on these questions about safety, we can also ask whether a
robot will safely complete a sensing
task---whether~\eqref{eq:time_sensitive_sensing} is feasible.

\subsubsection{Feasibility, failure, and completeness}

Our definition of time-sensitive sensing~\eqref{eq:time_sensitive_sensing} sets
a high bar for producing even just an approximate solution.
A feasible policy must:
\begin{itemize}
  \item Complete the sensing task (or else the completion time is infinite),
  \item Guarantee that the robot will remain in the safe set,
  \item And must do so for all of the environments that remain
    consistent with prior observations~\eqref{eq:possible_environments}.
\end{itemize}
Given an instance of~\eqref{eq:time_sensitive_sensing}, we may ask several
relevant questions about feasibility:
\begin{itemize}
  \item Does a feasible solution exist?
  \item Given a policy, does that policy constitute a feasible solution?
    (A policy that is feasible and thereby always eventually finishes the
    sensing task is said to be \emph{complete})
  \item Does any of the above hold for an individual environment and starting
    position pairing or a restricted collection of such pairs?
  \item Does any of the above hold for a given partial history
    $(\states_{0:t}, \observations_{0:t})$?
  \item Is the converse true for any of the above?
    (e.g. Is there no feasible solution?)
\end{itemize}
We will answer a small subset of this class of questions.
Questions about safety already answer parts of the converse forms
(e.g. Is failure imminent?
Can a given policy violate the safety constraint?).
Classical frontier-based exploration~\citep{yamauchi1997} approaches are also
complete given reasonable assumptions:
The robot will navigate to and observe unknown space until there is
nothing left to observe.
We will then attempt to replicate aspects of such classical approaches
when considering more nuanced problems involving dynamics, unstructured
environments, and cameras with limited fields of view.

\subsubsection{Decision processes and adaptivity}
\label{sec:problem_adaptivity}

Time-sensitive sensing problems form \emph{decision processes} by nature
of control via policies $\policy(\states_{0:t}, \observations_{0:t})$,
functions of prior decisions and observations, rather than fixed sequences of
control actions.
However, does \eqref{eq:time_sensitive_sensing} readily fit the form
of a classical Partially Observable Markov Decision Process
(POMDP)~\citep{kaelbling1998}?
At this point, it is important to realize that the reward (or, more accurately,
cost) in \eqref{eq:time_sensitive_sensing} is the completion time.
Although time is easy to compute, completion depends on the sensing quality
$\sensingquality(\states_{0:t}, \observations_{0:t})$
which is a function of the sequence of states.
Adapting these problems to the form of a traditional POMDP then requires an
exponential number of states for the same reasons as described by
\citet{golovin2011jair} regarding the \emph{adaptive submodular maximization}
problems which they define.
Alternatively, certain POMDP variants allow for more complex
rewards which we discuss in Sec.~\ref{sec:adaptation}

\subsection{The receding-horizon approximation}
\label{sec:receding_horizon_planning}
Because solving full time-sensitive sensing problems is challenging, this
thesis takes the approach of solving a simpler sub-problem.
Given prior states and actions $\states_{0:t}$ and $\observations_{0:t}$
we seek to maximize the sensing quality over the next $L$ steps for a fixed
sequence of future control actions $\control_{1:L}$
(known as the \emph{receding-horizon}) and adopt the policy:
\begin{align}
  \pi(\states_{0:t},\observations_{0:t}) = \argmax_{\control_{1:L}}~&
  \E_\environment[ \sensingquality(\states_{0:t+L},\observations_{0:t+L}) ]
  \nonumber \\
  \mathrm{s.t.}~
              & \state_{t+l} \in \safespace(\environment)
  \nonumber \\
              & \state_{t+l} = \dynamics(\state_{t}, \control_{l})
  \nonumber \\
              & \observation_{t+l} = \observationfunction(\state_{t+l}, \environment)
  \nonumber \\
              &\mbox{for all } l \mbox{ in } \{1\ldots L\}
              \label{eq:receding_horizon_sensing}
\end{align}
Receding-horizon methods such as this are common in robotics and
control~\citep{rawlings1993tac}
and for sensing problems like those we discuss in this
thesis~\citep{charrow2015rss,lauri2015ras,bircher2018receding,schlotfeldt2018ral}.
We adopt this approach as a heuristic and would not expect to obtain
approximation guarantees.
However, we will be able to guarantee certain properties of the resulting
system such as safe operation.
Furthermore, this sub-problem does not involve adaptation and therefore no
longer requires specialized analysis for adaptation~\citep{golovin2011jair}.
This feature can then enable application of informative path planning techniques
for non-adaptive problems.

\subsubsection{Informative path planning}
The solution to \eqref{eq:receding_horizon_sensing} corresponds to a path
(as $\state_{t:t+L}$ is a function of $\control_{1:L}$) instead of a policy.
This path may be directed or undirected as discussed earlier.
Assuming that $\E_\environment[ \sensingquality(\states_{0:t+L},\observations_{0:t+L}) ]$
satisfies certain properties, existing methods for informative path
planning can solve~\eqref{eq:receding_horizon_sensing} with various
guarantees~\citep{chekuri2005,chekuri2012talg,singh2009,hollinger2014ijrr,zhang2016aaai}.
Much of this thesis will focus on how to apply those and similar
methods to large instance of~\eqref{eq:receding_horizon_sensing} where the
states and dynamics represent \emph{teams of robots}.

\subsection{Exploration}
\label{sec:exploration}

In this thesis, \emph{exploration} refers to the process of mapping some
environment.
This section presents a simplified but also useful model of the exploration
process.
To begin, we model the environment as a sequence of $\numcells$ cells
$\environment=[c_1,\ldots,c_{\numcells}]$ which respectively represent free and
occupied space
$c_i \in \{0,1\}$.
This environment is drawn from some probability distribution, and we will make
no assumptions about that distribution at this point.

While navigating this environment the robot must avoid occupied cells while
observing cell occupancy values with a camera or other sensor.
For some state $\state$, the camera function
$\camera(\state, \environment) \subseteq \{1,\ldots,\numcells\}$
returns the set of cells which the robot observes from that state and given the
specific environment.
This camera function may then refer an arbitrary projective camera
model.\footnote{%
  An appropriate sanity condition for $\camera$ would be to require the set
  of cells which the robot observes $A=\camera(\state, \environment)$ to depend
  only on the state and the values of the cells that the robot observes.
  Consider the occupancy values of those cells $\environment_A$.
  We can require for all environments $\environment'$ that whenever the values of the
  observed cells are the same $\environment_A'=\environment_A$, then set of
  cells that the robot observes $A=\camera(\state, \environment')$ must also be
  the same.
  This avoids inadvertently providing information about cells outside of the
  field of view.
}
Then, the observation function indicates which cells the robot observes along
with their values
$\observationfunction(\state, \environment) = \{(i, c_i) : i \in \camera(\state,
\environment)\}$.
Similarly, the occupancy function
$\occupancy(\state) \subseteq \{1,\ldots,\numcells\}$
determines which cells that the robot occupies.
A robot is in the safe set if all the cells it occupies are free
$\safespace(\environment) =
\{ \state : c_i = 0 \mbox{ for all } i \in \occupancy(\state) \}$.
We will reward the robot for the number of cells that it observes.
So, the sensing quality in exploration for given states $\states_{0:t}$ and
observations $\observations_{0:t}$ is
\begin{align}
  \sensingquality(\states_{0:t}, \observations_{0:t}) =
  \left|
  \bigcup\nolimits_{i=0}^t \camera(\state_i, \environment)
  \right|.
  \label{eq:exploration_coverage_problem_chapter}
\end{align}
When written as a function of a
set~\eqref{eq:exploration_coverage_problem_chapter} is a kind of coverage
objective and has useful monotonicity properties
(Chapter~\ref{chapter:background} discusses both coverage and monotonicity in
detail).
Additionally, the expectation, which appears in the receding-horizon
problem~\eqref{eq:receding_horizon_sensing}, retains similar properties.
Yet, unlike some coverage objectives~\citep{golovin2011jair},
\eqref{eq:exploration_coverage_problem_chapter} is not necessarily adaptive
submodular
(see Appendix~\ref{appendix:not_adaptive_submodular}).

\todo{Discuss exploration in the context of time-sensitive sensing}

\section{Time-average tracking}

Chapter~\ref{chapter:target_tracking} deviates from the rest of this thesis by
focusing on sensor planning for observation of \emph{dynamic systems}.
Unlike exploration where mobile (dynamic) robots seek to completely map a static
environment, (target) tracking problems are ongoing and can be evaluated based
on average performance.
In this case, we can consider tracking a target whose state $\targetstate$
evolves according to a Markov process so that
$\targetstate_t \sim \P(\targetstate_t | \targetstate_{t-1})$,
(here, the super-script ``t'' refers to target states)
although more general processes are also relevant to this discussion.
In this case, the sensing quality
$\sensingquality(\states_{0:t}, \observations_{0:t})$
might correspond to the number of times that the robot has observed a target or
a measure of the uncertainty in the target's entire trajectory
$\targetstates_{0:t}$.
Adapting \eqref{eq:time_sensitive_sensing} as appropriate, the time-average
performance is
\begin{align}
  \max_{\policy}~& \lim_{t \rightarrow \infty}
  \frac{ \E[\sensingquality(\states_{0:t}, \observations_{0:t})]
  }{ t },
  \label{eq:target_tracking_problem}
\end{align}
given some observation model and subject to robot dynamics.
Then, given the focus on time-average performance, the tracking problems
we study (unlike exploration) are not time-sensitive sensing problems as
described in Sec.~\ref{sec:time_sensitive_sensing}.

We will now discuss these problems briefly because they are of secondary
importance to this thesis and because many of the same properties and challenges
apply.

\subsection{Long duration Markov processes}
The target tracking problem \eqref{eq:target_tracking_problem}, as described, is
a kind of long
duration Markov Decision Process (MDP)~\citep{arapostathis1993}, and if the
target states $\targetstates$ are not known exactly, such tracking problems are
also \emph{partially observable} as for time-sensitive sensing problems.
Here, the key feature of the problem is that \eqref{eq:target_tracking_problem}
focuses on average performance for all time and does not favor near-term gains.
Although some solution methods exist for these problems, the problems that we
study also incorporate features that make these problems difficult to solve such
as very large state and action spaces.

\subsection{Ergodicity}
One important property in such problems is \emph{ergodicity}.
For an ergodic system, the distribution of states for a single trajectory, taken
over a very long duration, is the same as the distribution  for many
trajectories.
In short, ergodicity states that no transitions in a system have irrevocable
effects.
This could refer to termination conditions or a system that falls off a ledge
and cannot get back up again.

Long duration problems are typically studied under various sorts of ergodicity
conditions~\citep{arapostathis1993}.
In this sense, ergodicity conditions determine
whether~\eqref{eq:target_tracking_problem} describes rewards for an
individual trajectory or whether rewards could vary greatly across even very
long trials.
Unsurprisingly, ergodicity has also been used in heuristics to solve similar
problems~\citep{mathew2011,miller2015tro,ayvali2017iros}.

\subsection{Information dynamics}
We briefly note that concepts from information theory and information dynamics
are highly relevant to target tracking problems and will go into more detail on
related concepts later in this text~\citep{cover2012,bossomaier2016}.
Information theory can describe how quickly uncertainty in target states can
grow~\citep{cover2012} and can provide further connections to ergodicity or
characterize interactions between system components~\citep{bossomaier2016}.

\subsection{Stability}
Stability properties are also relevant to tracking problems.
In this case stability can describe whether robots may travel far from the
targets that they are tracking or whether uncertainty can grow
indefinitely~\citep{ryan2010ras}.
Such conditions will generally produce worst case rewards
in~\eqref{eq:target_tracking_problem}
(such as zero or infinitely negative reward).
Although we will not seek to characterize stability in the systems we study,
this property characterizes the kind of behavior that could arise in results.

\section{Ramifications and challenges of multi-robot sensing}

In an abstract sense, the discussion so far applies to both individual
robots and multi-robot teams when viewed as a collective.
Yet, this chapter has not explicitly addressed the ramifications of multi-robot
sensing.
To begin, the following chapters will describe various methods to decompose and
plan for multi-robot sensing problems.
These methods are all much more efficient than applying planners for individual
robots to the joint state space of the entire team.

Additional challenges and constraints also arise when considering multi-robot
teams.
For example, safety constraints (Sec.~\ref{sec:safety_and_stopping}) can be
amended to account for inter-robot collisions.
We also encounter new challenges when considering distributed planning (via
processors onboard each robot) and accounting for communication between robots.
Regarding distributed planning, robots should be able to obtain solutions in a
timely manner, preferably while taking advantage of parallel computation.
Likewise, a distributed planner may require systems for sharing sensor data or
maintaining consistent beliefs via communication between robots.
This, in turn, leads to challenges with respect to communication as robots may
have to maintain connectivity across the team or respect bandwidth constraints.
We will begin to explore these challenges next with the background and
related work (Chapter~\ref{chapter:background}), and later
Chapter~\ref{chapter:distributed_communication} will address challenges related
to communication more directly.

\chapter{Background and Related Work}
\setdatapath{./fig/background}
\label{chapter:background}

This chapter will establish the theoretic and practical foundations for this
thesis and for sensing and information gathering in unknown
environments.
We will begin with a high-level discussion of sensor planning, exploration, and
navigation in unknown environments (Sec.~\ref{sec:background_autonomy}).
This section builds somewhat on the challenges for sensing problems
(Chapter~\ref{chapter:problem}) and introduces related works in
this area.
The foundations for questions about information and uncertainty in
robotics come from information theory (Sec.~\ref{sec:information_theory}).
More directly, there is a wide body of literature related to robotics that
applies information theory to planning and control for sensing and information
acquisition (Sec.~\ref{sec:robotic_information_gathering}).
And for the purpose of this thesis, we are particularly interested in
applying techniques for sensing and information gathering to multi-robot
settings (Sec.~\ref{sec:multi-robot_coordination}).
Many sensing objectives, especially those based on information theory, share
useful monotonicity properties.
One of the best known monotonicity properties is submodularity, and many single-
and multi-robot planning problems can be solved efficiently and near-optimally
using techniques for submodular maximization
(Sec.~\ref{sec:submodular_maximization}).
In particular, we are interested in applying methods for submodular maximization
to distributed planning for multi-robot teams.
There are already a number of works on parallel and distributed
submodular maximization (Sec.~\ref{sec:distributed_parallel_submodular}).
However, not all of these works are readily applicable to multi-robot settings.
Further, no existing works are able to scale to large numbers of robots
due to increasing planning time.
Therefore, in addition to developing methods for aerial autonomy and
exploration, this thesis also presents new methods for submodular maximization
that can scale to arbitrary numbers of robots.

\section{Autonomy and exploration in unknown environments}
\label{sec:background_autonomy}

While sensing goals and objectives can vary significantly, we expect urban
emergency and disaster response tasks to feature cluttered and at least
partially unknown environments and refer to scenarios where mapping is the
primary sensing task as exploration problems.
We may then cast navigation and mapping in unknown environments as a
sensing problem~\citep{charrow2015icra,julian2014} and plan to maximize some
notion of information gain.
However, operation in unknown environments is distinguished by several unique
challenges:
\begin{itemize}
  \item Robots must avoid collision with objects and each other;
  \item The navigable subset of the environment is revealed incrementally while
    the robot observes free and occupied space;
  \item Robot states and positions are uncertain;
    and
  \item Environments may be large and informative observations distant.
\end{itemize}
Further, the non-trivial dynamics of aerial robots exacerbate these challenges.
The rest of this section will address information-based formulations and each of
these challenges in turn.

\subsection{Safe navigation}

Safety reduces to planning collision-free paths for slow-moving ground robots
with large fields of view~\citep{yamauchi1997,stachniss2005rss,lauri2015ras},
and authors frequently overlook challenges related to safety constraints.
On the contrary, the aerial robots that we study can move quickly and cannot
stop instantaneously.
These same robots can also move side-to-side, outside of the camera field of
view, and may not be able observe what they are moving toward.
Constraints on reachability that ensure set-invariance~\citep{rawlings1993tac}
can ensure that robots always avoid collisions with the environment and each
other or can ensure that robots can return to a starting
position~\citep{fridovich2019icra}.
Safety requirements also constrain how quickly robots move
toward~\citep{goel2019tr,goel2019fsr} and plan through~\citep{janson2018rss}
unmapped space.
As such, learning to avoid collisions~\citep{richter2017rss} can also improve
speeds to the extent that the learner can generalize about the environment.

\subsection{State uncertainty}

Thus far, we have emphasized observations of the state of the
environment rather than the states of the robots themselves.
As studied in robotics, uncertainty in the states of robots
leads to
simultaneous localization and mapping (SLAM) problems
and some additional challenges: estimates of the
robot state can drift dangerously over time, and uncertainty in the
state history contributes to uncertainty in the map.
Direct observation of position by GPS or motion capture can mitigate or
eliminate these challenges.
Except, GPS access can also be limited when operating in urban
environments~\citep{murphy2008} which motivates development of reliable SLAM
systems for the problems we study.

Robots can also plan to minimize uncertainty
via information-theoretic methods for sensor
planning~\citep{stachniss2005rss,indelman2014icra,atanasov2015icra} forming
methods for what is known as active SLAM.
While classical SLAM algorithms that apply particle filters to two-dimensional
SLAM readily admit joint optimization to minimize uncertainty in the map and
robot state~\citep{stachniss2005rss}, methods for sensing with modern
smoothing-based frameworks~\citep{indelman2014icra,atanasov2015icra} are
somewhat more abstract and often represent uncertainty in environment geometry
indirectly.
Additionally, some works in this area employ the same greedy algorithms for
submodular maximization~\citep{atanasov2015icra,indelman2014icra} which we study
in this thesis.
However, submodularity generally only
applies to robots' histories of states and observations
but not to future states, due to dependence on the robots' own control
actions~\citep{atanasov2015icra}.

As such,
even though this thesis does not address state uncertainty explicitly, methods
similar to those we propose could play a role in systems for active SLAM.
Alternatively, designers may also wish to address uncertainty in future states
through constraints on uncertainty as a matter of safety, and such constraints
could be incorporated into the problem formulations we propose.

\subsection{Information-based exploration}

The mobility of aerial robots exploring some environment contrasted against
limitations on field of view
motivate methods that can predict information gain at different view points
such as with mutual information~\citep{charrow2015icra,julian2014,zhang2019icra}
and coverage-like~\citep{delmerico2018auro,burgard2005tro} objectives.
Authors have also applied a variety of such objectives along with a numerous
techniques for planning paths and
actions~\citep{charrow2015rss,lauri2015ras,bircher2018receding}.

We will apply methods based on motion
primitives~\citep{tabib2016iros,nelson2018auro} and Monte-Carlo tree
search~\citep{lauri2015ras,chaslot2010}, and, more recently, our work has
addressed motion primitive design in this
framework~\citep{goel2019tr,goel2019fsr}.
Additionally, Chapter~\ref{chapter:time_sensitive_sensing} will revisit
objective design and draw connections between a number of relevant objectives
for robotic exploration.

\todo{Consider discussing learning}

\subsection{Frontiers and the distribution of information}
\label{sec:background_frontiers}

Some of the earliest works on exploration~\citep{yamauchi1997} emphasize the
geometry of the environment: observations of unknown space by a robot in known
free space all pass through
the \emph{boundary between the known free space and unknown space} which is
called the \emph{frontier}.
Providing the weak assumption that robots are always able to observe nearby
frontiers, navigating toward the nearest frontier can serve
as a simple and reliable technique for exploring an environment to completion.
Although frontiers can be useful for selecting local control
actions~\citep{cieslewski2017iros,delmerico2018auro}, they also model the
distribution of information sources in exploration tasks:
after mapping one
part of an environment, a robot may traverse a significant distance to
reach the next nearest frontier.
In this sense, we can generalize frontiers as sets of sufficiently informative
points in the robot state space~\citep{corah2019ral}, and robots can alternate
between maximizing information gain locally and navigating to new
information-rich regions of the environment.
Moreover, in either case, designers may consider routing robots between
information-rich regions or frontiers~\citep{kulich2011icra} to reduce travel
distance or completion time.

\subsection{Multi-robot exploration}

Most works on multi-robot exploration focus on
assignment of frontiers~\citep{burgard2005tro,butzke2011planning} and
environment regions~\cite{solanas2004iros}
or maintaining maps via communication~\citep{corah2019ral,chedly2018}.
Although authors do not typically apply arguments based on submodularity in this
domain, greedy assignment mechanisms
are common~\citep{burgard2005tro,butzke2011planning} but at the level of task
assignment rather than receding-horizon planning as in our approach
(which we would expect to run at faster rates than global assignment processes).

As such, designers seeking to map an environment quickly may deploy large
numbers of robots that interact on short time scales, and this thesis addresses
the concomitant challenges via coordination over finite horizons.
Our approach then complements works on exploration that focus on
larger spatial and temporal
scales~\citep{burgard2005tro,butzke2011planning,mitchell2019icra}, and
such methods may run in outer loops or on centralized nodes to
supplement the distributed planners we propose
(as in Fig.~\ref{fig:system_diagram}).

\subsection{Connectivity and communication}

Exploration scenarios, such as for search and rescue~\citep{murphy2016}
have the potential to impair communication between robots via damage to
infrastructure and harsh environments (e.g. due to stone and concrete).
Teams of robots may then jointly plan to explore the environment and to
maintain or regain communication links to communicate data to the operator and
each other~\citep{banfi2018tro,banfi2018auro,tatum2020ms}.
Such, methods for maintaining communication are diverse:
\citet{banfi2018auro} allow robots to become disconnected but enforce recurrent
connectivity constraints to enable periodic communication;
\citet{tatum2020ms} proposes a method where robots jointly explore and place
communication nodes with emphasis on subterranean environments;
or designers may simply require the team to maintain connectivity at all
times~\citep{michael2009iser,wang2016cdc}.

This thesis does not address connectivity directly.
However, the planners we propose are robust to loss of communication,
and our approach could reasonably be extended by applying any of the
cited connectivity-aware methods as part of a high-level planning component.

\section{Information theory}
\label{sec:information_theory}

Having discussed a few sensing problems in robotics, let us delve into the
foundations for quantifying uncertainty in states and information gained via
sensing actions.
To begin,
entropy quantifies the uncertainty in a random variable $X$---such as the
position of a robot or a target---with the
average number of bits necessary to encode a realization of that variable
and is denoted as
\begin{align}
  \H(X) = \sum_{i} -\P(X=i)\log_2\P(X=i).
  \label{eq:entropy}
\end{align}
The joint entropy then expresses the combined uncertainty of two (or more)
variables $X$ and $Y$ and simply involves another sum:
\begin{align}
  \H(X,Y) = \sum_{i}\sum_{j} -\P(X\!=\!i,Y\!=\!j)\log_2\P(X\!=\!i,Y\!=\!j).
  \label{eq:joint_entropy}
\end{align}
Then, conditional entropy encodes the expected uncertainty in $X$
given that $Y$ will be observed:
\begin{align}
  \H(X|Y) &= \sum_{i}\sum_{j} -\P(X=i,Y=j)\log_2\P(X=i|Y=j).
  \label{eq:conditional_entropy}
  \intertext{This can be written as an expectation over possible outcomes
  for $Y$:}
          &= \sum_{i}\sum_{j} -\P(Y=j)\P(X=i|Y=j)\log_2\P(X=i|Y=j)
          \nonumber
          \\
          &= \E_{j\sim Y}\left[
            \H(X|Y=j)
          \right]
          \label{eq:conditional_entropy_expectation}
  \intertext{or in terms of the joint entropy:}
          &= \H(X,Y) - \H(Y)\nonumber.
\end{align}
The goal of exploration and other sensor planning problems is often to
reduce uncertainty and therefore entropy of a target variable (say $X$) by
obtaining observations $Y$.

Mutual information quantifies that expected reduction
of the entropy of $X$ given an observation $Y$
\begin{align}
  \MI(X;Y)
  &=\sum_{i}\sum_{j}
  -\P(X=i,Y=j)
  \log_2\frac{\P(X=i)\P(Y=j)}{\P(X=i,Y=j)},
  \label{eq:mutual_information}
  \intertext{and can be written in terms of entropies}
  &= \H(X) - \H(X|Y) = \H(X) + \H(Y) - \H(X,Y).
  \label{eq:mutual_information_entropy}
\end{align}
In some cases, we will also apply the conditional mutual information which
simply involves modifying the definition to include conditioning on the given
variable
\begin{align}
  \MI(X;Y|Z) &= \H(X|Z) - \H(X|Y,Z)
  \label{eq:conditional_mutual_information}
  \\
  &= -\H(Z) + \H(X,Z) + \H(Y,Z) - \H(X,Y,Z).\nonumber
\end{align}

Please refer to \citet{cover2012} for further detail on information theory and
the properties of entropy and mutual information and in reference to further
discussion in this section, except when otherwise noted.

\subsection{Continuous variables}
The discussion at the beginning of this section focuses on information theory
for the simple case of discrete variables.
Analogous expressions exist for continuous variables and probability densities.
These collectively refer to differential information.
For example, the differential entropy can be written as
\begin{align}
  \h(X) = \int -\p(X=x) \log_2 \p(X=x) \dd{x}
  \label{eq:differential_entropy}
\end{align}
where $X$ is a continuous variable, and $\p(X=x)$ is its probability density.

The expression for \eqref{eq:differential_entropy} is nearly identical to the
original expression for entropy in \eqref{eq:entropy} aside from replacing the
sum with an integral.
Likewise, differential information satisfies many of the same properties as
the discrete form so that the two are frequently interchangeable.
In particular, this thesis will make use of monotonicity properties such as
submodularity which are discussed later in this chapter;
both forms of information satisfy these properties equally.
As such, this thesis will not typically disambiguate differential and discrete
information, except in cases where the two behave differently.
One such difference is that differential entropy may be negative so that
$\h(X) \ngeq 0$.
By extension, $\MI(X;Y) = \h(X) - h(X|Y) \nleq \h(X)$ so that mutual information
is also not upper-bounded by differential entropy.
Fortunately, even though differential entropy may be negative, mutual
information between continuous variables is always non-negative.

\subsection{Properties of mutual information}
Information theoretic expressions exhibit a number of intuitive properties
from which this theory gains much of its elegance and generality.

Entropy and mutual information satisfy \emph{chain
rules}~\citep[Theorem~2.5.1--2]{cover2012}.
For entropy
\begin{align}
  \H(X_1,\ldots, X_n) = \H(X_n|X_1,\ldots,X_{n-1}) + \H(X_1,\ldots,X_{n-1}),
  \label{eq:chain_rule_entropy}
\end{align}
and for mutual information
\begin{align}
  \MI(X;Y_1,\ldots, Y_n)
  = \MI(X;Y_n|Y_1,\ldots,Y_{n-1}) + \MI(X;Y_1,\ldots,Y_{n-1}).
  \label{eq:chain_rule_information}
\end{align}
These equations allow us to write both kinds of quantities in terms of sums of
marginal gains (the conditional term in each expression).
We will later find that we are able to take advantage of the behaviors of these
marginal gains when optimizing information-theoretic quantities.

Entropy also decreases under conditioning so that $\H(X|Y) \leq \H(X)$ which
expresses that new information cannot increase the uncertainty of a random
variable.
If we write the mutual information in terms of entropies, we can rearrange this
inequality to prove that the mutual information between two variables is
non-negative
$\MI(X;Y) = \H(X) - \H(X|Y) \geq 0$.

Also, independent variables do not carry information about each other.
If $X$ and $Y$ are independent $\H(X) = \H(X|Y)$ and therefore $\MI(X;Y)=0$.
Similar statements hold for conditioning when variables are conditionally
independent.
Additionally, these properties can be used to prove that common cases of mutual
information are submodular, and we will come back to that later.

Having an upper bound on mutual information is often useful such as when we
would like to establish when a sensing task is almost complete.
For \emph{discrete} random variables $\H(X) \geq 0$ is non-negative
(with or without conditioning).
As a result of non-negativity, mutual information for discrete variables can be
upper-bounded by entropy as
\begin{align}
  \MI(X;Y) = \H(X) - \H(X|Y) \leq \H(X).
\end{align}
However, entropy for continuous variables may be negative because, with enough
observations, we may learn the value of a continuous variable with infinite
precision and thereby gain infinite information.
However, if observations share a noisy channel, mutual information between
the observation and target variable is bounded by the mutual information between
the target variable and the output of the channel.
Specifically, the \emph{Markov inequality} establishes that if
$X \rightarrow Y \rightarrow Z$\footnote{%
  The expression $X \rightarrow Y \rightarrow Z$ is a Bayes graph
  and states that the joint distribution of the random variables can be factored
  as
  $\P(X=x,Y=y,Z=z) = \P(X)\P(X=x|Y=y)\P(Z=z|Y=y)$.
}
\begin{align}
  \MI(X;Y) \leq \MI(X;Z)
  \label{eq:markov_inequality}.
\end{align}
This will be important later for target tracking problems where we will discuss
channel capacities between robots and targets.

\section{Robotic information gathering}
\label{sec:robotic_information_gathering}

This thesis applies techniques from information theory and combinatorial
optimization to robot sensing and information gathering.
As a robot moves through an environment it may obtain sensor data, reduce
uncertainty in its environment model, and gain information rewards.
Problems related to selecting actions to maximize quality of sensor data are
prominent in robotics and control and studied with a wide variety of tools.
The methods we propose for coordinating teams of robots complement algorithms
for individual robots as both can contribute to improving completion times in
time-sensitive sensing tasks.
Here, we provide an overview of such techniques moving from general applications
(Sec.~\ref{sec:active_sensing}), to more formal informative path planning
problems (Sec.~\ref{sec:informative_path_planning}), and theory that
characterizes how and why robots benefit from replanning and updating their
decisions based on incoming sensor data (Sec.~\ref{sec:adaptation}).

\subsection{Active sensing and information-based control}
\label{sec:active_sensing}

Sensing and information gathering problems span a wide spectrum that includes
tracking and localization problems~\citep{charrow2014ijrr,dames2015tase}, robot
state estimation~\citep{atanasov2015icra,indelman2014icra}, sensing problems
based on manipulation and mechanical
interactions~\citep{bohg2017tro,corah2017icra,abraham2018rss}, and active
perception problems for control and reconstruction with image
data~\citep{bajcsy1988,roberts2017submodular}.
Although methods for decision-making vary, submodular functions and
information-theoretic objectives not only generalize across these domains but
are also applied
regularly~\citep{charrow2014ijrr,dames2015tase,atanasov2015icra,
indelman2014icra,corah2017icra,roberts2017submodular}.
While we focus on time-sensitive mapping and sensing problems such as urban
disaster response, the methods we propose may also be applied more broadly.
Likewise, we desire planning techniques that apply across multiple sensing tasks
as aspects of different tasks such as
localization~\citep{charrow2014ijrr,dames2015tase} and
mapping~\citep{julian2014,charrow2015icra} may arise simultaneously.

\subsection{Informative path planning}
\label{sec:informative_path_planning}
Algorithms for informative path planning seek to plan a path for a robot to
maximize information gain or some other submodular function, subject to a
limited travel budget.
Such problems have been studied extensively.
Performance guarantees exist for algorithms on
graphs~\citep{chekuri2005,singh2009,zhang2016aaai}, and sampling-based methods
exist for
optimization of paths in continuous spaces~\citep{hollinger2014ijrr}.

These and related algorithms can also serve as suboptimal oracles to greedily
optimize paths for individual robots in multi-robot problems~\citep{singh2009},
and performance guarantees can be rewritten in terms of the performance of
the given planner.
In this sense, path planners can be chosen independently of methods for
coordinating the multi-robot team.

\subsection{Decision processes and adaptation}
\label{sec:adaptation}
Decision processes such as Markov decision process (MDPs), partially-observable
Markov decision process (POMDPs), and variants thereof directly model continuing
processes of selecting actions, obtaining observations, and selecting new
actions to maximize some reward.
Although rewards are typically functions of states and actions, specialized
formulations permit rewards in terms of beliefs e.g. based on mutual information
or entropy~\citep{araya2010pomdp,spaan2015aamas}.
Greedy algorithms have been applied to solve combinatorial subproblems when
actions are sets of sensors~\citep{satsangi2018auro} and in multi-agent
problems to optimize sequences of policies~\citep{kumar2017aaai}, and those
lines of work are complementary to our own.

Instead, adaptive submodularity~\citep{golovin2011jair}
seeks to directly extend techniques for submodular optimization to decision
processes.
However, such approaches require strict regularity on objectives and
the sets of actions to maintain monotonicity and diminishing returns.
Mobile robots generally violate these conditions as sets of available sensing
actions change while the robot moves through the environment.
Other problems in robotics, such as manipulation, are more amenable to the
requirements of adaptive submodularity~\citep{javdani2013icra,chen2015aaai}.

Few methods for decision processes are applicable to mobile robots and
large-scale problems.
However, Monte-Carlo techniques for POMDPs provide an avenue to addressing
large-scale decision processes~\citep{silver2010nips}.
These techniques and applications in robotic exploration~\citep{lauri2015ras}
inspired our use of Monte-Carlo tree search for single-robot planning in this
thesis.

\begin{figure}
  \begin{subfigure}[b]{0.3\linewidth}
    \includegraphics[width=\linewidth]{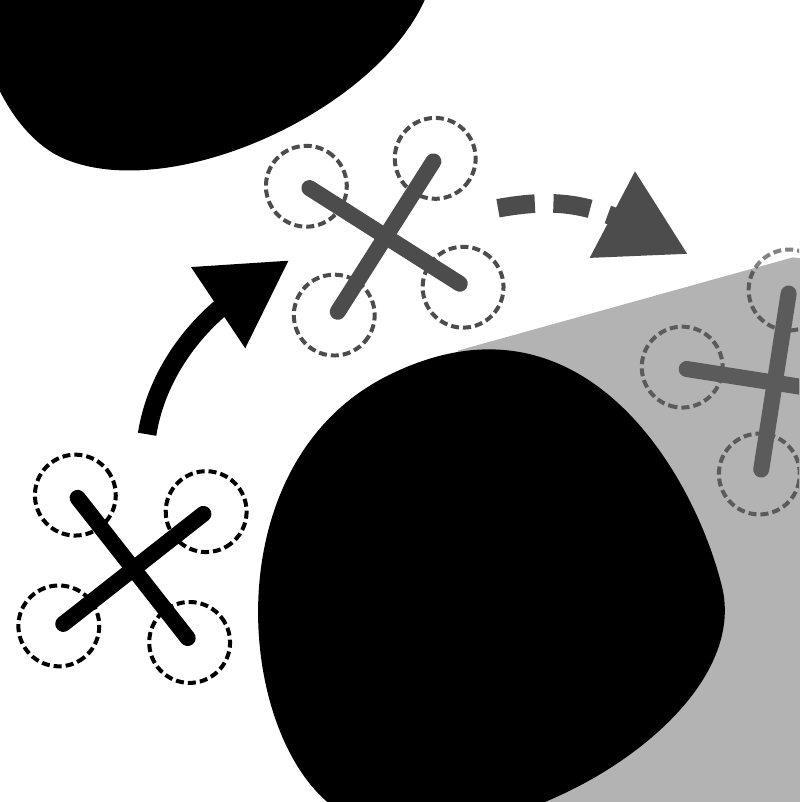}
    \caption{No obstacle}\label{subfig:no_obstacle}
  \end{subfigure}
  \hskip 0.13\linewidth
  \begin{subfigure}[b]{0.3\linewidth}
    \includegraphics[width=\linewidth]{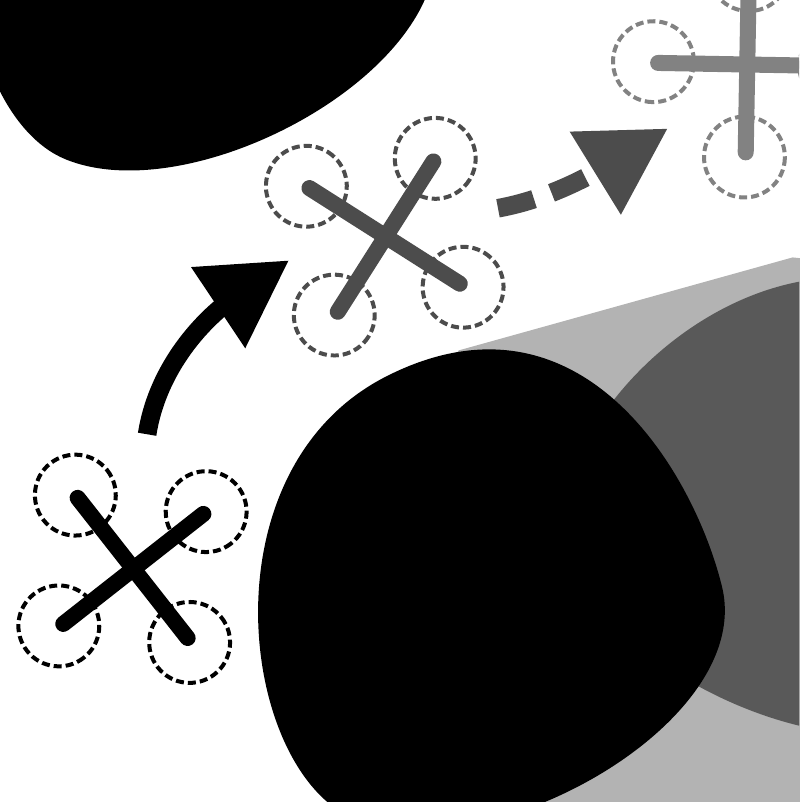}
    \caption{Obstacle}\label{subfig:obstacle}
  \end{subfigure}
  \caption[The benefit of adaptation in exploration]{%
    The illustrations above provide a typical example of adaptation
    (or replanning) during robotic exploration.
    Here, a robot will observe unknown space behind a boulder
    only as that robot navigates around to the back.
    This robot may benefit from \emph{quick adaptation}
    by (\subref{subfig:no_obstacle})
    navigating into the unknown space if there is no obstacle.
    Else, (\subref{subfig:obstacle}) the robot must be prepared to remain in
    free space if that unknown space turns out to be occupied.
  }%
  \label{fig:adaptation}
\end{figure}

Otherwise, rather than seeking to optimize adaptation via policies---which is
often intractable---we may instead consider the benefit of adaptation and the
cost of not adapting~\citep{hollinger2013ijrr,asadpour2015,goemans2006}.
Operation in unknown environments, as in urban disaster response, also induces
rapidly changing beliefs.
Robots moving through cluttered environments may benefit from quickly responding
to observations of occluded objects~\citep{goel2019tr,goel2019fsr}
(see the example in Fig.~\ref{fig:adaptation}), and further, teams of robots
should maintain this ability to adapt and replan as appropriate for the
time-scales of a given sensing problem.


\section{Multi-robot sensing and coordination}
\label{sec:multi-robot_coordination}

The sensing problems that we have discussed to this point also come with
multi-robot counterparts.
We are interested in scenarios where robots interact locally while operating in
close proximity and observing nearby regions of an environment, and our methods
seek to enable robots to react to new observations both individually and
collectively over short time spans.
The techniques we develop leverage simple greedy algorithms and are scalable
while providing strong performance guarantees.
However, various other techniques exist and are applicable to the problems we
consider.
\citet{best2019ijrr} propose a multi-robot planner that applies Monte-Carlo tree
search as we do but uses a different mechanism for multi-robot
coordination---probability collectives---a generic optimization approach based
on game theory.
In this case, our approach will provide stronger guarantees on solution quality
for planning in finite time.
We also note that \citet{sung2018icra} propose a max-min formulation for
tracking problems that addresses similar short time scales that satisfies
performance guarantees before rounding.
\citet{sung2018icra} also address a similar challenge as we do by developing an
algorithm with computation time that is independent of the problem size.
Although we study a slightly different class of sensing problems, we note that
our mechanisms for coordination, based on sharing incremental solutions, are
somewhat simpler and more general than the linear programming methods that they
employ.

\subsection{Sequential greedy maximization for multi-robot sensing}
\label{sec:sequential_multi-robot_sensing}

\begin{figure}
  \begin{center}
    \includegraphics[width=0.27\linewidth]{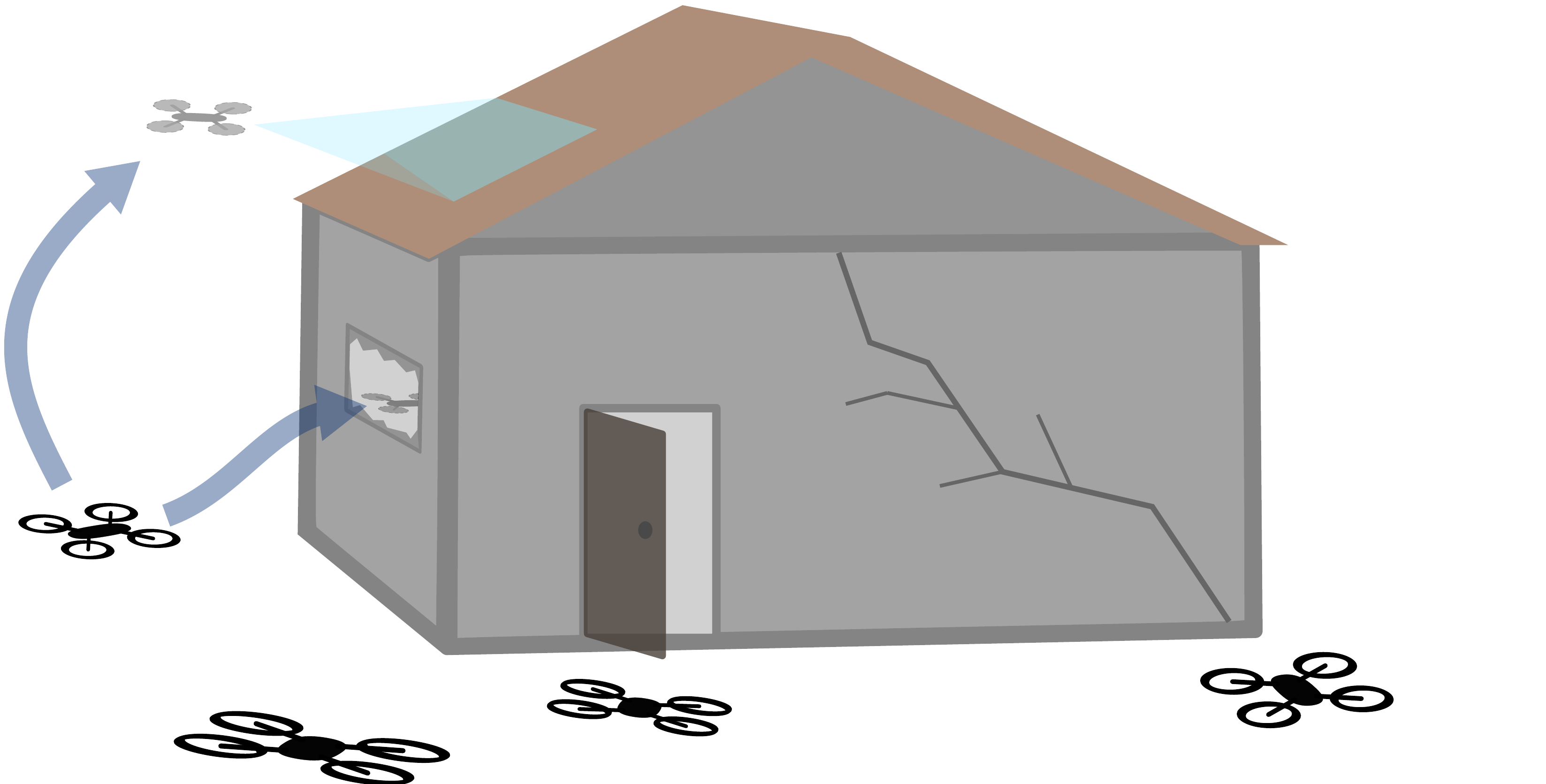}%
    \hspace{-2.5ex}%
    \includegraphics[width=0.27\linewidth]{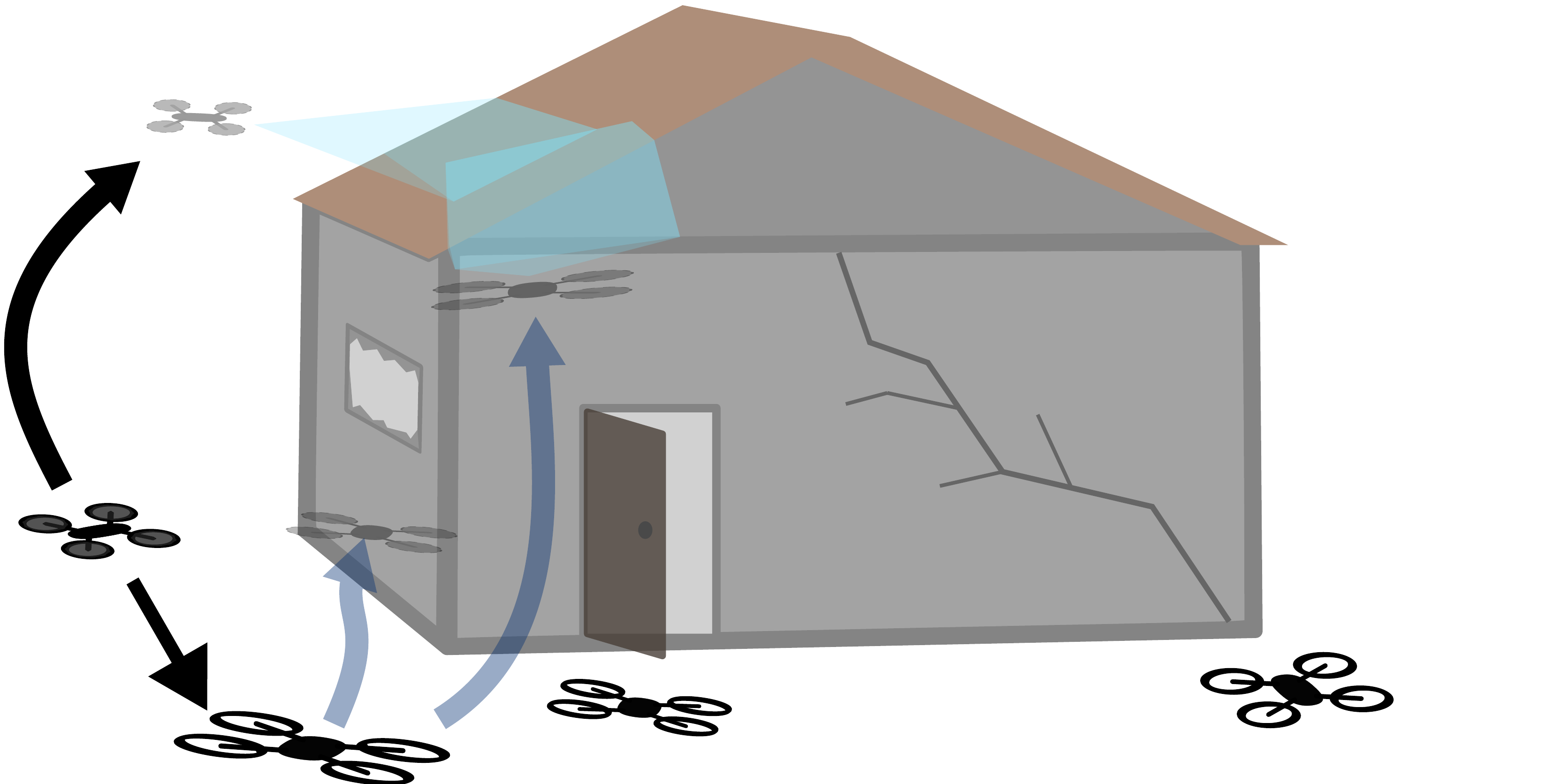}%
    \hspace{-2.5ex}%
    \includegraphics[width=0.27\linewidth]{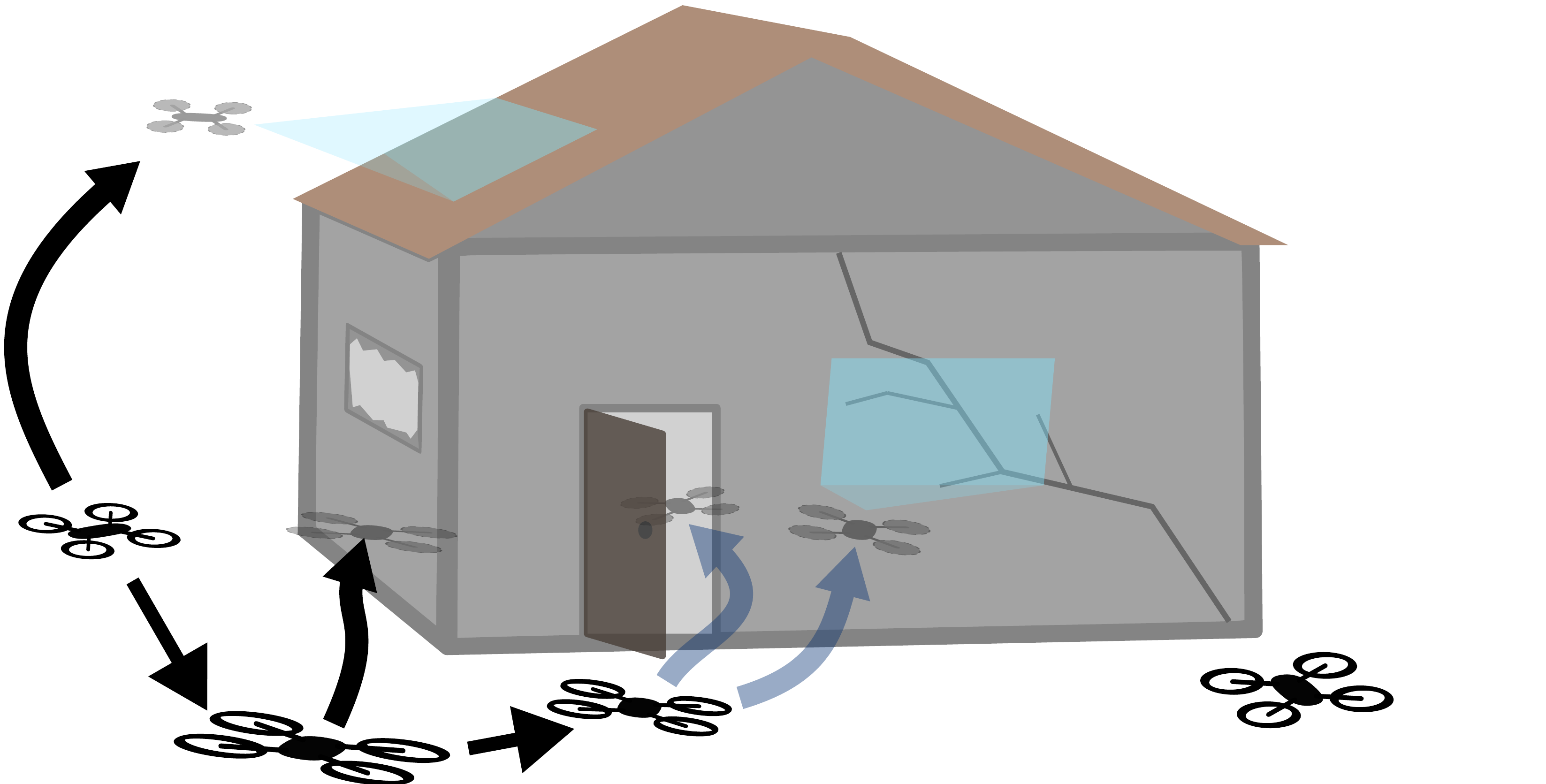}%
    \hspace{-2.5ex}%
    \includegraphics[width=0.27\linewidth]{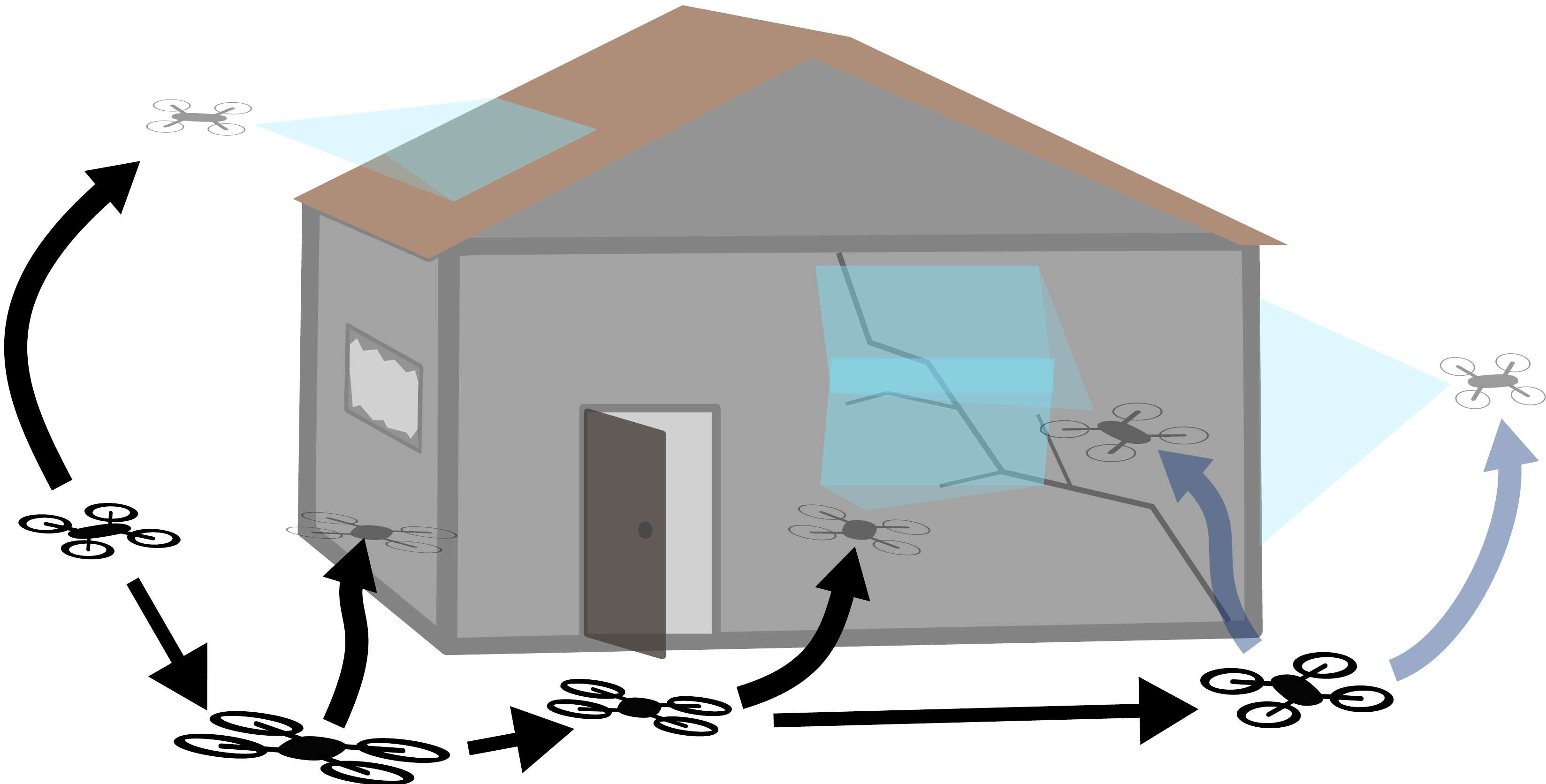}%
  \end{center}
  \caption[Sequential planning]{%
    Four robots plan sequentially. Each robot selects an action to maximize
    a sensing objective given the decisions by the prior robots.%
  }%
  \label{fig:sequential_illustration}
\end{figure}

\citet{singh2009} are largely responsible for developing
sequential planning methods for multi-robot sensing (based on methods for
submodular maximization which we will discuss next in
Sec.~\ref{sec:submodular_maximization}) and
informative path planning.\footnote{%
  Historical note: \citet{singh2009} are most commonly cited for
  specializing matroid-constrained submodular maximization to partition matroids
  for sensor planning and, more specifically, multi-robot path planning.
  \citet{goundan2007} observed around the same time that this kind of approach
  recreates a variant of the locally greedy algorithm described by
  \citet{fisher1978}.
  In fact, \citet{singh2009}, \citet{goundan2007}, and even also
  \citet{williams2007} all appear to have independently (re)developed similar
  specializations to partition matroids (implicitly or explicitly) at around the
  same time.
}
In these methods, robots plan in a sequence,
(as illustrated in Fig.~\ref{fig:sequential_illustration})
and each robot seeks to maximize a sensing objective given all prior decisions.
For many objectives, solutions by sequential planning are also guaranteed to
be within half of the optimal objective value, whereas obtaining optimal
solutions is NP-Hard.

While \citet{singh2009} studied a mutual information objective, sequential
planners can provide suboptimality guarantees for wide variety of objective
functions (which we will discuss in the next section).
As a result, their work influenced not only a variety of works on informative
planning~\citep{atanasov2015icra,regev2016iros}
but also problems with more varied objectives such as
search for a moving target~\citep{hollinger2009ijrr}
and survivability-aware problems where robots may fail en-route to their
destinations~\citep{jorgensen2017iros}.
While most of these authors study similar multi-robot path planning problems,
\citet{jorgensen2017iros} and \citet{williams2017icra}
also apply related greedy planning methods to more complex assignment problems
such as with constraints on coverage, resources (such as a limited number of
runways), and network connectivity.

\subsubsection{Single-robot sub-problems in sequential planning}
\label{sec:single-robot_planning}

A useful feature of sequential planners is that specialized planners like those
in Sec.~\ref{sec:informative_path_planning} can be used to implement the
maximization step while retaining similar approximation
guarantees~\citep{singh2009}
(that is, they serve as \emph{maximization oracles}).
This is particularly important because the alternative is to iterate over all
possible solutions for a given robot---doing so is typically intractable for
path planning problems as the number of possible solutions is generally
exponential in both the time horizon and the number of available actions at each
step.
Moreover, this approach enables designers to take advantage of advances for
planning for individual robots and to apply those methods to planning for
multiple robots with minimal effort.

\subsubsection{Adaptation in sequential planning}
\label{sec:sequential_adaptive}

In non-adaptive settings, (where robots do not replan after obtaining new
information)
solutions can be found offline and without strict constraints on time or
computation.
For example, \citet{singh2009} plan paths for a sensing task on a lake and
compute the entire paths beforehand.
However, this thesis studies adaptive settings
(Sec.~\ref{sec:problem_adaptivity}) where the plan or sensing actions change in
response to incoming sensor data during execution of the sensing task,
and sequential planning forms a component of a solver for a receding-horizon
optimization problem
(Sec.~\ref{sec:receding_horizon_planning})~\citep{%
charrow2015wafr,bircher2018receding,lauri2015ras}.

Sequential planners for these receding horizon problems then must satisfy
time constraints for replanning.
When planning with large numbers of robots, meeting time constraints may require
improving efficiency via parallel versions of these planners or by parallelizing
the single-robot planners.
The following sections will describe the theory for these sensing problems
and provide more detail regarding parallel and distributed planning.

\section{Submodular maximization}
\label{sec:submodular_maximization}

The sequential planning and informative path planning problems that arise in
this thesis broadly fall into the framework of so-called submodular maximization
problems.
Specifically, many sensing objectives
\emph{increase monotonically} and exhibit \emph{diminishing returns}
(submodularity).
These classes of functions have been studied extensively from the perspective of
combinatorial optimization, and there are numerous results for exact and
approximate optimization in various settings.
In this way, monotonic and submodular functions studied in combinatorial
optimization are analogous to the convex functions studied in nonlinear
optimization:
each has been studied intensively and optimization problems involving either
class of functions can be solved efficiently\footnote{
  There are connections between submodular functions and both convex and concave
  functions, but those are unrelated to this analogy.
}.
Objective functions with these properties appear frequently in robotics and
especially in sensing problems.
For example, notions of monotonicity or diminishing returns intuitively
describe how the value of a collection of sensing actions changes as actions are
added or removed.
Later, we will discuss submodular objective functions such as common cases of
mutual information objectives and different notions of sensor coverage.

Going a step further, we are also interested in different kinds of constraints
on these optimization problems, just as constraints are common in other kinds of
optimization problems.
Whereas this section discusses optimizing functions of sets, the constraints in
question will determine which sets to consider.
These constraints model concepts such as which robots can perform an action, how
many actions a robot can perform at once, or which sets of actions will result
in collisions between robots.
In this sense, these constraints will describe which sets of actions teams
of robots can feasibly execute.

\subsection{Set functions and submodularity}
\label{sec:submodularity}

Consider a finite---but possibly extremely large---set of sensing actions
$\ground$, commonly referred to as the \emph{ground set}.
The ground set describes the universe of set elements that are available in a
given optimization problem e.g. the set of all sensing actions or finite-horizon
plans available to any robot.
A \emph{set function} $\setfun : 2^\ground\rightarrow\real$ maps a collection of
such actions to a real value---often a reward---where $2^\ground$ is the power
set of the ground set, the set of all subsets of $\ground$.


Most objectives have zero value at the empty set, or else that value can be
subtracted to normalize the objective.
\begin{definition}[Normalized]
  A set function $\setfun$ is normalized if $\setfun(\emptyset) = 0$.
\end{definition}
A function is monotonically increasing\footnote{%
  Some texts describe the functions that we call monotonically increasing as
  being ``non-decreasing'' to emphasize the weak inequality in the definition.
}
(or monotonic) if its value cannot decrease.
\begin{definition}[Monotonically increasing]
  A set function $\setfun$ is monotonically increasing if for any
  $A \subseteq B \subseteq \ground$
  then
  \begin{align}
    \setfun(A) \leq \setfun(B)
    \label{eq:monotonicity}
  \end{align}
\end{definition}
Next, a set function is submodular if it exhibits diminishing returns.
\begin{definition}[Submodular (supermodular)]
A set function $\setfun$ is submodular if
for any $A\subseteq B \subseteq \ground$ and
$C\subseteq \ground \setminus B$ then
\begin{align}
  \setfun(A\cup C) - \setfun(A) \geq \setfun(B\cup C) - \setfun(B).
  \label{eq:submodularity}
\end{align}
Here, the marginal gain for $C$ decreases after adding elements to $A$
to obtain $B$.
The negation $-\setfun$ is supermodular.
\end{definition}

\subsubsection{Relevant submodular functions}
\label{sec:relevant_submodular_functions}

Objective functions for robot sensing are frequently monotonic and submodular.
Mutual information \eqref{eq:mutual_information} and entropy \eqref{eq:entropy}
naturally characterize reduction of uncertainty with respect to an unknown
environment, and robots may select sensing actions which reduce that
uncertainty.
Mutual information with respect to a target variable satisfies monotonicity and
submodularity
when observations are conditionally independent~\citep{krause2005uai}.
Specifically, if we associate elements of a ground set $a\in\ground$
with conditionally independent observations $Y_a$, the set function
$\setfun(A)=\MI(X;Y_A)$ is normalized, monotonic, and
submodular~\citep{krause2005uai}.
However, designers may consider other kinds of objectives, either for different
tasks or because mutual information can be expensive to evaluate.
For example, we will also study simpler coverage-like objectives which satisfy
the same properties and can be thought of as representing sums of rewards for
observing objects in subsets of the environment.

\subsubsection{Notation for sets and set functions}
For brevity, we will treat set functions as multi-variate functions with
implicit unions so that $\setfun(A, B) = \setfun(A\cup B)$ where
$A, B \subseteq \ground$.
We will also implicitly convert elements of the ground
set to subsets so that $\setfun(x) = \setfun(\{x\})$ for $x\in \ground$.
We also write incremental changes as
$\setfun(A|B) = \setfun(A,B) - \setfun(B)$ which is
analogous to conditioning for mutual
information~\eqref{eq:conditional_mutual_information}
and also expresses the discrete derivative of $\setfun$ at $B$ with respect to
$A$ (discussed next in Sec.~\ref{sec:derivatives}).
Using this notation, the expression for submodularity \eqref{eq:submodularity}
can be written more concisely as
\begin{align}
  \setfun(C|A) \geq \setfun(C|B),
\end{align}
recalling that $A\subseteq B$.

When dealing with sets and set functions, being able to concisely specify and
index into sets will be advantageous.
Often sets will be specified with an implied ordering as in
$X=\{x_1,\ldots,x_n\}$.
Subscripts will then be used for indexing subsets as in the following range
$X_{1:i}=\{x_1,\ldots,x_i\}$ for $i\leq n$
or sometimes using sets of indices so that
$X_{A} = \{x_j \mid j \in A\}$.

\subsubsection{Discrete derivatives of set functions}
\label{sec:derivatives}

\begin{figure}
  \centering
  \begin{subfigure}[b]{0.3\linewidth}
    \includegraphics[width=\linewidth]{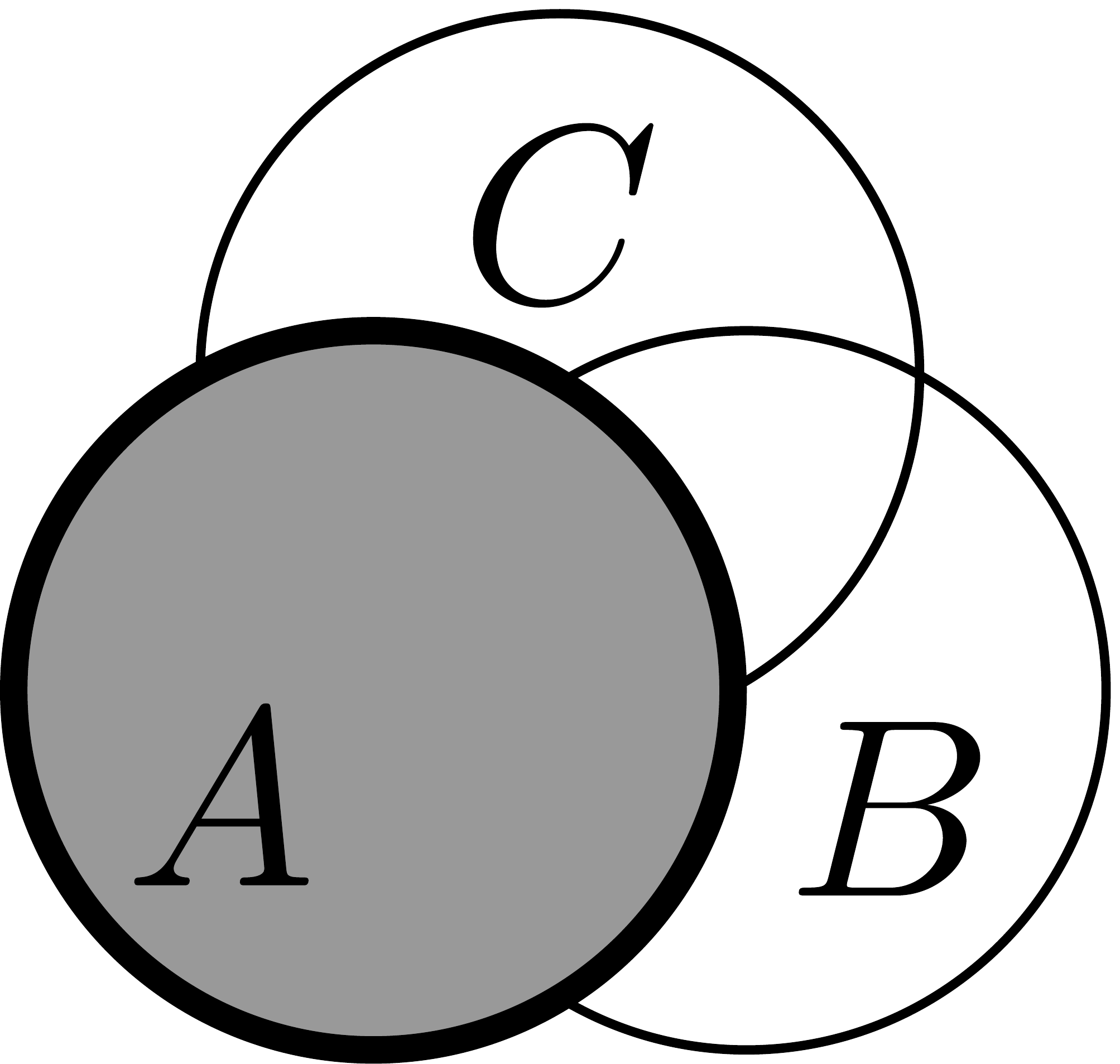}
    \caption{$\setfun(A) = \setfun(A|\emptyset)$}%
    \label{subfig:zeroth_derivative}
  \end{subfigure}
  \begin{subfigure}[b]{0.3\linewidth}
    \includegraphics[width=\linewidth]{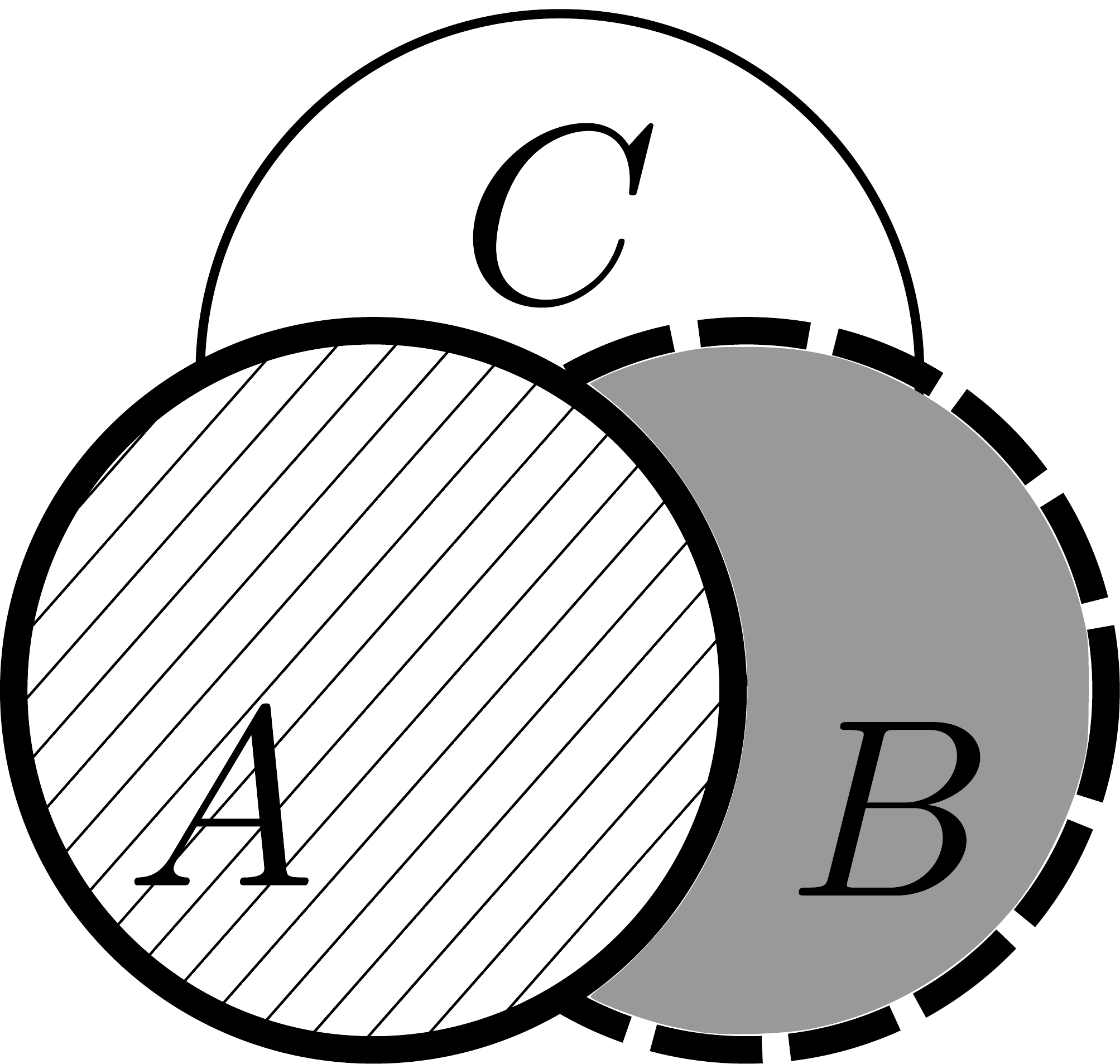}
    \caption{$\setfun(B|A)$}%
    \label{subfig:first_derivative}
  \end{subfigure}
  \begin{subfigure}[b]{0.3\linewidth}
    \includegraphics[width=\linewidth]{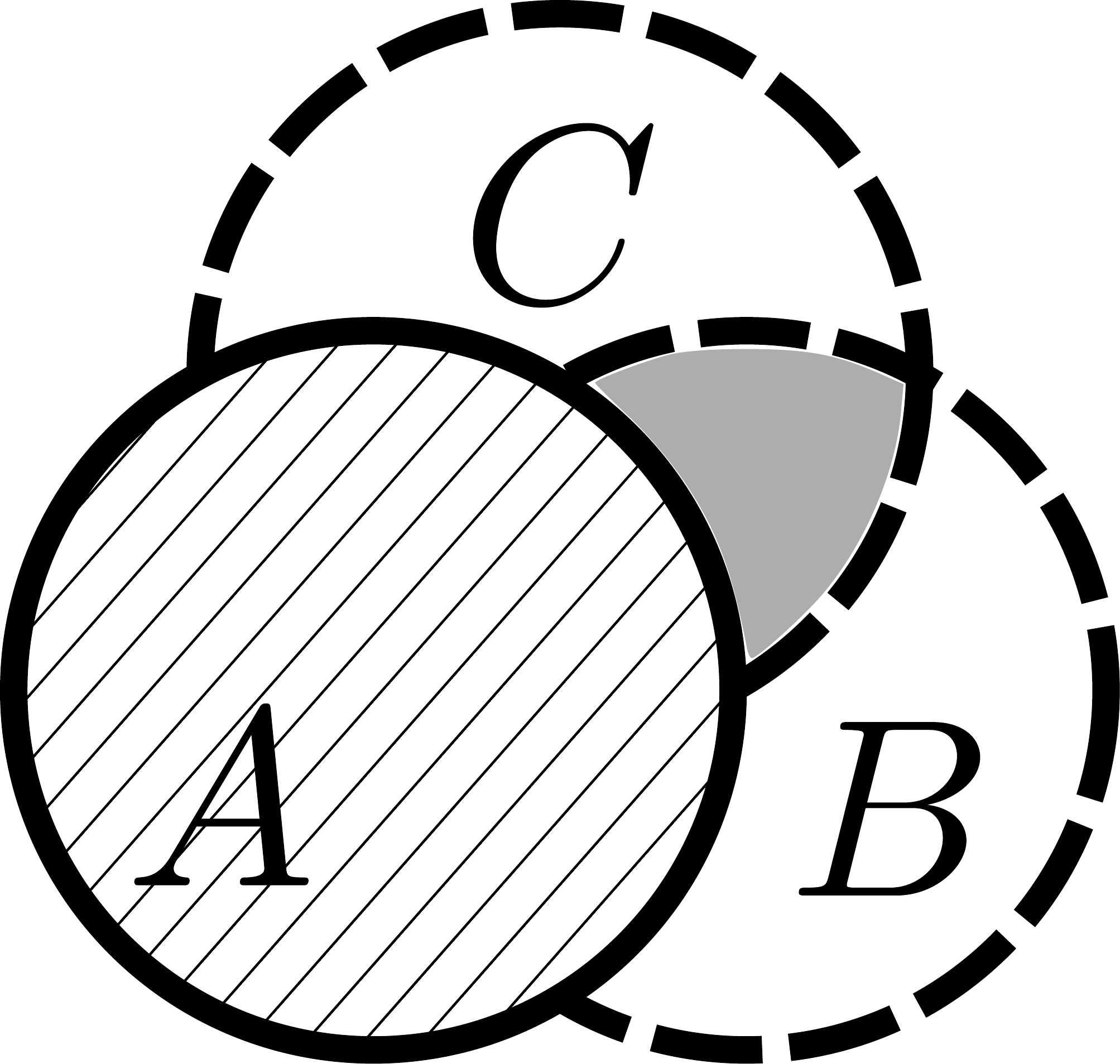}
    \caption{$-\setfun(B;C|A)$}%
    \label{subfig:second_derivative}
  \end{subfigure}
  \caption[Illustration of derivatives of set functions]{%
    In this figure, areas shown in gray are equal in size to
    (\subref{subfig:zeroth_derivative})
    zeroth,
    (\subref{subfig:first_derivative})
    first, and
    (\subref{subfig:second_derivative})
    second derivatives of a set function $\setfun$, each evaluated at $A$.
    As will be discussed in more detail, later, the derivatives of some set
    functions are monotonic such as in this illustration of area coverage.
    If we interpret $A$ as a subset of $\real^2$,
    (\subref{subfig:zeroth_derivative})
    area coverage increases monotonically in $A$;
    (\subref{subfig:first_derivative})
    marginal gains decrease monotonically;
    (\subref{subfig:second_derivative})
    and the second derivative increases monotonically
    (the negation of which, the highlighted area, decreases).
  }%
  \label{fig:set_function_derivatives}
\end{figure}

The definitions of monotonic~\eqref{eq:monotonicity}
and submodular~\eqref{eq:submodularity} functions are closely related.
The value of a monotonic function may only stay the same or increase, and
the values of incremental changes of submodular functions may only stay the same
or decrease.

We can go further and define higher derivatives of set functions and
corresponding monotonicity conditions.
To assist in understanding, Fig.~\ref{fig:set_function_derivatives} provides
examples of such derivatives for an area coverage objective.
Given disjoint variables $A,B,C \subseteq \ground$ the second derivative of
$\setfun$
at $C$ in directions $A$ and $B$ is
\begin{align*}
  \setfun(A;B|C) &= \setfun(A, B, C) - \setfun(A,C) - \setfun(B,C) + \setfun(C).
\end{align*}
Expressions of second derivatives will also arise with respect to the empty set.
Such expressions take the form of
\begin{align*}
  \setfun(A;B) = \setfun(A|B) - \setfun(A) =
  \setfun(A,B) - \setfun(A) - \setfun(B),
\end{align*}
given that $\setfun(\emptyset)=0$,
and being able to recognize them will be useful.
For higher derivatives, given $X\subseteq\ground$ and
$Y_1,\ldots,Y_n\subseteq\ground$, all disjoint, we can define the
$n^\mathrm{th}$ derivative of $\setfun$ with respect to $Y_1,\ldots,Y_n$
recursively as
\begin{align}
  \setfun(Y_1;\ldots;Y_n|X) &=
  \setfun(Y_1;\ldots;Y_{n-1}|X,Y_n) - \setfun(Y_1;\ldots;Y_{n-1}|X).
  \label{eq:recursive_derivative}
\end{align}
Note that, just as for a continuous derivatives, derivatives of set functions
are not sensitive to the order of the directions.

This notation for set functions and their discrete derivatives also
intentionally overlaps with notation for entropy and mutual information.
Continuing with this analogy, the mutual information could be thought of as the
\emph{negation} of the second derivative of entropy were it interpreted as a
set function as $\MI(A;B|C) = \H(A|C) - \H(A|B,C)$.
Therefore, one might adopt the colloquialism $\MI(A;B|C) = - \H(A;B|C),$
\emph{although no such notation will be used in this text}.

\citet{foldes2005} provide a slightly different definition of the derivative.
As this thesis will reference their work repeatedly, we provide a brief proof of
equivalence in Appendix~\ref{appendix:derivative_representation}.

\subsubsection{Higher-order monotonicity and derivatives}
\label{sec:higher-order_monotonicity}
\begin{figure}
  \centering
  \def\svgwidth{0.8\linewidth}
  \input{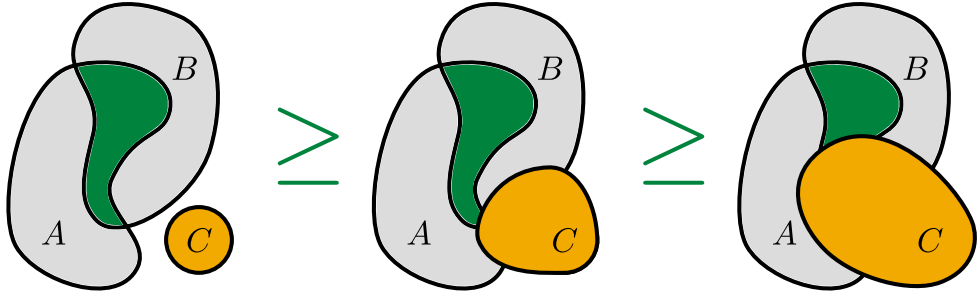}
  \caption[Illustration of the 3-increasing condition for coverage]{
    The above illustrates the 3-increasing condition for coverage.
    Let the area covered by a number of sets (or sensing actions) be a function
    $\setfun$.
    The green area corresponds to an amount of sensing redundancy between two
    sensor coverage actions $A$ and $B$ conditional on a third observation $C$.
    The size of that area in green is equal to
    the negation of a second derivative
    $-\setfun(A;B|C) = \setfun(A|C) + \setfun(B|C) - \setfun(A,B|C)$,
    and the area decreases monotonically as the region represented by $C$
    increases.
    Then, second derivatives $\setfun(A;B|C)$ \emph{increase}
    monotonically for coverage objectives so we say that $\setfun$ is
    3-increasing
    (as the first monotonicity condition corresponds to the zeroth derivative).
  }%
  \label{fig:monotonicity_of_redundancy}
\end{figure}

\citet{foldes2005} also describe monotonic and submodular set
functions as being, respectively, 1-increasing and 2-decreasing
and provide detailed discussion of these conditions.
More generally, if the $n^\mathrm{th}$ derivative of a function is non-negative,
then derivatives of order $n\!-\!1$ increase monotonically, and we say that the
function is $n$-increasing.
Then, if that derivative is non-positive, the function is instead
$n$-decreasing.
Later, we will apply this terminology in a requirement that certain objective
functions are 3-increasing.
In particular, weighted set coverage is 3-increasing, and we illustrate the
intuition in Fig.~\ref{fig:monotonicity_of_redundancy}.
Later, we will provide more direct analysis when we encounter this condition
again.

Higher-order monotonicity conditions, such as this, are not yet common in
the literature on optimizing set functions but are experiencing increasing
interest~\citep{wang2015,chen2018,gupta2020,korula2018}.
\citet{wang2015} apply the 3-increasing condition to develop a
variation of the continuous greedy algorithm, and \citet{chen2018} study
functions with alternating monotonicity conditions, something we will also
encounter.
Measures of redundancy~\citep{finn2020phd} and analogues of mutual
information~\citep{iyer2020} also fit the form of second derivatives and are
highly relevant to works on 3-increasing functions.

\subsubsection{Chain rule for derivatives of set functions}
\label{sec:chain_rules}

When proving bounds for submodular maximization techniques, we will often want
to decompose a derivatives with respect to sets (say $A\subseteq \ground$) in
terms of the contributions of individual elements (i.e. $a\in A$).
Chain rules allow us to do so.

Previously, we have discussed such chain rules in the context of information theory
(Sec.~\ref{sec:information_theory}).
Set functions and their derivatives satisfy analogous chain
rules.
Later, we will require chain rules for both first and second derivatives of set
functions.
Toward this end, the following defines a chain rule for any derivative of a set
function.
\begin{lemma}[Chain rule for derivatives of set functions]
  \label{lemma:chain_rule}
  Consider sets $Y_1,\ldots,Y_n,X\subseteq\ground$, all disjoint.
  Then, writing the elements of $Y_n$ as $Y_n=\{y_{n,1},\ldots,y_{n,|Y_n|}\}$,
  the derivative of $\setfun$ can be rewritten in terms of derivatives with respect
  to the individual elements of $Y_n$ as
  \begin{align}
    \setfun(Y_1;\ldots;Y_n|X)
    = \sum_{i=1}^{|Y_n|} \setfun(Y_1;\ldots;Y_{n-1};y_{n,i}|Y_{n,1:i-1},X).
    \label{eq:chain_rule}
  \end{align}
\end{lemma}
\noindent The proof of Lemma~\ref{lemma:chain_rule} is in
Appendix~\ref{appendix:chain_rule}.

\subsection{Matroids and independence systems}
\label{sec:independence}
This thesis focuses primarily on multi-robot planning problems that can be
described using combinatorial constraints.
Such constraints that describe feasible solution sets are called
\emph{set systems}.
However, a completely general constraint would not be useful for the purpose of
optimization.
Independence systems describes constraints that admit subsets of any feasible
set and which is the most general kind of set system that we will find useful.
\begin{definition}[Independence system]
  \label{def:independence_system}
  A tuple $(\ground, \independence)$
  where $\independence$ is composed of subsets of the ground set such that
  $I\in\independence \rightarrow I\subseteq\ground$
  is an independence system
  if $\independence$ satisfies a heredity property so that for all
  $I_2 \subseteq I_1\in\independence$ then $I_2\in\independence$
  and, additionally, $\independence$ is non-empty ($\emptyset\in\independence$).
\end{definition}
Multi-robot non-collision constraints form one example of an independence
system; eliminating a robot from a non-colliding multi-robot plan will not cause
the remaining robots to be in collision.

A matroid is a special case of an independence system and generalizes the
notions of linear independence from vector spaces to set systems.
Matroids have been studied extensively in the optimization literature (discussed
in more detail in following sub-sections) and find use in this work to model
product spaces in multi-robot planning problems.
\begin{definition}[Matroid]
  \label{def:matroid}
  An independence system $(\ground, \independence)$ that also satisfies the
  exchange property is a matroid, that is if for all
  $I_1,I_2\in\independence$
  such that $|I_1|>|I_2|$ there exists some $x\in I_1 \setminus I_2$ such that
  $I_2\cup \{x\} \in \independence$.
\end{definition}
The reader may think of elements that can be added to a given set as being
effectively perpendicular to that set.
For more detail on this topic,
the reader may wish to refer to \citet{schrijver2003}.
A consequence of the exchange property is that all maximal independent
sets---those to which we cannot add any more elements---have
the same cardinality which is referred to as the \emph{rank} of the matroid
or $\matroidrank(\independence)$.
Additionally, in this work, the rank will also be equal to the number of robots
or other sensing agents.

Although matroids are the focus of numerous related works on optimization,
a matroid is a somewhat more general structure than is necessary in this thesis,
and we will gain in efficiency from using slightly more specialized constraints.
A \emph{partition matroid} is a special case of a matroid and is nearly directly
analogous to a product space.
\begin{definition}[Partition matroid]
\label{def:partition_matroid}
Consider a partitioning of the ground set into blocks
$\block_i \subseteq \independence$ for all $i\in\{1\dots n\}$ such that
$\block_i \cap \block_j = \emptyset$ for all $i \neq j$.
Then $(\ground, \independence)$ is a partition matroid if
\begin{align*}
  \independence = \{ I \subseteq \ground \mid a_i \geq |I\cap \block_i|,
  ~\forall i \in \{1 \dots n\}\}
\end{align*}
where $a_i \in \integer_{>0}$ is the maximum number of elements that can be
selected from a given block.
\end{definition}

When $a_i=1$ for all $i$, the partition matroid will be referred to as
a \emph{simple partition matroid}~\citep[Sec.~39.4]{schrijver2003}.
These matroids will be used to model joint action spaces of multi-robot
teams where $\block_i$ is the space of trajectories or actions available to
robot $i$, assuming that there are no inter-robot constraints on action
selection.
Although the techniques developed in this thesis will all use simple partition
matroids, they all extend to general partition matroids with slight
modification.

Finally, we will discuss one more (and simpler) type of matroid, the uniform
matroid, which expresses a cardinality constraint on sets in the set system.
\begin{definition}[Uniform matroid]
  \label{def:uniform_matroid}
  A uniform matroid is an independence system $(\ground, \independence)$ that
  fits the form fits the form
  $\independence = \{I \subseteq \ground \mid n \geq |I| \}$
  where $n \in \integer_{>0}$.
\end{definition}



\subsection{Submodular maximization and greedy algorithms}
\label{sec:greedy_algorithms}

The problems studied in this thesis involve selecting control actions to
maximize notions of sensing quality such as mutual information with a map or
sensor coverage.
When such objectives are submodular
(and generally also monotonic and normalized),
the resulting optimization problems fit broadly into the realm of submodular
maximization problems.
Then, when these problems involve planning actions for a multi-robot team,
as hinted in Sec.~\ref{sec:independence}, we will often also introduce a
matroid constraint of some form.
\begin{problem}[Submodular (monotonic, normalized) maximization problem]
  \label{prob:submodular_general}
  Let $\setfun:2^\ground \rightarrow \real$ be a submodular, monotonic, and
  normalized set function and $(\ground,\independence)$ a set system.
  Together, these define a submodular maximization problem
  \begin{align*}
    X^\opt \in \argmax_{X \in \independence} \setfun(X)
  \end{align*}
  with optimal solution(s) $X^\opt$.
\end{problem}
Now, unless otherwise specified, the objectives in any submodular maximization
problems discussed in this text will also be monotonic and normalized.
We will primarily only refer to other cases when discussing related work on
other problems or in passing.
Instead, we now focus on the form of the constraint $\independence$ and,
eventually, further conditions on $\setfun$.


A general set system may present $2^{|\ground|}$ feasible solutions.
Although $|\independence|$ will generally be smaller, solving submodular
maximization problems by iteration is typically intractable.
Recall that when $(\ground,\independence)$ is an independence system
(Def.~\ref{def:independence_system}), subsets of feasible solutions are also
feasible.
Solutions can then be constructed incrementally by searching $\ground$ for new
elements to add to partial solutions.
In fact, algorithms that construct such solutions greedily often guarantee
solutions within a constant factor of the optimal solution.

\begin{algorithmdef}[Greedy submodular maximization]
  \label{alg:greedy_submodular}
  \addtocounter{algorithm}{-1}
  \captionlistentry[algorithm]{Greedy submodular maximization}
  Greedy submodular maximization produces feasible (suboptimal)
  solutions to submodular maximization problem
  (Prob.~\ref{prob:submodular_general}) where $(\ground,\independence)$ is an
  independence system.
  Define an incremental solution $X_i$ such that $X_0=\emptyset$ and
  \begin{align}
    X_i \in
    \argmax_{ \{x\}\cup X_{i-1} \in \independence} \setfun(x| X_{i-1}).
    \label{eq:greedy_independence}
  \end{align}
  This algorithm outputs a solution, designated $X^\mathrm{g} = X_n$, once
  no more solution elements can be added, i.e. once there is no $x\in\ground$
  such that $\{x\}\cup X_n \in \independence$.
\end{algorithmdef}

Solutions can be obtained far more efficiently than by iterating over
$2^\ground$ when using this algorithm, but keep in mind that solving
\eqref{eq:greedy_independence} may still be intractable when $\ground$ is large.
Developing methods that avoid incurring costs of such sequential maximization
steps is a recurring theme in related work and this thesis.

\subsubsection{Greedy submodular maximization with cardinality constraints}

Cardinality-constrained submodular maximization forms a simple and particularly
common submodular maximization problem.
In this case, the statement that the
cardinalities of feasible solutions are no greater than some value produces
a uniform matroid (Def.~\ref{def:uniform_matroid}) constraint.

\begin{problem}[Cardinality-constrained submodular maximization]
  \label{prob:submodular_cardinality}
  Any instance of Prob.~\ref{prob:submodular_general} where
  $(\ground,\independence)$ is a uniform matroid is an instance of
  cardinality-constrained submodular maximization.
\end{problem}

Cardinality-constrained problems arise frequently in
applications~\citep{clark2011cdc,jorgensen2017irc,carlone2018tro} due to the
simplicity of the constraint as well as because of strong tightness and hardness
guarantees.
These tightness and hardness guarantees are important to us because
uniform matroids are a special case of set systems such as partition
matroids and because many results for cardinality constrained problems
also apply to problems with more complex constraints
(such as by forming lower bounds).
\citet{nemhauser1978} proved that greedy solutions to cardinality-constrained
problems satisfy a constant-factor suboptimality guarantee
$\setfun(X^\mathrm{g}) \geq
(1-1/e) \setfun(X^\opt) \approx 0.632 \setfun(X^\opt)$.
Soon after, some of the authors of that work also proved that this approximation
guarantee is optimal in the value oracle model
(where $\setfun$ is treated as a black box)
for algorithms that evaluate $\setfun$ a polynomial number of
times~\citep{nemhauser1978best}.
Still, certain classes of these problems could be easier to solve or could be
solved more quickly given complete access to the problem representation.
\citet{feige1998} proved that no better bound can be achieved by polynomial time
algorithms for set coverage objectives\footnote{%
  Set coverage refers to ``covering'' a set $\coverset$.
  The elements of the ground set $x\in\ground$ are subsets of this set
  $x\subseteq\coverset$, and the objective is to maximize the cardinality of
  their union $\setfun(X) = \left|\bigcup_{x \in X} x\right|$.
}
unless P=NP.
Not only is this a stronger guarantee, but the simplicity of the set coverage
objective enables reductions to other relevant objectives.
This includes generalizations of coverage discussed later in this text and,
thanks to \citet{krause2005uai}, Shannon mutual information on
Bayes graphs.

\subsubsection{Greedy submodular maximization on general matroids}

While cardinality-constrained problems provide hardness results that are
relevant to this thesis, matroid-constrained problems will \emph{generalize}
the multi-robot planning problems that we will discuss.
\begin{problem}[Matroid-constrained submodular maximization]
  \label{prob:submodular_matroid}
  An instance of Prob.~\ref{prob:submodular_general} where
  $(\ground,\independence)$ is a matroid is an instance of
  matroid-constrained submodular maximization.
\end{problem}
\citet{fisher1978} proved that when $(\ground, \independence)$ is a matroid,
solutions by Alg.~\ref{alg:greedy_submodular} satisfy
$\setfun(X^\mathrm{g}) \geq 1/2 \setfun(X^\opt)$.
Although this thesis applies greedy algorithms,
more recent works~\citep{calinescu2011,filmus2012focs} propose more advanced
methods such as the continuous greedy
algorithm~\citep{vondrak2008,calinescu2011} which guarantee results within
$1-1/e$
which is the best possible guarantee given the hardness results for
cardinality-constrained problems.

\subsubsection{Locally greedy (sequential) planning on simple partition
matroids}
\label{sec:locally_greedy}

Although direct application of Alg.~\ref{alg:greedy_submodular} provides
solutions within half of optimal, as mentioned, the greedy step
\eqref{eq:greedy_independence} involves iteration over the entirety of
$\ground$ which may be expensive.
This can be avoided in cases such as when the matroid constraint is a partition
matroid.
For simplicity, we will focus on the special case of simple partition matroids.

\begin{problem}[Simple partition matroid-constrained submodular maximization]
  \label{prob:submodular_partition_matroid}
  An instance of Prob.~\ref{prob:submodular_general} where
  $(\ground,\independence)$ is a simple partition matroid is an instance of
  partition matroid-constrained submodular maximization.
\end{problem}

Problems constrained with simple partition matroids admit a relaxation of the
greedy algorithm: rather than computing the incremental solutions by maximizing
over the entire ground set, increments can be computed by maximizing over the
blocks of the partition.

\begin{algorithmdef}[Locally greedy algorithm for simple partition matroids]
  \label{alg:local_greedy}
  \addtocounter{algorithm}{-1}
  \captionlistentry[algorithm]{%
    Locally greedy algorithm for simple partition matroids}
  The locally greedy algorithm produces feasible (suboptimal)
  solutions to partition matroid-constrained maximization submodular
  maximization problems
  (Prob.~\ref{prob:submodular_partition_matroid}).
  Define solution elements $x_i$ such that
  \begin{align}
    x_i \in
    \argmax_{ x \in \block_i } \setfun(x | \{x_1,\ldots,x_{i-1}\})
    \label{eq:locally_greedy}
  \end{align}
  to produce the output $X^\mathrm{g} = \{x_1,\ldots,x_n\}$ where $n$ is the
  rank of the partition matroid.
\end{algorithmdef}

Notably, this algorithm implements the same sequential planning process
as in Sec.~\ref{sec:sequential_multi-robot_sensing} and will gain guarantees on
solution quality for submodular functions.
This algorithm is also similar to the prior algorithm for greedy submodular
maximization (Alg.~\ref{alg:greedy_submodular}).
However, because the maximization steps are now over individual blocks,
computation of a complete solution requires only a single pass over the ground
set.
Yet, crucially, solutions via Alg.~\ref{alg:local_greedy} satisfy the same bounds
as for Alg.~\ref{alg:greedy_submodular}:
$\setfun(X^\mathrm{g}) \geq 1/2 \setfun(X^\opt)$.

The form of the locally greedy algorithm is particularly desirable for
distributed implementations on multi-robot systems.
Recall that a simple partition matroid can be used to model the joint space in a
multi-robot planning problem.
In this case, $\block_i$ is the local set of actions available to the
$i^\mathrm{th}$ robot.
Robots can then plan for themselves by search over these local sets of
actions and given access to prior robots' plans but without access to the entire
ground set.
However, dependency on prior solution elements is a barrier to parallel
computation in distributed implementations.
One of the contributions of this thesis is to eliminate that barrier in
multi-robot sensing problems.

\subsubsection{Inequalities and intuition for submodular maximization}

Monotonicity serves the starting point in the proofs of many common performance
bounds.
If $X^\opt$ is an optimal solution to a submodular maximization problem
(Prob.~\ref{prob:submodular_general}) and $X\subseteq \ground$, then
\begin{align}
  \setfun(X^\opt) \leq \setfun(X^\opt,X).
  \label{eq:monotonicity_proofs}
\end{align}
Typically, $X$ is the full solution obtained by the greedy algorithm, but that
is not strictly necessary~\citep{grimsman2018tcns}.
We can exchange an element of the optimal solution $x^\opt \in X^\opt$ for an
element of a greedy solution by applying the principle of greedy choice
\begin{align}
  \setfun(x^\opt|\hat X) \leq \setfun(x^\mathrm{g}|\hat X)
  \label{eq:greedy_choice}
\end{align}
where $\hat X$ is generally a subset of the greedy solution and
$x^\opt\in B$ belongs to the feasible set $B\subseteq\ground$ of a greedy
maximization step that solves $x^\mathrm{g}=\max_{x \in B} \setfun(x|\hat X)$.
Observe that if we instead have an expression for $\setfun(x|\bar X)$ where
$\hat X \subseteq \bar X \subseteq \ground$ we can apply
submodularity to replace $\bar X$ with $\hat X$ because
$\setfun(x|\bar X) \leq \setfun(x|\hat X)$.
If the greedy solutions are also suboptimal, we may also apply the expression
for the suboptimality to \eqref{eq:greedy_choice} to replace the optimal greedy
solution elements with a suboptimal ones.

\subsubsection{Online bounds}
\label{sec:online_bounds}

As \citet{minoux1978} observes, this intuition can be applied more
generally to obtain online bounds~\citep{krause2008,golovin2011jair}
for solutions obtained by any algorithm such
as by applying \eqref{eq:monotonicity_proofs} and bounding the terms of the sum
in $\setfun(X^\opt,X) \leq \sum_{x^\opt\in X^\opt} \setfun(x^\opt|X)$
e.g. by maximizing
over ground set $\ground$ or an independence system $\independence$.
These online bounds have been used to demonstrate that real solutions by greedy
algorithms can often perform significantly better than worst case bounds in
practice~\citep{krause2008,golovin2011jair}.
Similar bounds will later prove to be useful for characterizing solution quality
for multi-robot exploration.

\section{Distributed and parallel algorithms for submodular maximization}
\label{sec:distributed_parallel_submodular}

An important challenge in this thesis is to reduce computation time for
submodular maximization in multi-robot sensor planning.
Ideally, this computation time would be independent of the number of robots in
the team, and we would achieve this by taking advantage of parallel
computation.
However, that is easier said than done due to sequential constraints on the
decision process and constraints on information access.
Further, although there is a growing body of work on parallel and distributed
submodular maximization, many existing methods are impractical for
multi-robot sensor planning.
Still, a few works propose methods that are relevant to this thesis,
and we will seek to take advantage of their contributions.

\subsection{Assumptions and model for distributed computation}
\label{sec:background_model_of_computation}

\begin{figure}
  \centering
  \def\svgwidth{0.8\linewidth}
\begingroup%
  \makeatletter%
  \providecommand\color[2][]{%
    \errmessage{(Inkscape) Color is used for the text in Inkscape, but the package 'color.sty' is not loaded}%
    \renewcommand\color[2][]{}%
  }%
  \providecommand\transparent[1]{%
    \errmessage{(Inkscape) Transparency is used (non-zero) for the text in Inkscape, but the package 'transparent.sty' is not loaded}%
    \renewcommand\transparent[1]{}%
  }%
  \providecommand\rotatebox[2]{#2}%
  \newcommand*\fsize{\dimexpr\f@size pt\relax}%
  \newcommand*\lineheight[1]{\fontsize{\fsize}{#1\fsize}\selectfont}%
  \ifx\svgwidth\undefined%
    \setlength{\unitlength}{1220.18009696bp}%
    \ifx\svgscale\undefined%
      \relax%
    \else%
      \setlength{\unitlength}{\unitlength * \real{\svgscale}}%
    \fi%
  \else%
    \setlength{\unitlength}{\svgwidth}%
  \fi%
  \global\let\svgwidth\undefined%
  \global\let\svgscale\undefined%
  \makeatother%
  \begin{picture}(1,0.42799704)%
    \lineheight{1}%
    \setlength\tabcolsep{0pt}%
    \put(0,0){\includegraphics[width=\unitlength,page=1]{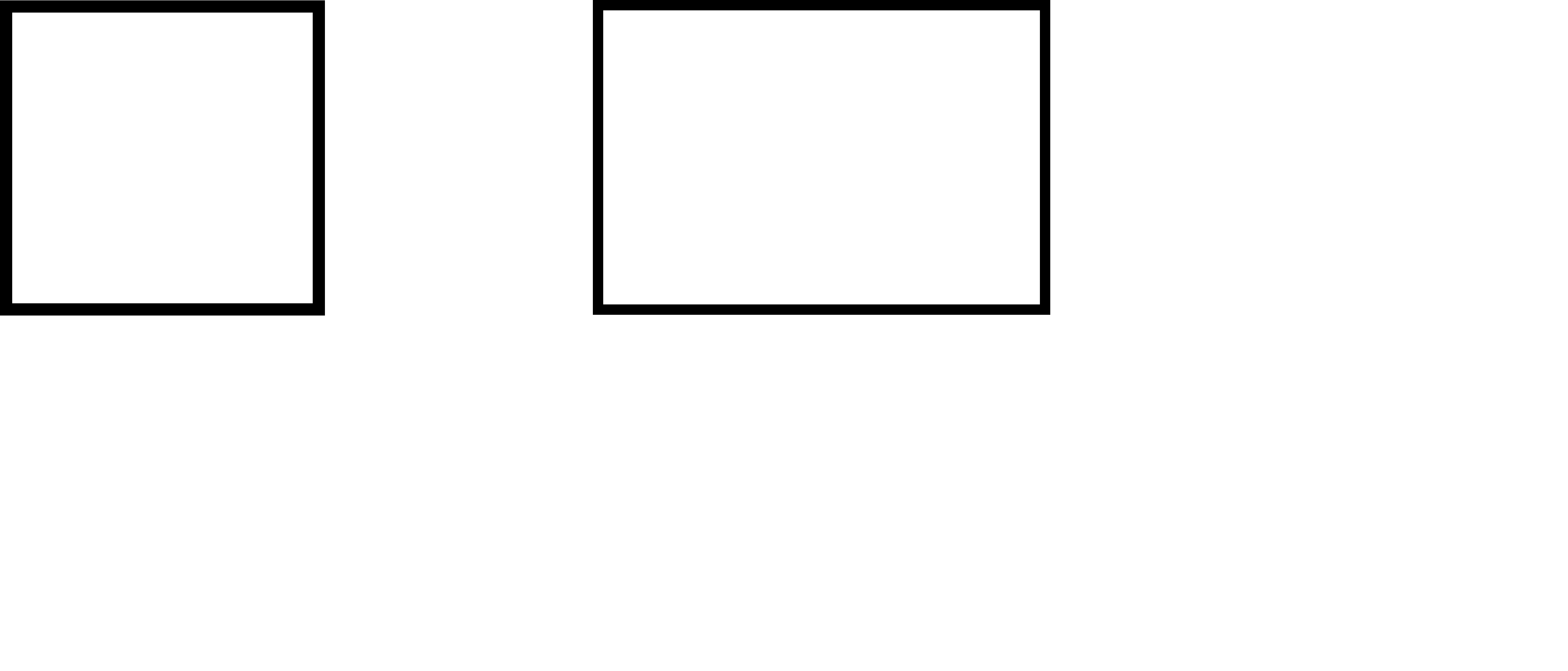}}%
    \put(0.39495327,0.39010043){\color[rgb]{0,0,0}\makebox(0,0)[lt]{\lineheight{0.5}\smash{\begin{tabular}[t]{l}Receding-Horizon\\Planning\end{tabular}}}}%
    \put(0,0){\includegraphics[width=\unitlength,page=2]{computational_model.pdf}}%
    \put(0.01714962,0.39010043){\color[rgb]{0,0,0}\transparent{0.98000002}\makebox(0,0)[lt]{\lineheight{0.5}\smash{\begin{tabular}[t]{l}Distributed\\Perception\end{tabular}}}}%
    \put(0,0){\includegraphics[width=\unitlength,page=3]{computational_model.pdf}}%
  \end{picture}%
\endgroup%

  \caption[Illustration of the model of computation
  for distributed sensor planning]{
    Illustration of the model of computation for distributed sensor planning.
    Each robot $i \in \robots$ has access to a block of a partition matroid
    $\block_i$ which represents that robot's space of control actions.
    The perception system then provides a locally consistent model $\theta_i$.
    Given this model, a robot can accurately approximate marginal gains
    in the sensing objective
    $\setfun_{\theta_i}(x|Y)$
    for its own actions $x \in \block_i$ given a set of others' actions $Y$
    (e.g. numbers of newly observed cells in an occupancy map for $x$ or the
    reduction in uncertainty of positions of nearby targets).
    But, that robot
    \emph{cannot compute marginal gains for other, possibly distant, robots}
    (and distant targets or regions of an occupancy map).
  }%
  \label{fig:computational_model}
\end{figure}

Figure~\ref{fig:computational_model} illustrates the distributed setting for the
submodular maximization problems that we will study.
In this work, robots will solve submodular maximization problems
(Prob.~\ref{prob:submodular_partition_matroid})
where each robot $i \in \robots$ selects an action from
a block of a partition matroid $x \in \block_i$.
We assume robots have access to a locally approximate model of the environment
which we refer to here as $\theta_i$
(and which we will omit for most of this text)
via a distributed perception system.
With this model a robot can approximate marginal gains for its own actions
$\setfun_{\theta_i}(x|Y)$ given others $Y \subseteq \ground$.
However, a robot may not evaluate marginal gains for assignments to
other robots because that robot may not have an up-to-date model of
distant regions of an occupancy map or estimates of positions of distant
targets.

Additionally, while evaluating different parallel and distributed algorithms, we
will
sometimes write results in terms of the number of robots $\numrobot = |\robots|$
based on
the following assumption:
\begin{assumption}[Proportional problem size]
  \label{assumption:proportionality}
  Consider a team of $\numrobot$ robots and a submodular sensor planning
  problem (in the form of Prob.~\ref{prob:submodular_partition_matroid}).
  The size of the ground set is proportional to the number of robots
  $|\ground| \in \Theta(\numrobot)$;
  the rank of the matroid constraint is $\numrobot$;
  and the blocks of the partition matroid have constant size
  $|\block_i| \in \Theta(1)$ for all $i\in\robots$.
\end{assumption}

\subsection{Quantities for evaluating of parallel and distributed algorithms}
\label{sec:quantify_parallel}

A number of properties of parallel and distributed submodular can contribute to
increases in computation time in multi-robot sensor planning.
Consider a team of $\numrobot$ robots that plan using $O(\numrobot)$ onboard
processors.\footnote{%
  Or, for the purpose of this section, the robots may be supported by a
  centralized resource with the same number of processors.
}
We are interested in the following quantities:
\begin{itemize}
  \item \textbf{Adaptivity:} The number of sequential calls to the objective
    $\setfun$.
    See the next section (Sec.~\ref{sec:low_adaptivity_algorithms})
  \item \textbf{Parallelism:} The number of parallel queries to the objective
    $\setfun$ that an algorithm assumes.
  \item \textbf{Query number:} The maximum number of queries to the objective
    $\setfun$.
    Given an algorithm that runs with $k$ queries, the $\numrobot$ processors
    require at least $O(k/\numrobot)$ time.
  \item \textbf{Query size:} This is the cardinality of the largest set $|X|$
    that an algorithm uses to query the objective $\setfun$. For derivatives
    \eqref{eq:recursive_derivative},
    (such as the marginal gain) this includes elements of the sets representing
    the derivative directions.
\end{itemize}

Previously, Sec.~\ref{sec:submodular_maximization} mentioned that different
algorithms take different numbers of passes over the ground set.
In the same direction, the general greedy algorithm
(Alg.~\ref{alg:greedy_submodular}) involves $\Theta(\numrobot^2)$ queries because
it maximizes over the entire ground set at each step.
On the other hand, the locally greedy algorithm (Alg.~\ref{alg:local_greedy})
only completes a single pass for $\Theta(\numrobot)$ queries.
Parallelizing the general greedy algorithm over $\numrobot$ processors may then
obtain similar computation times as the local algorithm on a single processor.

The total number of queries for the continuous greedy algorithm is more subtle.
However, we can note that \emph{each integration step} requires at least
$\Omega(\numrobot)$ queries
(technically, this scales with the rank of the matroid, but the rank is
$\numrobot$ in all cases that we encounter).
Algorithms that solve subproblems with the continuous greedy algorithm on single
processors require at least $\Omega(\numrobot)$ time.

Additionally, we are not aware of any algorithm (other than what we propose)
that obtains constant-factor suboptimality with a maximum query size that is
less than the number of robots.\footnote{
  The query size for \rrsp{} depends on the size of the largest communication
  neighborhood.
}
Instead, we are able to achieve constant query size and suboptimality in certain
settings by allowing robots to select actions for themselves given limited
access to prior selections.

\subsection{Adaptivity and bounds on numbers of sequential steps}
\label{sec:low_adaptivity_algorithms}

Nominally, greedy methods (as in Alg.~\ref{alg:local_greedy}) operate
sequentially (e.g. with one step per robot) because each decision in the
sequence takes all prior decisions into account.
A few works, that we will discuss later in more detail, consider variations of
such greedy algorithms where agents can ignore some prior
decisions~\citep{gharesifard2017,grimsman2018tcns}.
However, if we fix the degree of suboptimality, the best algorithms in this
class require a number of sequential steps that is
\emph{proportional to the number of robots}, just as for other greedy
algorithms.

More broadly: How much can parallel computation reduce planning time
for submodular maximization with any algorithm?
To answer this question, recent works have begun to study the
\emph{adaptive complexity} of submodular maximization problems.
\citet{balkanski2018stoc} describe the adaptive complexity as
``the minimal number of sequential rounds required to achieve a constant factor
approximation when polynomially-many queries [to the submodular objective] can
be executed in parallel at each round.''
Moreover, \emph{adaptivity} is the worst-case number of sequential steps for a
given algorithm;
an algorithm that executes in $k$ steps is $k$-adaptive.
By virtue of parallel execution, none of the queries to the objective during a
given round can depend on another, and the set of queries for a given
adaptive round is known at the beginning of the round.

\paragraph*{Caution:} Adaptivity is related to but distinct from computational
complexity and the (parallel) duration of computation.
Adaptivity provides a \emph{lower bound on the growth of the computation
time}, but the actual computation time may grow more quickly,
depending on factors such as the number of available processors and the
application context.
Nevertheless, decreasing adaptivity is a necessary first step toward
developing constant-time algorithms in this thesis.

\paragraph*{}
Looking toward applications in multi-robot sensing,
the best known algorithm for submodular maximization with a matroid constraint
has adaptivity:
$O(\log|\ground |\log (\matroidrank\independence))$%
~\citep{balkanski2019,chekuri2019}.
Further, \citet{balkanski2018stoc} proved that no algorithm
can obtain adaptivity better than $\tilde\Omega(\log|\ground|)$
(lower-bounded by $\log|\ground|$ up to lower-order logarithmic terms).
Critically, these results improve on prior algorithms for submodular
maximization that obtain constant factor guarantees, including the continuous
greedy algorithm~\citep{calinescu2011}, which has adaptivity of at least
$\Omega(|\ground|)$~\citep[Sec.~3.1]{balkanski2019} in general.\footnote{
  The adaptivity of the continuous greedy algorithm depends on the number of
  discrete steps in an integral that are necessary to obtain a performance
  bound.
  \citet{balkanski2019} (informally) provide a \emph{lower} bound on
  this number.
  Additionally, suboptimality guarantees~\citep{calinescu2011} typically
  produce upper bounds on the adaptivity.
}
However, we caution that the
continuous greedy algorithm
may have better or even constant
adaptivity on the class of problems that we study---the same analysis techniques
that we apply to sequential planning in this thesis might also be applicable
to the continuous greedy algorithm to obtain performance bounds where the number
of integration steps is independent of the number of robots.

Further, one limitation of low-adaptivity methods is that converging to
non-trivial constant factor bounds (e.g. to obtain a $1-1/e-\epsilon$ for small
$\epsilon$) can require a large (polynomial) number of steps in both theory and
practice as discussed by~\citep{breuer2019} and~\citep{li2020}.
However, \citet{breuer2019} presented an algorithm for cardinality-constrained
problems which obtains both low-adaptivity and efficiency in practice and is the
first to do so.
Still, no such algorithm exists for more general matroid constraints.
Likewise, although low-adaptivity results characterize what is possible for
submodular maximization algorithms,
they also depend on shared memory models of computation\footnote{%
  Parallel Random Access Machines (PRAM)
}
which makes them difficult to adapt to distributed computation on multi-robot
teams.

Table~\ref{tab:adaptivity} summarizes the results we have discussed in this
section in terms of the number of robots $\numrobot$ in a sensing problem.
Moreover, the adaptive complexity of submodular maximization with a
matroid constraint lies in $\tilde\Omega(\log\numrobot)$.
In order to develop algorithms with constant adaptivity, we will need to restrict
the problem class further and rely on additional structural properties of the
sensing problems we study.


\begin{table}
  \caption[Adaptivity results for submodular maximization on partition
  matroids]{%
    Adaptivity results (numbers of sequential steps) for submodular maximization
    on partition matroids within a constant factor of optimal
    in terms of the number of robots $\numrobot$.
    This thesis presents algorithms that achieve $O(1)$ adaptivity on a slightly
    restricted class of submodular maximization problems.
  }%
  \label{tab:adaptivity}
  \centering
  \renewcommand{\tabularxcolumn}[1]{>{\arraybackslash}m{#1}}
  \begin{tabularx}{\linewidth}{Xcc}
    Description
    &
    \pbox{4.5cm}{Lower bound  \\ (best possible \\ adaptivity)}
    &
    \pbox{4.0cm}{Upper bound \\ (best known \\ algorithm)}
    \\\belowtoprule
    For \emph{any} algorithm that achieves constant-factor performance
    (general adaptive complexity)~\citep{balkanski2018stoc,balkanski2019}*
    &
    $\tilde\Omega(\log \numrobot)$
    &
    $O(\log^2 \numrobot)$
    \\
    For variations of the local greedy algorithm (Alg.~\ref{alg:local_greedy})
    that allow robots to ignore prior
    decisions~\citep{gharesifard2017,grimsman2018tcns}
    &
    $\Omega(\numrobot)$
    &
    $O(\numrobot)$\\\bottomrule
  \end{tabularx}
  *{%
    \footnotesize
    The tilde in $\tilde\Omega$ or ``Soft-Omega'' indicates that the expression
    absorbs lower-order logarithmic terms.
  }
\end{table}

\subsection{Parallel algorithms for submodular maximization}
\label{sec:parallel_submodular}

Many parallel algorithms for submodular maximization
apply the popular \emph{MapReduce}\footnote{%
  The MapReduce model~\citep{dean2008} is popular due to its simplicity, and
  roughly consists of three steps:
  1) Distribute data across processors
  2) Execute (map) some function on that data in parallel
  (say by squaring values)
  3) Combine (reduce) the outputs from the prior step
  (e.g. by summing the squares).
}
model to solve problems with
cardinality~\citep{mirzasoleiman2013nips}
and matroid~\citep{barbosa2016a,barbosa2015icml,kumar2015} constraints.
Although these algorithms may appear widely different at first glance, they
share features that make them less suitable for multi-robot sensing.
Each of these approaches~\citep{%
mirzasoleiman2013nips,barbosa2016a,barbosa2015icml,kumar2015}
seeks to reduce computation time by running greedy algorithms
(sequential greedy algorithms or the continuous greedy algorithm)
on small subsets of the ground set $\ground$.
In doing so, they distribute such subsets across processors and run some
greedy algorithm on each subset in parallel to produce candidate solutions.
Although the mechanisms for producing subsets differ,
those that run sequential greedy algorithms~\citep{%
mirzasoleiman2013nips,barbosa2015icml,kumar2015%
}
on those subsets all have
high adaptivity (at least $O(\numrobot)$) which means that computation times
will inevitably increase and become intractable for large enough numbers of
robots.
Likewise, all approaches, including those that apply the continuous greedy
algorithm~\citep{barbosa2016a},
produce full rank solutions when solving subproblems and so have
$\Omega(\numrobot)$ queries and computation time on each processor.

Moreover, these approaches are also generally impractical in the multi-robot
settings that we consider in this thesis as each robot would have to compute
candidate solutions for the entire team.
Additionally, distributing necessary information for that computation and
collecting the candidate solutions would involve a significant amount of
communication.
By contrast, this thesis proposes approaches with low adaptivity where robots
plan for themselves and communicate individual solution elements rather than
full solutions.

\subsection{Distributed algorithms for sensing and submodular maximization}

A number of works have proposed distributed algorithms that are applicable to
submodular maximization for different kinds of networked multi-robot and
multi-agent systems~\citep{%
choi2009tro,williams2017icra,mokhtari2018,xie2019,gharesifard2017,grimsman2018tcns}.
However, none of these are able to provide suboptimality guarantees for in
constant time as we would like for the problems we study.

Additionally, not all networked systems are relevant to the multi-robot settings
that we study.
\citet{mokhtari2018} and \citet{xie2019} present algorithms based on a
setting where the objective is distributed across processors (e.g.
$\setfun(X)=\sum_{i=1}^\numrobot \setfun_i(X)$) such as if the objective is
derived from data that is spread out across all processors.
Their approaches are interesting due to how they apply continuous
consensus techniques~\citep{olfati2004,francesco2018} to distribute the
continuous greedy algorithm across processors.
On the other hand, all processors operate on the entire ground
set and constraint structure which requires $\Omega(\numrobot)$ time
(much like algorithms in Sec.~\ref{sec:parallel_submodular}).

Instead, \citet{robey2019} extend the work of \citet{mokhtari2018} to the
same setting that we apply for multi-robot sensing with the ground set
and partition matroid distributed across the computation nodes.
However, convergence still slows with increasing numbers of robots as evident
from inspection of the error bound~\citep[Theorem~1]{robey2019} and due to
limitations on adaptivity of the continuous greedy algorithm
(refer to Sec.~\ref{sec:low_adaptivity_algorithms}, \citep{balkanski2019}).
Although we will not present novel variants of the continuous greedy
algorithm, the methods for analysis of redundancy that we develop
could possibly also be applied to algorithms like what~\citet{robey2019}
describe to provide constant time guarantees for the problems we study.
Otherwise, the primary challenge for applying distributed versions of the
continuous greedy algorithm~\citep{robey2019,segui2015iros} to the problems we
study is adapting the continuous greedy algorithm to use single-robot planners
as maximization oracles.
Doing so would require solving path planning problems on the
multilinear extension of the objective~\citep{calinescu2011} which requires
costly sampling.
Additionally, the large solution spaces endemic to path planning problems may
adversely impact suboptimality as error bounds for the continuous greedy
algorithm depend on the size of the ground set.

\subsubsection{Distributed task assignment for submodular maximization}

Multi-robot sensor planning is also closely related to task assignment as both
can often written in the form of similar submodular maximization problems.
Specifically, there is a special case of submodular maximization with a partition
matroid constraint
(Prob.~\ref{prob:submodular_partition_matroid})
that is relevant to task assignment, the submodular welfare
problem.\footnote{%
  \citet{vondrak2008} provides an example of the reduction from a welfare
  problem to Prob.~\ref{prob:submodular_partition_matroid}.
}
Seeking to solve such problems, \citet{choi2009tro} developed
the consensus-based bundle algorithm (CBBA) which implements a
distributed version of the general greedy algorithm
(Alg.~\ref{alg:greedy_submodular}).
Later, \citet{williams2017icra} demonstrated that this approach also applies to
problems with more complex constraints such as intersections of matroids.
Most importantly, several works have extended~\citep{%
johnson2017jais,shin2019%
}
CBBA, and others have demonstrated distributed implementations have been
demonstrated on aerial robots~\citep{ponda2012}.
Although CBBA-like approaches can provide significant speedups on
relevant problems compared to the general greedy algorithm (which these
approaches implement), they still have $O(\numrobot)$ adaptivity in the worst
case.
However, CBBA can also converge more quickly in practice which is an important
point for comparison.

\subsubsection{Game theory for distributed submodular maximization}
\label{sec:game_theory}

Distributed submodular maximization has also been studied from the perspective
of game theory.
Specifically, various equilibria satisfy constant factor
suboptimality guarantees~\citep{roughgarden2009}.
In fact, the strategy sets for games with submodular utility form a partition
matroid, and Nash equilibria satisfy the same performance guarantees as greedy
algorithms on matroid-constrained problems~\citep{vetta2002focs}.

When agents employ algorithms that converge to such equilibria, game theoretic
methods can be interpreted as producing algorithms for distributed
optimization.
Moreover, \citet{goundan2007} observed that one method of doing so would
implement the locally greedy algorithm (Alg.~\ref{alg:local_greedy}) exactly.
Additionally, this line of work has been applied to planning and sensing
problems~\citep{kumar2017aaai} and has produced interesting results for coverage
on Voronoi decompositions~\citep{marden2015}.
Still, game theoretic methods do not yet provide tools for
reducing the time to find a near-optimal solution which is a key challenge for
this thesis.

\subsubsection{Distributed planning on directed acyclic graphs}
\label{sec:dag_models}

Increasing numbers of robots, unreliable and constrained communication, and
limited planning time all motivate a closer look at algorithms and models of
computation for submodular maximization applied to multi-robot systems.
A few works~\citep{gharesifard2017,grimsman2018tcns} have begun to study the
locally greedy algorithm (Alg.~\ref{alg:local_greedy}) on distributed
computation models that are relevant to multi-robot sensor planning (where each
agent in a network selects an action from a block of a partition matroid).
Specifically, \citet{gharesifard2017} consider variants of the locally greedy
algorithm where agents have limited information about others' decisions
according to a directed acyclic graph (DAG) Fig.~\ref{fig:dag_models}).
Crucially, this framework can describe planners that employ parallel
computation~\citep{sun2020acc}, and for this reason, we apply similar models
throughout this thesis.
\citet{gharesifard2017} also provided worst-case analysis on this model (see
also Sec.~\ref{sec:low_adaptivity_algorithms}) which \citet{grimsman2018tcns}
then extended with tighter bounds.
However, planners based on these worst-case results have $\Omega(\numrobot)$
adaptivity when the suboptimality is held constant.
To address this, we will focus on developing planners that can take advantage of
features of problem structure beyond submodularity in order to provide
near-optimal results with a constant number of sequential steps.

\begin{figure}
  \newcommand{\quadblack}{\includegraphics[width=1.5cm]{../quadrotors/quadrotor_black.pdf}}
  \centering
  \begin{subfigure}[t]{0.8\linewidth}
    \centering
    \tikzsetnextfilename{complete_dag}
    \begin{tikzpicture}[xscale=2.5]
      \tikzset{vertex/.style = {shape=circle,draw,minimum size=1.5em}}
      \tikzset{edge/.style = {->,> = latex, ultra thick}}

      \node (1) at (0,0) {\quadblack};
      \node (2) at (1,0) {\quadblack};
      \node (3) at (2,0) {\quadblack};
      \node (4) at (3,0) {\quadblack};
      \node (5) at (4,0) {\quadblack};
      \node (6) at (5,0) {\quadblack};

      \draw[edge] (1) to                               (2);
      \draw[edge] (1) to[bend right=60, looseness=1.5] (3);
      \draw[edge] (1) to[bend  left=60, looseness=1.5] (4);
      \draw[edge] (1) to[bend right=60, looseness=1.5] (5);
      \draw[edge] (1) to[bend  left=60, looseness=1.5] (6);

      \draw[edge] (2) to                               (3);
      \draw[edge] (2) to[bend right=60, looseness=1.5] (4);
      \draw[edge] (2) to[bend  left=60, looseness=1.5] (5);
      \draw[edge] (2) to[bend right=60, looseness=1.5] (6);

      \draw[edge] (3) to                               (4);
      \draw[edge] (3) to[bend right=60, looseness=1.5] (5);
      \draw[edge] (3) to[bend  left=60, looseness=1.5] (6);

      \draw[edge] (4) to                               (5);
      \draw[edge] (4) to[bend right=60, looseness=1.5] (6);

      \draw[edge] (5) to                               (6);
    \end{tikzpicture}
    \caption{A full sequential communication graph}%
    \label{subfig:full_sequential_illustration}
  \end{subfigure}
  \begin{subfigure}[t]{0.49\linewidth}
    \centering
    \tikzsetnextfilename{cliques_dag}
    \begin{tikzpicture}[xscale=2.5]
      \tikzset{vertex/.style = {shape=circle,draw,minimum size=1.5em}}
      \tikzset{edge/.style = {->,> = latex, ultra thick}}

      \node (1a) at (0,0) {\quadblack};
      \node (1b) at (1,0) {\quadblack};
      \node (1c) at (2,0) {\quadblack};

      \draw[edge] (1a) to                (1b);
      \draw[edge] (1b) to                (1c);
      \draw[edge] (1a) to[bend right=60, looseness=1.5] (1c);

      \node (2a) at (0,2.5) {\quadblack};
      \node (2b) at (1,2.5) {\quadblack};
      \node (2c) at (2,2.5) {\quadblack};

      \draw[edge] (2a) to                (2b);
      \draw[edge] (2b) to                (2c);
      \draw[edge] (2a) to[bend left=60, looseness=1.5] (2c);
    \end{tikzpicture}
    \caption{A graph consisting of two cliques}%
    \label{subfig:cliques_illustration}
  \end{subfigure}
  \begin{subfigure}[t]{0.49\linewidth}
    \centering
    \tikzsetnextfilename{maximal_parallel_dag}
    \begin{tikzpicture}[xscale=2.5]
      \tikzset{vertex/.style = {shape=circle,draw,minimum size=1.5em}}
      \tikzset{edge/.style = {->,> = latex, ultra thick}}

      \node (1a) at (0,0) {\quadblack};
      \node (1b) at (1,0) {\quadblack};
      \node (1c) at (2,0) {\quadblack};

      \draw[edge] (1a) to                (1b);
      \draw[edge] (1b) to                (1c);
      \draw[edge] (1a) to[bend right=60, looseness=1.5] (1c);

      \node (2a) at (0,2.5) {\quadblack};
      \node (2b) at (1,2.5) {\quadblack};
      \node (2c) at (2,2.5) {\quadblack};

      \draw[edge] (2a) to                (2b);
      \draw[edge] (2b) to                (2c);
      \draw[edge] (2a) to[bend left=60, looseness=1.5] (2c);

      \draw[edge] (1a) to (2b);
      \draw[edge] (1a) to (2c);
      \draw[edge] (1b) to (2c);
      \draw[edge] (2a) to (1b);
      \draw[edge] (2a) to (1c);
      \draw[edge] (2b) to (1c);
    \end{tikzpicture}
    \caption{A maximal graph for parallel planning}%
    \label{subfig:parallel_illustration}
  \end{subfigure}
  \caption[Directed acyclic graphs for sequential planning]{%
    This figure illustrates several directed acyclic communication graphs
    for submodular maximization using the model proposed by
    \citet{gharesifard2017}.
    (\subref{subfig:full_sequential_illustration})
    The standard locally greedy algorithm (Alg.~\ref{alg:local_greedy})
    corresponds to a completed directed acyclic graph;
    (\subref{subfig:cliques_illustration})
    and although analysis by \citet{gharesifard2017} and
    \citet{grimsman2018tcns} leads to planners consisting of cliques,
    (\subref{subfig:parallel_illustration})
    algorithms proposed in this thesis use subsets of edges from maximal
    communication graphs with parallel computation.
  }%
  \label{fig:dag_models}
\end{figure}


\chapter{Toward Distributed Multi-Robot Exploration}
\setdatapath{./fig/distributed_multi-robot_exploration}
\label{chapter:distributed_multi-robot_exploration}

This chapter (which originally appeared in~\citep{corah2017rss,corah2019auro})
introduces an early version of the methods developed in this thesis in the
context of a team of robots exploring an unknown environment.
We will continue to use Monte-Carlo tree search for single-robot planning
throughout this thesis as we also do in other related
works~\citep{corah2019ral,goel2019fsr}.
The parallel algorithm for submodular maximization (\dgreedy{}) that we describe
is also a precursor to methods based on Randomized Sequential Partitions
(\rsp) which we will introduce later on.
Although possibly interesting in its own right, an actual distributed
implementation of \dgreedy{} would be more complex than the distributed
implementation of \rsp{} which we introduce in
Chapter~\ref{chapter:distributed_communication}.
Moreover, \dgreedy{} requires more communication than \rsp{} methods and does
not fully eliminate certain elements of sequential computation.

However, some of the contributions of this chapter are unique in this thesis.
For example, this is the only chapter to address inter-robot collisions.
Although this introduces some additional complexity, similar methods could be
applied to the \rsp{} planners that we discuss later.
Similarly, this chapter provides the only results involving physical robots.

\section{Introduction to exploration}

We pose multi-robot exploration as the problem of actively mapping environments
by planning actions for a team of sensor-equipped robots to maximize informative
sensor measurements.
In this work, we address the problem of planning for exploration with
large teams of robots using distributed computation and emphasize online
planning and operation in confined and cluttered environments.

\begin{figure}
  \begin{center}
    \begin{subfigure}{0.25\linewidth}
      \includegraphics[width=\linewidth, trim={660 190 490 70}, clip]{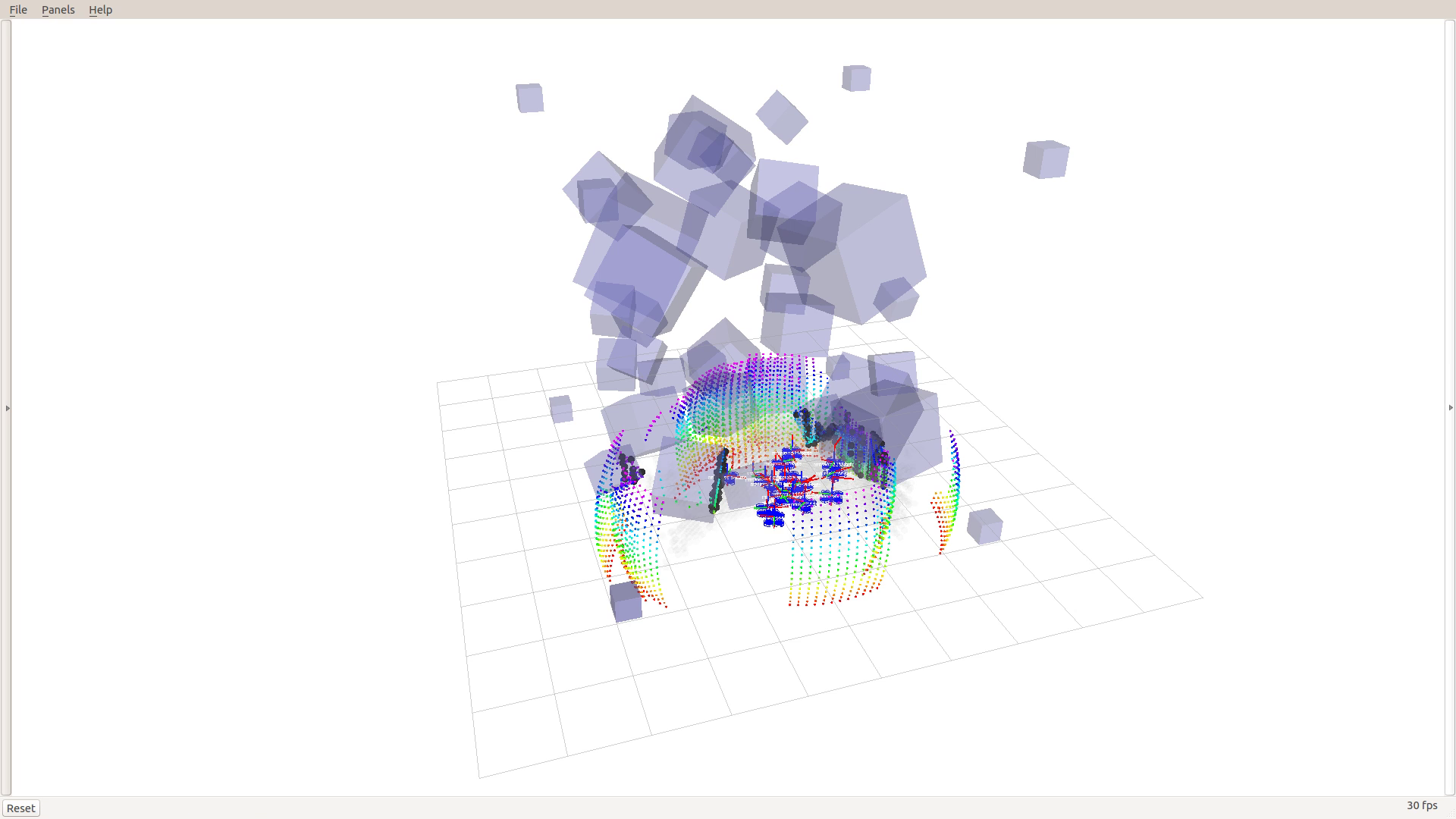}%
      \caption{}\label{subfig:intro1}
    \end{subfigure}%
    \begin{subfigure}{0.25\linewidth}
      \includegraphics[width=\linewidth, trim={660 190 490 70}, clip]{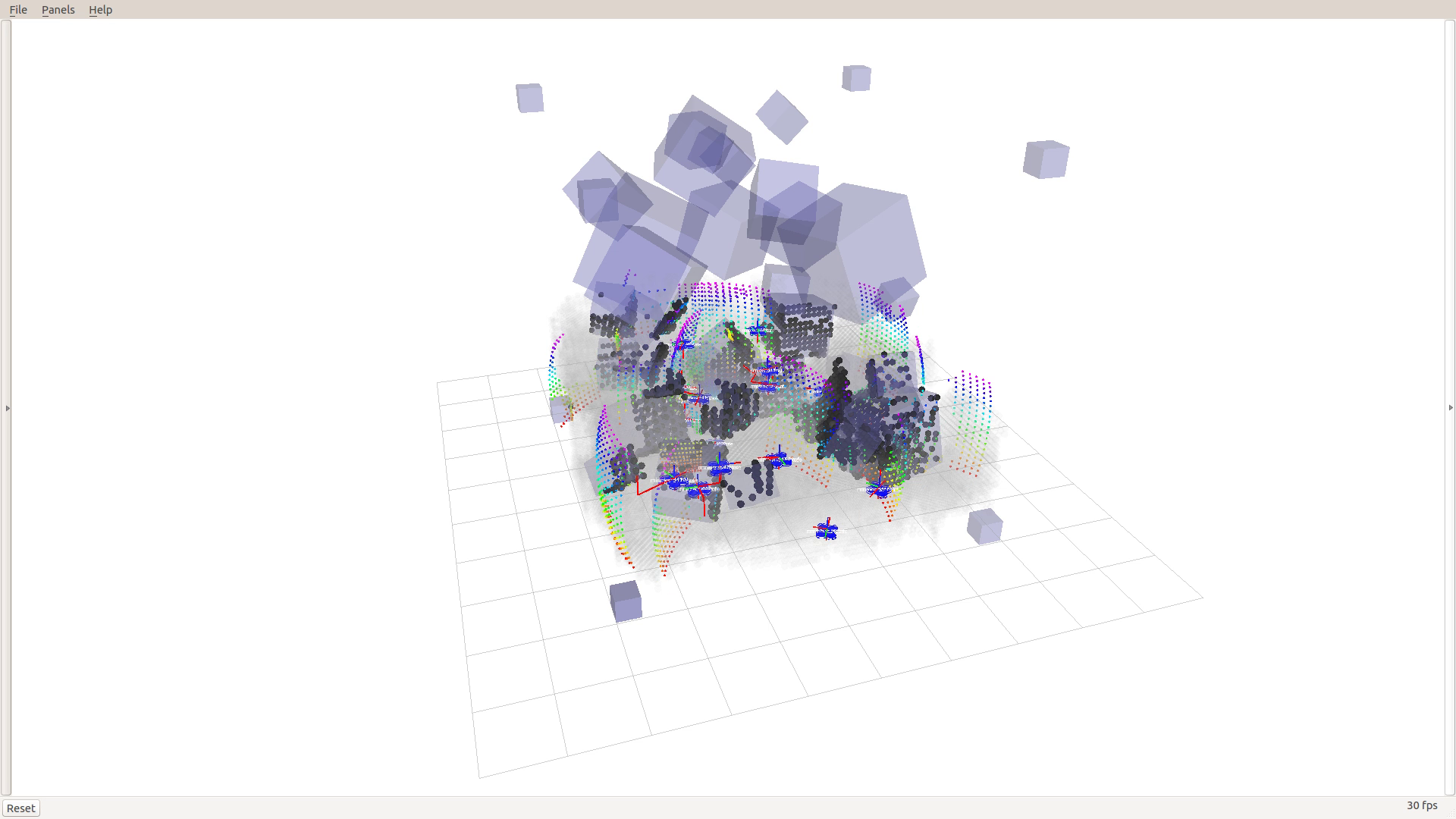}%
      \caption{}\label{subfig:intro2}
    \end{subfigure}%
    \begin{subfigure}{0.25\linewidth}
      \includegraphics[width=\linewidth, trim={660 190 490 70}, clip]{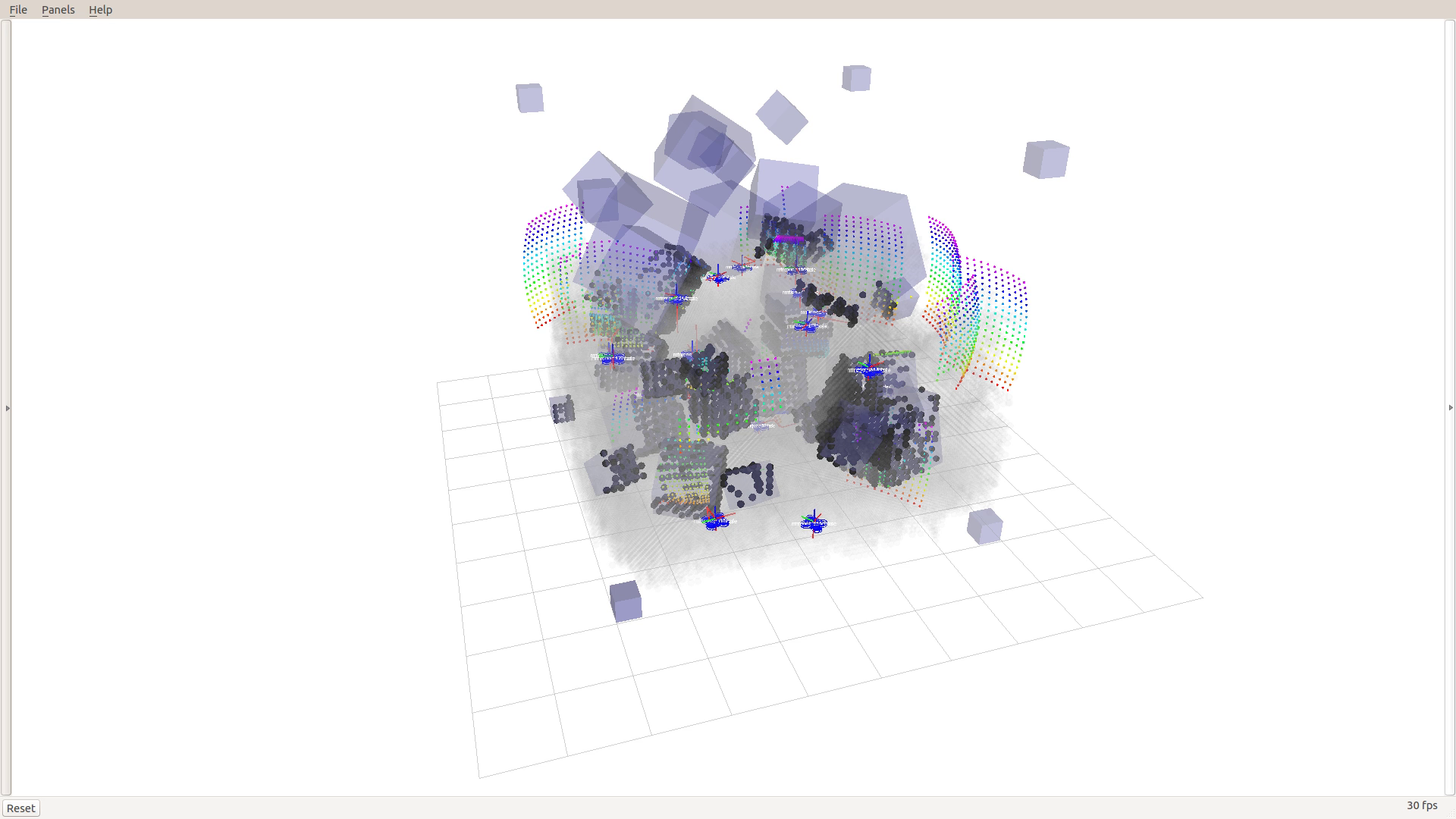}%
      \caption{}\label{subfig:intro3}
    \end{subfigure}%
    \begin{subfigure}{0.25\linewidth}
      \includegraphics[width=\linewidth, trim={660 190 490 70}, clip]{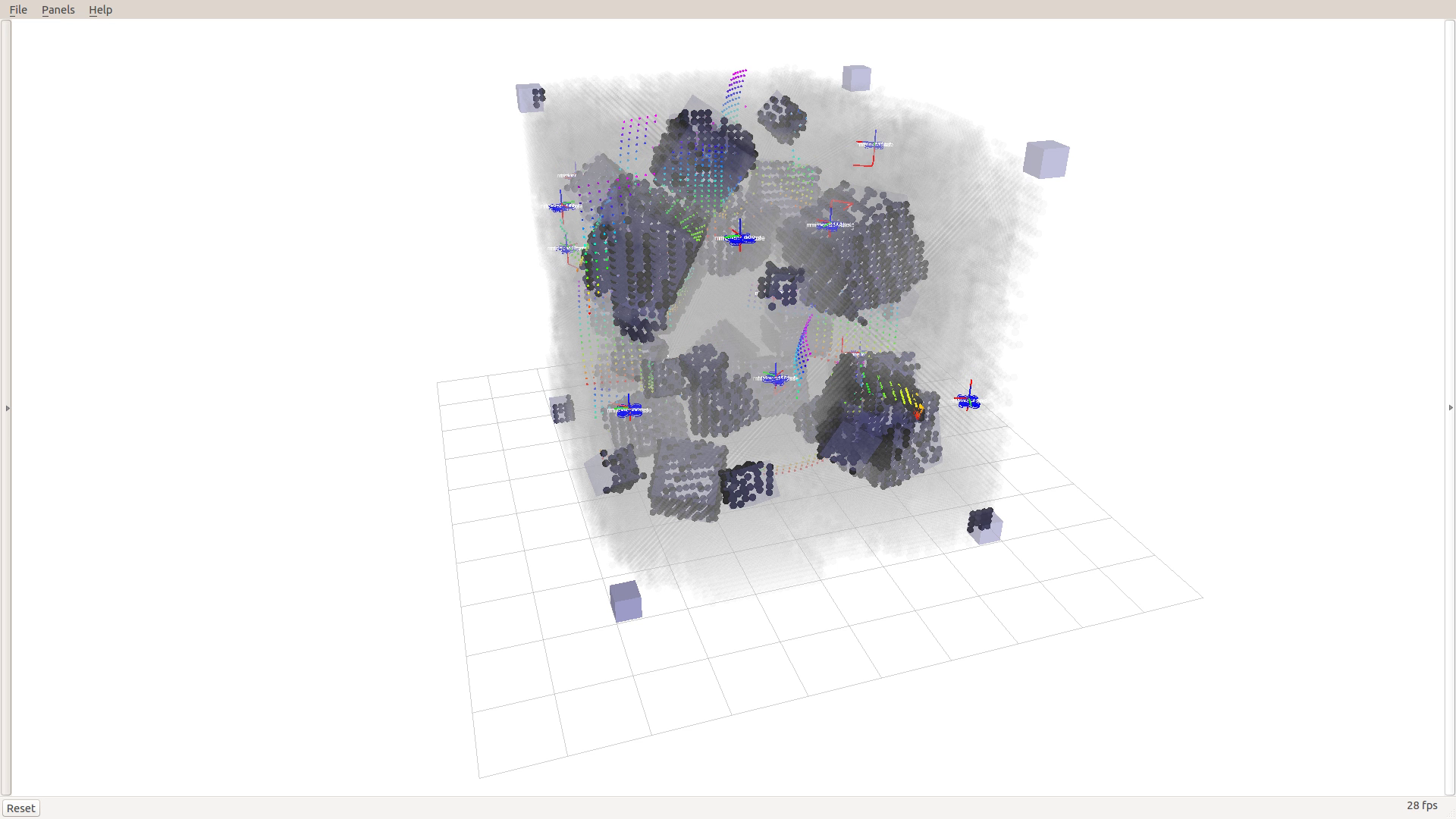}%
      \caption{}\label{subfig:intro4}
    \end{subfigure}
  \end{center}
  \caption[An example exploration experiment]
  {%
    An example exploration experiment.
    A multi-robot team explores
    a three-dimensional environment cluttered with numerous obstacles (cubes)
    while using an online distributed planner.
    Known empty space is gray, and occupied space is black.
    Robots are shown in blue with red trajectories and obtain rainbow-colored
    point-cloud observations from their depth cameras.
    (\subref{subfig:intro1})
    The robots begin with randomized initial positions near a lower
    edge of the exploration environment which is bounded by a cube.
    (\subref{subfig:intro2})
    After entering the environment robots spread out to cover the
    bottom of the cubic environment
    (\subref{subfig:intro3})
    and then proceed upward to
    cover more of the volume.
    (\subref{subfig:intro4})
    Given enough time the robots explore the entire
    environment.
  }\label{fig:intro_image}
\end{figure}

Informative planning problems of this form are
known to be NP-Hard~\citep{krause2008jmlr}.
Rather than attempt to find an optimal solution in possibly
exponential time, we seek approximate solutions with bounded suboptimality that
can be found efficiently in practice.
Sequential planning techniques (Alg.~\ref{alg:local_greedy}) are common in
active sensing and
exploration~\citep{atanasov2015icra,charrow2014auro,regev2016iros}.
In this chapter, we will refer to Alg.~\ref{alg:local_greedy} as performing
sequential greedy assignment (\greedy{}).
Here, \greedy{} assigns plans to robots in sequence using a
single-robot planner to maximize mutual information between a robot's future
observations and the explored map given knowledge of plans assigned prior robots
in the planning sequence.
Recall that the properties of mutual information objectives ensure that
techniques such as \greedy{} achieve suboptimality within half of
optimal~\citep{fisher1978} and that suboptimal single-robot planners achieve
similar bounds~\citep{singh2009}.

The sequential nature of \greedy{} implies that planning time increases at least
linearly with the number of robots and precludes online planning for large
teams.
Increasing computation time is especially relevant to exploration problems
as new information about occupancy significantly affects both feasible and
optimal plans and necessitates reactive planning to enable high rates of
exploration.
We propose a modified version of \greedy{}, distributed sequential
greedy assignment (\dgreedy{}), which consists of a fixed number of sequential
planning rounds.
At the beginning of each round, robots plan in parallel using a single-robot
planner.
A subset of those plans is chosen to minimize the difference between the
information gain for the entire subset and the sum of the
information gains for each robot individually, which does not consider
redundancy between robots.
We obtain a performance bound in terms of the result by~\citet{singh2009} that
explicitly describes the additional suboptimality due to parallel planning as
the accrual of differences between actual and computed objective values during
the planning process.
In doing so, we reduce the distributed planning problem to selection of
subsets of plans after each parallel planning phase that minimize these
differences.
Thus, we are able to take advantage of decoupled observations and distributed
computation to improve computation times for online planning without
compromising exploration performance.

However, robots may collide with each other and the environment, and aerial
robots have non-trivial dynamics.
Therefore, we also introduce inter-robot collision constraints and modify our
algorithms to guarantee collision-free operation.
We apply this approach to teams of aerial robots and demonstrate collision-free
exploration both in simulation and experimentally in a motion capture arena.

\section{Multi-robot exploration formulation}
This section describes the problem of distributed multi-robot exploration.
We begin by describing the system and environment models and then introduce
the planning problem as a finite-horizon optimization.

\subsection{System model}
Consider a team of robots, $\robots=\{r_1,\ldots,r_{\numrobot}\}$, engaged in exploration of
some environment $m$.
The dynamics and sensing are described by
\begin{align}
  x_{t} &= \dynamics(x_{t-1},u), \label{eq:dynamics}\\
  y_{t} &= \observationfunction(x_t, m) + \nu \label{eq:observation}
\end{align}
where $x_t$ represents a robot's state at time $t\in\integer_{\geq0}$ and
$u\in\controls$ is a control input selected from the finite set $\controls$.
For aerial robots as studied here $x_t\in\SEthree$.
The observation, $y_t$, is a function of both the state and the environment and
is corrupted by noise, $\nu$.
We use capital letters to refer to random variables and lowercase
for realizations so $M$ and $Y_t$ represent random variables
associated with the environment and an observation, respectively.

\subsection{Occupancy grids and ranging measurements}
The environment itself is unknown and can be with the random variable $M$
which is modeled as an occupancy grid \citep{elfes1989} and with an associated
approximations of mutual information for ranging
sensors~\citep{charrow2015icra,julian2014,zhang2019icra}.
This occupancy grid is in turn discretized into cells,
$M=\{C_1,\ldots,C_{\numcells}\}$, that are either occupied or free with some
probability.
Cells occupancy is independent such that the probability of a realization $m$ is
$\P(M\!=\!m)=\prod_{i=1}^{\numcells}\P(C_i\!=\!c_i)$ where $c_i\in\{0, 1\}$.
The conditional probability of $M$ given previous states and observations is
then written
\begin{align}
  \P(M\!=\!m|x_{1:T},y_{1:T})
  &= \frac{\prod_{t=1}^T\P(y_{t}|M\!=\!m,x_{t})\P(M\!=\!m)}
  {
    \sum_{m'\in\mathcal{M}}\prod_{t=1}^T
    \P(y_{t}|M\!=\!m',x_{t})\P(M\!=\!m')
  }
\end{align}
where $\mathcal{M}\in\{0,1\}^{\numcells}$ is the set of possible environments.
As representing an unconstrained joint distribution between cells is
intractable, the conditional probabilities of the cells given previous
observations are also treated as being independent with probability
$p_{i,t}$
such that the conditional probability is
\begin{align}
\P(M\!=\!m|x_{1:T},y_{1:T})
= \prod_{i=1}^{\numcells}p_{i,t}
\end{align}
We denote the collection of probabilities as the belief,
$b_t=\bigcup_{i=1}^{\numcells} p_{i,t}$.

The robots considered in this work are equipped with depth cameras, a form of
ranging sensor.
Each sensor observation consists of a collection of range observations which
each provide the distance along a ray from the sensor origin to the
nearest occupied cell in the environment, subject to additive Gaussian
noise.
For more detail, we refer the reader to work by~\citet{charrow2015icra}.

\subsection{Problem description and objective}
For one robot and one time-step the optimal control input
in terms of entropy reduction is
\begin{align}
  u_1^\star &= \argmax_{u\in\controls}~\MI(Y_{t+1};M|b_t,x_t).
\end{align}
subject to the dynamics~\eqref{eq:dynamics} and observation
model~\eqref{eq:observation}.
Consider an $l$-step horizon.
The problem becomes a belief-dependent, partially observable Markov
decision process (POMDP) as is discussed in more detail by~\citet{lauri2015ras}
and is an optimization problem over policies.
We instead optimize over a fixed series of actions
(see Sec.~\ref{sec:receding_horizon_planning})
which results in a simpler problem.
To simplify notation, let $\mathcal{Y}_i$ indicate the space of possible
observations available to robot $i$ over the finite horizon induced by the
finite set of control inputs, the dynamics~\eqref{eq:dynamics}, and the
observation model~\eqref{eq:observation}.
The optimal multi-robot, finite-horizon informative plan
is then
\begin{align}
  Y_{t+1:t+l,1:\numrobot}^\star &=
  \argmax_{Y_{1:l,1:\numrobot}\in
  \mathcal{Y}_{1:\numrobot}}~\MI(Y_{1:l,1:\numrobot};M|b_{t}, x_{t,1:\numrobot})
  \label{eq:objective}
\end{align}
where the indexing $x_{1:t,1:\numrobot}$ represents values at times $1$ through
$t$ and for robots $1$ through $\numrobot$.
In the following sections, we will drop the time and robot index as well as
robot states and belief when appropriate.

Solutions to~\eqref{eq:objective} will be constructed incrementally using greedy
algorithms.
Using the definitions in Sec.~\ref{sec:independence},
the constraints in~\eqref{eq:objective} can be interpreted as simple partition
matroid where
$\ground = \bigcup_{i=1}^{\numrobot} \mathcal{Y}_i$
and
$\independence
= \{Y\subset\ground\mid |Y\cap \mathcal{Y}_i|\leq 1,~\forall i\in\robots\}$.

\subsection{Assumptions}
We make the following assumptions regarding the exploration scenario:
\emph{1) all robots have the same belief state, operate synchronously, and
communicate via a fully connected network};
\emph{2) incremental motions via $\dynamics$ are bounded}; and
\emph{3) the sensor range is bounded}.
The first assumption simplifies analysis in the context of this work.
Here we emphasize scenarios where large numbers of robots operate in close
proximity leading to redundant observations.
Extending the proposed algorithm to incorporate additional considerations such
as communication constraints is left to future work.
The second and third assumptions
ensure that the mutual information between observations made by
distant robots is zero.
These assumptions simplify the problem structure and are the key reason
that the proposed efficient algorithm comes with little to no reduction in
solution quality.

\section{Single-robot planning}
\label{section:single_robot}
We employ Monte-Carlo tree search~\citep{chaslot2010,browne2012} for the
single-robot planner as previously proposed for active perception and
exploration~\citep{lauri2015ras,best2019ijrr,patten2017} and
in multi-robot active perception \citep{best2019ijrr}.

In order to ensure bounded and similarly scaled rewards,
constant terms from~\eqref{eq:objective} are dropped when planning for the
$i^\mathrm{th}$ robot to obtain a local objective
\begin{align}
  \MI(Y_{t+1:t+l,i};M|Y_{t+1:t+l,A}),
  \label{eq:single_robot_objective}
\end{align}
the mutual information between $Y_{t+1:t+l,i}$ and the map
conditional on observations $Y_{t+1:t+l,A}$ obtained by some set of robots $A$
such that $A\cap \{i\}=\emptyset$.

Denote solutions obtained from the Monte-Carlo tree search single-robot
planner maximizing~\eqref{eq:single_robot_objective} as
\begin{align}
\hat Y_i=\mathrm{SingleRobot}(i,Y_{A})
\label{eq:single_robot_planner}
\end{align}
and assume this planner has suboptimality
$\eta\geq 1$, such that
\begin{align}
  \eta \MI(M;\hat Y_i|Y_{A}) &\geq
  \max_{Y\in\mathcal{Y}_i} \MI(M;Y| Y_{A})
  \label{eq:approximation}
\end{align}
as in to the approach of~\citet{singh2009}.

Although Monte-Carlo tree search does not come with a suboptimality guarantee,
some existing algorithms for informative path
planning~\citep{chekuri2005,singh2009} do.
Monte-Carlo tree search is used instead on account of limited computation time
and later for ease in incorporation of inter-robot collision constraints which
are time-varying.

\section{Multi-robot planning}
\label{section:multi_robot}
The main contribution of this work is the design and analysis of a new
distributed multi-robot planner that extends the single-robot planner discussed
in Sec.~\ref{section:single_robot} or any planner
satisfying~\eqref{eq:approximation} to multi-robot exploration.
In development of a distributed algorithm,
we first present a commonly used algorithm, \greedy{}, which provides
suboptimality guarantees \citep{singh2009} but requires robots to plan
sequentially.
We then propose a similar distributed algorithm, \dgreedy{}, and analyze its
performance in terms of time and suboptimality.

\subsection{Sequential greedy assignment}
Consider an algorithm that plans for each robot in the team by
maximizing~\eqref{eq:single_robot_objective} given all previously assigned
plans and continues in this manner to sequentially assign plans to each robot.
We will refer to this as sequential greedy assignment (\greedy{}).
\citet{singh2009} use the properties of mutual information
discussed in Sec.~\ref{sec:submodularity} to establish that \greedy{}
obtains an objective value within $1+\eta$ of the
optimal solution.
The greedy solution using an optimal single-robot planner
can be defined inductively as $Y^\mathrm{g}=Y^\mathrm{g}_{0:\numrobot}$ using a
suboptimal planner as in~\eqref{eq:single_robot_planner}
to obtain the solution $Y^\mathrm{g}=Y^\mathrm{g}_{0:\numrobot}$ such that
\begin{align}
  \begin{split}
    Y^\mathrm{g}_{0} &= \emptyset \\
    Y^\mathrm{g}_{i} &= \mathrm{SingleRobot}(i, Y_{1:i-1}^\mathrm{g})
    \label{eq:suboptimal_greedy}
  \end{split}
\end{align}
This algorithm satisfies the following suboptimality bound.
\begin{theorem}[Suboptimality bound of sequential assignment \citep{singh2009}]
  \label{theorem:greedy_assignment}
  \greedy{} obtains a suboptimality bound of
  \begin{align}
    \MI(M;Y^\star) \leq& (1+\eta) \MI(M;Y^\mathrm{g})
  \end{align}
\end{theorem}

This multi-robot planner is formulated as an extension of a generic single-robot
planner and depends only on the suboptimality of the single-robot planner.
As robots plan sequentially, this leads to large computation times as the number
of robots grows.

\subsection{Distributed sequential greedy assignment}

\begin{algorithm}[t]
  \caption[Distributed sequential greedy assignment (\dgreedy{})]{%
    Distributed sequential greedy assignment (\dgreedy{}) from the
  perspective of robot $i$}\label{alg:distributed_greedy}
  \begin{minipage}{\linewidth}
    \begin{algorithmic}[1]
      \State $\numrounds \gets \text{number of planning rounds}$
      \State $\numrobot \gets \text{number of robots}$
      \State $Y_F \gets \emptyset$ \Comment{set of fixed trajectories}
      \For{$1,\ldots,\numrounds$} \label{line:rounds}
      \State $Y_{i|Y_F} \gets \mathrm{SingleRobot}(i,Y_F)$ \label{line:single-robot}
        \State $I_{i,0} \gets \MI(M;Y_{i|Y_F}|Y_F)$ \Comment{planner reward}
        \label{line:initial_objective}
        \State $I_{i,F} \gets I_{i,0}$ \Comment{updated reward}
        \label{line:updated_objective}
        \For{$k=1,\ldots, \lceil \frac{\numrobot}{\numrounds} \rceil$}
        \label{line:selection}
        \State $j \gets \argmin\limits_{j\in 1:\numrobot} I_{j,0} - I_{j,F}$
        \Comment{computed by distributed reduction across robots}
        \label{line:reduction}
          \If{$i=j$}
            \State Transmit: $Y_{i|Y_F}$
            \label{line:transmit}
            \State \Return $Y_{i|Y_F}$
          \Else
          \State Receive: $Y_{j|Y_F}$
          \State $Y_F \gets Y_F \cup Y_{j|Y_F}$
          \label{line:fixed_plans}
          \State $I_{i,F} \gets \MI(M;Y_{i|Y_F}|Y_F)$ \Comment{update reward}
          \label{line:update}
          \EndIf
        \EndFor
      \EndFor
    \end{algorithmic}
  \end{minipage}
\end{algorithm}

Consider a scenario with spatially distributed robots such that the
mutual information between any observations obtainable within a
finite horizon by any pair of robots is zero.
The union of solutions obtained for individuals independently is then equivalent
to a solution to the combinatorial problem over all robots, $Y^\star$.
A weaker version of this idea applies such that if the plans returned for a
subset of robots are conditionally independent, those plans are optimal over
that subset of robots regardless of the inter-robot distances.
The distributed planner, \dgreedy{}, is designed according to this principle and
allows all robots to plan at once and then selects a subset of those plans
while minimizing suboptimality.

\dgreedy{} is defined in Alg.~\ref{alg:distributed_greedy} from the
perspective of robot $i$.
Planning proceeds in a fixed number of rounds, $\numrounds$ (line~\ref{line:rounds}).
Each round begins with a planning phase where
each robot plans for itself given the set of plans that are
assigned in previous rounds (line~\ref{line:single-robot}), stores the initial
objective value, $I_{i,0}$ (line~\ref{line:initial_objective}), and copies this to a
variable, $I_{i,F}$ (line~\ref{line:updated_objective}) that represents the
updated value as more plans are assigned.
The round ends with a selection phase (line~\ref{line:selection})
during which a subset of $\lceil \frac{\numrobot}{\numrounds} \rceil$ plans are
assigned to robots.
The plans are assigned greedily to minimize the decrease in the objective
values, $I_{j,F}-I_{j,0}$, and the robot whose plan is to be assigned is
selected using a distributed reduction across the multi-robot team
(line~\ref{line:reduction})~\citep{ladner1980}, and ties are broken arbitrarily.
The chosen robot sends its plan to the other robots (line~\ref{line:transmit}),
and these robots store this plan (line~\ref{line:fixed_plans}) and update
their objective values (line~\ref{line:update}).

Denote a planner with $\numrounds$ planning rounds as \dgreedy[\numrounds].
Let $D_i$ be the set of robots whose trajectories are assigned during the
$i^\mathrm{th}$ distributed planning round and let $F_i=\bigcup_{j=1}^i D_j$ be
the set of all robots with trajectories assigned by that round.
Denote solutions to this new distributed algorithm as $Y^\mathrm{d}$.
Then, let $\hat Y_{r|Y^\mathrm{d}_{F_i}}$ represent the approximate solution returned by
the single-robot planner given previously assigned trajectories.
The result of \dgreedy{} can then be written as
$Y^\mathrm{d}_{D_{i,j}}=\hat Y_{D_{i,j}|Y^\mathrm{d}_{F_{i-1}}}$ where $D_{i,j}$ is the
$j^\mathrm{th}$ robot assigned during round $i$.
\dgreedy{} achieves a bound related to
Theorem~\ref{theorem:greedy_assignment} with an additive term based on the
decrease in objective values from initial planning to assignment that
\dgreedy{} seeks to minimize
(Alg.~\ref{alg:distributed_greedy}, line~\ref{line:reduction}).

\begin{theorem}
  \label{theorem:distributed_suboptimality}
  The excess suboptimality of the distributed algorithm
  (Alg.~\ref{alg:distributed_greedy}) compared to the bound for greedy
  sequential assignment is given by the sum of conditional mutual informations
  between each selected observation and prior selections for the same
  planning round\footnote{%
    Although we address multi-robot exploration, this result applies
    generally to informative planning problems and general monotone submodular
    maximization with partition matroid constraints (aside from notation and
    problem specialization).
  }%
  \begin{align}
    \begin{multlined}
      \MI(M;Y^\star)
      \leq (1+\eta)\MI(M;Y^\mathrm{d}) + \eta\psi
    \end{multlined}
  \end{align}
  where $\psi=\sum_{i=1}^{\numrounds} \sum_{j=1}^{|D_i|}
    \MI(Y^\mathrm{d}_{D_{i,j}};Y^\mathrm{d}_{D_{i,1:j-1}}|Y^\mathrm{d}_{F_{i-1}})$
  represents excess suboptimality. The proof is provided in
  Appendix~\ref{appendix:distributed_proof}.
\end{theorem}

This is an online bound in the sense that it is parametrized by the planner
solution.
However, as will be shown in the results, $\psi$ tends to be small in practice
indicating that \dgreedy{} produces results comparable to \greedy{}.
In this sense, small values of $\psi$ serve to certify the greedy bound of
$1+\eta$ empirically without needing to obtain the objective value returned by
\greedy{} explicitly.

\subsection{Worst-case suboptimality}
The main focus of this work is to investigate how aspects of problem structure
can be used to achieve efficient planning with minimal impact on suboptimality.
This contrasts starkly to the worst-case suboptimality which is inversely
proportional to the number of robots.
While worst-case results by \citet{gharesifard2017} and
\citet{mirzasoleiman2013nips} are relevant, \citet{grimsman2018tcns} recently
proved a somewhat tighter bound that is readily applicable to
Alg.~\ref{alg:distributed_greedy}.

\begin{theorem}[Worst-case suboptimality]
  \label{theorem:worst-case}
  The following worst-case bound for Alg.~\ref{alg:distributed_greedy} holds:
  \begin{align}
    \MI(M;Y^\star) \leq (1+\eta \lceil \numrobot/\numrounds \rceil) \MI(M;Y^\mathrm{d}).
    \label{eq:worst_case}
  \end{align}
  The proof can be found in Appendix~\ref{appendix:worst_case_proof}.
\end{theorem}

According to this bound, the planner may entirely fail to take advantage of
additional robots if additional planning rounds are not introduced.
However, this does not account for locality which can lead distant robots
to become decoupled in the solution.
This bound can then be interpreted as a limiting scenario for when locality does
not hold such as due to extremely close proximity or complex interactions
between observations and the environment model.

\subsection{Subset selection strategies}
The subset selection step of \dgreedy{} itself uses a greedy strategy.
Looking at this more directly,
the negation of the contribution of a single round is
\begin{align}
  \MI(M;Y^\mathrm{d}_{D_{i}}|Y^\mathrm{d}_{F_{i-1}})
  -\sum_{j=1}^{|D_i|}
  \MI(M;Y^\mathrm{d}_{D_{i,j}}|Y^\mathrm{d}_{F_{i-1}}),
  \label{eq:subset}
\end{align}
found by application of the chain rule of mutual information
(see Sec.~\ref{sec:information_theory})
to~\eqref{eq:planner_difference}.
Equation~\eqref{eq:subset} is submodular and monotonically decreasing unlike the
monotonically increasing objectives considered previously.
While we apply a heuristic approach and evaluate results empirically,
other works have evaluated this setting more directly~\citep{gharan}.

More generally, approaches that seek to minimize~\eqref{eq:subset} over the
course of individual rounds may fail for sufficiently unbalanced problems.
For example, a large number of robots with zero contribution could cause
all robots with non-zero objective values to be selected at once at
significant detriment to solution quality.
While such scenarios are not encountered in this work, this is an important
direction for future work.

\subsection{Algorithm run time analysis}
We compare the run time of
\dgreedy{} to \greedy{} for variable numbers of robots
with run time defined as the time elapsed from when the first robot
begins computation until the last robot is finished.
We assume point-to-point communication over a fully connected network requiring
a fixed amount of time per message.
The messages have fixed sizes and correspond either to a finite-horizon plan
or a difference in mutual information.
Given these assumptions, broadcast and reduction steps each require $O(\log
\numrobot)$ time~\citep{ladner1980}.
Let $\alpha(\numrobot, b_t)$ bound the run time of the selected single-robot
planner, noting the dependence on the number of robots and the environment.
Similarly, let $\beta(\numrobot, b_t)$ bound cost of the mutual information
computation in Alg.~\ref{alg:distributed_greedy}, line~\ref{line:update}.

\greedy{} consists of $\numrobot$ planning steps, each with one broadcast step.
The computation time of sequential greedy assignment is then
\begin{align}
  \greedy\!:\quad O(\numrobot\alpha(\numrobot, b_t) + \numrobot \log \numrobot).
  \label{eq:greedy_runtime}
\end{align}

Each of the $\numrounds$ rounds in \dgreedy{} begins with a single planning phase,
coming to a cost of $O(\numrounds\alpha(\numrobot, b_t))$.
The rest of the algorithm consists of subset selection, broadcast of the
chosen plans, and computation of mutual information which
cumulatively occur once per robot for a total cost of
\begin{align}
  \dgreedy[\numrounds]\!:\quad
  O(\numrounds\alpha(\numrobot, b_t) + \numrobot\beta(\numrobot, b_t) + \numrobot\log \numrobot).
  \label{eq:distributed_runtime}
\end{align}
If $\numrounds$ grows slowly or is constant, the performance of \dgreedy{}
relative to \greedy{} depends on the relative cost of the mutual information
computation ($\beta$).

For the approximation developed by \citet{charrow2015icra}, evaluation of mutual
information is linear in the number of cells of the map that are observed.
Given the assumption of bounded sensor range, evaluation of mutual
information scales linearly in the number of robots so that
$\beta(\numrobot,b_t)\in O(\numrobot)$.
When planning time is dominated by a constant number of mutual information
computation evaluations, such as when using a Monte-Carlo tree search planner
with a fixed number of sample trajectories, then $\alpha(\numrobot,b_t)\in O(\numrobot)$ so
that both algorithms are quadratic.
However, in practice, $\alpha$ includes a large constant factor leading to
a significant speedup for \dgreedy[\numrounds]{} compared to \greedy{}.
In general, $\alpha$ may also depend non-trivially on factors such as the length
of the plan or the scale or complexity of the environment.
This further emphasizes the significance of the of eliminating the $\numrobot$
coefficient from the single-robot planning step.

\section{Persistent safe exploration given vehicle dynamics}
\label{sec:safety_and_independence}
Up to this point, the analysis has considered maximization of the
mutual information objective over a joint space of finite-horizon trajectories
which has the form of a partition matroid constraint.
However, inter-robot collision constraints cannot be modeled using a matroid,
and further challenges arise if the planner can fail to provide a solution or if
no solutions exist.

\subsection{Independence systems, collision constraints, and suboptimality}
\label{sec:collisions_as_independence}
After including inter-robot collision constraints, the space of feasible joint
plans no longer has the structure of a partition matroid or even a general
matroid.
Consider, a trajectory that is free of collisions with one assignment of
trajectories to a subset of robots.
This trajectory is not necessarily free of collisions with
another assignment of trajectories to the same subset of robots and neither can a
robot in the subset be assigned an additional trajectory from the first
assignment.
This violates the exchange property for matroids (Def.~\ref{def:matroid}).
Instead the constraints can be modeled using an independence system
(Def.~\ref{def:independence_system}), given that subsets of collision-free
trajectories are also collision-free.
In general, greedy algorithms perform arbitrarily poorly for general
independence systems~\citep{fisher1978}.
However, we can recover a lower bound of $1/(\eta \numrobot)$ by maximizing mutual
information using the first selection in each planning round which is otherwise
under-determined because the differences in Alg.~\ref{alg:distributed_greedy},
line~\ref{line:reduction} are uniformly zero for the first plan selected in each
round.
The bound follows by observing that the maximal feasible subsets have
cardinality of at most $\numrobot$ and by applying suboptimality of the single-robot
planner.
This result will be used later during discussion of system liveness in
Sec.~\ref{sec:liveness}.

Although suboptimality bounds change with the constraint structure,
this is not necessarily indicative of typical performance in exploration.
Mutual information objectives as used in exploration often favor configurations
with large inter-robot distances that minimize redundancy in sensor
observations.
Robots are also often small compared to their sensor footprints
which further mitigates detrimental effects of collision constraints.


\subsection{Sufficient requirements for safety}
\label{sec:safety}
Safety is interpreted as meaning that robots do not enter any occupied part
of the environment and do not collide with each other, and each condition is
implemented using thresholds on distances as discussed in more detail in
Sec.~\ref{sec:results}.
Safety is guaranteed by enforcing the invariant that the current joint plan is
safe and terminates in a controlled invariant state%
\footnote{%
  In the implementation, robots are required to come to a stop and
  trajectories are then temporally extended as necessary.
  Although these conditions can be relaxed, such as to invariant sets,
  there do not appear to be clear or significant benefits from doing so.
}
as is common in model predictive control~\citep{rawlings1993tac}.
Plans are executed in a receding-horizon fashion so that robots do not
necessarily enter the planned invariant states whereas in the case of planner
failure, one or more robots may enter invariant states and remain in those
states until a feasible plan can be found.

Safety will be guaranteed under the following assumptions:
\emph{
  1) the initial state, at the start of the exploration process, is an invariant
  state%
}; and
\emph{
  2) states that are believed not to be in collision with the environment are
  free of collisions with the environment and remain so for all time%
}.
The first assumption ensures safety by induction.
The second prevents arbitrary changes in the conditions for safety.
This latter assumption does not always hold in practice.
Approaches, such as selectively relaxing constraints, can address this if
necessary.

Assume that the team of robots is tracking a plan that is
safe for all time (i.e. meets the stated requirements for all time).
We modify Alg.~\ref{alg:distributed_greedy} to only commit to a
single-robot plan if the resulting joint plan that results from swapping the old
single-robot plan with the new one is also safe for all time.
The following constraints are also implemented in the single-robot planner:
\emph{
  1) plans terminate in an invariant state%
}; and
\emph{
  2) plans respect collision constraints with other robots and the environment
  according to the full and current joint plan at planning time%
}.
The joint plan which results from replacing a robot's plan with the output
of the single-robot planner may not be safe either due to failure to meet the
above constraints or due to updates to other robots' plans.
In this case, the system falls back to the robot's prior plan which is known to
be safe.
As a result, joint plan is always safe for all time.

\subsection{Liveness in multi-robot exploration}
\label{sec:liveness}

Informally, liveness properties refer to guarantees that a system will make
progress toward a goal.
In exploration, a natural statement of liveness for some system state
is that the multi-robot team will eventually select an action that reduces the
entropy of the map.
Although liveness is not guaranteed in general, the proposed system design does
admit a limited liveness guarantee.
This prevents scenarios such as when a small number of conflicting robots can
create a persistent state of deadlock in the entire system.

\emph{Assume that some robot finds a feasible single-robot plan
with a mutual information reward of at least $\epsilon$.}
Then, using the results and modifications described in
Sec.~\ref{sec:collisions_as_independence},
the resulting joint plan also provides a mutual information reward of at least
$\epsilon$ which in turn corresponds to $\epsilon$ entropy reduction in
expectation.
This condition guarantees liveness in exploration with high probability
(i.e. barring infinite sequences of disinformative observations) so long as some
robot can find an action that reduces the entropy of the map.

\section{Results and discussion}
\label{sec:results}
The results are divided into two main parts.
Section~\ref{sec:partition_experiments} does not incorporate inter-robot
collisions and addresses scalability and performance of
Alg.~\ref{alg:distributed_greedy}
as well as the suboptimality in terms of
Theorem~\ref{theorem:distributed_suboptimality} in the intended context of
submodular maximization over a partition matroid.
Section~\ref{sec:collision_experiments} introduces inter-robot collisions
and the application to a team of aerial vehicles both in simulation and on real
hardware.
These experiments incorporate the approach described in
Sec.~\ref{sec:safety_and_independence} in order to address additional challenges
related to dynamics and collision prevention.

\subsection{Exploration with large numbers of kinematic quadrotors}
\label{sec:partition_experiments}
The proposed approach is first evaluated in a simplified scenario with a team
of kinematic quadrotors, disregarding dynamics and inter-robot collisions, so
the discussion in Sec.~\ref{sec:safety_and_independence} does not come
into play.
Performance is evaluated in terms of planner objective values, computation time,
and entropy reduction with respect to the map.

\begin{remark}
  Chapter~\ref{chapter:time_sensitive_sensing} revisits the results from this
  chapter.
  While developing that work we found that relative transformations for control
  actions were being applied a second time before computing the information
  gain which affects the results in this chapter for the kinematic model
  (but not results for the dynamic model or with physical robots).
  Applying transforms twice has little effect for translations
  (as the field of view does not change much)
  but significant effect for rotations.
  Although we include the original results,
  note that fixing this error improved completion time
  (Sec.~\ref{sec:completion_and_rates}) by about 18\% for that version of the
  exploration system and eliminated some aberrant behavior.
\end{remark}

\subsubsection{Exploration scenario}
We test the exploration methodology in a confined and cluttered environment
with obstacles (cubes) of various sizes and with robot positions initialized
randomly near a lower edge as depicted in Fig.~\ref{fig:intro_image}.
The environment is bounded by a
$6\,\si\metre\times6\,\si\metre\times6\,\si\metre$ cube, and the robots model
this environment using a 3D occupancy grid with a $0.1\,\si\metre$ resolution.
The confines and clutter ensure that robots remain proximal to each other.
This leads to redundant observations and potential for suboptimal joint
plans over the planning horizon.
The experiments compare \dgreedy[1] through \dgreedy[3]
(noting that \dgreedy[\numrounds] represents Alg.~\ref{alg:distributed_greedy} with
$\numrounds$ planning rounds)
to \greedy{} for 4, 8, 16, and 32 robots over 3000 iterations (time-steps).
Tests evaluating exploration performance are each run twenty times each with
randomized initializations.
Experiments that evaluate computation time use specially-instrumented planner
and single trial for each configuration.

\subsubsection{Implementation details}
\label{sec:kinematic_implementation}
We evaluate the proposed algorithm in simulation and run
experiments on a computer equipped with an Intel Xeon E5-2450 CPU (2.5~GHz).
The planner and other components of the system are implemented using C++ and
ROS~\citep{quigley2009icra}.
When timing the distributed planner, planning steps that would normally be
performed in parallel (single-robot planning during each planning round
and information propagation) are executed serially.
The duration of each such step is taken as the maximum duration of the serially
computed steps in order to mimic distributed computation.
In practice, computing the reduction to find the minimum excess term
(Alg.~\ref{alg:distributed_greedy}, line~\ref{line:reduction})
requires an insignificant amount of time.
So, although the analysis assumes a logarithmic-time parallel, we
compute this by iteration over all elements in the implementation.

The simulated robots emulate kinematic quadrotors
moving in a three-dimensional environment with
planning, mapping, and mutual information computation each performed in 3D.
The single-robot Monte-Carlo tree search planner is run for a fixed number of
samples (200) using a discrete set of actions consisting of the choice of
translations of $\pm0.3\,\si\metre$ in the $x$--$y$--$z$ directions or heading
changes of $\pm\pi/2\,\si\radian$.
Each robot is equipped with a simulated time-of-flight camera with a range of
$2.4\,\si\metre$ which is similar to typical depth cameras and having a
$19\times12$ resolution and a $43.6\si\degree\times34.6\si\degree$ field of view
which is oriented with the long axis aligned vertically to promote effective
sweeping motions.
For efficient evaluation of the mutual information objective, we substitute the
approximation of Cauchy-Schwarz mutual information (CSQMI) described by
\citet{charrow2015phd}
(which is not necessarily submodular)
for the Shannon mutual information~\eqref{eq:mutual_information} and downsample
rays by two in each direction.
Rather than the uniform prior for cell occupancy ($50\%$) which is commonly used
in mapping, we introduce a prior of $12.5\%$ probability of occupancy during
evaluation of mutual information~\citep{tabib2016iros}.
This encourages selection of actions that observe larger volumes of unknown
space.
This set of experiments is not run in real-time although the next is.
The planner maintains the robot states internally and triggers the camera after
each iteration.

\begin{table}
  \caption[Exploration performance results for the kinematic exploration scenario]
  {%
    Exploration performance results for the kinematic exploration scenario:
    The reduction rate is computed with respect to the map
    over the first 250 robot-iterations and is representative of nominal
    performance.
    The total entropy reduction shows the entropy reduction with respect to the
    map at the end of the experimental trial (3000 robot-iterations).
  }\label{tab:exploration_performance}
  \resizebox{\linewidth}{!}{%
    \begin{tabular}{lr|ll|ll|ll|ll}
      Alg. &
      $\numrobot$ &
      \multicolumn{2}{c|}{Objective $\left(\frac{\text{bits}}{\text{robot}}\right)$} &
      \multicolumn{2}{c|}{$\psi$ $\left(\frac{\text{bits}}{\text{robot}}\right)$} &
      \multicolumn{2}{c|}{Entropy red. rate
      $\left(\frac{\text{bits}}{\text{robot}\cdot\text{iter.}}\right)$} &
      \multicolumn{2}{c}{Total entropy red.
      $\left(\text{bits}\right)$} \\
      &
      & avg.&std. dev.%
      & avg.&std. dev.%
      & avg.&std. dev.%
      & avg.&std. dev.
      \\\hline
      \begin{filecontents*}{\datapath/kinematic_stats.csv}
algorithmnames, colnumrobots, rewardaverage, rewarddeviation, excessaverage, excessdeviation, entropyreductionrateaverage, entropyreductionratestandarddeviation, entropyreductionaverage, entropyreductionstandarddeviation
$\dgreedy[1]$,  4,    2.84e+01,    1.29e+01,    2.76e+00,    3.90e+00,    2.28e+02,    4.83e+01,    1.54e+05,    1.21e+04
$\dgreedy[2]$,  4,    2.99e+01,    1.30e+01,    3.39e-01,    8.50e-01,    2.36e+02,    2.96e+01,    1.61e+05,    2.04e+03
$\dgreedy[3]$,  4,    2.99e+01,    1.34e+01,    3.15e-01,    8.75e-01,    2.32e+02,    2.31e+01,    1.61e+05,    2.41e+03
    $\greedy$,  4,    3.06e+01,    1.34e+01,    0.00e+00,    0.00e+00,    2.30e+02,    3.68e+01,    1.60e+05,    2.85e+03
$\dgreedy[1]$,  8,    2.63e+01,    1.04e+01,    6.49e+00,    5.80e+00,    2.07e+02,    1.50e+01,    1.60e+05,    2.23e+03
$\dgreedy[2]$,  8,    2.82e+01,    1.08e+01,    1.23e+00,    1.66e+00,    2.10e+02,    1.78e+01,    1.63e+05,    1.89e+03
$\dgreedy[3]$,  8,    2.81e+01,    1.10e+01,    2.67e-01,    6.30e-01,    2.04e+02,    1.41e+01,    1.64e+05,    1.75e+03
    $\greedy$,  8,    2.84e+01,    1.06e+01,    0.00e+00,    0.00e+00,    2.05e+02,    1.53e+01,    1.64e+05,    2.80e+03
$\dgreedy[1]$, 16,    2.22e+01,    7.47e+00,    1.24e+01,    8.14e+00,    1.64e+02,    1.11e+01,    1.60e+05,    1.77e+03
$\dgreedy[2]$, 16,    2.46e+01,    8.08e+00,    3.04e+00,    2.58e+00,    1.77e+02,    9.52e+00,    1.66e+05,    1.42e+03
$\dgreedy[3]$, 16,    2.52e+01,    8.17e+00,    9.67e-01,    1.30e+00,    1.78e+02,    8.97e+00,    1.67e+05,    1.67e+03
    $\greedy$, 16,    2.48e+01,    7.86e+00,    0.00e+00,    0.00e+00,    1.74e+02,    7.60e+00,    1.68e+05,    1.65e+03
$\dgreedy[1]$, 32,    1.69e+01,    4.64e+00,    2.00e+01,    1.00e+01,    1.26e+02,    4.60e+00,    1.61e+05,    1.75e+03
$\dgreedy[2]$, 32,    1.94e+01,    4.99e+00,    6.00e+00,    3.59e+00,    1.27e+02,    5.45e+00,    1.67e+05,    1.44e+03
$\dgreedy[3]$, 32,    2.02e+01,    5.30e+00,    2.44e+00,    1.79e+00,    1.29e+02,    7.60e+00,    1.69e+05,    9.94e+02
    $\greedy$, 32,    2.03e+01,    4.63e+00,    0.00e+00,    0.00e+00,    1.31e+02,    8.73e+00,    1.72e+05,    6.51e+02
\end{filecontents*}
\csvreader[head to column names]{\datapath/kinematic_stats.csv}{}
      {
        \algorithmnames & \colnumrobots &
        \rewardaverage & \rewarddeviation &
        \excessaverage & \excessdeviation &
        \entropyreductionrateaverage & \entropyreductionratestandarddeviation &
        \entropyreductionaverage & \entropyreductionstandarddeviation \\
      }
    \end{tabular}
  }
\end{table}

\begin{figure}
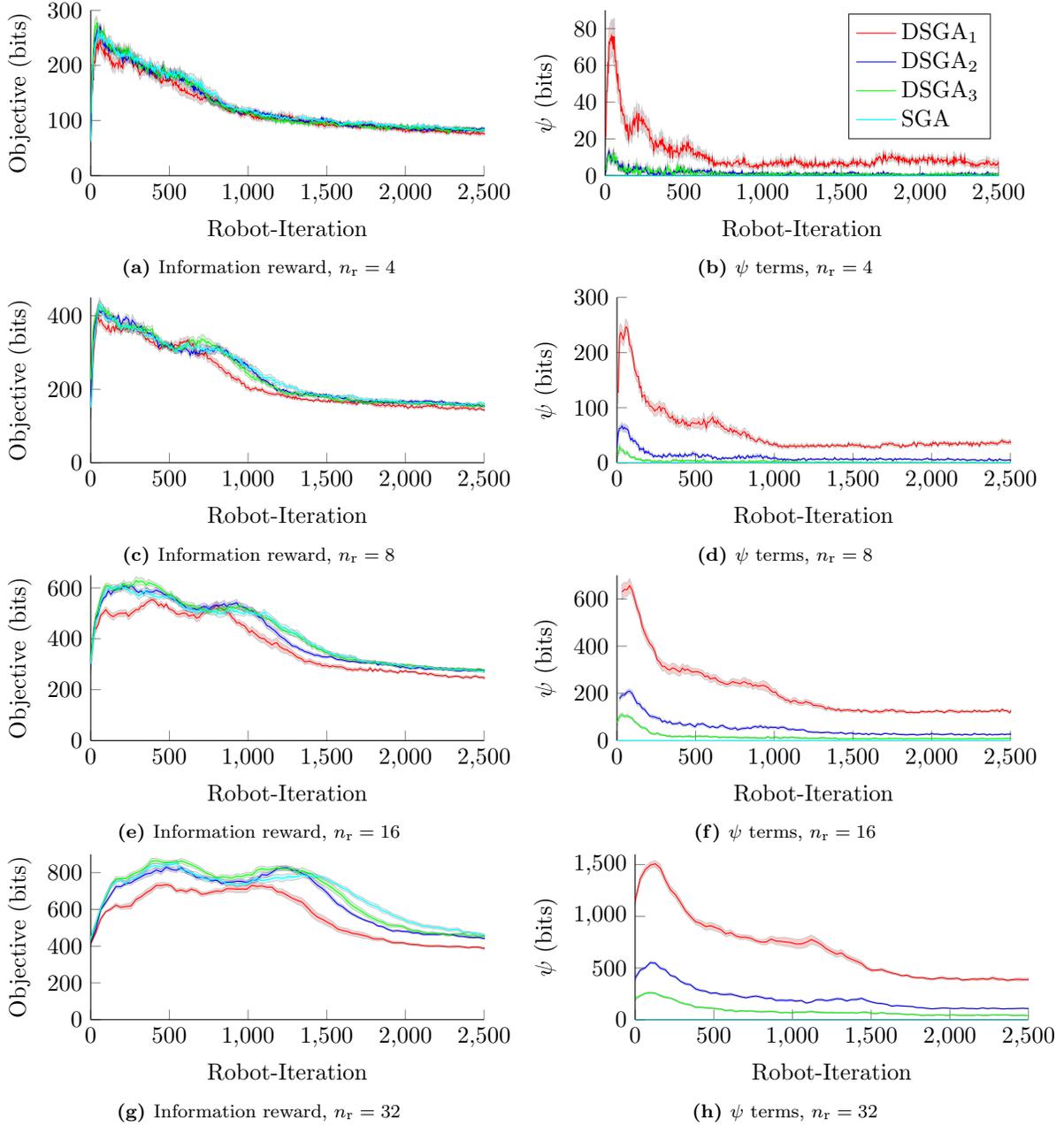

  \begin{center}
    \begin{subfigure}[b]{0.48\linewidth}
      \setlength{\figurewidth}{0.8\linewidth}
      \setlength{\figureheight}{0.40\figurewidth}
      \footnotesize
      \inputfigure[csqmi_4]{\datapath/csqmi_4.tex}
      \caption{Information reward, $\numrobot=4$}\label{subfig:csqmi_4}
    \end{subfigure}
    \begin{subfigure}[b]{0.48\linewidth}
      \footnotesize
      \setlength{\figurewidth}{0.8\linewidth}
      \setlength{\figureheight}{0.40\figurewidth}
      \inputfigure[loss_4]{\datapath/loss_4.tex}
      \caption{$\psi$ terms, $\numrobot=4$}\label{subfig:loss_4}
    \end{subfigure}
    \begin{subfigure}[b]{0.48\linewidth}
      \setlength{\figurewidth}{0.8\linewidth}
      \setlength{\figureheight}{0.40\figurewidth}
      \footnotesize
      \inputfigure[csqmi_8]{\datapath/csqmi_8.tex}
      \caption{Information reward, $\numrobot=8$}\label{subfig:csqmi_8}
    \end{subfigure}
    \begin{subfigure}[b]{0.48\linewidth}
      \footnotesize
      \setlength{\figurewidth}{0.8\linewidth}
      \setlength{\figureheight}{0.40\figurewidth}
      \inputfigure[loss_8]{\datapath/loss_8.tex}
      \caption{$\psi$ terms, $\numrobot=8$}\label{subfig:loss_8}
    \end{subfigure}
    \begin{subfigure}[b]{0.48\linewidth}
      \setlength{\figurewidth}{0.8\linewidth}
      \setlength{\figureheight}{0.40\figurewidth}
      \footnotesize
      \inputfigure[csqmi_16]{\datapath/csqmi_16.tex}
      \caption{Information reward, $\numrobot=16$}\label{subfig:csqmi_16}
    \end{subfigure}
    \begin{subfigure}[b]{0.48\linewidth}
      \footnotesize
      \setlength{\figurewidth}{0.8\linewidth}
      \setlength{\figureheight}{0.40\figurewidth}
      \inputfigure[loss_16]{\datapath/loss_16.tex}
      \caption{$\psi$ terms, $\numrobot=16$}\label{subfig:loss_16}
    \end{subfigure}
    \begin{subfigure}[b]{0.48\linewidth}
      \setlength{\figurewidth}{0.8\linewidth}
      \setlength{\figureheight}{0.40\figurewidth}
      \footnotesize
      \inputfigure[csqmi_32]{\datapath/csqmi_32.tex}
      \caption{Information reward, $\numrobot=32$}\label{subfig:csqmi_32}
    \end{subfigure}
    \begin{subfigure}[b]{0.48\linewidth}
      \footnotesize
      \setlength{\figurewidth}{0.8\linewidth}
      \setlength{\figureheight}{0.40\figurewidth}
      \inputfigure[loss_32]{\datapath/loss_32.tex}
      \caption{$\psi$ terms, $\numrobot=32$}\label{subfig:loss_32}
    \end{subfigure}
  \end{center}
  \caption[Objective and $\psi$ values for varying numbers of robots and
  distributed planning rounds]
  {%
    Objective and $\psi$ values for varying numbers of robots and distributed
    planning rounds:
    (left)
    Mutual information objective values for \dgreedy[2] and \dgreedy[3] closely
    track \greedy{} while the performance of \dgreedy[1]{}
    (which ignores inter-robot interactions) degrades with increasing numbers of
    robots.
    (right)
    Variations in the excess, $\psi$, terms correspond with the objective
    values---although $\psi$-values increase with the numbers of robots,
    they also decrease with increasing numbers of planning rounds and remain
    small compared to the objective for \dgreedy[3]{}.
    Transparent patches indicate standard-error.
  }\label{fig:objective}
\end{figure}

\subsubsection{Evaluation of planner performance and objective values}
\label{sec:kinematic_objective}

Figure~\ref{fig:objective} and Table~\ref{tab:exploration_performance}
show results for exploration experiments comparing \dgreedy[1]{} through
\dgreedy[3]{} to \greedy{}.
The excess ($\psi$) is largest at the beginning of each exploration run as all
robots are start near the same location.
As the robots spread out, all planners approach approximately
steady-state conditions in terms of both excess suboptimality and objective
values before decaying once the environment is mostly explored.
The $\psi$ terms remain relatively large for
\dgreedy[1]{}---which assigns all plans in a single round and does not reason
about conditional dependencies.
However, the $\psi$ terms decrease monotonically with increasing numbers of
planning rounds.
These values remain small for \dgreedy[3]{} and are approximately
one-eighth of the objective value on average for 32 robots whereas values for
\dgreedy[1]{} exceed the value of the objective for the same number of robots.
Decreasing values of $\psi$ are reflected in plots of the mutual information
objective.
Whereas \dgreedy[2]{} and \dgreedy[3]{} closely track \greedy{},
objective values for \dgreedy[1]{} degrade with increasing numbers of robots.

When interpreting these results and entropy reduction results to be shown in
Sec~\ref{sec:kinematic_entropy_reduction}, it is helpful to note that the scale
of the environment remains constant and that increasing the number of robots
serves to increase a notion of the density of the distribution of robots in the
environment.
In this sense, similar performance can be expected of larger teams operating in
larger environments up to constraints on computation and communication.
Toward this end, the next subsection evaluates computation times for
\dgreedy{} and demonstrates that it scales much better than \greedy{} in
practice.


\subsubsection{Computational performance}
\label{sec:time}

\begin{figure}
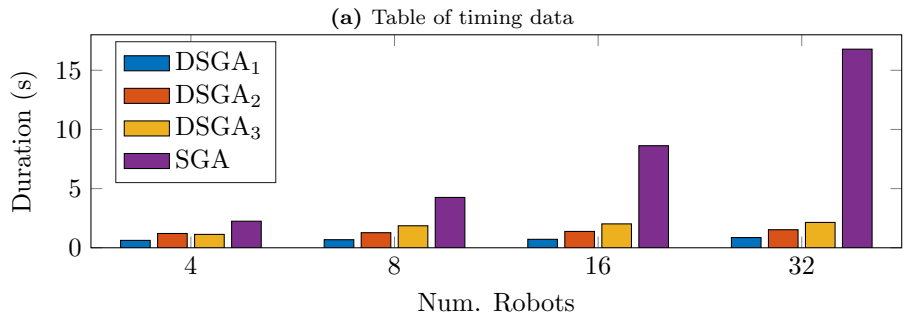

  \begin{subfigure}[b]{\linewidth}
    \centering
      \begin{tabular}{lr|ll|ll|ll}
        Alg. &
        $\numrobot$ &
        \multicolumn{2}{c|}{S.R. Planning} &
        \multicolumn{2}{c|}{Prop.} &
        \multicolumn{2}{c}{Total} \\
        &
        & avg.&std.%
        & avg.&std.%
        & avg.&std.
        \\\hline
        \begin{filecontents*}{\datapath/timing.csv}
$\dgreedy[1]$, 4, 0.600055, 0.0602648, 0.0251158, 0.00529355, 0.625171, 0.0610056
$\dgreedy[2]$, 4, 1.18958, 0.0928361, 0.0180527, 0.00421721, 1.20764, 0.0934141
$\dgreedy[3]$, 4, 1.1124, 0.106173, 0.0189353, 0.00414887, 1.13134, 0.107314
$\greedy$, 4, 2.22711, 0.149984, 0.0184086, 0.0025256, 2.24621, 0.151089
$\dgreedy[1]$, 8, 0.629707, 0.0561645, 0.0524025, 0.00664196, 0.68211, 0.0576995
$\dgreedy[2]$, 8, 1.22491, 0.0863074, 0.049133, 0.00695618, 1.27405, 0.0885664
$\dgreedy[3]$, 8, 1.81644, 0.105251, 0.0428036, 0.00708299, 1.85924, 0.107885
$\greedy$, 8, 4.21768, 0.243191, 0.0341722, 0.00381133, 4.25351, 0.24497
$\dgreedy[1]$, 16, 0.60937, 0.0564739, 0.105684, 0.00916346, 0.715055, 0.0616293
$\dgreedy[2]$, 16, 1.27597, 0.0867042, 0.105994, 0.009542, 1.38197, 0.0911598
$\dgreedy[3]$, 16, 1.91283, 0.091098, 0.106181, 0.00985369, 2.01902, 0.0923332
$\greedy$, 16, 8.55414, 0.329374, 0.0667325, 0.00542727, 8.62516, 0.331594
$\dgreedy[1]$, 32, 0.648619, 0.0564943, 0.216785, 0.0105305, 0.865404, 0.0596426
$\dgreedy[2]$, 32, 1.3039, 0.0547071, 0.221512, 0.0110486, 1.52541, 0.0583348
$\dgreedy[3]$, 32, 1.9237, 0.0949012, 0.218274, 0.0123685, 2.14197, 0.101272
$\greedy$, 32, 16.6402, 0.592438, 0.13057, 0.00815857, 16.7851, 0.596317
\end{filecontents*}
\csvreader[column count=8, no head]{\datapath/timing.csv}{
          1=\algorithmnames, 2=\colnumrobots,
          3=\singlerobotaverage, 4=\singlerobotdeviation,
          5=\propagationaverage,  6=\propagationdeviation,
          7=\totalaverage,  8=\totaldeviation
        }
        {
          \algorithmnames & \colnumrobots &
          \num\singlerobotaverage & \num\singlerobotdeviation &
          \num\propagationaverage & \num{\propagationdeviation} &
          \num\totalaverage & \num\totaldeviation \\
        }
      \end{tabular}
    \caption{Table of timing data}\label{subfig:timing}
  \end{subfigure}
  \begin{subfigure}[b]{\linewidth}
    \begin{center}
      \setlength{\figurewidth}{0.7\linewidth}
      \setlength{\figureheight}{0.25\figurewidth}
      \footnotesize
      \inputfigure[distributed_timing_bar]{\datapath/bar.tex}
    \end{center}
    \caption{Timing for \greedy{} and \dgreedy[3]{}}\label{subfig:bar}
  \end{subfigure}
  \caption[Computational performance in terms of total computation time]
  {%
    Computational performance (seconds) in terms of total computation
    time (time elapsed from when the first robot starts planning until the last
    robot
    stops).
    (\subref{subfig:timing}) Time per
    iteration spent in the single-robot planner, propagation of the information
    reward (\dgreedy{} only), and total computation time.
    (\subref{subfig:bar}) Comparison of the timing differences between \greedy{}
    and \dgreedy{}.
  }\label{fig:exploration_timing}
\end{figure}

While planning performance is largely consistent, \dgreedy{} is able to take
advantage of distributed computation in to provide significantly
improved computation times (Fig.~\ref{fig:exploration_timing}).
Computation times for \greedy{} scale approximately linearly in our simulation
experiments.
The average computation time for \greedy{} increases more than seven times from
$\numrobot=4$ to $\numrobot=32$ robots from $2.25\,\si\second$ to $16.8\,\si\second$
while times for \dgreedy{} remain small and decrease slowly.
As a result, \dgreedy[3]{} provides a 2 to 8 times speedup compared to \greedy.
With 32 robots, the computation time increases by only a factor of 2.5
from \dgreedy[1]{} to \dgreedy[3]{}, despite tripling the number of planning
rounds.
These results indicate that \dgreedy{} is able to provide multi-robot
coordination for large numbers of robots while accommodating requirements for
online performance well after doing the same with \greedy{} becomes intractable.

\subsubsection{Entropy reduction performance}
\label{sec:kinematic_entropy_reduction}

\begin{figure}
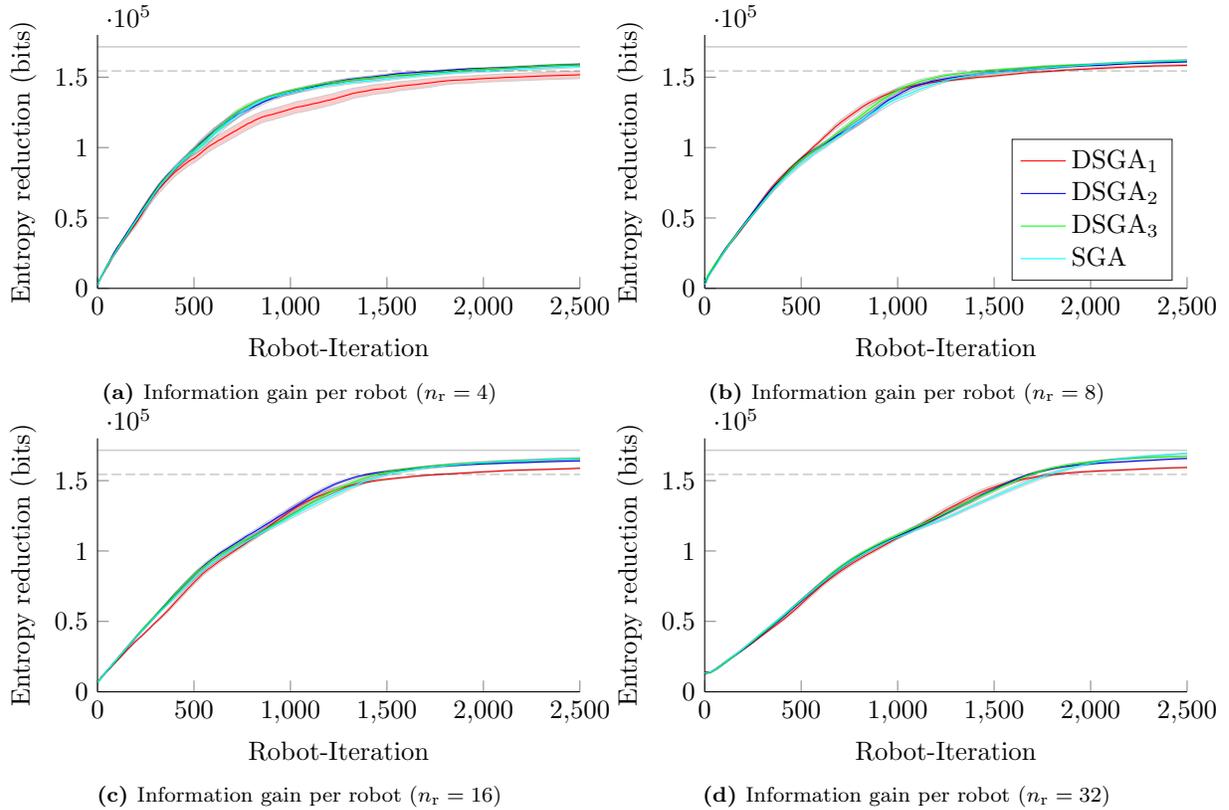

  \begin{center}
    \begin{subfigure}[b]{0.49\linewidth}
      \setlength{\figurewidth}{0.85\linewidth}
      \setlength{\figureheight}{0.50\figurewidth}
      \footnotesize
      \inputfigure[normalized_4]{\datapath/normalized_4.tex}
      \caption{Information gain per robot ($\numrobot=4$)}\label{subfig:normalized_4}
    \end{subfigure}
    \begin{subfigure}[b]{0.49\linewidth}
      \setlength{\figurewidth}{0.85\linewidth}
      \setlength{\figureheight}{0.50\figurewidth}
      \footnotesize
      \inputfigure[normalized_8]{\datapath/normalized_8.tex}
      \caption{Information gain per robot ($\numrobot=8$)}\label{subfig:normalized_8}
    \end{subfigure}
    \begin{subfigure}[b]{0.49\linewidth}
      \setlength{\figurewidth}{0.85\linewidth}
      \setlength{\figureheight}{0.50\figurewidth}
      \footnotesize
      \inputfigure[normalized_16]{\datapath/normalized_16.tex}
      \caption{Information gain per robot ($\numrobot=16$)}\label{subfig:normalized_16}
    \end{subfigure}
    \begin{subfigure}[b]{0.49\linewidth}
      \setlength{\figurewidth}{0.85\linewidth}
      \setlength{\figureheight}{0.50\figurewidth}
      \footnotesize
      \inputfigure[normalized_32]{\datapath/normalized_32.tex}
      \caption{Information gain per robot ($\numrobot=32$)}\label{subfig:normalized_32}
    \end{subfigure}
  \end{center}
  \caption[Entropy reduction performance with different numbers of robots and
  planner configurations]
  {%
    Entropy reduction performance with different numbers of robots and planner
    configurations:
    \dgreedy[2]{} and \dgreedy[3]{} closely track \greedy{} in all cases
    while \dgreedy[1]{}, which does not incorporate inter-robot coordination,
    performs worst for $\numrobot=4$ and has reduced total entropy reduction for
    $\numrobot=16$ and $\numrobot=32$.
    Transparent patches show standard-error.
    Gray lines indicate the maximum mean entropy reduction of
    $1.72\cdot 10^5$ bits
    (data continues through 3000 robot-iterations)
    and a 90\% threshold for task completion
    which facilitates approximate comparison to results in
    Chapter~\ref{chapter:time_sensitive_sensing}.
  }\label{fig:entropy_reduction}
\end{figure}

Figure~\ref{fig:entropy_reduction} compares \dgreedy{} to \greedy{} for various
numbers of robots and is summarized in Tab.~\ref{tab:exploration_performance}.
Exploration rates early in exploration runs remain consistent across planners
and degrade by approximately 43\% as the numbers of robots increase due to
crowding.
This translates to significant improvements in the total rate of exploration
even as efficiency decreases.

The entropy reduction performance of \dgreedy[2]{} and \dgreedy[3]{} closely
matches \greedy{} for each number of robots.
Interestingly, the performance of \dgreedy[1] is worst for $\numrobot=4$ robots where
the rate of entropy reduction degrades after initially performing well.
For larger numbers of robots, differences in entropy reduction performance are
most apparent in the total entropy reduction.
This is likely a result of the planners involving inter-robot coordination
being able to distribute robots across the environment more effectively.
Because we use local finite-horizon planner, \dgreedy[1] then occasionally fails
to allocate robots to and explore some portions of the environment.
Given a spatially global planning strategy, longer exploration
times could then be expected for \dgreedy[1] due to the cost of traveling to
these unexplored regions of the environment.

\subsection{Simulation and experiments of dynamic robots with
inter-robot collision constraints}
\label{sec:collision_experiments}
This section addresses more realistic systems and involves inter-robot collision
avoidance and non-trivial dynamics in the form of teams of real and simulated
quadrotors.
Unlike the previous section, robots now track dynamic trajectories while
planning is performed in real-time.
Simulation results for a team of six robots demonstrate that \dgreedy{}
retains performance comparable to \greedy{}.
Experiments using a three real quadrotors serve to ground the simulation results
by demonstrating comparable results on a similarly configured system in
practice.

\subsubsection{Implementation details}
As these experiments now involve inter-robot collision constraints, we now
incorporate the details Sec.~\ref{sec:safety_and_independence} into both the
\greedy{} and \dgreedy{} planners.
Rather than discrete steps, the Monte-Carlo tree search planner now uses a
library of polynomial motion primitives similar to that described
by~\citet{tabib2016iros} to provide actions.
Because trajectories are executed in real-time, the Monte-Carlo tree search
planner is run in a fully any-time fashion at a rate of $1/3\,\si\hertz$
rather than for a fixed number of samples, and time in single-robot planning is
budgeted according to the number of sequential planning steps in the given
multi-robot planner.
The planner uses a finite $6\,\si\second$ horizon with motion primitives whose
durations vary from  $0.5\,\si\second$ to $10\,\si\second$ ($2\si\second$ is
typical), and at each planning step as many as 162 motion-primitive actions are
available although some may be infeasible due to collision constraints.
Robot-environment collision checks are implemented use a truncated distance
field for efficient lookups, and inter-robot collision checks use horizontal
distances to prevent any robot from flying in another's downwash.
This set of experiments uses a custom and simulation and control framework
that implements a standard quadrotor dynamics model and a quadrotor controller
that tracks polynomial trajectories in the differentially-flat
outputs~\citep{mahony2012} and a team of custom quadrotors equipped with
Structure brand depth sensors.

\subsubsection{Simulation results}
\label{sec:dynamic_simulation}
\begin{figure}
  \begin{center}
    \includegraphics[width=0.3\linewidth, trim={420 100 420 70}, clip]{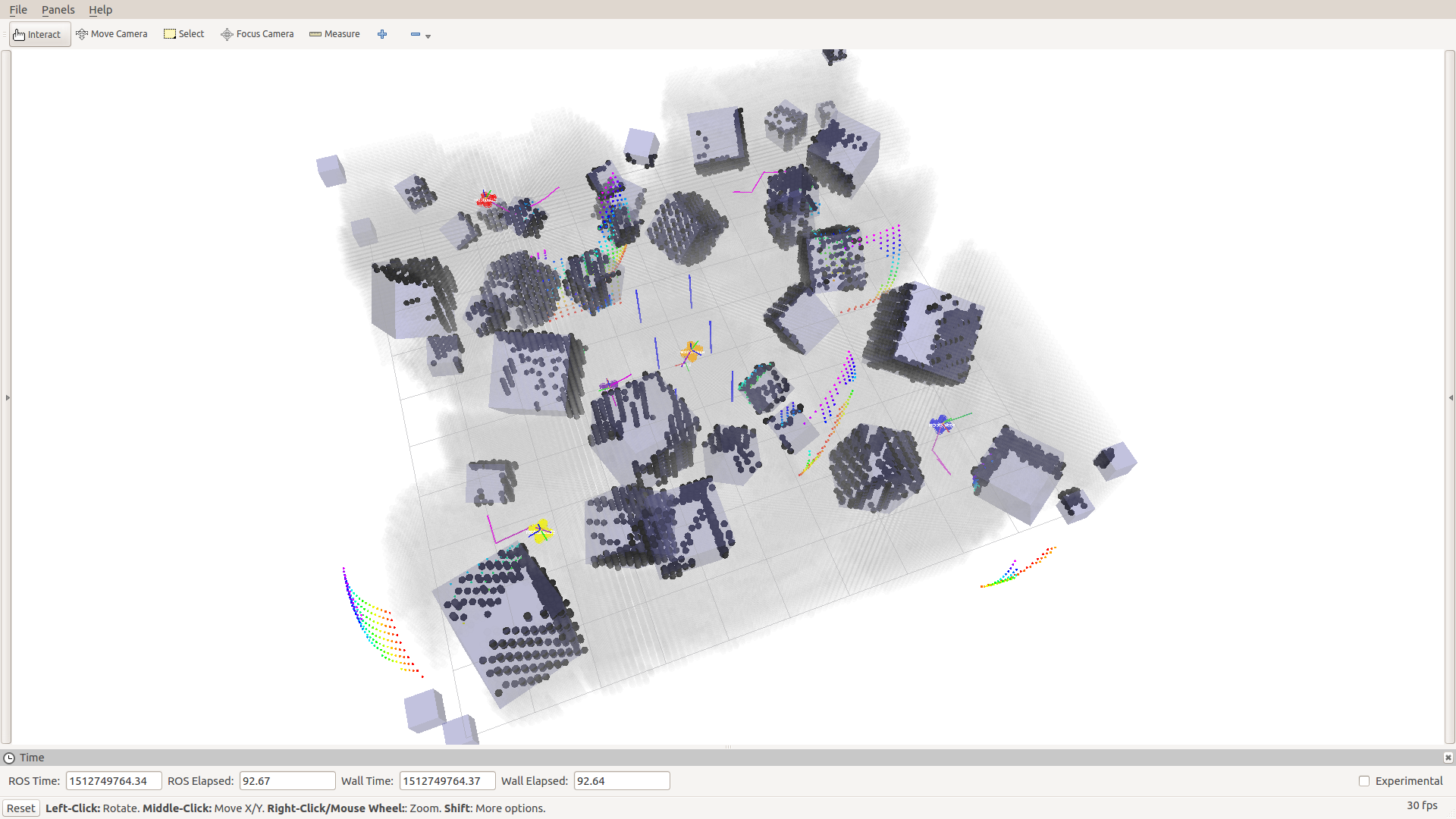}
    \includegraphics[width=0.3\linewidth, trim={780 480 850 350}, clip]{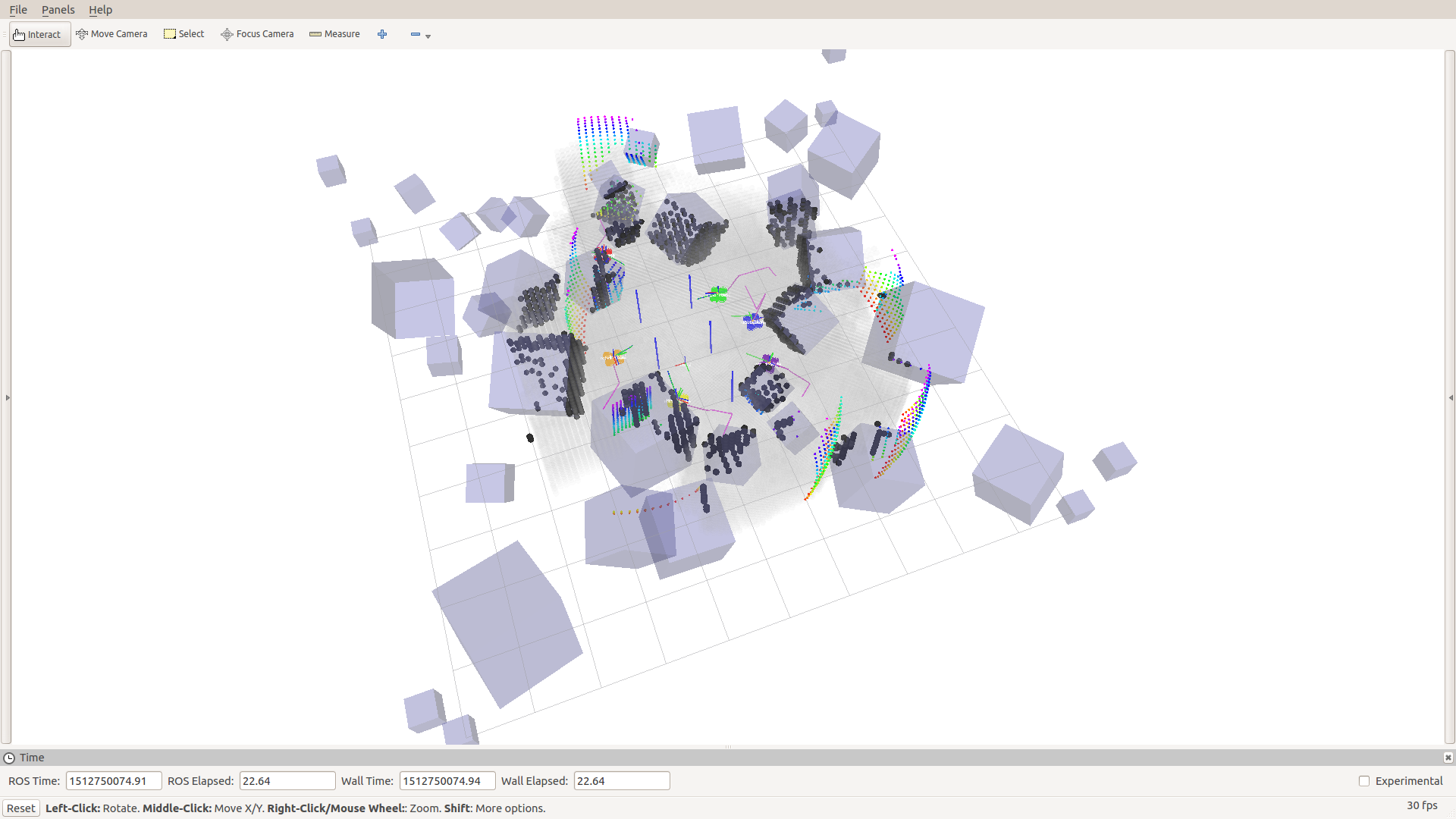}
    \setlength{\figurewidth}{0.26\linewidth}
    \setlength{\figureheight}{\figurewidth}
    \footnotesize
    \inputfigure[dynamic_cumulative_rounds]{\datapath/dynamic_cumulative_rounds.tex}
  \end{center}
  \caption[Simulated exploration results with dynamic quadrotors]
  {%
    (left) Six robots explore a pseudo-planar environment using
    motion-primitive trajectories.
    (middle) Robots (such as the green and blue) plan informative and
    collision-free trajectories using \dgreedy[3]{} despite operating in close
    proximity.
    The portion of each robot's receding-horizon plan that is currently being
    executed is colored in green, and the rest of the horizon is magenta.
    (right) Planners perform similarly in terms of entropy reduction although
    the initial performance of \dgreedy[1]{} is impaired as it frequently
    resorts to using fallback trajectories due to conflicts in planned
    trajectories.
  }\label{fig:dynamic_simulation}
\end{figure}

The simulation trials (Fig.~\ref{fig:dynamic_simulation}) evaluate a team of
six robots exploring a pseudo-planar rearrangement of the environment shown in
Figure~\ref{fig:intro_image} using sequential planning (\greedy{}), myopic
planning (\dgreedy[1]{}), and distributed planning with three rounds
(\dgreedy[3]{}) with twenty trials per each planner.
Single-robot planning steps (Monte-Carlo tree search) were run in parallel
on the same computer as in Sec~\ref{sec:kinematic_implementation}.

\begin{figure}
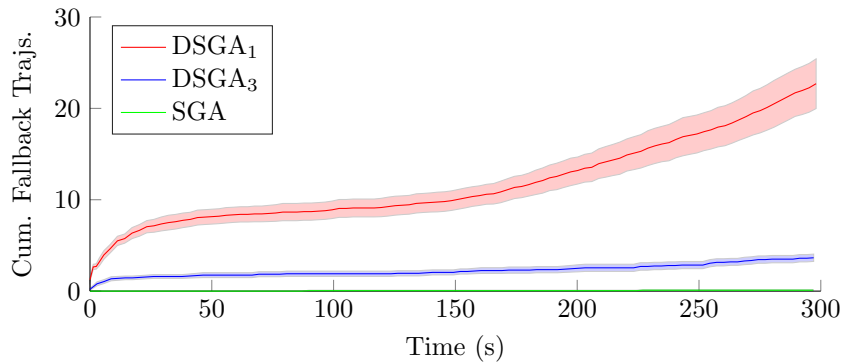

  \begin{center}
    \setlength{\figurewidth}{0.6\linewidth}
    \setlength{\figureheight}{0.40\figurewidth}
    \footnotesize
    \inputfigure[fallback]{\datapath/fallback.tex}
    \vspace{-0.5cm}
  \end{center}
  \caption[Cumulative numbers of fallback trajectories selected by multi-robot
  planner in simulation results with dynamic quadrotors]
  {%
    Cumulative numbers of fallback trajectories selected by
    multi-robot planner in simulation results with dynamic quadrotors:
    The multi-robot planners resort to fallback trajectories when the
    single-robot planner fails or returns a plan that would result in
    inter-robot collisions.
    \dgreedy[1] uses fallback trajectories frequently early in trials when
    robots are near each other and later after most of the environment has been
    explored.
  }\label{fig:dynamic_fallback}
\end{figure}

The simulation environment and results for entropy reduction are evident shown
in Fig.~\ref{fig:dynamic_simulation}.
The exploration performance is largely similar across planners except that
\dgreedy[1]{} exhibits a slight delay at the start due to a tendency for
robots to choose conflicting trajectories.
Because robots are initialized near each other, toward the center of the
environment, (with blue lines marking the locations where robots took off)
the single-robot planner returns plans that would result in inter-robot
collisions most often early in each trial.
This is evident in Fig.~\ref{fig:dynamic_fallback} which plots the cumulative
number of times that the multi-robot planner has had to fallback to the existing
safe trajectory due to either inter-robot collisions or failure of single-robot
planning as described in Sec~\ref{sec:safety}.
This is most pronounced for \dgreedy[1]{} due to the lack of coordination in the
planner.
Using the fallback trajectories is most detrimental at the very beginning of the
trial when the fallback consists of staying stationary at the starting position,
causing the delay.
Later, once they have been populated with prior planning results,
resorting to fallback trajectories becomes less significant.

\subsubsection{Experimental results}
\begin{figure}
  \begin{center}
    \includegraphics[width=0.3\linewidth, trim={290 10 300 220}, clip]{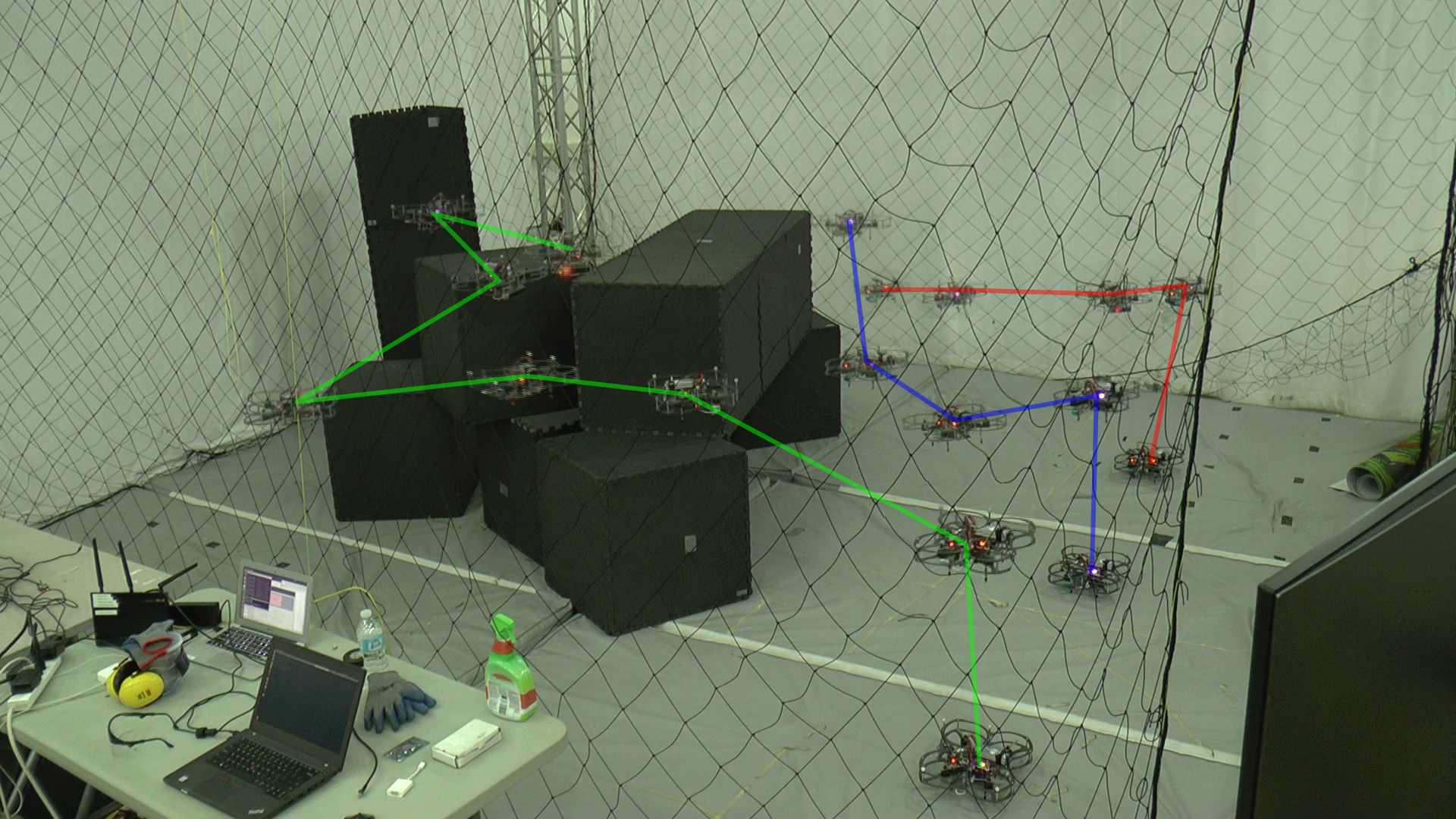}%
    \includegraphics[width=0.3\linewidth, trim={2370 380 680 200}, clip]{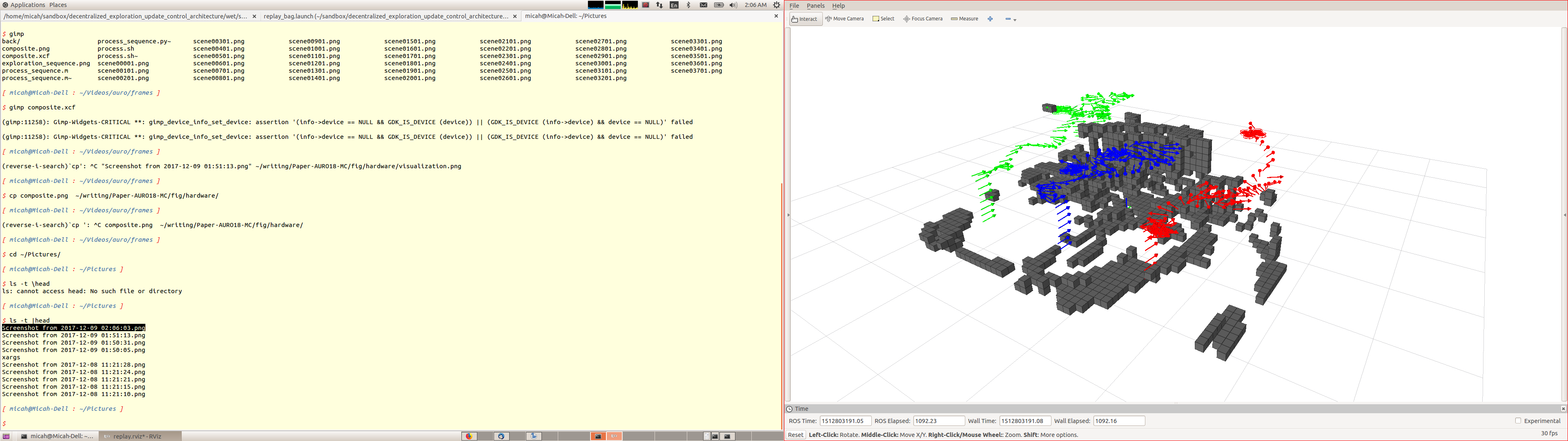}%
    \setlength{\figurewidth}{0.3\linewidth}%
    \setlength{\figureheight}{0.45\figurewidth}%
    \footnotesize%
    \inputfigure[hardware_entropy]{\datapath/hardware_entropy.tex}
  \end{center}
  \caption[Results for hardware experiments]
  {%
    (left) A team of three robots explores
    a motion capture arena occupied by a geometric object consisting of foam
    boxes using Monte-Carlo tree search and motion-primitive trajectories and
    coordination via \greedy{} and (middle) produce a voxel grid map of the
    environment
    (shown from a different vantage-point to also detail the robot in red which
    explores behind the obstacle).
    (right) The entropy of the map decreases as the robots move through the
    environment.
  }\label{fig:experimental_collisions}
\end{figure}

An experimental trial with a team of three quadrotors is also included as
depicted in Fig.~\ref{fig:experimental_collisions}.
This experiment serves to ground the simulation results described in the prior
subsections on a real system.
Due to the small size of the team, only \greedy{} is used for coordination
as \dgreedy{} would be expected to provide little benefit
(\dgreedy[3]{} is essentially equivalent to \greedy{} for three robots).
Planning is performed offboard on a laptop with a Intel i7-5600U CPU (2.6~GHz).
Otherwise, the system configuration is identical to prior set of simulation
results.
Whereas the simulation results demonstrate application of the approach described
in Sec.~\ref{sec:safety_and_independence} for exploration with collision
constraints and dynamic robots using a significant number of robots, these
hardware results connect the implementation and its configuration to a real
system.

The experiment is performed in a motion capture flight arena populated with a
pile of foam blocks of various shapes and sizes.
During the course of the exploration trial, the team of robots is able to
successfully navigate and map the environment as evident in the trajectories
and cumulative entropy reduction.
The rate of entropy reduction per robot at the beginning of the trial
is similar to Fig.~\ref{fig:experimental_collisions} at approximately
$400\frac{\text{bits}}{\si\second \cdot \text{robot}}$.
While the simulation results demonstrate efficiency of \greedy and \dgreedy
with and without collision constraints these experimental results demonstrate
that the planner design and configuration is also applicable to exploration with
a real multi-robot team and performs similarly.

\section{Conclusions and future work}
\label{sec:conclusion}

The proposed distributed algorithm (\dgreedy{}) efficiently approximates
sequential greedy assignment (\greedy{}) and is
appropriate for implementation on large multi-robot teams using online planning.
We have applied this algorithm to the problem of multi-robot exploration, and
demonstrated consistent entropy reduction performance in simulation for large
numbers of robots exploring and mapping a complex three-dimensional environment.
The results demonstrate that \dgreedy{} is able to effectively take advantage of
parallel computation without degradation in solution quality at scales where
application of \greedy{} becomes intractable.
We expect that this result will be instrumental in development of physical
multi-robot systems that take advantage of distributed computation
for exploration and other online informative planning problems.

In order to address realistic scenarios, we incorporated robot dynamics and
inter-robot collision constraints into the problem formulation.
Although collision constraints result in increased problem complexity, we were
able to modify the planning approach to guarantee safety and liveness under
weak assumptions.
The simulation results demonstrated consistent exploration performance and
indicate that collision constraints have a relatively minor impact on
exploration performance.
Further, including collision constraints makes this approach viable for
application to real robots, and we demonstrated this on a team of aerial
robots flying in a motion capture arena.

Although \dgreedy{} can take advantage of parallel computation, the actual
assignments in the subset selection step are fully sequential, and the
asymptotic computation time is identical to sequential greedy assignment even
though we obtained significant improvements in practice.
The bound on suboptimality
(Theorem~\ref{theorem:distributed_suboptimality}) also only provides a post-hoc
guarantee, and optimization performance \emph{could} degrade for
large numbers of robots (Theorem~\ref{theorem:worst-case}).
Instead, the \rsp{} methods that we introduce in the following chapters
eliminate all sequential computation involving the entire multi-robot team
and obtain strong suboptimality guarantees for any number of robots.

\chapter{Scalable and Near-Optimal Planning for Multi-Agent Sensor Coverage}
\setdatapath{./fig/scalable_multi-agent_coverage}
\label{chapter:scalable_multi-agent_coverage}


Many objective functions that arise in sensor planning problems
such as mutual information objectives~\citep{singh2009}
(and as in Chapter~\ref{chapter:distributed_multi-robot_exploration}),
objectives for sensing in hazardous environments~\citep{jorgensen2017iros},
and various notions of area,
set, and sensor coverage~\citep{roberts2017submodular} are submodular.
Intuitively, submodularity implies diminishing returns when constructing sets of
sensing actions.
This work explores multi-agent planning problems with submodular objective
functions and especially variants of set and sensor coverage.
We emphasize problems that feature networks of large or even unspecified
numbers of agents seeking to maximize a global submodular objective, and
the planners we propose obtain constant-factor performance under mild
conditions such as for agents with limited sensor range and spatially local
sensing actions.

Chapter~\ref{chapter:distributed_multi-robot_exploration} began to address
the challenge that sequential planning scales poorly as the number of
agents increases.
Further, solving agents' local planning subproblems can be time-consuming on
its own as the space of actions may be nearly
infinite~\citep{singh2009}.
At the same time, we have already seen that dynamic environments and beliefs
motivate real-time planning so that efficient multi-agent coordination is
critical when scaling to large numbers of agents.
We address this issue by proposing efficient distributed planners that consist
of fixed numbers of sequential planning steps and approach existing
constant-factor performance bounds (in expectation) when known pairwise
interactions between agents are proportional to objective values.

\begin{figure}[!t]
  \begin{subfigure}[b]{0.38\linewidth}
    \includegraphics[width=\linewidth]{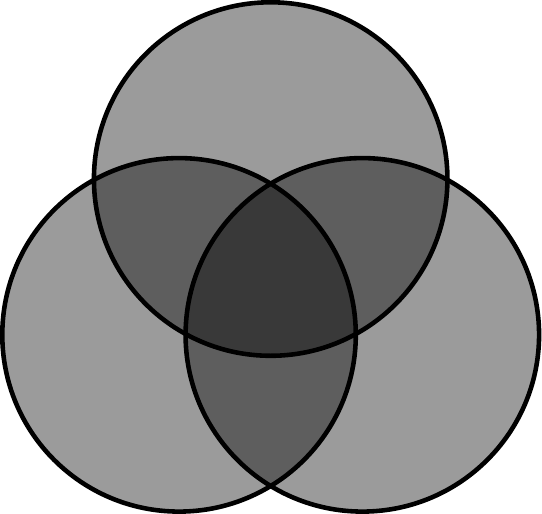}
    \caption{Area coverage}
    \label{subfig:coverage}
  \end{subfigure}
  \begin{subfigure}[b]{0.58\linewidth}
    \def\svgwidth{0.95\linewidth}
    \input{\datapath/inter-agent_redundancy.pdf_tex}
    \caption{Inter-agent redundancy}
    \label{subfig:inter-agent_redundancy}
  \end{subfigure}
  \caption[Illustration of area coverage and inter-agent redundancy]
  {%
    Consider a team of robots maximizing (\subref{subfig:coverage}) sensor area
    coverage.
    Intuitively, distant agents may make decisions independently with no loss in
    optimality.
    We exploit such conditions to enable efficient distributed planning.
    Specifically,
    (\subref{subfig:inter-agent_redundancy})
    inter-agent redundancy quantifies the maximum coupling between two agents.
    The distributed algorithm that we propose requires planning time that is
    independent of the number of agents and guarantees near-optimal performance
    when redundancy decreases monotonically subject to prior observations as is
    true for coverage objectives.
  }
  \label{fig:introduction}
\end{figure}

Several recent works~\citep{gharesifard2017,grimsman2018tcns}
also address the core challenge of this work: design of parallel variants of
sequential planners for multi-agent systems.
\citet{gharesifard2017} define a class of distributed planners based on
directed acyclic graphs where agents perform greedy planning steps using only
a subset of the decisions made by prior agents.
Although they provide worst-case bounds on suboptimality,
\citet{grimsman2018tcns} provide tighter results for the same framework.
However, both works obtain bounds that are inversely proportional to the maximum
number of agents that can plan in parallel.
In contrast, Chapter~\ref{chapter:distributed_multi-robot_exploration}
demonstrated that such planners can be effective when agents that plan in
parallel can find sets of decoupled actions.
However, that approach also only provided post-hoc bounds and
did not scale to arbitrary numbers of agents as some steps remained fully
sequential.

We continue in the direction of the previous chapter by seeking to develop
efficient distributed planners that exploit problem structure.
As discussed in Fig.~\ref{fig:introduction}, our approach is inspired by the
intuition that decisions by distant agents are often decoupled.
We model this notion with a concept of inter-agent redundancy which describes how
much one agent's marginal gain can decrease as a result of ignoring another
agent.
Our approach requires such redundancy to decrease monotonically in the presence
of actions selected by previous agents in the planning sequence.
This condition is equivalent to requiring the objective to satisfy the next
higher-order monotonicity property after submodularity.

Together, these properties enable use of pairwise redundancy to bound the effect
of ignoring an agent at any step of the planning process which relates total
redundancy to suboptimality.
We propose an algorithm---Randomized Sequential Partitions (\rsp)---that
randomly partitions agents and obtains a constant-factor bound when the optimal
solution is proportional to the cumulative pairwise redundancy between all
agents.
This condition is generally satisfied by problems involving limited sensing
range and distributions of agents with bounded density, and the approach further
admits features such as local adaptation and limits on communication range.
We refer to the latter case as Range-limited \rsp (or \rrsp).
Finally, we prove that a generalized variant of weighted set coverage satisfies
higher-order monotonicity conditions and provide simulation results for two
cases, area coverage and a probabilistic detection scenario.

This chapter originally appeared in~\citep{corah2018cdc}.

\section{Background discussion}

\subsection{Properties of set functions and the 3-increasing condition}
\label{section:set_functions}

Consider a set function $\setfun : 2^\ground \rightarrow \real$ where $\ground$
is the ground set.
Just as for the previous chapter, the functions we study in this one
are \emph{normalized}, \emph{monotonically increasing}, and \emph{submodular}
according to the definitions in Sec.~\ref{sec:submodularity}.

We now introduce the next higher-order monotonicity property
(see Sec.~\ref{sec:higher-order_monotonicity})~\citep{foldes2005}: the
objectives in this chapter are 3-increasing.\footnote{%
Previously, we referred to the 3-increasing condition as
\emph{supermodularity of conditioning}~\citep{corah2018cdc}.
\citet{wang2015} also refer to the combination of this condition with
submodularity as ``strong submodularity.''
}
Given the definition of the derivative of a set function
\eqref{eq:recursive_derivative} and for $A\subseteq B \subseteq \ground$ and
disjoint subsets $X, Y \subseteq \ground \setminus B$, 3-increasing functions
satisfy
\begin{align}
  \setfun(X;Y|A) &\leq \setfun(X;Y|B).
  \label{eq:3_increasing}
\end{align}
When expanded and negated, this expression takes the form
$\setfun(X|B) - \setfun(X|B, Y) \leq \setfun(X|A) - \setfun(X|A, Y)$.
Given that $A\subseteq B$, this can be interpreted as stating that conditioning
reduces redundancy, and expressions of the form
$-\setfun(A;C) = \setfun(A) - \setfun(A|C)$
will be referred to as expressing the pairwise redundancy between $A$ and $C$.
Such higher-order monotonicity properties have not been used extensively
in the literature on optimization of submodular functions although a few works
study the same and similar properties%
~\citep{chen2018,foldes2005,ramalingam2017,korula2018}.

\begin{figure}
  \begin{center}
    \includegraphics[width=\linewidth]{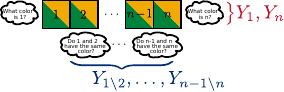}
  \end{center}
  \caption[Example of a submodular objective that is not 3-increasing]
  {%
    This illustration depicts an example of a submodular objective
    that is not 3-increasing and generalizes common examples for when mutual
    information increases under conditioning~\citep{cover2012}.
    In this example, boxes in the set $C=\{1,\ldots,n\}$ are colored
    green or gold independently and with equal probability.
    Sensors may observe either end directly
    $\{\textcolor{endcolor}{Y_1},\textcolor{endcolor}{Y_n}\}$
    and differences colors of adjacent boxes
    $\{\textcolor{differencecolor}{Y_{1\setminus 2}},\ldots,
    \textcolor{differencecolor}{Y_{n-1\setminus n}}\}$,
    to obtain a submodular mutual information reward $\setfun(Y) = \MI(C;Y)$
    where $Y$ is a set of such observations.
    Each individual observation provides one bit of information, and
    pairs obtain two bits and have no redundancy.
    This trend only changes once we consider all of the observations together:
    \emph{we can determine the color of the last box in the sequence by observing
    the first box and each subsequent change in color.}
    This violates the 3-increasing condition \eqref{eq:3_increasing} as the
    addition of last observation signifies an increase in redundancy and
    equivalently a decrease in the second derivative of the objective.
    That is
    $\setfun(\textcolor{endcolor}{Y_1};\textcolor{endcolor}{Y_n})=0$
    but
    $\setfun(\textcolor{endcolor}{Y_1};\textcolor{endcolor}{Y_n}|
    \textcolor{differencecolor}{Y_{1\setminus 2}},\ldots,
    \textcolor{differencecolor}{Y_{n-1\setminus n}}
    )=-1$.
  }
  \label{fig:mutual_information}
\end{figure}

Weighted set cover is an example of a submodular function that is 3-increasing,
and such objectives have been studied extensively and used to prove hardness
results for submodular maximization~\citep{feige1998} and tightness results for
distributed planning~\citep{grimsman2018tcns}.
Because existing results already use functions that are 3-increasing, requiring
this condition does not impact hardness of an optimization problem.
At the same time, some common submodular objectives are not necessarily
3-increasing.
Figure~\ref{fig:mutual_information} describes one such scenario for a submodular
mutual information objective.%

\subsection{Partition matroids}
\label{section:independence}
Recall from Sec.~\ref{sec:independence} that \emph{partition matroids} describe
multi-agent problems where the joint action space is a product of local action
spaces for each agent.
The ground set of the partition matroid $(\ground, \independence)$
is partitioned by a set of blocks $\{\block_1,\ldots,\block_n\}$, and
the admissible sets of actions are
$\independence = \{X \subseteq \ground \mid |X \cap \block_i|\leq \ell_i\}$ for
$\ell_i \geq 0$.

\section{Problem statement}
\label{section:problem_statement}

Consider a multi-agent planning problem with agents
$\agents = \{1,\ldots,\numagent\}$
where each agent $i\in \agents$ is associated with a set of actions
$\block_i$ which is also a block of the partition matroid.
Agents may select at most one action so that
$|X \cap \block_i| \leq 1$ for each admissible set of actions
$X\in\independence$.
Further, the agents are engaged in a sensing task with an objective $\setfun$ that
satisfies the conditions outlined in Sec.~\ref{section:set_functions} and so
seek to solve
\begin{align}
  X^\star \in \argmax_{X\in\independence} \setfun(X).
  \label{eq:problem_definition}
\end{align}
Previously, Sec.~\ref{sec:locally_greedy} described
the local greedy heuristic
(sequential maximization, Alg.~\ref{alg:local_greedy})
which obtains approximate solutions
by recursively applying a greedy maximization step.
Further, solutions via this approach are guaranteed to be within half of
optimal.

\section{Greedy planning on directed acyclic graphs}
\label{sec:dag_planning}

Sequential planning (Alg.~\ref{alg:local_greedy}) on a large network of agents
is time-consuming as each agent must wait to receive the incremental solution
from the previous agents before beginning computation.
\citet{gharesifard2017} propose a related class of planners where
agents may ignore the decisions of previous agents
according to a directed acyclic graph.
Rather than planning with respect to all prior decisions as in
Alg.~\ref{alg:local_greedy}, these planners obtain the solution
$X^\mathrm{d} = \{x_1^\mathrm{d},\ldots,x_{\numagent}^\mathrm{d}\}$
by evaluating
\begin{align}
  x^\mathrm{d}_i \in \argmax_{x\in\block_i}
  \setfun(x|X_{\neighbor_i}^\mathrm{d})
  \label{eq:dag_greedy}
\end{align}
using incremental solutions from $\neighbor_i \subseteq \{1,\ldots,i-1\}$, the
set of in-neighbors of agent $i$ in the directed acyclic graph.
This model can be used to design distributed planners where distant agents do
not communicate or where subsets of agents execute their planning steps in
parallel.
Prior works studying such planners examine worst-case behavior for objectives
that are submodular and monotonic~\citep{gharesifard2017,grimsman2018tcns}.
These fail to obtain constant-factor suboptimality when given only a fixed
number of sequential planning steps but an arbitrary number of agents.
Instead, this work examines sufficient conditions for a distributed planner with
a fixed number of sequential planning steps to approach the performance of a
sequential planner (half of optimal, see Sec~\ref{sec:locally_greedy})
We begin by analyzing \eqref{eq:dag_greedy} based on the redundancy of
sensing actions between pairs of agents.

\section{Analysis using inter-agent redundancy}
\label{section:redundancy}
The performance of the distributed planner will be analyzed by bounding
decreases in marginal gains due to failure to condition on choices by prior
agents i.e.
$\setfun(x^\mathrm{d}_i|X^\mathrm{d}_{\neighbor_i})
- \setfun(x^\mathrm{d}_i|X^\mathrm{d}_{1:i-1})$.
The 3-increasing condition enables derivation of bounds on such changes in
marginal gains using pairwise redundancies between elements.

Define the \emph{inter-agent redundancy graph} as a weighted, undirected graph
$\redundancygraph=(\agents, \edges, \weights)$
with agents as vertices,
edges $\edges=\{(i,j) \mid i,j\in\agents, i \neq j\}$,\footnote{%
  Being undirected, $(i,j)$ and $(j,i)$ are the same edge.
}
and weights
\begin{align}
  \weights(i,j) = w_{ij} = \max_{x_i \in \block_i, \, x_j \in \block_j}
  - \setfun(x_i ; x_j).
  \label{eq:redundancy_weights}
\end{align}
This connects the notion of redundancy to the multi-agent planning problem
via maximum (and undirected) inter-agent redundancies.

Decreases in marginal gains can be bounded using the pairwise redundancies
from the inter-agent redundancy graph using the following lemma.
\begin{lemma}[Pairwise redundancy bound]
  \label{lemma:redundancy}
  Consider disjoint subsets $A, B, C \subseteq \ground$.
  Then
  \begin{align}
    \setfun(A|B,C) - \setfun(A|C) \geq \sum_{b \in B} \setfun(A;b)
    \label{eq:redundancy_modified}
  \end{align}
\end{lemma}
\begin{proof}
  Lemma~\ref{lemma:redundancy} follows from the chain rule
  (Lemma~\ref{lemma:chain_rule}) and
  $\setfun$ being 3-increasing.
  Given an ordering $B=\{b_i,\ldots,b_{|B|}\}$, construct
  a telescoping sum
  \begin{align*}
    \setfun(A|B,C) - \setfun(A|C) &= \setfun(B|A,C) - \setfun(A|C) \\
                      &=
    \sum_{i=1}^{|B|} \setfun(b_i|B_{1:i-1},A,C) - \setfun(b_i|B_{1:i-1},C) \\
    &=
    \sum_{i=1}^{|B|} \setfun(b_i;A|B_{1:i-1},C)
  \end{align*}
  Then, \eqref{eq:redundancy_modified} follows because $\setfun$ is
  3-increasing~\eqref{eq:3_increasing}.
  Finally, note that, although the expression for \eqref{eq:redundancy_modified}
  is tailored to its usage, the left hand side has the form of a second
  derivative of $\setfun$ so that
  $\setfun(A;B|C) = \setfun(A|B,C) - \setfun(A|C)$.
\end{proof}
\noindent Although produced independently, the above lemma is equivalent
to~\citep[Lemma~1]{wang2015}.

The total weight of the redundancy graph will characterize suboptimality for our
approach.
We will refer to a problem defined according to~\eqref{eq:problem_definition}
as \emph{$\alpha$-redundant} for $\alpha > 0$ if
\begin{align}
  \alpha \setfun(X^\star) \geq \sum_{(i,j)\in\edges} w_{ij}.
  \label{eq:standard_weight}
\end{align}
Instances of \eqref{eq:problem_definition} with finite objective values and
numbers of agents are all $\alpha$-redundant for some $\alpha$ although specific
values are not guaranteed in general.
We will use $\alpha$-redundancy to absorb additive terms proportional to graph
weights into constant-factor multiplicative bounds in terms of $\alpha$.

\subsection{Analysis of distributed planners using inter-agent redundancy}
\label{section:dag_bound}

The inter-agent redundancy graph defined in the previous section can be
applied in the analysis of distributed planners \eqref{eq:dag_greedy} using a
similar approach as in the previous chapter
(Chapter~\ref{chapter:distributed_multi-robot_exploration}).
Let $\hat\neighbor_i=\{1,\ldots,i-1\}\setminus \neighbor_i$ be the set of
preceding agents that are ignored at step $i$ of the assignment process
according to \eqref{eq:dag_greedy}
so that the set of all deleted edges is
$\deletededges = \{(i,j) \mid i\in\agents,\,j\in \hat \neighbor_i\}$.
Then the planner suboptimality can be bounded in terms of
the weights of these deleted edges.
\begin{theorem}[Suboptimality of distributed planning]
  \label{theorem:graphical_bound}
  The suboptimality of a planner obeying
  \eqref{eq:dag_greedy}
  can be bounded using the cumulative weight of deleted edges as
  \begin{align}
    \setfun(X^\star) &\leq 2\setfun(X^\mathrm{d})
                           + \sum_{(i,j) \in \deletededges} w_{ij}.
    \label{eq:graphical_bound}
  \end{align}
\end{theorem}
\begin{proof}
  Theorem~\ref{theorem:graphical_bound} follows from application of pairwise
  redundancy on the inter-agent redundancy graph to the standard proof technique
  for sequential maximization,
  \begin{align}
    \setfun(X^\star)
    &\leq \setfun(X^\star,X^\mathrm{d}) \nonumber \\
    &= \setfun(X^\mathrm{d})
       + \sum_{i=1}^\numagent \setfun(x^\star_i|X^\mathrm{d},X^\star_{1:i-1}) \nonumber \\
    &\leq \setfun(X^\mathrm{d})
          + \sum_{i=1}^\numagent
              \setfun(x^\star_i|X^\mathrm{d}_{\neighbor_i}) \nonumber \\
    &\leq \setfun(X^\mathrm{d})
          + \sum_{i=1}^\numagent
              \setfun(x^\mathrm{d}_i|X^\mathrm{d}_{\neighbor_i}) \nonumber \\
    &= 2\setfun(X^\mathrm{d})
       + \sum_{i=1}^\numagent
           \left(\setfun(x^\mathrm{d}_i|X^\mathrm{d}_{\neighbor_i})
           - \setfun(x^\mathrm{d}_i|X^\mathrm{d}_{1:i-1})\right).
    \label{eq:excess}
  \end{align}
  The first equality is a telescoping sum.
  The inequalities follow first from $\setfun$ being monotonically increasing,
  second
  from submodularity, and third by greedy choice according to
  \eqref{eq:dag_greedy}.
  The last equality is another telescoping sum.
  The main result \eqref{eq:graphical_bound} follows from
  application of \eqref{eq:redundancy_modified} and \eqref{eq:redundancy_weights}
  to the sum in \eqref{eq:excess} and the definition of $\deletededges$.
\end{proof}

\section{Randomized distributed planners}
\label{section:distributed_planning}

Let us now apply the analysis from the previous section to design of randomized
distributed planners.
A set of agents can execute planning steps in parallel if no pair of agents in
the set shares an edge in the planner model \eqref{eq:dag_greedy}.
Considering this, we construct distributed planners by partitioning the agents
and eliminating edges within blocks of the partition and then bound
suboptimality for randomly assigned partitions.
Finally, we present conditions for such planners to scale to an arbitrary number
of agents and to admit features such as limited communication range.

\subsection{Distributed planning on partitioned agents}
\label{section:partition_planner}
Consider a planner where subsets of agents plan in parallel according to a
partition $\{D_1,\ldots,D_{\numrounds}\}$ of agents $\agents=\cup_{i=1}^{\numrounds} D_i$.
In such a planner, $\numrounds$ corresponds to the maximum number of sequential
planning steps.
Let $d_i$ map each agent $i$ to its block in the partition so that
$i \in D_{d_i}$,
and let the total ordering of agents respect a partial ordering induced by
ordering the blocks of the partition so that $i < j$ implies $d_i \leq d_j$.
We construct a planner \eqref{eq:dag_greedy}
from the partition and ordering of agents and blocks
by eliminating neighbors that share the same block from the complete directed
acyclic graph
\begin{align}
\neighbor_i = \{1,\ldots,i-1\}\setminus D_{d_i}, \quad
\hat\neighbor_i = \{1,\ldots,i-1\} \cap D_{d_i}.
\label{eq:partition_neighbors}
\end{align}
Ideally, the partition would minimize the cumulative weight of edges eliminated
in the subgraphs of the blocks.
However, that is equivalent to maximizing the weight of edges outside of the
subgraphs which is the Max $k$-Cut problem on the inter-agent redundancy graph.
Finding exact solutions is intractable because Max $k$-Cut is
NP-Complete~\citep{karp1972}.
Therefore, the next section proposes randomized approaches that produce
approximate solutions.

\subsection{Planning with random partitions}
\label{section:randomized_planner}

As observed by \citet{andersson1999}, a random partition obtains a trivial
$\frac{\numrounds-1}{\numrounds}$ by observing that edges are removed uniformly at random.
The approach presented here is similar and is presented from the perspective
of individual agents.

Consider a distributed planner as defined by \eqref{eq:dag_greedy}
where agents share partial solutions with neighbors given a
partition of the agents as in \eqref{eq:partition_neighbors}.
Let each agent select its partition index $d_i$ independently and
uniformly at random from $\{1,\ldots,k_i\}$ so that
$\numrounds = \max_{i\in\agents} k_i$.
We refer to such planners as implementing
Randomized Sequential Partitions (\rsp[\numrounds]) as this
approach partitions agents into $\numrounds$ groups which plan sequentially
over the same number of sequential steps.

We consider two policies for selection of $k_i$ based on the weights of the
redundancy graph \eqref{eq:redundancy_weights}
and a per-agent budget for additive suboptimality $\budget > 0$.
For \emph{global adaptive} planners agents $i\in\agents$ select from a fixed
number of partition indices proportional to the total redundancy so that
\begin{align}
  k_i = \numrounds =
  \left\lceil\frac{1}{\numagent\budget} \sum_{(i,j) \in \edges} w_{ij}\right\rceil.
  \label{eq:global_adaptive}
\end{align}
With \emph{local adaptive} planners $k_i$ is
proportional instead to the cumulative redundancy for that agent
which may be large compared to other agents but involves less global knowledge
\begin{align}
  k_i = \left\lceil\frac{1}{2\budget}
  \sum_{j\in\agents\setminus \{i\}} w_{ij}\right\rceil.
  \label{eq:local_adaptive}
\end{align}
Note that the factor of \emph{two} in arises because the sum in
\eqref{eq:local_adaptive} counts each edge twice unlike
\eqref{eq:global_adaptive}.
Both local and global planners respect the following bound.
\begin{theorem}[Suboptimality of \rsp{} planning]
  \label{theorem:adaptive_partitioning}
  Given a budget $\budget > 0$ for per-agent suboptimality and a planner defined
  according to \eqref{eq:dag_greedy} which partitions agents according to
  \eqref{eq:partition_neighbors} by drawing partition indices $d_i$ uniformly
  from $\{1,\ldots,k_i\}$ using \eqref{eq:global_adaptive} or
  \eqref{eq:local_adaptive} suboptimality is bounded in expectation as
  \begin{align}
    \setfun(X^\star) \leq 2 \E[\setfun(X^\mathrm{d})] + \numagent\budget.
    \label{eq:adaptive_partitioning}
  \end{align}
\end{theorem}

\begin{proof}
The expectation of the cumulative weight of deleted edges
for either planner is bounded as
\begin{align}
  \E\left[\sum_{(i,j) \in \deletededges} w_{ij}\right]
  &=
  \frac{1}{2}\sum_{i=1}^{\numagent}
  \E\left[\sum_{j\in D_{d_i}\setminus \{i\}} w_{ij}\right]
  \nonumber \\
  &\leq
  \sum_{i=1}^{\numagent}
  \frac{1}{2k_i}\sum_{j\in\agents\setminus \{i\}} w_{ij}
  \label{eq:local_suboptimality}
\end{align}
where the inequality accounts for when $d_j > k_i$.
That is, when another agent $j$ selects a partition index $d_j > k_i$
outside of the set considered by $i$, the corresponding edge cannot be
deleted from the perspective of agent $i$.
For global adaptive planners,
$k_i=\numrounds$ for all $i$ and \eqref{eq:local_suboptimality} simplifies to
$\frac{1}{\numrounds} \sum_{(i,j) \in \edges} w_{ij}$ and holds with equality.
Then \eqref{eq:adaptive_partitioning} follows by applying
\eqref{eq:global_adaptive} or \eqref{eq:local_adaptive} and substituting
into the expectation of \eqref{eq:graphical_bound} over partitions of agents.
\end{proof}

Restating in terms of $\alpha$-redundancy provides a stronger statement that
is useful when varying the number of agents.

\begin{corollary}[Constant factor suboptimality]
  \label{corollary:constant_bound}
  Problems with fixed $\alpha$-redundancy satisfy the constant-factor bound
  \begin{align}
    \frac{1-\epsilon}{2} \setfun(X^\star) \leq \E[\setfun(X^\mathrm{d})]
    \label{eq:constant_factor}
  \end{align}
  for $\epsilon > 0$ and a budget of
  \begin{align}
    \budget=\frac{\epsilon}{\alpha \numagent} \sum_{(i,j)\in\edges} w_{ij}
    \label{eq:budget}
  \end{align}
  by substituting \eqref{eq:budget} into \eqref{eq:adaptive_partitioning},
  applying \eqref{eq:standard_weight}, and rearranging.
\end{corollary}

\begin{corollary}[Fixed $\numrounds$ for global planning]
  \label{corollary:fixed_steps}
  Given fixed $\alpha$-redundancy, global adaptive planners
  \eqref{eq:global_adaptive} provide constant-factor suboptimality for
  $\numrounds = \lceil \frac{\alpha}{\epsilon}\rceil$
  sequential planning steps which follows by rearranging
  \eqref{eq:budget} to match \eqref{eq:global_adaptive}.
\end{corollary}

\subsection{Near-optimality for varying numbers of agents}
\label{section:scalability}

In this section, we present sufficient conditions to preserve these guarantees
when increasing the number of agents.
These conditions correspond intuitively to scenarios where agents have access to
local actions and where the environment volume and rewards scale with the number
of agents.

We say problems \eqref{eq:problem_definition} with $\numagent$ agents
exhibit \emph{$\beta$-linear scaling} for $\beta > 0$ if
\begin{align}
  \setfun(X^\star) \geq \beta \numagent
  \label{eq:linear_scaling}
\end{align}
which expresses the condition that rewards scale with the number of agents.

In order to express the relationship between inter-agent distances---or
the distribution of agents---and inter-agent redundancy, define a function
of inter-agent distance $r : \real_{\geq 0} \rightarrow \real_{\geq 0}$
so that
\begin{align}
  w_{ij} \leq r(||\vec{p}_i-\vec{p}_j||)
  \label{eq:distance_bound}
\end{align}
where $||\cdot||$ is some norm and $\vec{p}_i,\vec{p}_j\in \real^d$ are
appropriately defined agent positions associated with the blocks of the
partition matroid.
The following theorem identifies sufficient conditions for problems to have
finite $\alpha$ on average and in turn to satisfy
Theorem~\ref{theorem:adaptive_partitioning} and corollaries
which implies a constant expected number of sequential steps ($\numrounds$) for the
global planner design \eqref{eq:global_adaptive} and any number of agents.

\begin{theorem}[Finite average redundancy]
  \label{theorem:scalability}
  Consider a class of problems \eqref{eq:problem_definition}
  with a distribution of agents in $\real^d$ with finite density of at most
  $\agentdensity$
  that satisfies
  linear scaling \eqref{eq:linear_scaling} and has redundancy bounded in terms
  of inter-agent distance \eqref{eq:distance_bound} for fixed $\beta$ and $r$.
  If
  $\int_0^\infty r(s) s^{d-1} \dd{s}$
  is finite,
  the average value of $\alpha$, interpreted as a random variable, is also
  finite.
\end{theorem}
\begin{proof}
  Let $\agentdensity$ be an upper bound on the marginal density of agents in
  $\real^d$,
  and let $A_d$ be the surface area of the unit sphere under the chosen norm.
  Because the distribution of agents has fixed maximum density
  $\agentdensity$, we may obtain results for arbitrarily large numbers of agents
  by taking the limit as this
  distribution of agents covers the entire Euclidean space.
  By applying \eqref{eq:distance_bound} and integrating over spheres centered
  on $\vec{p}_i$ for an arbitrarily large environment, the
  expected redundancy for a given agent is at most
  $\maxredundancy = \agentdensity A_d \int_0^\infty r(s) s^{d-1} \dd{s}
  \geq
  \E\left[\sum_{j\in\agents\setminus\{i\}} w_{ij}\right]$
  which is proportional to the integral in Theorem~\ref{theorem:scalability}.
  Treating $\alpha$ as a random variable so that
  \eqref{eq:standard_weight} is tight and applying \eqref{eq:linear_scaling}
  we get
  $\E[\alpha] \leq
  \E\left[\frac{\sum_{(i,j)\in\edges} w_{ij}}{\setfun(X^\star)}\right]
  \leq \frac{\maxredundancy}{\beta}$ which is finite if $\maxredundancy$ is also
  finite given that $\beta > 0$.
\end{proof}

\subsection{Limited communication range}
\label{section:range}

Similar analysis can be used to analyze or design limits on
communication range.
Let $\communicationrange$ be the maximum communication range
so that the set of agents that are in range is
$\rangeneighbor_i=\{j \mid \communicationrange > ||\vec{p}_i-\vec{p}_j|| \}$.
Then, any existing communications graph can be readily modified by intersecting
the set of in-neighbors with the set of agents that are in range to obtain
a new set of neighbors
$\hat \neighbor_i =  \rangeneighbor_i \cap \neighbor_i$.
Doing so significantly reduces messaging: instead of each robot sending a
number of messages proportional to the total number of robots, range limits
reduce the number of messages per robot to the average size of $\rangeneighbor$.
To emphasize this difference, we refer to this approach as Range-limited \rsp{}
(or \rrsp).

To account for ignoring agents past a given range, applying
\eqref{eq:distance_bound} to Theorem~\ref{theorem:graphical_bound}, demonstrates
that each agent incurs at most
\begin{align}
\sum_{j\in \agents \setminus (\{i\} \cup \rangeneighbor_i)}
\frac{w_{ij}}{2}
\leq \sum_{j\in \agents\setminus (\{i\} \cup \rangeneighbor_i)}
\frac{r(||x_i-x_j||)}{2}
\label{eq:range_cost}
\end{align}
additional suboptimality (i.e. increase to $\budget$ in
\eqref{eq:adaptive_partitioning}).
Then, as in Sec.~\ref{section:scalability},
the expectation of \eqref{eq:range_cost} over the distribution of agents is
upper-bounded by $\rho A_d \int_{\communicationrange}^\infty r(s)
s^{d-1}\dd{s}$.
This reduces the problem of limited communication range to a question of whether
the additional suboptimality is acceptable and whether techniques such as
multi-hop communication are necessary to extend the communication range.
Note that even though the communication range can be designed to incur
arbitrarily little additional suboptimality,
limiting the communication range is not sufficient to reduce the number of
sequential planning steps as those agents within range of any one agent may, in
turn, depend on agents that are out of range of that agent and so on.

\section{Probabilistic coverage objectives}
\label{section:objective}

Although submodular set functions have been studied extensively, set functions
with higher-order monotonicity properties have received relatively little
interest~\citep{chen2018,foldes2005,ramalingam2017}.
Before moving on to present simulation results, let us examine one such
objective which satisfies the conditions presented in
Sec.~\ref{section:set_functions}.
The two scenarios that we will study in simulation involve special-cases of this
following objective which is a mild extension of weighted set cover.

Consider a general event detection or identification problem with independent,
probabilistic failures.
We define a set of events $\events$ and let each event $e\in\events$
have value $v_e\geq 0$.
Each event $e\in\events$ and element of the ground set $x\in\ground$ is
associated with an independent failure probability
$0 \leq p^e_x \leq 1$.
The expected value of identified events given a set of sensing actions
$X\subseteq\ground$ is then
\begin{align}
  \setfun(X) &=
  \sum_{e\in\events}\bigg(1-\prod_{x\in X} p^e_x\bigg)v_e
  \label{eq:coverage_objective}
\end{align}
and is equivalent to the well-known weighted set cover objective in
the deterministic case ($p_x^e\in\{0,1\}$).

The probabilistic sensor coverage objective is monotonically increasing,
submodular, and 3-increasing
and even satisfies alternating monotonicity conditions on higher
derivatives~\citep{bach2013,chen2018}.\footnote{
  See discussion of the M\"obius inversion~\citep[Sec.~6.3]{bach2013}.
}
These conditions follow inductively by demonstrating that differences are
similar in form to the original function.
We note that similar results exist in the literature for
coverage~\citet[Section 4.1]{wang2015}
and related generalizations thereof~\citep[Theorem 13]{salek2010}.
\begin{theorem}
  [Probabilistic coverage is monotonic, submodular, and 3-increasing]
  \label{theorem:sensor_failure}
  Coverage with sensor failure \eqref{eq:coverage_objective} and, by extension,
  weighted set cover satisfy alternating monotonicity conditions and are
  $i$-increasing and $i$-decreasing respectively for odd and even $i$.
  As such, each is monotonic, submodular, and 3-increasing.
\end{theorem}
\begin{proof}
  The probabilistic coverage objective \eqref{eq:coverage_objective} can be
  written in the form
  \begin{align}
    \setfun(X) =
    a - \sum_{e\in\events}\prod_{x\in X\setminus A} p^e_x \hat v_e
  \end{align}
  for $a\in\real$, $A\subseteq\ground$, $\hat v_e \in \real_{\geq0}$,
  and $0\leq p^e_x\leq 1$ for $e\in \events$ and $x\in\ground$.
  Observe that $\setfun$ is monotonic due to the form of the product and because
  the terms of the product are probabilities.
  Recall the recursive definition of the derivative
  \eqref{eq:recursive_derivative}, and
  consider the derivative $\setfun(Y|X)$ in the direction $Y\subseteq\ground$
  evaluated at $X\subseteq\ground\setminus Y$ which has the form
  \begin{align}
    \setfun(Y|X)
    &=
    \setfun(Y,X) - \setfun(X)
    \nonumber \\
    &=
    - \sum_{e\in\events}\prod_{j\in (X\cup Y)\setminus A} p^e_j \hat v_e
    + \sum_{e\in\events}\prod_{k\in X\setminus A} p^e_k \hat v_e
    \nonumber \\
    &=
    \sum_{e\in\events}\bigg(1-\prod_{j\in Y\setminus A} p^e_j\bigg)
    \prod_{k\in X\setminus A} p^e_k \hat v_e.
  \end{align}
  Because $-\setfun(Y|\cdot)$ has the same form as $\setfun$, this first
  derivative $\setfun(Y|\cdot)$ is decreasing; that is, $\setfun$ is submodular.
  Applying the recursive definition of the derivative
  \eqref{eq:recursive_derivative} to $\setfun(Y|\cdot)$ then produces
  the second derivative which is, in turn, a monotonically increasing function
  with the same form as $\setfun$.
  Therefore, $\setfun$ is 3-increasing \eqref{eq:3_increasing}.
  Moreover, by induction $\setfun$ also satisfies alternating monotonicity
  conditions on higher derivatives (Sec.~\ref{sec:higher-order_monotonicity}).
  Finally, probabilistic coverage $\eqref{eq:coverage_objective}$, having the
  same form, satisfies the same monotonicity conditions.
\end{proof}

\section{Results and discussion}
The proposed distributed planning approach is evaluated through two sets of
simulation experiments, each using a variant of the objective function
analyzed in Sec.~\ref{section:objective}.
The first evaluates the performance of distributed planners (\rsp{}) that use
various numbers of sequential planning steps (agent partition size $\numrounds$)
in an area coverage task.
The second set of experiments evaluates adaptive planning and limits on
communication range (\rsp{} and \rrsp{}) in a more complex problem with spatially
varying rewards and probabilistic sensing.

\begin{table}
  \caption{%
    Agent ($\agentrange$) and sensor ($\sensorrange$)
    radii as a function of the number of agents
    ($\numagent\!=\!50$ in this chapter).
  }
  \label{tab:radii}
  \centering
  \begin{tabular}{rclcl}
          & \multicolumn{2}{c}{Area Coverage}
          & \multicolumn{2}{c}{Probabilistic Sensing} \\
          & By $\numagent$ & $\numagent=50$
          & By $\numagent$ & $\numagent=50$
          \\\belowtoprule
    $\sensorrange$ & $\sqrt{2/(\numagent \pi)}$   & 0.113
                   & $\sqrt{0.6/(\numagent \pi)}$ & $6.18 \cdot 10^{-2}$
                   \\
    $\agentrange$  & $2\sensorrange$      & 0.226
                   & $4\sensorrange$      & 0.247
  \end{tabular}
\end{table}

\subsection{Common parameters of experiment designs}
Several aspects of experiment design are kept constant in each scenario.
Each scenario is evaluated in 50 random trials%
\footnote{
  Unless otherwise specified, each trial features both random planners and
  scenarios according to their respective designs.
}
and features a large number of
agents ($\numagent\!=\!50$) so that the proposed distributed planners utilize
many times fewer
sequential planning steps than a standard sequential planner
(Alg.~\ref{alg:local_greedy}).
Agent positions are distributed uniformly at random over the unit square, and
each agent has a choice of 10 available sensing actions ($\block_i$)
which are sampled from a uniform distribution over a disk with radius
$\agentrange$ centered on the agent position.
Although the two sets of simulation experiments do not use the same sensor
model, each is a function of sensor radius $\sensorrange$.
The sensor and agent radii used in each experiment%
\footnote{
  Agent and sensor radii are set according to a normalization over the number of
  agents and by using parameter search to minimize the ratio of the average
  performance of myopic and sequential planning to identify hard problem cases.
}
are listed in
Tab.~\ref{tab:radii}.
In each case, the objective is designed to take on values no greater than one.

\begin{figure}[!tb]
  \begin{center}
    \includegraphics[width=0.6\linewidth]{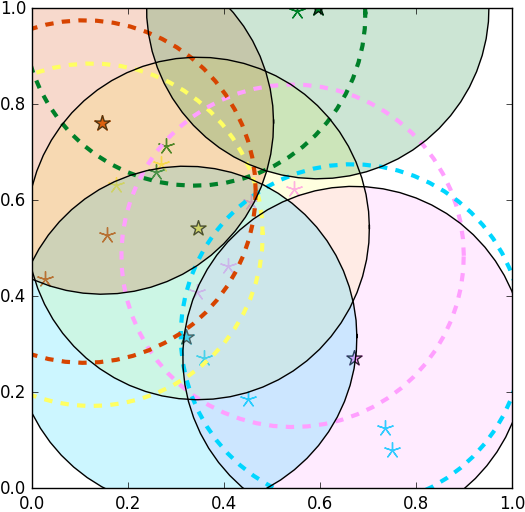}%
  \end{center}
  \caption[Maximum area coverage problem and sequential solution]
  {%
    This figure depicts a maximum area coverage problem and a sequential
    solution.
    Agents each have a unique color and are distributed uniformly throughout the
    environment.
    Sensing actions ($*$) are in turn distributed uniformly within agent
    radii (dashed lines).
    The solution consists of one selected action ($\star$) for each agent;
    these actions are centered on the translucent disks which make up the
    covered area.
  }
  \label{fig:coverage_example}
\end{figure}

\begin{figure}[!tb]
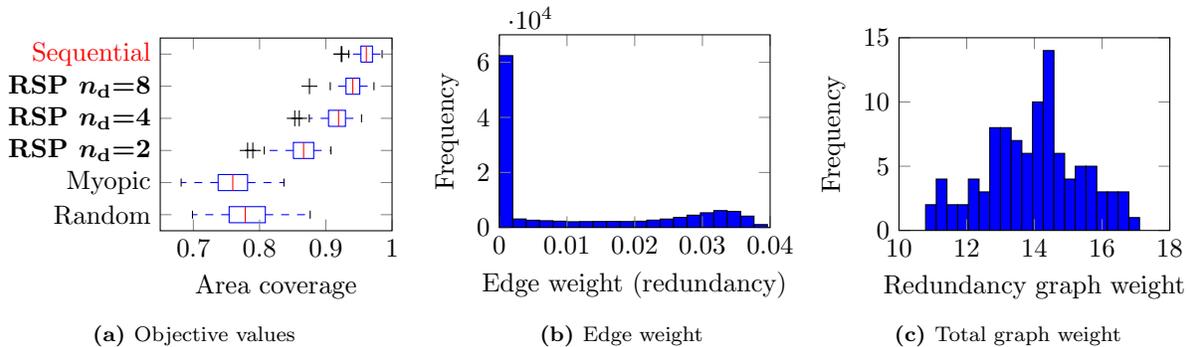

  \begin{center}
    \begin{subfigure}[b]{0.32\linewidth}
      \setlength{\figurewidth}{0.9\linewidth}
      \setlength{\figureheight}{0.8\linewidth}
      \footnotesize
      \inputfigure[coverage_results]{\datapath/compare_solvers/results.tex}%
      \caption{Objective values}%
      \label{subfig:coverage_results}
    \end{subfigure}
    \hskip 2ex
    \begin{subfigure}[b]{0.32\linewidth}
      \setlength{\figurewidth}{1.0\linewidth}
      \setlength{\figureheight}{0.8\linewidth}
      \footnotesize
      \inputfigure[edge_weight_frequency]{\datapath/compare_solvers/edge_weight_frequency.tex}
      \caption{Edge weight}%
      \label{subfig:coverage_edge_weight_frequency}
    \end{subfigure}
    \hskip -1ex
    \begin{subfigure}[b]{0.32\linewidth}
      \setlength{\figurewidth}{1.0\linewidth}
      \setlength{\figureheight}{0.8\linewidth}
      \footnotesize
      \inputfigure[total_graph_weight_frequency]{\datapath/compare_solvers/total_graph_weight_frequency.tex}
      \caption{Total graph weight}%
      \label{subfig:coverage_graph_weight}
    \end{subfigure}
  \end{center}
  \caption[Results for area coverage problem]
  {%
    Results for the area coverage problem
    (Fig.~\ref{fig:coverage_example}):
    (\subref{subfig:coverage_results})
    Objective values for myopic and random planning (no coordination),
    variations of the
    \textbf{distributed \rsp[][\textbf] planner} with $\numrounds$
    sequential planning steps,
    and fully \textcolor{red}{sequential planning}
    (intractable for large numbers of agents).
    The performance of the proposed distributed planner approaches sequential
    planning given many times fewer sequential planning steps.
    (\subref{subfig:coverage_edge_weight_frequency}) Redundancies are computed
    for each pair of agents.
    Sensing actions (disks) for distant agents cannot overlap resulting in many
    zero weighted edges, and remaining edges are distributed according to
    varying degrees of overlap in potential sensing actions.
    (\subref{subfig:coverage_graph_weight})
    The total weight of the redundancy graph is largely between 13 and 17.
    Even though the planners perform well, our suboptimality bounds would no
    longer be meaningful as deleted edge weights would exceed maximum objective
    values.
  }
  \label{fig:coverage_results}
\end{figure}

\subsection{Area coverage and evaluation of distributed planning}
\label{sec:area_coverage_results}

The reward for the area coverage task is the area of the union of discs,
each with radius $\sensorrange$, intersected with the unit square.
In terms of the sensor model defined in Sec.~\ref{section:objective},
this is equivalent to having a failure probability of one outside the disk and
zero inside.
An example of one simulation trial (using parameters tuned for visualization
purposes) is depicted in Fig.~\ref{fig:coverage_example}.
The experiments compare distributed planning with \rsp{} and fixed partition
sizes, $k_i = \numrounds\in \{2, 4, 8\}$, to sequential planning
(Alg.~\ref{alg:local_greedy})
and two naive planners: completely random action selection and myopic
maximization of the objective over the local space of sensing actions
(wherein agents ignore each entirely, equivalent to $\numrounds=1$).
Figure~\ref{fig:coverage_results} shows the results of these experiments.

Our \rsp{} planners perform well, although the performance bounds
(Theorem~\ref{theorem:adaptive_partitioning})
are technically degenerate for these instances because the deleted edge weight
generally exceeds the maximum possible objective value (one unit of area) which
is evident from the cumulative weights of the inter-agent redundancy
graph.\footnote{
  Note that the bound ceases to be degenerate for larger values of $\numrounds$
  and that all results for suboptimality and scaling still apply.
}
However, the trend in performance is similar to what would be expected for
increasing $\numrounds$ as the objective values of our \rsp{} planners approach
the performance of sequential planning approximately with $1/\numrounds$:
the difference in area coverage compared to sequential planning decreases by
approximately half each time $\numrounds$ is doubled and by 9.9 times from
$\numrounds=1$ to $\numrounds=8$.
Overall, the performance of the distributed \rsp{} planner represents a
significant improvement over sequential planning.
Given the number of agents, even the greatest value of $\numrounds$ provides a
$6$-times improvement in the number of sequential planning steps.
Then, by the scalability analysis in Sec.~\ref{section:scalability}, similar
performance can be expected for larger problems given similar densities of
agents.

\begin{figure}[!tb]
  \begin{center}
    \begin{subfigure}[b]{0.48\linewidth}
      \includegraphics[width=1.0\linewidth]{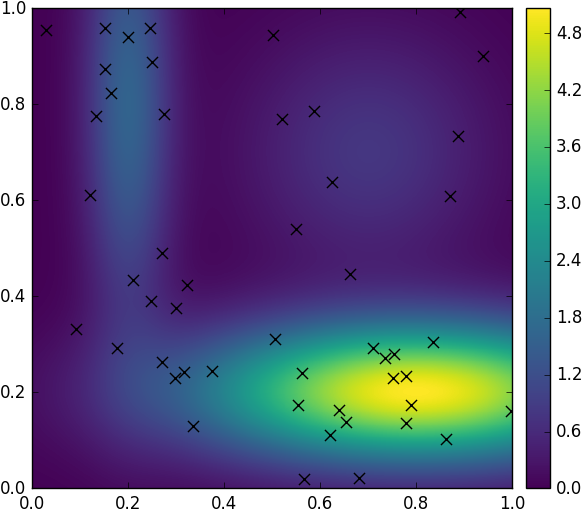}%
      \caption{Scenario}
      \label{subfig:probabilistic_scenario}
    \end{subfigure}
    \begin{subfigure}[b]{0.48\linewidth}
      \includegraphics[width=1.0\linewidth]{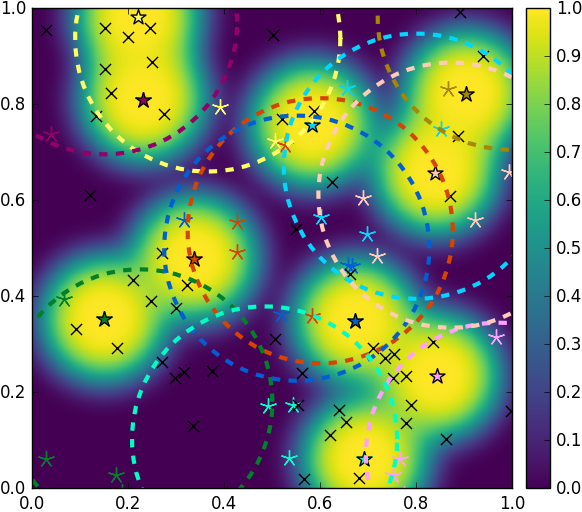}%
      \caption{Planner result}
      \label{subfig:probabilistic_example}
    \end{subfigure}
  \end{center}
  \caption[Example of a probabilistic sensing scenario and a sequential
  solution]
  {%
    This figure shows an example of a probabilistic sensing scenario and a
    sequential solution.
    Parameters for the scenario are identical to the experimental trials,
    but the parameters for the agents have been tuned for purpose of
    visualization.
    The goal of this task is to maximize the expected number of successful
    detections or identifications.
    (\subref{subfig:probabilistic_scenario})
    For each trial, events (x) are sampled from a mixture of Gaussians
    and are identified correctly with some probability dependent on the
    sensing actions.
    (\subref{subfig:probabilistic_example})
    Agents, each shown in a different color, are distributed uniformly
    throughout the environment.
    Sensing actions ($*$) are distributed uniformly
    within the agent radius (dashed lines), and
    each agent selects a single sensing action ($\star$) and
    successfully identifies events according to a soft-coverage sensing
    model.
    The resulting identification probability given selected actions is shown in
    the background; identification probability is high (yellow) near selected
    sensing actions and, for good selections, near the events.
  }
  \label{fig:probabilistic_example}
\end{figure}

\begin{figure}[!tb]
  \centering
  \begin{subfigure}[b]{0.48\linewidth}
    \setlength{\figurewidth}{0.8\linewidth}
    \setlength{\figureheight}{0.8\linewidth}
    \scriptsize
    \inputfigure[probabilistic_coverage_results]{\datapath/compare_adaptive_solvers/results.tex}
    \caption{Objective values}
    \label{subfig:probabilistic_results}
  \end{subfigure}
  \begin{subfigure}[b]{0.48\linewidth}
    \setlength{\figurewidth}{\linewidth}
    \setlength{\figureheight}{0.8\linewidth}
    \scriptsize
    \inputfigure[global_partition_size_frequency]{\datapath/compare_adaptive_solvers/global_partition_size_frequency.tex}
    \caption{Partition sizes: global adaptive \rsp{}}
    \label{subfig:global_partition_size_frequency}
  \end{subfigure}\\
  \begin{subfigure}[b]{0.48\linewidth}
    \setlength{\figurewidth}{\linewidth}
    \setlength{\figureheight}{0.8\linewidth}
    \scriptsize
    \inputfigure[local_partition_size_frequency]{\datapath/compare_adaptive_solvers/local_partition_size_frequency.tex}
    \caption{Partition sizes: local adaptive \rsp{}}
    \label{subfig:local_partition_size_frequency}
  \end{subfigure}
  \begin{subfigure}[b]{0.48\linewidth}
    \setlength{\figurewidth}{0.8\linewidth}
    \setlength{\figureheight}{0.8\linewidth}
    \scriptsize
    \inputfigure[deleted_edge_weights]{\datapath/compare_adaptive_solvers/deleted_edge_weights.tex}
    \caption{Deleted edge weight}
    \label{subfig:deleted_edge_weight}
  \end{subfigure}
  \caption[Results for a probabilistic sensing problem]
  {%
    Results for the probabilistic sensing problem
    (Fig.~\ref{fig:probabilistic_example}):
    (\subref{subfig:probabilistic_results})
    The proposed \textbf{local} and \textbf{global adaptive \rsp[][\textbf]{}
    planners}
    along with their \textbf{Range-limited \rsp[][\textbf]{} counterparts}
    outperform myopic planning and approach the performance of the
    fully \textcolor{red}{sequential planner}
    (intractable for large numbers of agents) in terms of objective values.
    (\subref{subfig:global_partition_size_frequency})
    The global adaptive planner uses 4 to 10 partitions in all trials while
    (\subref{subfig:local_partition_size_frequency})
    the local adaptive planner occasionally uses local partitions sizes
    exceeding 20 for agents with high redundancy.
    (\subref{subfig:deleted_edge_weight})
    The cumulative weight of deleted edges is on the same order as the objective
    value and is below the desired limit of 0.4.
    Range-limited planners are obtained by deleting edges from the associated
    adaptive planner at the cost of relatively little additional deleted edge
    weight.
  }
  \label{fig:probabilistic_results}
\end{figure}

\subsection{Adaptive planning with probabilistic sensing and non\hyph{}uniform
events}

The goal of the probabilistic sensing task is to maximize the value of correctly
identified events
(e.g. correct classification of objects moving through the environment).
For each trial $\numevents=50$ events, each worth a value of $1/\numevents$, are
sampled from a fixed Gaussian mixture, rejecting samples outside the unit
square.
An example is shown in Fig.~\ref{fig:probabilistic_example} although using agent
parameters more appropriate for visualization purposes but the same Gaussian
mixture and number of events.
This results in a spatially varying distribution of reward and redundancy.
The success probability of the sensor model is
$e^{-x^2/\sensorrange^4}$,
where $x$ is the distance from sensor to event location.
This success probability effectively amounts to area coverage with soft edges.

This set of experiments evaluates adaptive planning and limited communication
range.
The budget for deleted edge weight per-agent for the local and global
planners is set to $\budget=0.4/n=8 \cdot 10^{-3}$.
Range-limited \rsp{} planners are obtained by deleting edges from the
respective instances of the distributed planners with a communication range
of $\communicationrange=2\agentrange$ which allows for a small amount of
redundancy at the limits of the sensor range $\sensorrange$.
Random planning is not included in this set of experiments
because it vastly under-performs myopic planning as the problem design ensures
that a large fraction of sensing actions provide little value.

Fig.~\ref{fig:probabilistic_results} shows the results of these experiments.
Local and global adaptation each perform almost identically in terms of
distributions of objective values and with distribution only slightly below that
of sequential planning.
Because the objective and actions are highly local, enforcing limits on the
communication range has little impact on the planner performance in terms of
either objective value or cumulative weight of deleted edges.
The global adaptive \rsp{} planner obtains consistent partition sizes by
averaging over all agents.
In contrast with the local planners, agents sometimes select from as many as 33
planning steps.
Sec.~\ref{section:scalability} provides some discussion of mitigation
strategies to avoid such long planning times.
More generally, mixing the extremes of the local and global \rsp{} planning
may be desirable and avoid computing averages over all agents.

\section{Conclusion}
Efficiently solving submodular maximization problems on sensor networks is
challenging due to the inherent sequential structure of common planning
strategies.
Whereas prior works~\citep{gharesifard2017,grimsman2018tcns}
have shown that worst-case performance degrades rapidly when reducing the number
of sequential planning steps, we show that constant-factor performance
approaching that of the standard sequential algorithm can be obtained via
randomized planning (\rsp{}) so long as cumulative redundancy between agents is
at most proportional to the objective values.
Toward this end, the inter-agent redundancy graph expresses the degree of
coupling between agents in the submodular maximization problem, and functions
that are 3-increasing admit performance bounds in terms of
this graph structure.
Further, we have demonstrated that the bounds we obtain with this approach are
readily applicable for planner design such as for adapting the number of
sequential planning steps or via range limits on communication.

Later chapters will apply these results to the design of online anytime
receding-horizon planners for target tracking tasks and exploration (as in
Chapter~\ref{chapter:distributed_multi-robot_exploration}).
Likewise, this same approach is applicable to other multi-robot sensing tasks
which the reader might encounter.
Ultimately, planning in real time will also require increased attention to
timing for planning steps such as reasoning about the impact of available
planning time on anytime planning performance.

Additionally, we have noted that submodular objectives such as mutual
information are not necessarily 3-increasing.
Identifying when objective functions are 3-increasing (exactly or approximately)
is central to broad application of the results in this chapter, and we will find
ways to obtain similar results for mutual information objectives for both target
tracking and exploration.

Finally, high-order monotonicity conditions have been useful in other problems
involving submodular objectives~\citep{salek2010,wang2015,korula2018,gupta2020},
and developing these applications would be an interesting direction for future
study.

\chapter{Receding-Horizon Planning for Target Tracking with Sums of Submodular Functions}
\setdatapath{./fig/target_tracking}
\label{chapter:target_tracking}

Planners for robotics and sensor planning problems frequently introduce notions
of locality, coupling, or redundancy amongst robots, sensors, or
locations~\citep{krause2006,brown2020}.
Chapter~\ref{chapter:scalable_multi-agent_coverage} introduced one such notion
of redundancy using properties of 3-increasing functions in order to design
scalable planners.
In this chapter, we expand on the idea of redundancy as a tool
for planner design by analyzing a class of factored mutual information
objectives that are relevant to target tracking and form sums of submodular
functions.

Chapter~\ref{chapter:scalable_multi-agent_coverage} provided performance bounds
for a slight generalization of weighted set coverage but also presented a
counterexample for mutual information with conditionally independent
observations (i.e. for typical submodular mutual information objectives,
see Sec.~\ref{sec:relevant_submodular_functions}).
Mutual information is particularly important for
perception
and control~\citep{schwager2011isrr,julian2014,charrow2015icra} and is the focus
of Chapter~\ref{chapter:distributed_multi-robot_exploration}.
As such, generalizing to a variety of objective functions, as we will begin to
do here, is an important part of this thesis.

Additionally, this chapter applies the \rsp{} and \rrsp{} planners from the
previous chapter to produce anytime, receding-horizon planners and includes
additional analysis for this setting.

\section{Introduction for target tracking}

In target tracking problems, robots seek to observe a number of discrete targets
whose states may evolve in time,
such as for surveillance, monitoring wildlife~\citep{cliff2015rss}, and
intercepting rogue UAVs~\citep{shah2019ras}.
Such sensing tasks typically involve planning to observe a number of discrete
objects, \emph{targets}, whose states change with time.
When a large quantity of targets are spread over more space than a
single robot can cover, deploying teams of robots can improve tracking
performance.

Even simple target tracking problems, such as with noisy range sensors, produce
multi-modal posterior distributions that do not have closed-form solutions;
planning and tracking systems may then approximate both the posterior and
sensing utility~\citep{charrow2014auro,charrow2014ijrr}.
Realistic environments can also induce complex motion models which arise in
search on road networks~\citep{piacentini2019jair} and in indoor
environments~\citep{hollinger2009ijrr}.
This motivates selection of sufficiently general objectives and path planners to
capture the degree of complexity in these problems.
To address this, we provide analysis that applies to general problems and even
adversarial problem selection so long as the targets' dynamics do not depend on
the tracking robots.\footnote{%
  The latter condition excludes
  pursuit-evasion problems~\citep{chung2011auro,shah2019ras}.
}

Systems for target tracking in multi-robot settings often rely
on greedy algorithms~\citep{tokekar2014iros,zhou2019tro} for submodular
maximization.
However, existing results and those in
Chapter~\ref{chapter:scalable_multi-agent_coverage}
that seek to develop scalable planners that are relevant to target tracking
problems are also limited to coverage-like objectives~\citep{sung2019auro}.
Analysis for sequential planners also typically assumes individual robots obtain
either exact~\citep{fisher1978,atanasov2015icra} solutions or solutions within a
constant factor of optimal~\citep{singh2009,jorgensen2017iros}.
Although that analysis is appropriate for single-robot planners with guarantees on
solution
quality~\citep{chekuri2005,singh2009,jorgensen2017iros}, assuming
constant-factor suboptimality is less suited for
approximate, anytime, and sampling-based
planning~\citep{hollinger2014ijrr,lauri2015ras,atanasov2015icra} which we will
address in this chapter.

\subsection{Contributions}
This chapter presents a distributed planning algorithm, analysis that
accounts for common algorithmic approximations, and design of such a planner
along with simulation results.

\subsubsection{Analysis of pairwise redundancy for distributed planning}
The analysis in this chapter demonstrates that \rsp{} planners are applicable to
target tracking problems by providing guarantees on solution quality
in terms of pairwise redundancy between robots' actions via an extension to sums
of submodular functions.
This extends our prior results for coverage-like objectives
(Chapter~\ref{chapter:scalable_multi-agent_coverage})
to include more general objectives such as for mutual information%
\footnote{%
  Fig~\ref{fig:mutual_information}
  describes a mutual information objective that
  violates requirements on ``coverage-like'' objectives.
}
which can represent information gain with respect to targets that have complex
states and dynamics.

\subsubsection{Analysis of an approximate, anytime planning}
We also account for the contributions of common sources of
suboptimality \emph{(approximation of the objective and suboptimal
single-robot planning)}.
Our results affirm that methods for submodular maximization are applicable for
target tracking in the presence of approximate objective values and anytime
planners that may occasionally produce poor results or fail.

\subsubsection{Design of a planner for multi-robot multi-target tracking}
Finally, we apply these results to develop a planner for multi-robot
multi-target tracking with a mutual information objective, and
we show that distributed planning for target tracking with
Range-limited \rsp{} \emph{in constant time, independent of the number of
robots} can guarantee suboptimality approaching that of fully sequential
planning.
Finally, simulation results support our claim that \rrsp{} maintains consistent
solution quality while planning for up to 96 robots.
This produces over a $20\times$ reduction in the number of sequential planning
steps and an even further improvement in computation time
(sequential planning is intractable at this scale).

\section{Target tracking problem}

\begin{figure}
  \floatbox[{\capbeside\thisfloatsetup{capbesideposition={right,top},capbesidewidth=0.27\linewidth}}]{figure}[0.7\textwidth]
  {\caption[Target tracking illustration]{%
      A team of aerial robots $\robots$ (black) plan over a receding horizon to
      track a number of targets $\targets$ (red).
      In doing so, robots select sensing actions to minimize uncertainty in the
      target states which evolve independent of each other and the robots.
  }%
  \label{fig:target_tracking}
  }
  {\includegraphics[width=\linewidth]{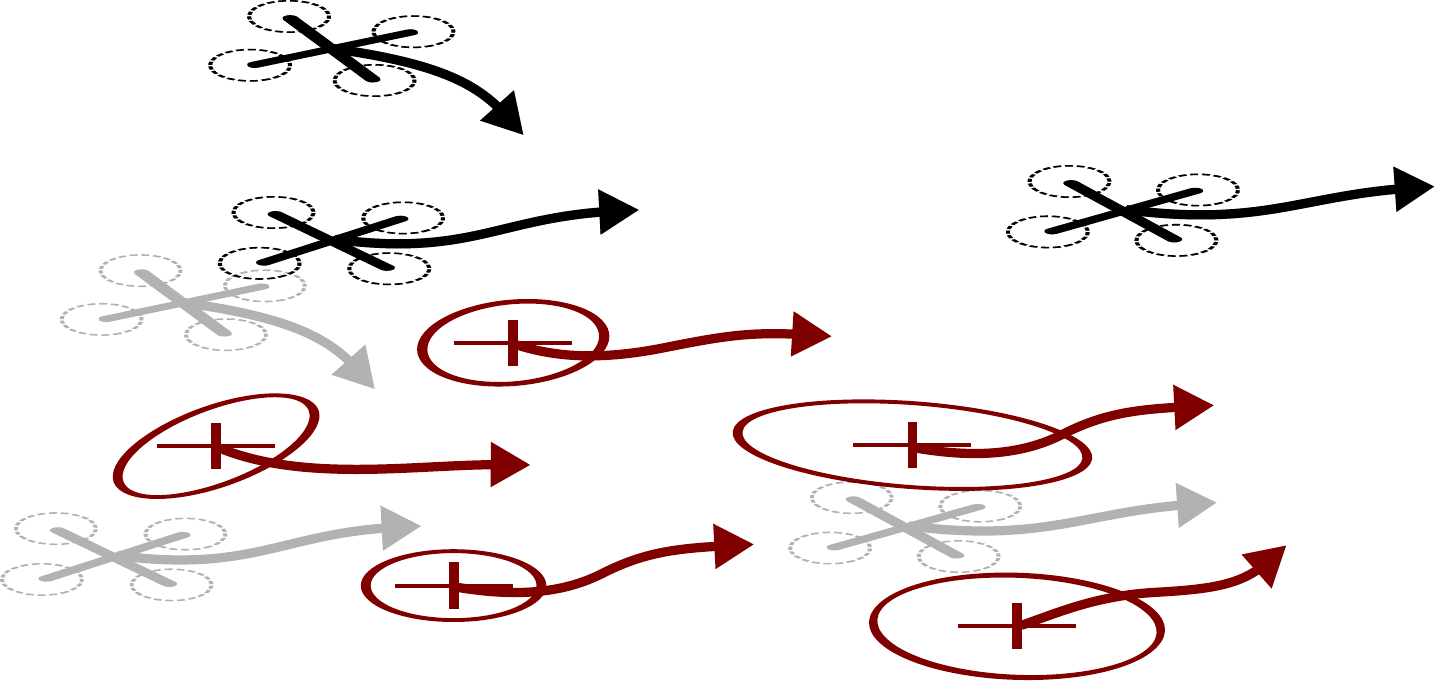}}
\end{figure}

Consider a set of moving targets $\targets=\{1,\ldots,\numtargets\}$
and robots tracking those targets $\robots=\{1,\ldots,\numrobots\}$, seeking to
minimize uncertainty (entropy~\citep{cover2012}), as illustrated in
Fig.~\ref{fig:target_tracking}.
Let $\robotstate_{i,t}\in\robotspace$ and
$\targetstate_{j,t}\in\targetspace$ be the respective states of robot
$i\in\robots$ and target $j\in\targets$ at time $t\in\{0,\ldots,T\}$.
The states of each evolve in discrete time, according to known dynamics
\begin{align}
  \robotstate_{i,t+1} &= \robotdynamics(\robotstate_{i,t},\control_{i,t}),
    &
  \targetstate_{j,t+1} &=
  \targetdynamics(\targetstate_{j,t},\targetnoise_{j,t}),
  \label{eq:tracking_dynamics}
\end{align}
where $\control_{i,t}\in\controlspace$ is a control input from the finite set of
control inputs $\controlspace$
and $\targetnoise_{i,t}$ is
a random disturbance.
The robots receive noisy observations $\observation_{i,j,t}$ of the target
states according to
\begin{align}
  \observation_{i,j,t} &=
  \observationfunction(
  \robotstate_{i,t},\targetstate_{j,t},\observationnoise_{i,j,t})
  \label{eq:observations}
\end{align}
where $\observationnoise_{i,j,t}$ is observation noise.
We refer to states and observations collectively using boldface capitals
as $\robotstates_t$, $\targetstates_t$, and $\observations_t$, each at
time $t$.

\subsection{Receding-horizon optimization problem}
\label{sec:optimization_problem}

The robots plan to jointly maximize a mutual information (MI) objective over a
receding horizon, starting at time $t$ with duration $l$.
Specifically, robots maximize a submodular, monotonic, and normalized
objective $\setfun$ subject to a partition matroid constraint, and the optimal
set of control actions is
\begin{align}
  X^\opt \in \argmax_{X\in\independence} \setfun(X)
  \label{eq:optimization_problem}
\end{align}
where $\independence$ is a partition matroid that represents assignment of
sequences of control actions to robots such that
\begin{align}
  \block_i &= \{(i, \control_{1:l})\mid \control_{1:l}\in \controlspace^l\}
  \quad \forall i\in\robots,
\end{align}
and $\setfun$ is the mutual information between observations and target states given
the choice of control actions.
Interpreting future states $\targetstates_{t+1:t+l}$ and observations
$\observations_{t+1:t+l}$ as random variables (induced by the
process~\eqref{eq:tracking_dynamics} and observation~\eqref{eq:observations} noise),
and write $\observations_{t+1:t+l}(X)$ for $X\subseteq\ground$\footnote{%
  More formally, $\observations_{t+1:t+l}(X)$ is a random variable that encodes
  the noisy observations \eqref{eq:observations} that robot $i$ receives after
  executing $\control_{1:l}$ according to \eqref{eq:tracking_dynamics} for each
  $(i, \control_{1:l}) \in X$.
  When $X\subseteq\ground$ includes multiple hypothetical assignments to the
  same robot---which arises in the analysis---duplicates
  obtain unique observations and observation noise $\targetnoise$.
}\footnote{%
  All analysis applies to discrete and continuous mutual information and
  transformations of target states e.g. for MI with position but not
  derivatives.
}
\begin{align}
  \setfun(X) = \MI(\targetstates_{t+1:t+l}; \observations_{t+1:t+l}(X)
  |\observations_{0:t},\robotstates_{0:t})
  \label{eq:tracking_objective}
\end{align}
where $\MI(X;Y|Z)$ is the Shannon mutual information between $X$ and $Y$
conditional on $Z$ and quantifies the reduction in uncertainty (entropy) of one
random variable given another.
Although we discuss properties of
entropy and mutual information,
we refer interested readers to~\citet{cover2012} for detail and definitions.
Critically, mutual information objectives are normalized, monotonic, and
submodular (defined in Sect.~\ref{sec:submodularity}) when observations
are conditionally independent of target states~\citep{krause2005uai}
although higher-order monotonicity conditions may not
apply (see discussion in Fig.~\ref{fig:mutual_information}).
Because the robot states are known and deterministic, individual
targets and observations of the same are jointly independent so that
\eqref{eq:tracking_objective} can be written as a sum over the targets
\begin{align}
  \setfun(X) =
  \sum_{j\in\targets}
  \MI(\targetstates_{j,t+1:t+l};
  \observations_{j,t+1:t+l}(X) |\observations_{j,0:t},\robotstates_{0:t})
  \label{eq:objective_sum}
\end{align}
because the mutual information is a difference of
entropies~\citep[Eq. 2.45]{cover2012} which, in turn, decompose as sums over
targets~\citep[Theorem 2.6.6]{cover2012}.

\subsection{Spatial locality}
\label{sec:spatial_locality}

Spatial locality in target tracking problems arises when the capacity to sense a
target decreases with some measure of distance.
The variations in each robot's $i\in\robots$ ability to sense different targets
$j\in\targets$ take the form of channel capacities $\capacity_{i,j}$ from
information theory~\citep[Chapter 7]{cover2012} so that
\begin{align}
  C_{i,j} =
  \max_{x \in \block_i}
  \MI(\targetstates_{j,t+1:t+l};
  \observations_{j,t+1:t+l}(x) |\observations_{j,0:t},\robotstates_{0:t}).
  \label{eq:target_capacities}
\end{align}
This channel capacity is itself an informative planning problem although
designers may apply existing results for channel capacities by relaxing
\eqref{eq:target_capacities} such as given bounded travel distances on the robot
and target.

Uncertainty in target positions is closely related to spatial locality in target
tracking problems.
Even when sensing power falls off quickly with distance, distant robots may
expect to gain small amounts of information about sufficiently uncertain
targets.
Rather than introduce additional machinery to characterize suboptimality in
terms of uncertainty in target positions, we will quantify the effect of spatial
locality in terms of channel capacities \eqref{eq:target_capacities}.

\subsection{Computational model}

A common feature of distributed planning problems is limited
access to information.
The central assumption for our computational model is that each robot $i \in
\robots$ is able to approximate the objective for its own set of actions
$\block_i$.
That is, each robot has access to an approximation of marginal gains
$\approxsetfun_i(x_i | A)$ for each action $x_i \in \block_i$ in its local set
and for prior selections $A \subseteq \ground$.
This may reflect both approximate evaluation of mutual information and
constraints on access to sensor data such as allowing robots to ignore distant
targets.

Robots also have limited access to the ground set $\ground$: they produce
elements of their local sets $\block_i\subseteq\ground$ implicitly via local
planners and obtain access to other robots' actions when they communicate their
decisions.

\section{Distributed planning algorithm}

\begin{algorithm}
  \caption[Distributed  algorithm for receding-horizon target tracking with
  \rsp{}]{%
    Distributed  algorithm for receding-horizon target tracking with Randomized
    Sequential Partitions (\rsp{}) from the perspective of robot $i\in\robots$
    for execution starting at time $t$.%
  }\label{alg:distributed_target_tracking}
  \begin{minipage}{\linewidth}
    \begin{algorithmic}[1]
      \State $\inneighbor \gets \text{\emph{in} neighbors of robot } i$
      \State $\outneighbor \gets \text{\emph{out} neighbors of robot } i$
      \State $\data \gets \text{sensor data (or summary), accessible to
      robot } i$
      \vskip 1.5ex
      \State \method{Receive}: $X^\mathrm{d}_{\inneighbor}$ from $\inneighbor$
      \label{line:receive}
      \State
      $\approxsetfun \gets \text{approximation of }
      \setfun \text{ given } \data$
      \State $x^\mathrm{d} \gets
      \method{PlanAnytime}(\approxsetfun,
                           \robotstate_{t},
                           X^\mathrm{d}_{\inneighbor})$
      \label{line:anytime}
      \State \method{Send}: $x^\mathrm{d}$ to $\outneighbor$
      \label{line:send}
      \State \method{Execute}: $x^\mathrm{d}$ starting at time $t$ and until
      the beginning of the result of the next planning round
      \label{line:execute}
    \end{algorithmic}
  \end{minipage}
\end{algorithm}

The distributed planner that we present in this section extends methods from
Chapter~\ref{chapter:scalable_multi-agent_coverage}.
Algorithm~\ref{alg:distributed_target_tracking} provides pseudo-code for this
distributed
\rsp{} algorithm based on a directed acyclic graph where robots are
vertices with incoming edges $\inneighbor$ (and outgoing $\outneighbor$).
Although we do not emphasize timing, this algorithm runs in synchronous epochs
whereby the robots collectively maximize information gain to collectively solve
instances of the joint receding-horizon optimization problems
\eqref{eq:optimization_problem}.
The output of this planner is then the collection of the robots'
decisions $X^\mathrm{d} = \{x^\mathrm{d}_1,\ldots,x^\mathrm{d}_{\numrobots}$\}.

Each robot begins planning after receiving decisions from a set of in-neighbors
and approximates the objective based on available data ($\data$) and
computational resources
(lines~\ref{line:receive}--\ref{line:anytime}).
Once the planner exits or runs out of time, the robot commits to an action
which it sends to any out-neighbors and then executes in a receding-horizon
fashion (lines~\ref{line:send}--\ref{line:execute}).
In this exposition, we assume the in- and out-neighbors are known for brevity.
However, the planner graph can also be implicit based on the messages the robot
has received when it starts planning.
Through this process, the single-robot planners (\method{PlanAnytime})
collectively run in $\numrounds$ sequential steps.
We assume robots are randomly assigned to planning steps via the \rsp{} methods
presented in Chapter~\ref{chapter:scalable_multi-agent_coverage}.

\section{Cost model for approximate distributed planning}

Aside from applying \rsp{} to receding-horizon optimization,
we seek to account for approximations
(sampling based planning and objective evaluation and ignoring distant
targets)
pursuant to constraints on computation
time and information access.
Specifically, the analysis (Sect.~\ref{sec:analysis}) will account for costs of
approximations arising in \emph{distributed planning}, \emph{objective
evaluation}, and \emph{anytime (single-robot) planning}.
These costs will then relate the suboptimality of
Alg.~\ref{alg:distributed_target_tracking} to the nominal bound for sequential
planning.

\subsection{Generalized cost of suboptimal decisions for individual robots}

Before moving on to the individual costs, let us present a generalized
expression of costs for approximations in decision making.
Any of the approximations we encounter can inhibit exact evaluation of
maximization steps (\method{PlanAnytime}) in the distributed planner
(Alg.~\ref{alg:local_greedy}).
Rather than assume constant factor suboptimality at each
step~\citep{singh2009}---as would be appropriate if the local planner provided a
consistent performance guarantee---we present a more flexible cost model that
accounts for uncertain results that arise when planning in real time.
Given an instance of \eqref{eq:optimization_problem} and a distributed
solution $X^\mathrm{d}$, the cost to robot $i\in\robots$ for making a suboptimal
decision $x^\mathrm{d}_i$ is the difference between the value of that decision
and the true maximum over $\block_i$
\begin{align}
  \cost_i(\setfun, X) &=
  \max_{x \in \block_i} \setfun(x|X)
  - \setfun(x^\mathrm{d}_i|X)
  \label{eq:single_robot_suboptimality}
\end{align}
for the true objective $\setfun$ marginally specific prior decisions
$X \subseteq X^\mathrm{d}_{1:i-1}$.

\subsection{Cost of distributed planning on a directed acyclic graph}
\label{sec:graph_model}

In the nominal sequential greedy algorithm (Alg.~\ref{alg:local_greedy}),
robot $i\in\robots$ plans with access to all previous decisions by robots
$\{1,\ldots,i-1\}$.
However, in our approach (Alg.~\ref{alg:distributed_target_tracking}) robots
only have access to
a subset $\neighbor_i \subseteq \{1,\ldots,i-1\}$ of these decisions
($\inneighbor$ in the algorithm description), which induces a directed acyclic
graph with edges $(j,i)$ for each robot $j\in\neighbor_i$ whose decision $i$
uses while planning~\citep{gharesifard2017,grimsman2018tcns}.
In a sense, the robots ignore decisions by the remaining robots
$\ignore_i=\{1,\ldots,i-1\}\setminus\neighbor_i$, and the cost of doing so is
a second derivative~\eqref{eq:recursive_derivative}:
\begin{align}
  \distcost_i &=
  \setfun(x^\mathrm{d}_i|X^\mathrm{d}_{\neighbor_i})
  - \setfun(x^\mathrm{d}_i|X^\mathrm{d}_{1:i-1})
  =
  - \setfun(x^\mathrm{d}_i;X^\mathrm{d}_{\ignore_i}|X^\mathrm{d}_{\neighbor_i}).
  \label{eq:distributed_planning_cost}
\end{align}
Later, we will upper bound the right-hand-side in terms of $\ignore_i$.

\subsection{Cost of approximate evaluation of the objective}

Although we do not focus on the communication and representation of sensor data,
we assume robots have access to relevant data for exact or approximate
evaluation of the mutual information objective~\eqref{eq:tracking_objective}.
However, robots may ignore distant targets or approximate the objective via
sampling.
We say robot $i\in\robots$ has access to a local approximation
$\approxsetfun_i$ of the objective, and the cost of this approximation (treating
stochasticity implicitly) is at most the sum of the maximum over- and
under-approximation of $\setfun$
\begin{align}
  \begin{split}
    \objectivecost_i =
    \max_{x_1,x_2 \in \block_i}
    \left(
    \approxsetfun_i(x_1|X_{\neighbor_i}) -
    \setfun(x_1|X_{\neighbor_i})
    +
    \setfun(x_2|X_{\neighbor_i}) -
    \approxsetfun_i(x_2|X_{\neighbor_i})
    \right).
  \end{split}
  \label{eq:objective_cost}
\end{align}
Here, $x_1$ and $x_2$ are respectively the points where $\approxsetfun_i$
most under- and over-approximates $\setfun$ over
all decisions $\block_i$ available to robot $i$.

\subsection{Cost of approximate (anytime) single-robot planning}

Selecting sensing actions for individual robots produces informative path
planning problems~\citep{chekuri2005,singh2009}.
Each robot has a limited amount of time available for planning and must
terminate planning and transmit results soon enough so that later robots
can make their own decisions before the plans go into effect (at time $t$ in
Algorithm~\ref{alg:distributed_target_tracking}).

Although some existing methods provide performance
guarantees~\citep{chekuri2005,singh2009,zhang2016aaai},
designers who apply these methods may have to vary replanning rates
or tune problem parameters to satisfy constraints on planning time for operation
in real-time.
On the other hand, methods such as
randomized planning~\citep{lauri2015ras,hollinger2014ijrr} and gradient- and
Newton-based trajectory generation~\citep{charrow2015rss,indelman2014icra}
converge to local or global maxima but provide no specific guarantees on
solution quality before convergence for anytime planning.
Along these lines, we will now provide analysis based on empirical performance
and will apply Monte-Carlo tree
search~\citep{browne2012,chaslot2010} later for single-robot planning in the
simulation results as in
Chapter~\ref{chapter:distributed_multi-robot_exploration}.

The cost of approximate single-robot planning in terms of empirical performance
is then
\begin{align}
  \plannercost_i &= \cost_i(\approxsetfun_i, X^\mathrm{d}_{\neighbor_i}).
  \label{eq:planner_cost}
\end{align}
This approach captures this inherent uncertainty of anytime planning and enables
us to characterize collective performance in terms of the bulk suboptimality of
single-robot planning.

\section{Analysis of suboptimality of distributed planning}
\label{sec:analysis}

The distributed planner described in Alg.~\ref{alg:distributed_target_tracking}
achieves a
performance bound that approaches that of sequential
planning (Alg.~\ref{alg:local_greedy}) with additional suboptimality that arises
from evaluating the objective, planning for individual robots, and distributed
coordination of the team.

\begin{theorem}[Suboptimality of Alg~\ref{alg:distributed_target_tracking} for
  target tracking tasks]
  \label{theorem:sum_submodular}
  Consider an instance of \eqref{eq:optimization_problem}, any solution $X^d$
  that Alg.~\ref{alg:distributed_target_tracking} produces satisfies
  \begin{align}
    \setfun(X^\opt)
    &\leq
    2\setfun(X^d)
    +
    \sum_{i\in\robots}
    \left(
      \distcost_i
      + \objectivecost_i
      + \plannercost_i
    \right),
    \label{eq:costs_bound}
  \end{align}
  and the total cost of distributed planning is bounded by
  \begin{align}
    \sum_{i\in\robots} \distcost_i
    \leq \sum_{i\in\robots}\sum_{j\in\ignore_i} \approxweights(i,j)
    \label{eq:weights_bound}
  \end{align}
  where $\approxweights$ is a collection of edge weights which we will define later
  that describes redundancies that arise when pairs of robots may observe the
  same targets.
\end{theorem}

\noindent
The proof of Theorem~\ref{theorem:sum_submodular} is in
Appendix~\ref{appendix:proof_of_sum_submodular}, and
we provide a brief summary at the end of this section in
Sect.~\ref{sec:summary_of_proof_of_sum_submodular} after introducing some
preliminary results related to the first
(Sect.~\ref{sec:general_suboptimality_analysis}) and second
(Sect.~\ref{sec:analysis_of_weights})
parts of this theorem.

Regarding the structure of Theorem~\ref{theorem:sum_submodular} the
bounds describe the performance of practical implementations of
Alg.~\ref{alg:distributed_target_tracking}.
An idealized version of this distributed algorithm would have access to exact
objective values and maxima so that the associated costs $\objectivecost$ and
$\plannercost$ would all be zero.
From this perspective, \eqref{eq:costs_bound} describes how real implementations
may deviate from this ideal and states that the total suboptimality is a simple
accumulation of individual inefficiencies.

\subsection{General suboptimality in multi-robot planning}
\label{sec:general_suboptimality_analysis}

The following lemma expresses the suboptimality of any decision as a sum of
costs of suboptimal decisions.

\begin{lemma}[Suboptimality of general assignments]
  \label{lemma:approximate_suboptimality}
  Given a submodular, monotonic, normalized objective $\setfun$,
  \emph{any} basis (assignment of actions to all robots)
  $X^\mathrm{d}\in\independence$ on a simple partition matroid satisfies
  \begin{align}
    \setfun(X^\opt) &\leq    2\setfun(X^\mathrm{d})
    + \sum_{i=1}^{\numrobots} \cost_i(\setfun, X^\mathrm{d}_{1:i-1}).
    \label{eq:approximate_suboptimality}
  \end{align}
\end{lemma}

\noindent
The proof of Lemma~\ref{lemma:approximate_suboptimality} is in
Appendix~\ref{appendix:approximate_suboptimality}.

Observe that if we obtain $X^\mathrm{d}$ via exact sequential maximization, the
terms ($\cost$) of the sum in \eqref{eq:approximate_suboptimality} go to zero,
and we obtain the original result by \citet{fisher1978}.
Similarly, \cite[Theorem~1]{singh2009} on constant factor suboptimal solvers
follows after substituting the suboptimality into the cost
model~\eqref{eq:single_robot_suboptimality}.

\subsection{Bounding the cost of distributed planning for target tracking
problems}
\label{sec:analysis_of_weights}

This section provides tools for characterizing the cost of distributed planning
\eqref{eq:weights_bound} in target tracking problems.
We begin by discussing decomposition of the objective as a sum over targets and
derive a bound in terms of the result.
Applying this bound produces a collection of weights that relate the cost of
distributed planning to the channel capacities \eqref{eq:target_capacities}
between the robots and targets.

\subsubsection{Decomposing objectives as sums}
The objectives of sensing problems can often be written as sums is true for our
target tracking objective \eqref{eq:tracking_objective} which is a sum over information
sources \eqref{eq:objective_sum}.
Specifically, let $\setfuns = \{\setfun_1,\ldots,\setfun_\numtargets\}$
be a collection of set functions so that for $j\in\targets$ and
$X\subseteq\ground$ then
\begin{align}
  \setfun_j(X) =
  \MI(\targetstates_{j,t+1:t+l};
  \observations_{j,t+1:t+l}(X) |\observations_{j,0:t},\robotstates_{0:t}).
  \label{eq:target_sum_decomposition}
\end{align}
This decomposition will enable characterization of interactions between
distant robots in terms of each robot's capacity to sense near and distant
targets.
\begin{definition}[Sum decomposition]
  \label{def:sum_submodular}
  A set of submodular, monotonic, and normalized functions
  $\setfuns=\{\setfun_1,\ldots,\setfun_n\}$
  decomposes a set function $\setfun$ if
  \begin{align}
    \setfun(X) = \sum_{\hat \setfun \in \setfuns} \hat \setfun(X),
    \quad
    \text{for all } X\subseteq\ground.
    \label{eq:sum_submodular}
  \end{align}
\end{definition}

Closure over sums~\citep{foldes2005} ensures that $\setfun$ is
submodular, monotonic, and normalized if the same is true for each
$\hat\setfun\in\setfuns$ as in Def.~\ref{def:sum_submodular}.
Further, although some such sum decomposition always exists
($\setfuns=\{\setfun\}$), the choice of decomposition will determine the
tightness of our performance bounds; here, we are interested in decompositions
that express how interactions vary with distance.
Choosing $\setfuns$ according to \eqref{eq:target_sum_decomposition}
captures spatial locality arising out of distributions of robots and targets.

\subsubsection{Derivatives and the sum decomposition}
Given some $\setfuns$ that decomposes $\setfun$, the second
derivative~\eqref{eq:recursive_derivative} at $X\subseteq\ground$ with respect to
$A,B\subseteq\ground$, all disjoint, is
\begin{align}
  \setfun(A;B|X) = \sum_{\hat \setfun \in \setfuns} \hat \setfun(A;B|X)
  \label{eq:sum_derivative}
\end{align}
and likewise for all other derivatives which are linear combinations of values
of $\setfun$ at different points.

This derivative has the same form as the cost of ignoring prior decisions during
distributed planning $\distcost$~\eqref{eq:distributed_planning_cost}.
The rest of this section is devoted to obtaining upper bounds
relating each robot's decision to the decisions they ignore ($A$ and $B$)
while eliminating the decisions each robot takes into account ($X$).

\subsubsection{Bounding second derivatives using sum decompositions}
Applying monotonicity and submodularity respectively provides a trivial upper
bound on the second derivative of a set function
\begin{align}
  \setfun(A;B|X) = \setfun(A|B,X) - \setfun(A|X) \geq -\setfun(A|X) \geq
  -\setfun(A)
  \label{eq:incremental_bound}
\end{align}
where $A,B,X\subseteq\ground$ are disjoint subsets of the ground set.
By symmetry
\begin{align}
  \setfun(A;B|X) \geq -\min(\setfun(A),\setfun(B)).
  \label{eq:min_bound}
\end{align}
Then, expressing the second derivative of $\setfun$ in terms of the sum
decomposition~\eqref{eq:sum_derivative} and bounding the derivatives of
$\hat\setfun\in\setfuns$ yields
\begin{align}
  \setfun(A;B|X) \geq \sum_{\hat \setfun\in\setfuns} -\min(\hat \setfun(A),\hat
  \setfun (B)).
  \label{eq:derivative_bound}
\end{align}

\begin{remark}
  Chapter~\ref{chapter:scalable_multi-agent_coverage} relies on $\setfun(A;B|X)$
  increasing monotonically in $X$, and we could make a similar statement here by
  writing the right-hand-side of \eqref{eq:min_bound} in terms of marginal gains
  with respect to $X$.

  More broadly, generalizing \eqref{eq:derivative_bound} to include any relevant
  lower bounds on second derivatives of $\hat \setfun\in\setfuns$
  unifies the results we present here with those for 3-increasing functions
  in Chapter~\ref{chapter:scalable_multi-agent_coverage}.
  This enables development of multi-objective applications such as for robots
  covering an environment while simultaneously localizing objects.
  \label{remark:monotonic_upper_bound}
\end{remark}

\subsubsection{Quantifying inter-robot redundancy}
Bounding the second derivative of $\setfun$ \eqref{eq:derivative_bound}
leads to bounds on interactions between agents.
The upper bounds on these interactions form a weighted undirected graph
$\graph = (\robots,\edges,\weights)$ connecting the robots
with edges $\edges=\{(i,j)|i,j\in\robots, i \neq j\}$ and weights
\begin{align}
  \weights(i,j) &=
  \max_{x_i\in\block_i, x_j\in\block_j}
  \sum_{\hat \setfun\in\setfuns} \min(\hat \setfun(x_i),\hat \setfun (x_j)).
  \label{eq:weights}
\end{align}
Evaluating the right-hand-side of the above expression is difficult as doing so
involves search over the product of two robots' action spaces.
To make evaluation of the weights tractable, relaxing this expression by taking
the pairwise minimum of the maximum values of each objective component
produces an upper bound in terms of the channel capacities
\eqref{eq:target_capacities} that avoids search over a product space
\begin{align}
  \begin{split}
    \approxweights(i,j)
    &=
    \sum_{k \in \targets}
    \min\left(
      C_{i,k},
      C_{j,k}
    \right)
    =
    \sum_{\hat \setfun\in\setfuns}
    \min\left(
      \max_{x_i\in\block_i}\hat \setfun(x_i),
      \max_{x_j\in\block_j}\hat \setfun(x_j)
    \right) \\
    &\geq
    \max_{x_i\in\block_i, x_j\in\block_j}
    \sum_{\hat \setfun\in\setfuns} \min(\hat \setfun(x_i),\hat \setfun (x_j))
    =\weights(i,j),
  \end{split}
  \label{eq:weight_by_component}
\end{align}
recalling that in \eqref{eq:target_sum_decomposition} we chose $\setfuns$ to
decompose $\setfun$ by targets $\targets$ so that the second equality follows
from \eqref{eq:target_capacities}.
As the terms of the sum are the smaller channel capacities, this bound captures
the idea that if sensing quality decreases with distance so do interactions
between robots.


\subsection{Summary of the proof of Theorem~\ref{theorem:sum_submodular}}
\label{sec:summary_of_proof_of_sum_submodular}

Theorem~\ref{theorem:sum_submodular} consists of two parts.
The first, the effect of approximations on planning performance
\eqref{eq:costs_bound} follows by applying
Lemma~\ref{lemma:approximate_suboptimality}, on the suboptimality of general
assignments, and substituting the definitions of the costs
(\eqref{eq:distributed_planning_cost}, \eqref{eq:objective_cost}, and
\eqref{eq:planner_cost}).
The second part \eqref{eq:weights_bound} characterizes suboptimality due to
distributed planning and follows by applying the chain rule to the definition of
cost of distributed planning \eqref{eq:distributed_planning_cost} and
substituting \eqref{eq:derivative_bound}, \eqref{eq:weights}, and
\eqref{eq:weight_by_component}.
Please refer to Appendix~\ref{appendix:proof_of_sum_submodular} for the full proof.

\section{Run time and scaling}
\label{sec:scaling}

Algorithm~\ref{alg:distributed_target_tracking} requires a number of
sequential planning steps
that depends on the structure of the planner graph (Sect.~\ref{sec:graph_model})
and a constant number of sequential steps for \rsp{} planners.
Yet, the sequential greedy algorithm (Alg.~\ref{alg:local_greedy}) requires
one step per robot.
Further, when the optimum is proportional to the sum of weights
(ignoring the objective and planner costs) distributed planning \emph{with a
constant number of steps independent of the number of robots} can guarantee
constant-factor suboptimality in expectation, approaching half of
optimal (by extension of Theorem~\ref{theorem:adaptive_partitioning}).

Spatial locality---more specifically ensuring that the robot-target channel
capacities~\eqref{eq:target_capacities} decrease sufficiently quickly with
distance---can enable robots to ignore distant robots and targets with bounded
costs and provides optimization performance independent of the number of robots.
Incorporating range limits in planning by ignoring robots and targets past a
given distance and tracking targets using sparse filters can ensure
that computation time for the single-robot planner is also constant.
Likewise, regarding communication, robots send one message for each edge in the
directed planner graph (Sect.~\ref{sec:graph_model}), and ignoring distant
robots reduces this to a constant number of messages per
robot.
Further, Appendix~\ref{appendix:scaling_analysis} presents sufficient conditions
for the cost of distributed planning
$\distcost$~\eqref{eq:distributed_planning_cost}
to be bounded and so to provide constant optimization performance for any number
of robots.

\section{Results}

\begin{figure}
  \floatbox[{\capbeside\thisfloatsetup{capbesideposition={left,top},%
  capbesidewidth=0.38\linewidth}}]{figure}[0.58\textwidth]
  {%
    \caption[Visualization of target tracking simulation]{%
      Visualizations of eight and sixteen robots tracking same numbers of
      targets.
      Robots with dotted finite-horizon trajectories are blue and
      targets red.
      The background illustrates the sum of target probabilities at each grid
      space, increasing from purple to yellow.
    }\label{fig:target_tracking_visualization}
  }{
     \includegraphics[width=0.49\linewidth]{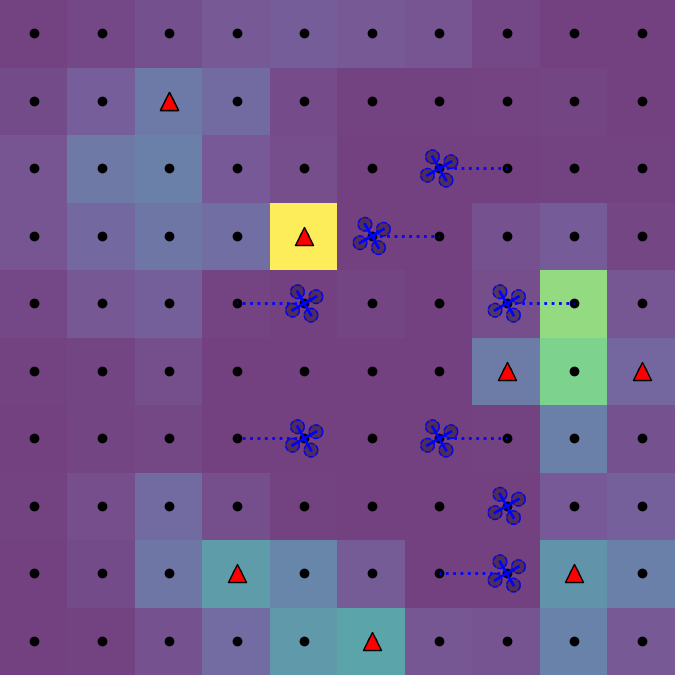}
     \includegraphics[width=0.49\linewidth]{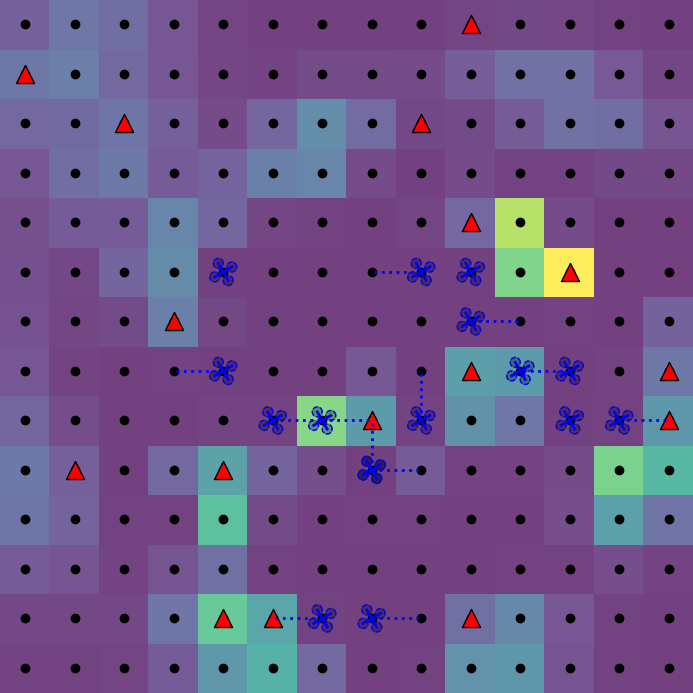}
   }
\end{figure}

To evaluate the approach, we provide simulation results
(visualized in Fig.~\ref{fig:target_tracking_visualization})
for teams of robots tracking targets (one target per robot),
robots moving according to planner output and targets via a random walk,
all on a four-connected grid (with side length $\sqrt{12.5\numrobots}$).
The robots estimate target locations using Bayesian filters given range
observations to each target with mean $\hat d = \min(d, 20)$ and variance
$0.25+0.5 \hat d^2$ where $d$ is the Euclidean distance to the target in grid
cells.
For the purpose of this chapter, all robots have access to all observations or,
equivalently, have access to centralized filters.
All simulation trials run for 100 time-steps; all initial states are uniformly
random; and initial target locations are known.
Additionally, all results ignore the first 20
steps of each trial to allow the system to converge to steady-state conditions.

Robots plan actions individually  using Monte-Carlo tree search
(MCTS, \method{PlanAnytime}) \citep{chaslot2010} with a two step horizon and
collectively according to the specified planner.
To ensure tractability we replace the original objective
\eqref{eq:tracking_objective} with a sum of mutual information for each time-step
\begin{align}
  \setfunsim(X) &=
  \sum_{i=1}^l \MI(\targetstates_{t+i}; \observations_{t+1:t+i}(X)
  |\observations_{0:t},\robotstates_{0:t}),\ X\subseteq\ground.
  \label{eq:simulation_objective}
\end{align}
This objective
is equivalent to \citep[(18)]{ryan2010ras}
and
can be though of as minimizing uncertainty at the time of each
planning step.
Being a sum, \eqref{eq:simulation_objective} remains submodular, monotonic, and
normalized so Theorem~\ref{theorem:sum_submodular} still applies.
Like \citet{ryan2010ras}, we evaluate this objective by simulating the
system and computing the sample mean of the filter entropy.
Because MCTS itself is sample-based, the planner estimates the objective
implicitly by simulating the system once per MCTS rollout; by sampling the more
valuable actions more often, MCTS produces increasingly accurate estimates for
nearly optimal trajectories.
Planners with sixteen or more robots use sparse filters with a threshold of
$1e-3$.

The experiments compare methods for multi-robot coordination including:
sequential planning
(Alg.~\ref{alg:local_greedy}, which has \emph{one step per robot});
the proposed distributed planner (Alg.~\ref{alg:distributed_target_tracking})
with each robot assigned randomly to one of $\numrounds$ sequential steps according to
the \rsp{} methods from Chapter~\ref{chapter:scalable_multi-agent_coverage};
myopic planning (MCTS without coordination, \emph{one step});
and random selection of actions.
Additionally, we provide some results for \emph{distributed planning with range
limits} (\rrsp{})
where robots plan while ignoring targets further than 12 units away (in
terms of mean position) and other robots further than 20 units.
Given the use of sparse filters, this latter planner runs in
\emph{constant time}.

\begin{figure}
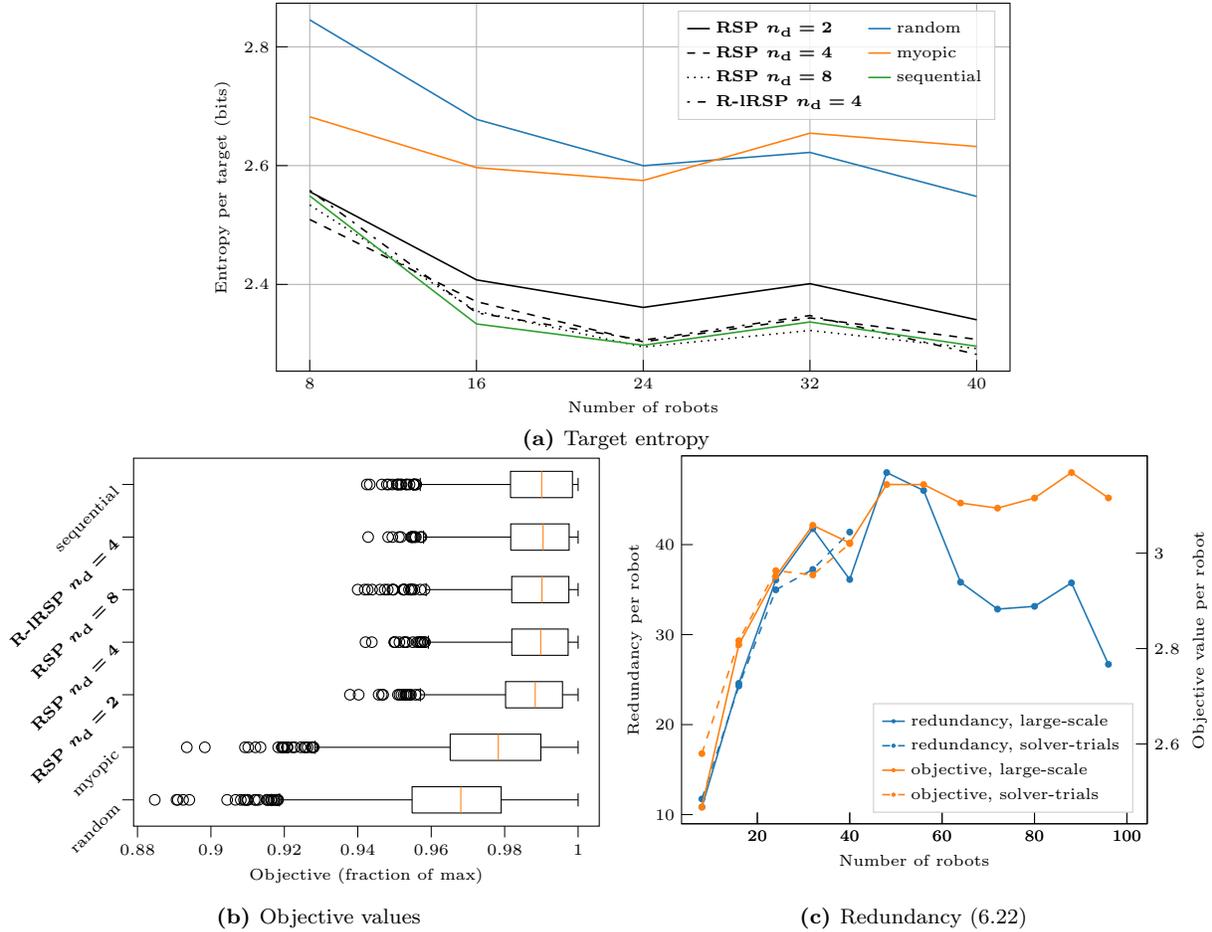

  \setlength{\subfiguresheight}{0.40\linewidth}
  \begin{subfigure}[b]{0.70\linewidth}
    \setlength{\figurewidth}{\linewidth}
    \setlength{\figureheight}{1.0\subfiguresheight}
    \centering
    \tiny
    \inputfigure[entropy_by_solver_plot]{\datapath/entropy_by_solver_plot.tex}
    \vspace{-0.1cm}
    \caption{
      Target entropy
    }\label{subfig:target_entropy}
  \end{subfigure}
  \begin{subfigure}[b]{0.48\linewidth}
    \setlength{\figurewidth}{1.0\linewidth}%
    \setlength{\figureheight}{1.0\subfiguresheight}%
    \centering%
    \tiny%
    \hskip-0.5cm%
    \inputfigure[normalized_solver_objectives]{\datapath/normalized_solver_objectives.tex}%
    \caption{
      Objective values
    }\label{subfig:solver_objectives}
  \end{subfigure}
  \begin{subfigure}[b]{0.48\linewidth}
    \setlength{\figurewidth}{1.0\linewidth}
    \setlength{\figureheight}{1.0\subfiguresheight}
    \centering
    \tiny
    \inputfigure[weights_by_number_of_robots]{\datapath/weights_by_number_of_robots.tex}
    \caption{
      Redundancy~\eqref{eq:weight_by_component}
    }\label{subfig:target_weights}
  \end{subfigure}
  \caption[Target tracking results]{
    (\subref{subfig:target_entropy})
    For target entropy (lower is better) which is the task performance criterion
    distributed planning with \rsp{} consistently improves upon myopic
    planning and approaches sequential planning given many times fewer
    sequential steps.
    (\subref{subfig:solver_objectives})
    Objective values on common 16-robot subproblems reflect a similar trend, and
    results for distributed planning (bold) are effectively equivalent to
    sequential planning given only four planning rounds.
    \eqref{eq:optimization_problem}.
    (\subref{subfig:target_weights})
    Average objective value and total redundancy
    weight~\eqref{eq:weight_by_component} per robot for \rsp{} with
    $\numrounds=4$.
    Traces depict values for trials from (\subref{subfig:solver_objectives})
    and five additional trials with \rrsp{} reaching up to 96 robots.
    Values of each initially increase but appear to approach an asymptote which
    indicates that the distributed planner approaches constant factor
    suboptimality at large scales.
  }\label{fig:performance}
\end{figure}

We evaluate the distributed planning approach in terms of
\emph{task performance} (average target entropy) for various numbers of robots
(Fig.~\ref{fig:performance}),
\emph{objective values} on a common set of subproblems
\eqref{eq:optimization_problem},
and the \emph{redundancy per robot}~\eqref{eq:weight_by_component} for
increasing numbers of robots which is proportional to $1/n$ times the bound on
cost of distributed planning~\eqref{eq:weights_bound} for randomized assignment
in $n$ rounds (see Chapter~\ref{chapter:scalable_multi-agent_coverage}).
The results for average target entropy---which captures the uncertainty in
target locations~\citep{cover2012}---are based on 20 simulations of target
tracking for each configuration.
Results for objective values and redundancy use
planning subproblems~\eqref{eq:optimization_problem} taken from the simulation
trials for distributed planning in four rounds.
The results for objective values by solver are normalized according to the
maximum values across solvers for each planning problem and reflect trials with
16 robots.
For the results on redundancy, an additional five trials for a four-round
distributed planner with range limits demonstrate behavior for up to 96
robots.\footnote{%
  Planning at this scale is intractable for other planner configurations.
}

Proposed distributed planners provide consistent improvements in target tracking
performance (average target entropy) (Fig.~\ref{subfig:target_entropy}) compared
to myopic planning;
distributed planning in eight rounds roughly matches sequential planning despite
requiring as much as five times fewer planning steps and produces 5--13\% better
(lower) target entropy than when planing myopically.
The objective values (Fig.~\ref{subfig:solver_objectives}) exhibit a similar
trend, and all distributed planners closely match sequential planning, even in
terms of the distributions of results.

Although the redundancy per robot (Fig.~\ref{subfig:target_weights}) initially
increases with the number of robots, that redundancy eventually levels off.
This is consistent with the analysis of scaling performance in
Appendix~\ref{appendix:approximate_suboptimality} which indicates that the
suboptimality approaches a constant factor bound.
Overall, the results indicate that a small amount of coordination
that does not scale with the number of robots is sufficient to produce
performance comparable to sequential planning in receding-horizon settings.

\section{Conclusions and future work}

This chapter has presented a distributed planner for mutual information-based
target tracking that mitigates the effects of the sequential structure of
existing methods for submodular maximization.
The analysis provides bounds on suboptimality for distributed planners that can
be designed to run with fixed numbers of planning steps.
By explicitly accounting for suboptimal local planning (e.g. anytime planning)
and approximation of the objective, we affirm that the proposed approach is
applicable to practical tracking systems.
The results demonstrate that distributed planning improves tracking
performance (in terms of target entropy) compared to planners with no
coordination and that distributed planning with little coordination can even
match fully sequential planning given a
constant number of planning rounds.

Although we focus on target tracking, the analysis applies to general
multi-objective sensing problems and generalizes the results in
Chapter~\ref{chapter:scalable_multi-agent_coverage} on coverage.
In this sense, we present a first positive result for mutual
information objectives where the second discrete derivative is only nearly
monotonic.

\chapter{Time-Sensitive Exploration of Unknown Environments}
\setdatapath{./fig/time_sensitive_exploration}
\label{chapter:time_sensitive_sensing}

The line of work that led to this thesis began with multi-robot exploration
of unknown environments
(Chapter~\ref{chapter:distributed_multi-robot_exploration}), and we will now
revisit this topic in greater depth.
Obtaining improvements in performance for exploration---in terms of
completion time---via submodular maximization had proven to be a challenge
as the earlier results exhibited little variation in task performance across
different methods for submodular maximization.
The methods for analysis of redundancy that we
have developed since then also do not immediately apply to mutual information
objectives for exploration
(see discussion in Fig.~\ref{fig:mutual_information}).
We address the former through more thorough development and evaluation of the
exploration system.
Toward the same end, we also consider the design of objectives for robotic
exploration.
In doing so, we also address the theoretic limitations related to mutual
information in order to apply our existing analysis for \rsp{} planning to
multi-robot exploration.

Unlike the target tracking problem that we studied in the previous chapter
(Chapter~\ref{chapter:target_tracking}), mutual information and submodular
maximization on their own do not address a number of key challenges for
multi-robot exploration.
First, environments for exploration do not reflect the common assumption that
cells are occupied with independent probability.
Such cell independence assumptions are central to existing
information-theoretic
objectives~\citep{charrow2015icra,julian2014,zhang2019icra}
which often strongly emphasize efficient
computation~\citep{charrow2015icra,zhang2019icra,li2019rss,henderson2020icra}.
Although these independence assumptions can be limiting, they admit changes
to the occupancy prior for unobserved space
(typically decreasing the prior)%
~\citep{henderson2019thesis,henderson2020icra,tabib2016iros}
(see also Sec.~\ref{sec:kinematic_implementation}).
Still, even though recent works~\citep{zhang2019icra,li2019rss}
establish that Shannon mutual information can be evaluated efficiently
and accurately for an individual range observation along a ray,
methods for evaluating the joint contributions of multiple rays and camera views
depend on approximation by summing over rays~\citep{charrow2015icra}.
Our analysis highlights connections between mutual information and coverage
objectives and suggests that coverage can accurately represent joint
contributions when the prior probability of occupancy is low.
Moreover, our results demonstrate that switching to a coverage objective can
improve completion times by as much as 16\%.

The same connections to coverage support applying \rsp{} planning
to provide efficient, distributed submodular maximization
for exploration with large numbers of robots.
Toward this end, we investigate the design of exploration systems and the
potential impacts of methods for submodular maximization on the time to explore
an environment.
First, we introduce a reward for reducing distance to sufficiently informative
views~\citep{corah2019ral} which ensures reliable completion of exploration
tasks and mimics methods that obtain similar effects via navigation toward
frontiers~\citep{yamauchi1997,charrow2015icra}.
We also provide simulation results for more than a thousand simulation trials
which compare methods for multi-robot coordination via submodular maximization
for different environments and numbers of robots.
The results demonstrate that sequential (greedy) and \rsp{} planning
substantially improve suboptimality for receding-horizon planning for
multi-robot exploration.
However, decreasing time to complete exploration tasks remains a challenge as
the improvements in suboptimality translate primarily into increases in coverage
rates early in the exploration process that are not often sustained through the
end.
Still, we remain optimistic that incorporating methods for planning
and task assignment at larger spatial
scales~\citep{simmons2000aaai,mitchell2019icra}
can realize sustained improvements in exploration performance.

\section{Time-sensitive exploration problem}
\label{sec:exploration_problem}

The following problem description parallels and reiterates
prior discussion in
Sec.~\ref{sec:time_sensitive_sensing} and Sec.~\ref{sec:exploration} which
include additional details on the properties of exploration and related
problems.

Consider a team of robots $\robots=\{1,\ldots,\numrobot\}$ seeking to map a
discretized environment $\environment=[\cell_1,\ldots,\cell_{\numcells}]$
consisting of
cells $\cells = \{1,\ldots,\numcells\}$
that are each either \emph{free} (0) or
\emph{occupied}
(1)
(that is $\cell_i \in \{0,1\}$ for $i\in\cells$)
according to some probability distribution.
Each robot $r\in\robots$ moves through the environment with state
$\state_{t,r}\in\real^3$
at some \emph{discrete} time $t$
according to the following dynamics
\begin{align}
  \state_{t,r} &= \dynamics(\state_{t-1,r}, \control_{t,r})
                     \label{eq:exploration_dynamics}
\end{align}
where $\control_{t,r}\in\controls$ belongs to a finite set of control inputs
$\controls$.
Robots must remain in free space which induces the constraint that they
remain in a safe set
\begin{align}
  \state_{t,r} \in \safespace(\environment).
  \label{eq:safe_space}
\end{align}
The robots observe the environment with depth cameras
(or other ray-based sensors such as lidar sensors).
For the purpose of this chapter we assume that measurements are deterministic,
without noise;
robots can infer definitely that cells in the path of each ray are free up to
the first occupied cell or the maximum range along the beam.
For brevity, we abstract this process of inference and state that, at each
time-step, robots observe sets of cells
$\camera(\state, \environment) \subseteq \cells$
with depth cameras
and obtain their occupancy values via observations
\begin{align}
  \observation_{t,r} &=
  \observationfunction(\state_{t,r}, \environment) =
  \{(i, \cell_i) : i \in \camera(\state, \environment)\}.
  \label{eq:exploration_observation}
\end{align}

Through the process of exploration, robots seek to maximize the number
of cells observed
\begin{align}
  \sensingquality(\states_{1:t}, \observations_{1:t}) =
  \left|
  \bigcup\nolimits_{t'\in\{1,\ldots,t\},r\in\robots}
  \camera(\state_{t',r}, \environment)
  \right|.
  \label{eq:environment_coverage}
\end{align}
We refer to this quantity as the \emph{environment coverage}.\footnote{%
  Previously, (in Chapter~\ref{chapter:distributed_multi-robot_exploration})
  we also used entropy to evaluate exploration performance.
  Appendix~\ref{appendix:entropy_vs_coverage} explains the reasoning behind this
  change.
}
Exploration is complete once the environment coverage reaches an
environment-dependent quota $\quota(\environment)$ when
\begin{align}
  \sensingquality(\states_{1:T},\observations_{1:T})
  \geq \quota(\environment),
\end{align}
and the robots
seek to minimize the amount of time $T$ required to reach that quota and
complete the exploration process.

\section{Planning for exploration}

Like previous chapters, we propose planning with a receding-horizon approach
whereby robots collectively maximize an objective $\setfun$ over a horizon with
$L$ steps
\begin{align}
  \max_{\control'_{1:L,1:\numrobot}}~& \setfun(\states_{t:t+L,1:\numrobot})
  \nonumber \\
  \mathrm{s.t.}~
              & \state_{t+l,r} \in \safespace(\observations_{1:t})
  \nonumber \\
              & \state_{t+l,r} = \dynamics(\state_{t+l,r}, \control'_{l,r})
  \nonumber \\
              & \observation_{t+l,r} = \observationfunction(\state_{t+l,r},
                                                            \environment)
  \nonumber \\
              &\mbox{for all } l \in \{1\ldots L\} \mbox{ and }
              r \in \robots,
              \label{eq:receding_horizon_exploration}
\end{align}
where $\safespace(\observations_{1:t})$ refers to the subset of the state space
that is \emph{known} to be safe given available observations
(unlike \eqref{eq:safe_space} which refers to the complete set of safe states).
After solving \eqref{eq:receding_horizon_exploration},
robots then execute the first control actions in the sequence
$\control'_{1,1:\numrobot}$.

Likewise, we can interpret $\setfun$ as a submodular, monotonic, and normalized
set function and can rewrite \eqref{eq:receding_horizon_exploration}
as a submodular maximization problem with a simple partition matroid constraint
(Problem~\ref{prob:submodular_partition_matroid}).
From this perspective, the ground set consists of assignments of control actions
to robots $\ground = \robots \times \controls^L$,
and the blocks of the partition matroid (Def.~\ref{def:partition_matroid})
are assignments $\{r\} \times \controls^L$ to robots $r \in \robots$.

We propose an objective that consists of two components $\viewvalue$ and
$\distancevalue$ so that, given some assignment of actions $X \subseteq
\ground$, then
\begin{align}
  \setfun(X) &=
  \viewvalue(X)
  +
  \sum_{r \in \robots}
  \distancevalue(X_r),
  \label{eq:exploration_objective}
\end{align}
where $X_r$ is the assignment to robot $r\in\robots$.
The \emph{view} (or information) reward seeks to capture the value of
observations from camera views over the planning horizon while the
\emph{distance} component provides reward for moving toward regions of the
environment (beyond the planning horizon) where valuable observations can be
obtained.

Note that $\setfun$ is normalized ($\setfun(\emptyset)=0$) will retain the
monotonicity properties of $\viewvalue$ so long as
$\distancevalue$ is positive and zero when trajectories are not assigned
because the distance terms are additive.\footnote{%
  Being additive, marginal gains for the distance reward are fixed and do not
  depend on other robots.
  As such, aside from increasing monotonically, \emph{all} higher-order
  monotonicity conditions hold because \emph{all} second and higher-order
  derivatives \eqref{eq:recursive_derivative} are zero.
}

\section{Spatially local volumetric reward}

The volumetric reward $\viewvalue$ seeks to capture the joint value of
observations (camera views) in terms of the amount of unknown space that the
robots will collectively observe.
Ideally, $\viewvalue$ would correspond directly to the increment in environment
coverage \eqref{eq:environment_coverage}.
Except, the environment $\environment$ and, in turn, future values of the
environment coverage are unknown.

To mitigate this issue, prior works predominantly either compute rewards
given an assumed~\citep{%
simmons2000aaai,butzke2011planning,delmerico2018auro,bircher2018receding}
(or possibly learned)
guess at the
environment instantiation
or else approximate information gain
(i.e. mutual information~\eqref{eq:mutual_information})
for observations given a Bayesian prior%
~\citep{charrow2015icra,charrow2015rss,julian2014,henderson2020icra}
(Chapter~\ref{chapter:distributed_multi-robot_exploration} applies such methods)
so far \emph{always} assuming
independent (Bernoulli) cell occupancy.

As well as presenting the approach for this chapter,
the following discussion unifies and contrasts different kinds of
exploration rewards and demonstrates that our analysis for \rsp{} planning
applies to a variety of volumetric rewards.

\subsection{Expected coverage}
\label{sec:expected_coverage_exploration}

We will connect different exploration rewards in terms of expected coverage.
Consider some possible distribution over possible environments
$\environmentsguess$.
The only restriction is that
any environment that $\environmentsguess$ assigns non-zero probability
$E'\sim\environmentsguess$ must be consistent with observations up to the
current time $\observations_{1:t}$
($\environmentsguess$ may even assign zero probability to all but one possible
environment).
Now, for convenience define the set of \emph{future} states that robots will
visit while executing a set of actions $X\subseteq\ground$ as
$\actionstostates(X)$.
Given non-negative weights on cells
$\cellweight : \cells \rightarrow \real_{>= 0}$
the expected weighted coverage is
\begin{align}
  \expectedcoverage(X) &=
  \E_{\environment'\sim\environmentsguess}\left[
    \sum_{i \in \covercells(X,\environment')} \cellweight(i)
  \right]
  \label{eq:expected_coverage}
\end{align}
where $\covercells$ is the hypothetical set of cells that the robots may
observe:
\begin{align}
  \covercells(X,\environment')=
  \bigcup\nolimits_{\state\in\Phi(X)}\camera(\state,\environment').
  \label{eq:covercells}
\end{align}
An obvious weighting scheme, which we apply often in this chapter,
is to provide a uniform (fixed) reward for each newly observed cells
\begin{align}
  \newcellweight(i) &=
  \begin{cases}
    0 &
      i \in
      \bigcup\nolimits_{t'\in\{1,\ldots,t\},r\in\robots}
      \camera(\state_{t',r}, \environment)
    \\
    1 & \text{otherwise (for newly observed cells)}
  \end{cases}.
  \label{eq:uniform_weighting}
\end{align}
However, other weighting schemes are also relevant.
For example, \citet{yoder2016fsr} propose a system for inspecting surfaces.
Their surface frontier approach would be similar in spirit to a scheme that
provides increased weight to unobserved cells that are near known occupied
cells.
Later, we will see that weighting cells by entropy produces a mutual information
objective.\footnote{
  If not for mutual information, we would define the weighted expected coverage
  \eqref{eq:expected_coverage}, concisely, in terms of unobserved cells and the
  weight \eqref{eq:uniform_weighting} as a constant for all cells.
  However, probabilistic occupancy does not lend itself to explicit distinctions
  between known and unknown values.
}

Now let us discuss the properties of the expected coverage and the relationship
to mutual information.

\subsubsection{Expected coverage retains monotonicity properties}
\label{sec:expected_coverage_monotonicity}

The expected coverage \eqref{eq:expected_coverage} is normalized and satisfies
alternating monotonicity conditions
(including being monotonic, submodular, and 3-increasing).
This follows because these conditions hold for weighted coverage
(Theorem~\ref{theorem:sensor_failure})\footnote{%
  In fact, the same argument is valid for Theorem~\ref{theorem:sensor_failure}
  which describes the probabilistic coverage objective.
}
and because the expectation forms a convex combination which preserves
monotonicity conditions\footnote{
  Moreover, the expectation is itself weighted coverage, covering a possibly
  exponentially larger set.
  This is evident by observing that the sum over weighted coverage functions is
  equivalent to duplicating the set being covered for each summand and adjusting
  weights according to probability.
}(and being normalized)~\citep{foldes2005}.

Likewise, all the analysis for \rsp{} planning in
Chapter~\ref{chapter:scalable_multi-agent_coverage} applies to receding-horizon
planning for exploration with expected coverage.\footnote{%
  Note that none of the complications that arise in
  Chapter~\ref{chapter:target_tracking} apply to this chapter
  either for the 3-increasing condition or for the behavior of
  bounds on suboptimality for large numbers of robots, the latter because
  robots have clearly defined sensor ranges and are not observing objects with
  uncertain positions.
}

Note that the expected coverage is not \emph{necessarily} adaptive
submodular~\citep{golovin2011jair}.
See Appendix~\ref{appendix:not_adaptive_submodular} for more detail.

\subsubsection{Well posed volumetric rewards}
\label{sec:exploration_reward_sanity}

The previous section described how expected coverage has favorable properties
for solving optimization problems.
We now comment briefly on a property related to the exploration process.
Specifically, the volumetric reward $\viewvalue$ should be non-zero if and only
if the actual environment coverage $\eqref{eq:environment_coverage}$
(given the true environment $\environment$)
will increase after taking an observation.
As such the robot will never be rewarded by $\expectedcoverage$ for visiting a
state where it will not observe any unknown cells or obtain zero reward when it
will observe unknown cells.

A condition for this to hold for depth sensors is for the weight
$\cellweight$ to be zero if and only if a cell $i \in \cells$ has already been
observed
($i \in \bigcup\nolimits_{t'\in\{1,\ldots,t\},r\in\robots}
\camera(\state_{t',r}, \environment)$).
Note that the uniform weighting scheme \eqref{eq:uniform_weighting} satisfies
this requirement by construction.

This sanity condition follows, \emph{specifically for depth sensors}, because
each ray either:
\begin{enumerate}
  \item Terminates
    (or reaches the maximum range) in known space for all environments that
    are consistent with prior observations, or
  \item Terminates in unknown space, providing the occupancy value of at least
    one cell (even if the first unknown cell in the path of the ray is
    occupied).
\end{enumerate}
This observation is intuitive and not particularly interesting on its own.
However, the consequence is that seemingly pathological assumptions,
such as assuming that all unknown cells are occupied, can still produce sane
rewards.\footnote{
  This does not exclude the limiting case for weights $\cellweight$
  approaching zero which will occur for mutual information objectives if an
  unobserved cell is marked as free or occupied with probability approaching
  one.
}

\subsection{Noiseless mutual information for depth sensors}
\label{sec:mutual_information_exploration}

Most existing works on mutual information for occupancy mapping assume noisy
measurements via either a simplified (often Gaussian)
~\citep{charrow2015icra,zhang2019icra} or more general~\citep{julian2014} noise
model.
However, \citet{zhang2019icra} find that the choice of information
metric and noise model has little impact on performance in exploration
experiments.
Likewise, \citet{henderson2020icra} observe that the sensor
noise for modern lidar sensors and depth cameras is typically small compared to
the maximum range.
For these reasons, we assume that sensor noise is negligible for the purpose of
evaluating mutual information for
exploration\footnote{%
  Alternatively, note that sensor noise is likely more relevant to perception
  tasks such as surface reconstruction.
}
and ignore sensor noise in this chapter.
Additionally, prior works on mutual information for mapping%
~\citep{charrow2015icra,julian2014,zhang2019icra}
typically assume
cell occupancy is independent according to the prior $\environmentsguess$ on the
environment.

The combination of cell independence and lack of sensor noise produces a special
case of an expected coverage objective:

\begin{theorem}[Noiseless mutual information with independent cells is
  3-increasing]
  \label{theorem:mapping_information_is_coverage}
  The mutual information
  $\MI(\environment; \observations(X))$
  between an environment $\environment$
  with uncertain occupancy and hypothetical future observations
  $\observations(X)$
  can be written as
  \begin{align}
    \MI(\environment; \observations(X)) &=
    \E_{\environment' \sim \environmentsguess}
    \left[
      \sum\nolimits_{i \in \covercells(X, \environment')} \H(\cell_i)
    \right].
    \label{eq:mapping_information_independent}
  \end{align}
  \eqref{eq:expected_coverage} given that:
  \begin{enumerate}
    \item Cell occupancy probabilities $\environmentsguess$ are independent, and
    \item There is no sensor noise.
  \end{enumerate}
  This expression \eqref{eq:mapping_information_independent}
  has the form of expected weighted coverage
  \eqref{eq:expected_coverage}
  and is therefore 3-increasing.
\end{theorem}

The proof is included in
Appendix~\ref{appendix:proof_that_mapping_information_is_coverage}.
Note that Theorem~\ref{theorem:mapping_information_is_coverage} implies
that our results for $\rsp$ planning for 3-increasing functions apply to noiseless
mutual information just as for expected coverage.
This holds even though mutual information is not 3-increasing in general (see
Fig.~\ref{fig:mutual_information}).

Whether the sanity condition that Sec.~\ref{sec:exploration_reward_sanity}
describes applies to mutual information depends on the updates to cell occupancy
probabilities as
observed cells would have to be marked as occupied with probabilities of either
zero or one to ensure zero entropy and reward.
However, this can be attributed to pedagogy as this chapter puts relatively
limited weight on probabilistic models.
Instead, we note that \citet[Theorem~2.5]{julian2014} proves similar properties
for mutual information objectives in general.

\subsubsection{Limitations of existing approximations for computing mutual
information}
\label{sec:mutual_information_limitations}

Our presentation of mutual information
\eqref{eq:mapping_information_independent}
(and also expected coverage \eqref{eq:expected_coverage})
does not yet address computational challenges.
While either could be evaluated via sampling, doing so would significantly
increase computational costs for systems that need to react quickly to operate
effectively~\citep{goel2019fsr,dharmadhikari2020icra}.
For this reason, many works on information-based
exploration~\citep{charrow2015icra,zhang2019icra,li2019rss,henderson2020icra}
emphasize computational contributions and approximate evaluation.
However, relatively little is know about how design decisions and approximations
can affect decision-making and exploration performance.

Moreover, Chapter~\ref{chapter:distributed_multi-robot_exploration}
applied a CSQMI objective~\cite{charrow2015icra} while this chapter will apply a
coverage-based objective instead.
Our results for a small study indicate that this choice significantly improved
exploration performance
(by as much as 16\%).
However, we caution that we do not intend to establish conclusively that one
objective is better than another and note that the optimistic coverage objective
we compare to is itself a limiting case for mutual information objectives
(see Sec.~\ref{sec:occupancy_priors}).
We believe that these connections between mutual information and coverage are
useful for evaluating and improving mutual information objectives, and we
highlight some of these points and provide additional results in
Appendix~\ref{appendix:csqmi_and_coverage}.

\subsection{Remarks on specializing exploration objectives}
\label{sec:objectives_remarks}

Let us now expand on the prior two sections
(Sec.~\ref{sec:expected_coverage_exploration} and
Sec.~\ref{sec:mutual_information_exploration})
which define volumetric exploration objectives and their
properties and discuss the ramifications of these observations for objective
design.

\subsubsection{Optimistic coverage}
\label{sec:optimistic_coverage}

For the purpose of this chapter, we select a degenerate prior over environments
by optimistically assuming that unobserved space is empty along with the uniform
weighting scheme \eqref{eq:uniform_weighting}.
This choice is similar in effect to numerous other works on
exploration~\citep{simmons2000aaai,butzke2011planning}
and is compatible with our definition and discussion of expected weighted
coverage, despite failing to produce a meaningful distribution.
Moreover, accurate evaluation of optimistic coverage objective is trivial, even
for joint observations by teams of robots.\footnote{
  While writing this thesis, optimistic coverage served an
  important role to characterize the performance of submodular maximization
  for exploration.
  Because the objective values for optimistic coverage
  are not approximate and provide oracle rewards in some cases,
  we can better attribute potential causes for deficiencies in exploration
  performance.
}

\subsubsection{Occupancy priors for mutual information and optimism}
\label{sec:occupancy_priors}

Mapping applications and even numerous papers on
exploration~\citep{julian2014,charrow2015icra,zhang2019icra}
frequently assume that unobserved cells are independent and occupied with a
probability of 0.5.
Given a prior on occupancy of $p$, a beam will terminate after
traversing $1/p$ cells which works out to \emph{two} for a prior of 0.5.
However, the environments that robots explore may, more predominantly, consist
of open space.
For this reason, Chapter~\ref{chapter:distributed_multi-robot_exploration}
and our prior work~\citep{tabib2016iros} both select priors with relatively
lower occupancy probability.\footnote{%
  Inspection reveals that some of this work~\citep{tabib2016iros}
  did not address the topic of priors in as much detail as remembered.
}
Similarly, \citet{henderson2019thesis} provides detailed
discussion and illustrations that demonstrate how the occupancy prior can
affect decisions.

In the limit, a prior with low occupancy probability assumes that all unobserved
cells are unoccupied like the optimistic coverage objective.
In fact, \emph{optimistic coverage is equivalent to mutual information}
in this case after applying a scaling factor to normalize entropies of the
unknown cells in
\eqref{eq:mapping_information_independent}
(and assuming entropies of observed cells are zero).
This is a useful observation, because it provides a special case of the mutual
information that can be evaluated exactly and efficiently.

\subsubsection{Learned models and oracle objectives}
\label{sec:oracle_objectives}

Another useful property is for the objective to provide rewards for observations
given access to the environment $\environment$ (i.e. ground truth)
and thereby the true increment in the environment coverage
\eqref{eq:environment_coverage}.
Comparing to an oracle is a useful tool for characterizing
limits on performance as having an oracle provides the planner with
additional information about the environment.\footnote{
  Note that we still constrain the robot to operating in the known safe set
  $\safespace(\observations_{1:t})$ despite having access to oracle rewards.
}%
\footnote{
  An optimal policy for time-sensitive exploration
  (Sec.~\ref{sec:exploration_problem})
  would obtain strictly better performance with access to an oracle.
  Given that we do not have access to an optimal policy,
  actual results may vary.
}
Prior works on problems related to exploration sometimes apply oracles
similarly~\citep{saroya2020iros} while \citet{choudhury2018ijrr} use such
oracles in a learning process to train an exploration policy.

Notably, \emph{the optimistic coverage objective is an oracle for empty
environments},
and we provide results for one such environment later in this chapter
(Fig.~\ref{tab:environments}).\footnote{%
  Our implementation of the optimistic coverage objective includes access to
  the bounding box for exploration so it is a true oracle for our Empty
  environment.
}

Likewise, a learned model may emulate an
oracle~\citep{wang2019ral,saroya2020iros} and serve in a similar role as part of
an expected coverage objective
(with a degenerate prior)
or a mutual information objective
(while introducing a small amount of uncertainty).
Our results where the objective is an oracle also provide insight into how well
an exploration system with a learned model could perform.

\subsubsection{General distributions over environments}

Finally, the prior $\environmentsguess$ might also encode a distribution over
environments such as produced by a generative model where cells are not
independent.
\citet{choudhury2018ijrr} study expected coverage in a similar setting.

For such general priors, \emph{mutual information and expected coverage may
differ}
because cell occupancy may not be independent.
This leads to a philosophical issue for robotic exploration:
\emph{Should robots seek to observe portions of the environment which they are
certain of but have not seen?}
In this case, expected coverage encodes the affirmative
(all unobserved cells are valuable),
and mutual information the negative
(only uncertain cells are valuable).
Of course, this question is rhetorical, and answers will depend on the setting.
However, we note that our analysis for \rsp{} planning applies to the
former, and the latter is a matter for future work.

\section{Spatially global distance reward}
\label{sec:global_distance_reward}

In addition to being able to evaluate the value of nearby views over the
duration of a planning horizon, robots should be able to reason about the value
of visiting view points.
A common approach is to navigate toward the nearest boundary of unoccupied and
unknown space~\citep{yamauchi1997}
(the frontier, Sec.~\ref{sec:background_frontiers})
which is much like approaches that have robots navigate toward view points
at or near frontiers~\citep{simmons2000aaai,butzke2011planning,charrow2015rss}.

The second term of the exploration reward \eqref{eq:exploration_observation}
seeks to address this issue.
Our distance reward $\distancevalue$ is based on methods developed in our prior
work~\citep{corah2019ral}.
This approach defines informative regions of the environment in terms of view
value via a level set
\begin{align}
  \goalspace &=
  \{\state \mid \viewvalue(\state) \geq \viewthreshold,\,
  \state\in\safespace(\observations_{1:t})\}.
  \label{eq:informative_space}
\end{align}
where $\viewthreshold > 0$ is a threshold on view value.
Rather than evaluating \eqref{eq:informative_space} exactly, we approximate this
set by sampling and updating states in $\safespace$
(which we call informative views)
similarly as described
by~\citep{corah2019ral}.\footnote{
  Some notable changes are that we replace the novelty threshold
  with thresholds on yaw and translation distance and only count new views
  toward the sample limit.
  We also compute distances over the entire grid to obtain more idealized
  results unlike our prior work which approximated values over a local
  sub-map~\citep{corah2019ral}.
}

\begin{figure}
  \includegraphics[width=0.8\linewidth]{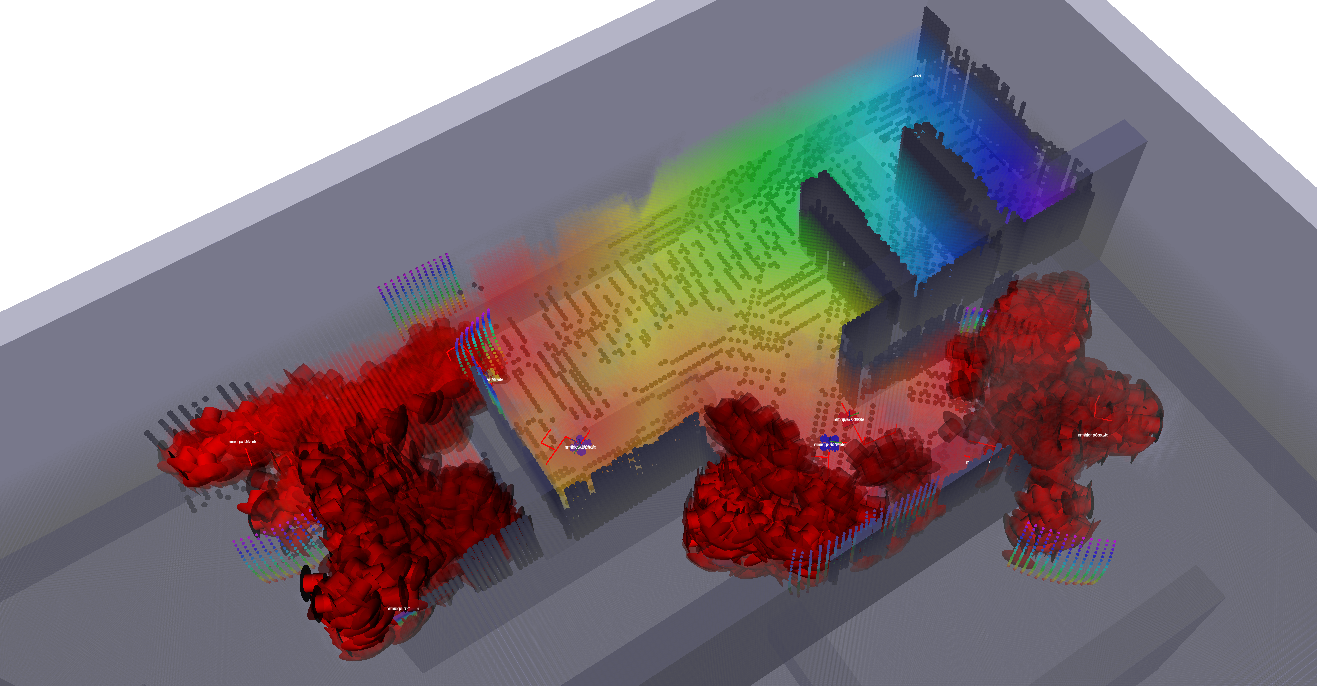}
  \caption[Visualization of sampled informative views and view distance]{%
    This visualization of sampled informative
    views in $\goalspace$ \eqref{eq:informative_space} (red arrows) and view
    distance (rainbow with red corresponding to least distance),
    demonstrates how sufficiently informative views, near the boundary of the
    unobserved space, (light gray) can play a similar role as frontiers for
    exploration.
  }\label{fig:view_distance}
\end{figure}

Next, solving for shortest path distances between \emph{each} point in
$\safespace$ to \emph{any} point in $\goalspace$ produces a distance field
$\goaldistance : \safespace \rightarrow \real_{\geq 0}$.
Figure~\ref{fig:view_distance} provides an example of such a distance field
along with the sampled informative views.
Reducing distances in $\goaldistance$ then brings robots closer to valuable
views in the goal region $\goalspace$.
This distance field can be computed efficiently
(for low-dimensional problems as we do on a three-dimensional grid)
using the fast marching method~\citep{valero-gomez2013} which generalizes
Dijkstra's algorithm.

The distance reward at time $t$ is proportional to the greatest reduction in the
path distance over the course of the trajectory
\begin{align}
  \distancevalue(X_r) =
  \distancefactor
  \cdot
  \max_{l \in \{1,\ldots,L\}}
  \big(
    \goaldistance(\state_t) - \goaldistance(\state_{t+l})
  \big).
  \label{eq:distance_reward}
\end{align}
Note that this value is non-negative.
Selecting the maximum along the trajectory is intended to provide sane
rewards for when robots are near the informative region
\eqref{eq:informative_space} and may exit the region by the end of the planning
horizon.

Also, just as for frontier methods,
this distance term provides the system with a weak notion of completeness.
So long as there are states with observations worth a given value,
robots will navigate to those states once local rewards from $\viewvalue$ have
decreased sufficiently.

Finally, observe that the distance to the nearest view in
\eqref{eq:distance_reward} could be replaced with the distance to a specific
goal location.
This provides an avenue for incorporating methods for goal assignment
(e.g. by solving a cardinality-constrained problem,
Prob.~\ref{prob:submodular_cardinality})
or scheduling which is common in other works related to multi-robot
exploration~\citep{simmons2000aaai,choi2009tro,mitchell2019icra}

\section{Experiment design}

\begin{figure}
  \begin{subfigure}[b]{0.24\linewidth}
    \includegraphics[width=\linewidth,trim={0 150 180 180},clip]{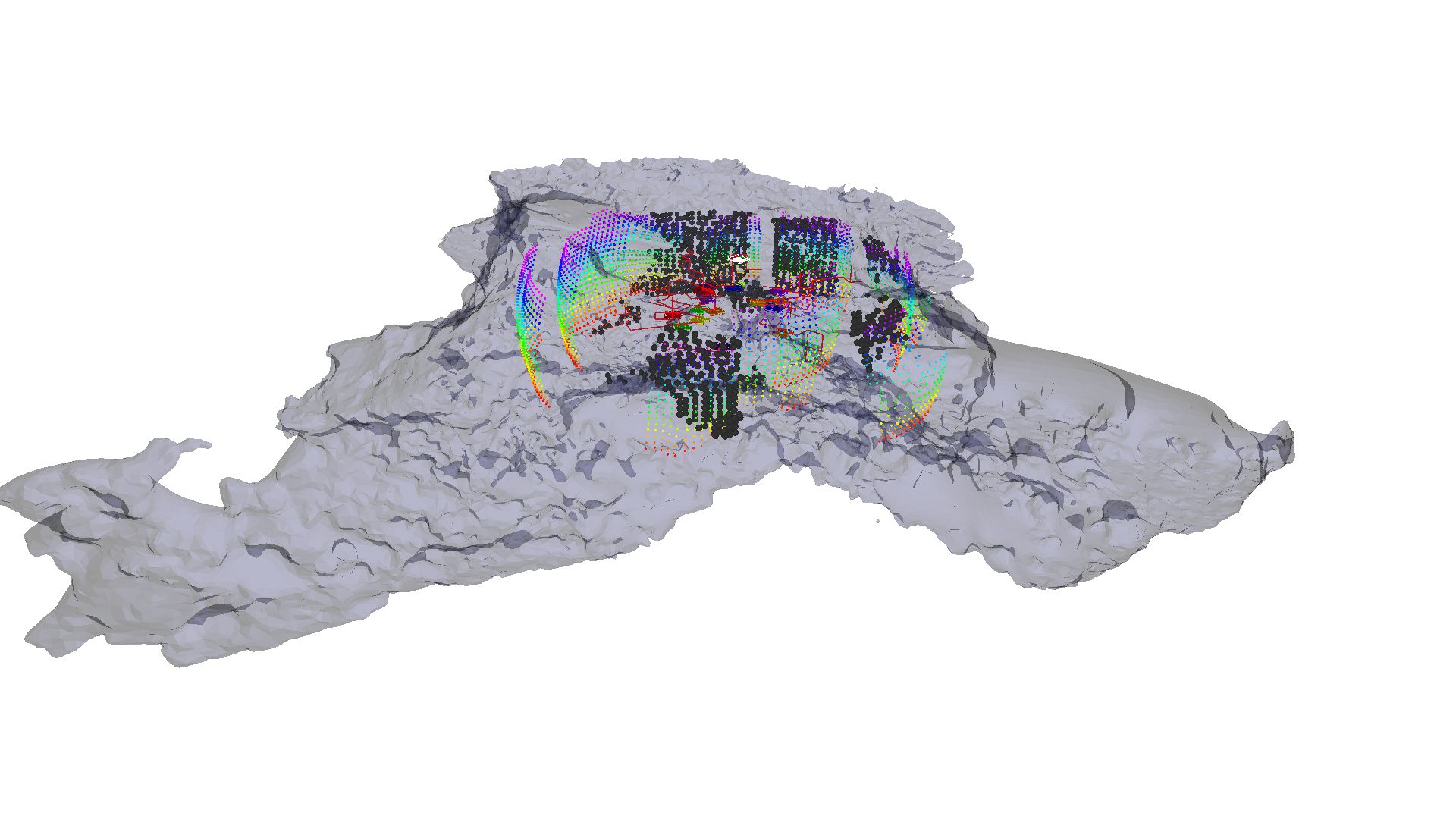}
    \caption{80 robot-iters.}
  \end{subfigure}
  \begin{subfigure}[b]{0.24\linewidth}
    \includegraphics[width=\linewidth,trim={0 150 180 180},clip]{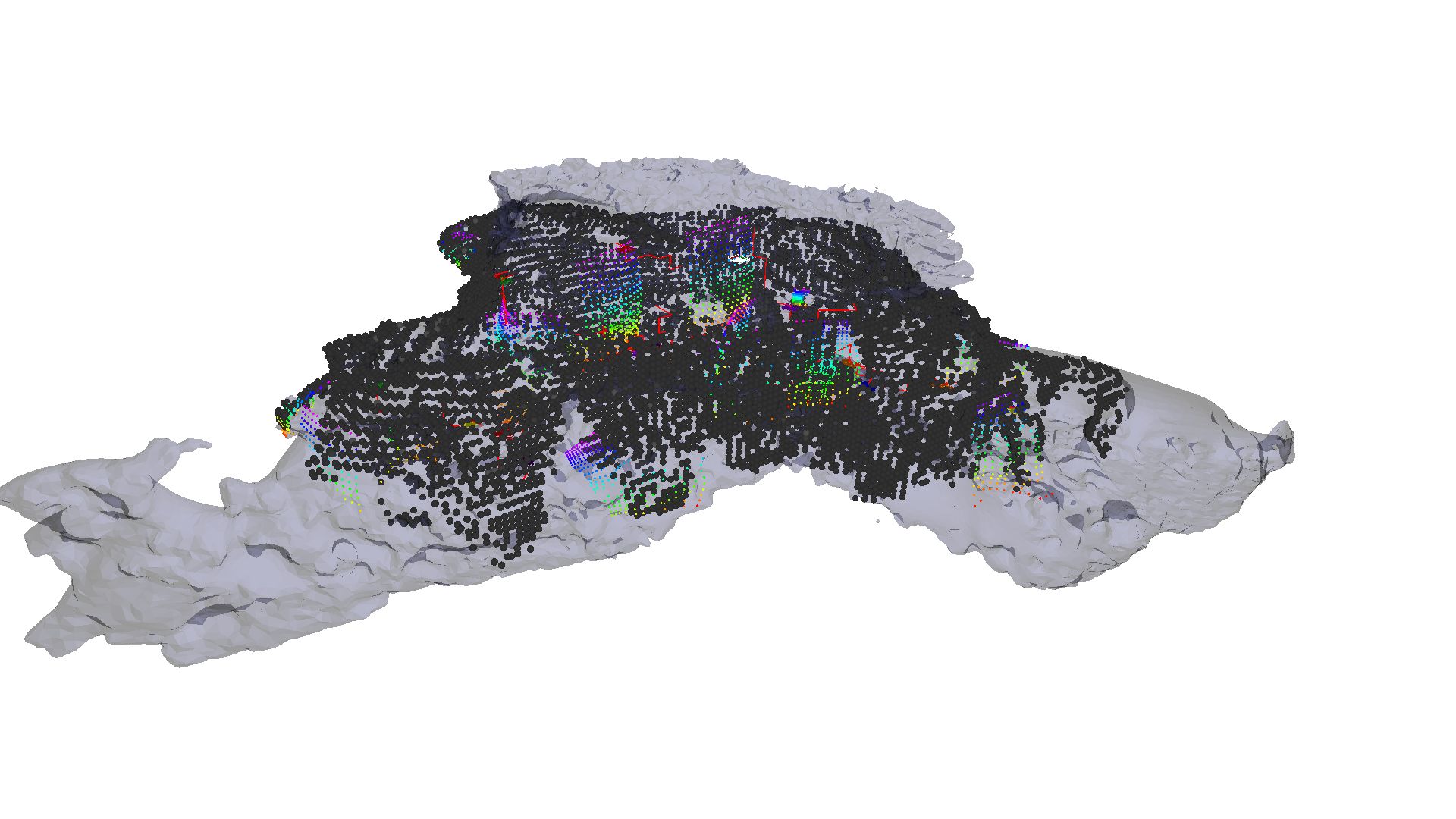}
    \caption{448 robot-iters.}
  \end{subfigure}
  \begin{subfigure}[b]{0.24\linewidth}
    \includegraphics[width=\linewidth,trim={0 150 180 180},clip]{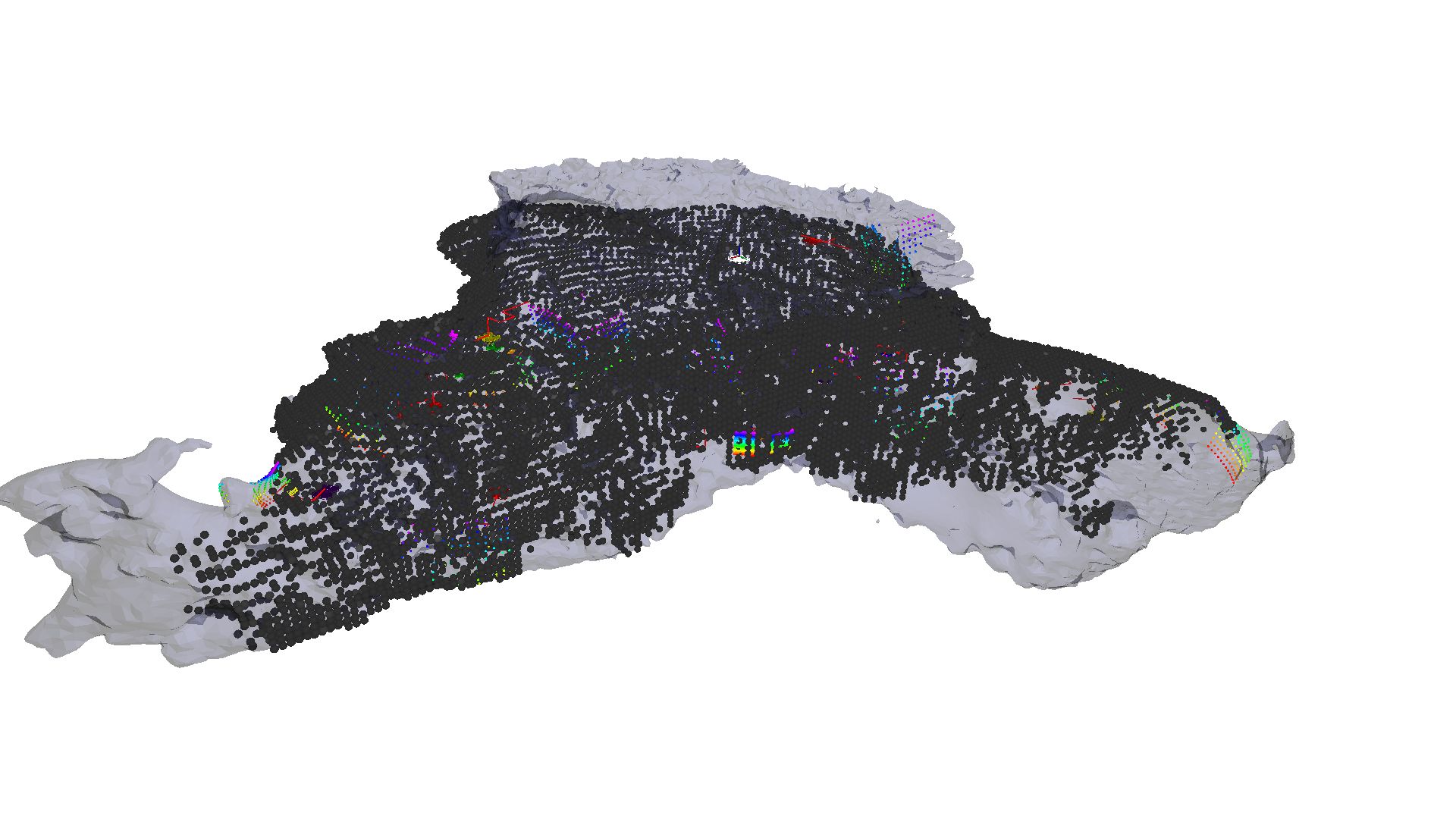}
    \caption{816 robot-iters.}
  \end{subfigure}
  \begin{subfigure}[b]{0.24\linewidth}
    \includegraphics[width=\linewidth,trim={0 150 180 180},clip]{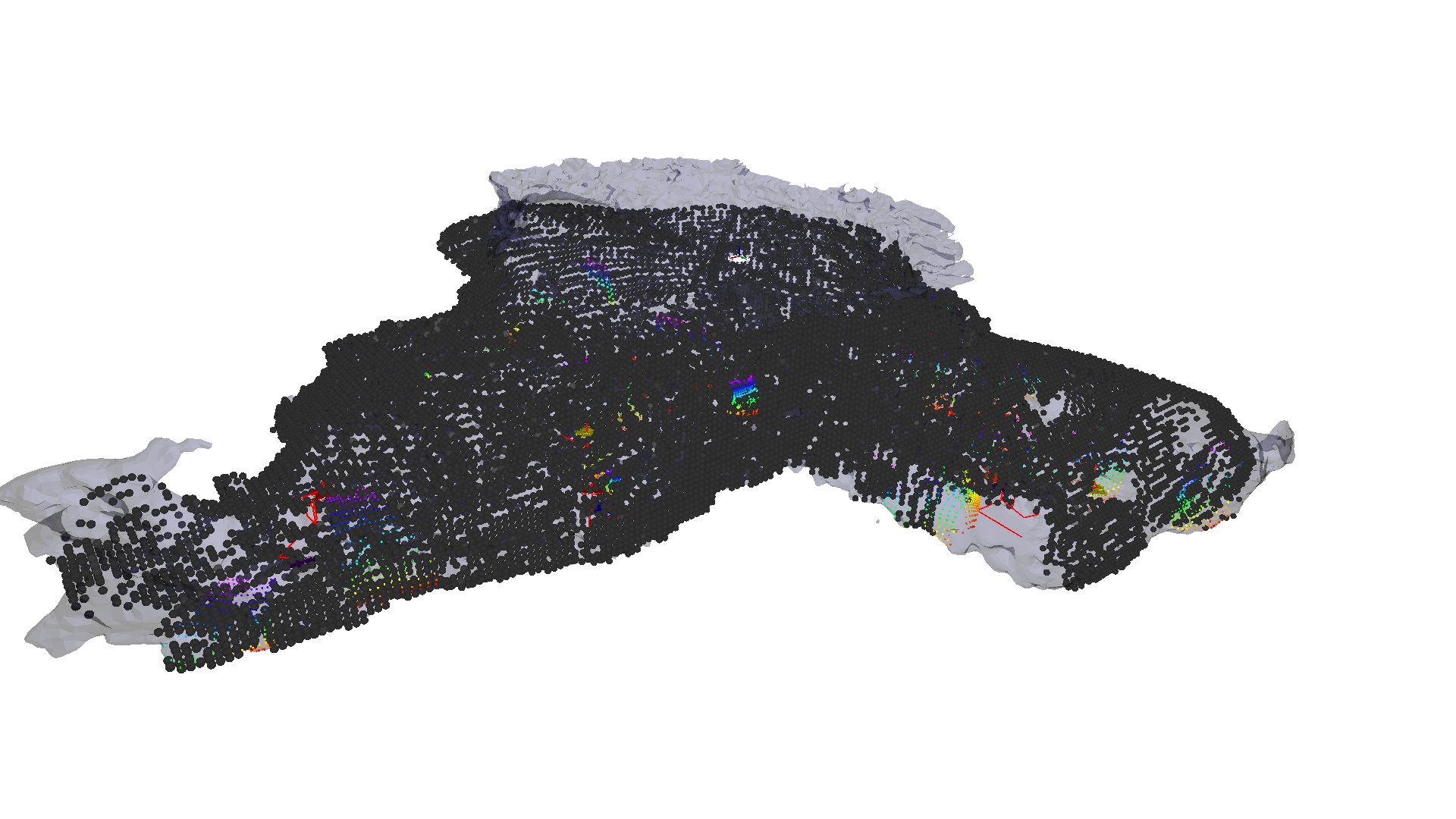}
    \caption{1184 robot-iters.}
  \end{subfigure}
  \caption[Visualization of exploration of the Skylight environment]{%
    The images above visualize an example of the process of exploration of the
    Skylight environment with 16 robots and \rsp{} planning with $\numrounds=6$.
    Additionally, a video providing examples of the exploration process for
    each environment and number of robots is available at:
    \url{https://youtu.be/B9j8LVIs384}
  }%
  \label{fig:time_sensitive_visualization}
\end{figure}

The following section describes the design of the exploration experiments for
this chapter.
We provide results for 10 trials per each configuration and environment
and for various numbers of robots (4, 8, 16, and 32).
For intuition, Fig.~\ref{fig:time_sensitive_visualization} visualizes an example
of the exploration process and provides a link to a video providing examples for
all environments and numbers of robots.

\subsection{Robot and sensor model}

The robot dynamics and sensor model are the same as in the kinematic model in
Sec.~\ref{sec:kinematic_implementation}.
The set of control actions consists of $0.3\,\si{\metre}$ translations in the
cardinal directions with respect to the body frame as well as
$\pi/2\,\si\radian$ yawing motions.
Robots obtain observations from depth cameras
with a range of $2.4\,\si\metre$, a resolution of
$19\times12$, and a field of view of $43.6\si\degree\times34.6\si\degree$,
facing forward, oriented with the long axis vertical.
Unlike Chapter~\ref{chapter:distributed_multi-robot_exploration}, we
\emph{do not} downsample rays when evaluating mutual information and coverage
objectives.

\subsection{Single- and Multi-robot planning}
\label{sec:planning_exploration}

\begin{table}
  \caption[Planner parameters for receding-horizon exploration]{%
    Planner parameters for receding-horizon exploration.
    The myopic and sequential planners were tuned separately to maximize
    performance for 16 robots in the Boxes and Empty environments
    (Table.~\ref{tab:environments}).
    All \rsp{} planners use parameters for sequential planning.
    The parameter $c_p$ belongs to the MCTS planner~\citep{browne2012}
    and is set to roughly half the typical value of the objective for a
    single robot.
  }
  \renewcommand{\tabularxcolumn}[1]{>{\arraybackslash}m{#1}}
  \footnotesize
  \label{tab:exploration_paramemters}
  \begin{tabularx}{\linewidth}{Xllllll}
    Planner
    & \pbox{15ex}{MCTS \\ samples}
    & $c_p$
    & Horizon ($L$)
    & \pbox{18ex}{View Value \\ Threshold ($\viewthreshold$)}
    & \pbox{15ex}{View Distance Factor ($\distancefactor$)}
    & Discount Factor
    \\\belowtoprule
    Sequential
               & 200 
               & 1500 
               & 10 
               & 900 
               & 500 
               & 0.7 
    \\
    Myopic
               & 200 
               & 1500 
               & 10 
               & 300 
               & 700 
               & 1.0 
    \\\bottomrule
  \end{tabularx}
\end{table}

Like prior chapters, robots plan by collectively solving receding-horizon
planning problems \eqref{eq:receding_horizon_exploration}%
---here with the optimistic coverage objective
Sec.~\ref{sec:optimistic_coverage}---and plan individually
via Monte-Carlo tree search (MCTS) and collectively via some method for
submodular maximization.
For submodular maximization,
we compare sequential planning (Alg.~\ref{alg:local_greedy}),
myopic planning (wherein robots plan via MCTS and ignore others' decisions),
and \rsp{} planning\footnote{%
  Appendix~\ref{appendix:dsga_and_rsp} compares suboptimality for
  \rsp{} to \dgreedy{}
  (proposed for exploration in
  Chapter~\ref{chapter:distributed_multi-robot_exploration}).
  Although \dgreedy{} generally outperforms \rsp{} for a given number of
  planning rounds, \rsp{} remains a more practical choice for
  distributed settings.
}
with
1, 3, and 6 rounds ($\numrounds$)
except for the larger Office environment
(see Sec.~\ref{sec:environments_description})
for which we only provide results for $\numrounds=6$.

We selected parameters by iteratively varying values of individual parameters
in simulation trials in the Boxes and Empty environments.
Parameters were selected separately for myopic and sequential planners with
\rsp{} inheriting parameters for sequential planning.\footnote{
  Note that \rsp[1]{} is equivalent to myopic planning but will use the same
  parameters as sequential planning so that any adverse impacts of parameter
  selection for the myopic planner will be evident.
}
Table~\ref{tab:exploration_paramemters} lists the parameters for planning.

\subsubsection{Discounting rewards}

We also discount rewards and treat each robot as if having an independent
probability of failure after each time-step.
This discounting strategy
produces a distribution over the states which robots will visit that is
independent of the realization of the environment and
is compatible with the theory for the objectives we study
Sec.~\ref{sec:expected_coverage_exploration}
and~\ref{sec:mutual_information_exploration}.
Although we will not go into detail, evaluation of the optimistic coverage
objective also remains straightforward.

Ideally, discounting prevents pathological behaviors where robots indefinitely
put off rewards to a future time-step and would weight one robots' early horizon
view above another's overlapping  late horizon view
(which is ostensibly more
uncertain).
However, preliminary experiments demonstrated relatively minor impacts on
performance.

\subsection{Environments and simulation scenarios}
\label{sec:environments_description}

The simulation results evaluate performance across a variety of environments
(listed in Table~\ref{tab:environments}).
In each case, robots start with random yaw and slightly perturbed positions near
a fixed starting location.\footnote{%
  Aside from the initial configuration, the planners also introduce
  stochasticity into the results.
}
We determined maximum coverage values and the lengths of the
simulation trials (iterations per robot) through longer preliminary
experiments with a low view value threshold ($\viewthreshold=100$)
to encourage more complete exploration.
Additionally, all maps use a $0.1\,\si{\metre}$ discretization so that
a volume of $1\,\si{\cubic\metre}$ contains 1000 grid cells.

The environments that we study include synthetic environments that encourage
different kinds of motions and behaviors,
(e.g. upward motion is often slow because robots cannot easily observe space
above them while moving upward)
an empty environment which seeks to characterize maximal steady-state
performance 
(and where optimistic coverage provides oracle rewards
(Sec.~\ref{sec:oracle_objectives})),
and more complex office-like and subterranean environments.

\begin{table}
  \caption[List of exploration environments]{
    This table lists details and descriptions for the different test
    environments.
    The \emph{bounding box volume} provides the fixed volume of the bounding box
    for the experiment while the \emph{exploration volume} lists the approximate
    maximum environment coverage volume for exploration.
    Images portray partially explored environments with unknown space
    (excluding subterranean environments)
    in gray
    and occupied in black.
  }%
  \label{tab:environments}
  \centering
  \setlength{\figurewidth}{0.20\linewidth}
  \renewcommand{\tabularxcolumn}[1]{>{\arraybackslash}m{#1}}
  \footnotesize
  \begin{tabularx}{\linewidth}{lp{5em}ccX}
    Image & Name
          & \pbox{12ex}{Bounding Box \\ Volume}
          & \pbox{12ex}{Exploration Volume} & Description
    \\\belowtoprule
    %
    %
    \raisebox{-0.5\height}{%
      \includegraphics[width=0.8\figurewidth]{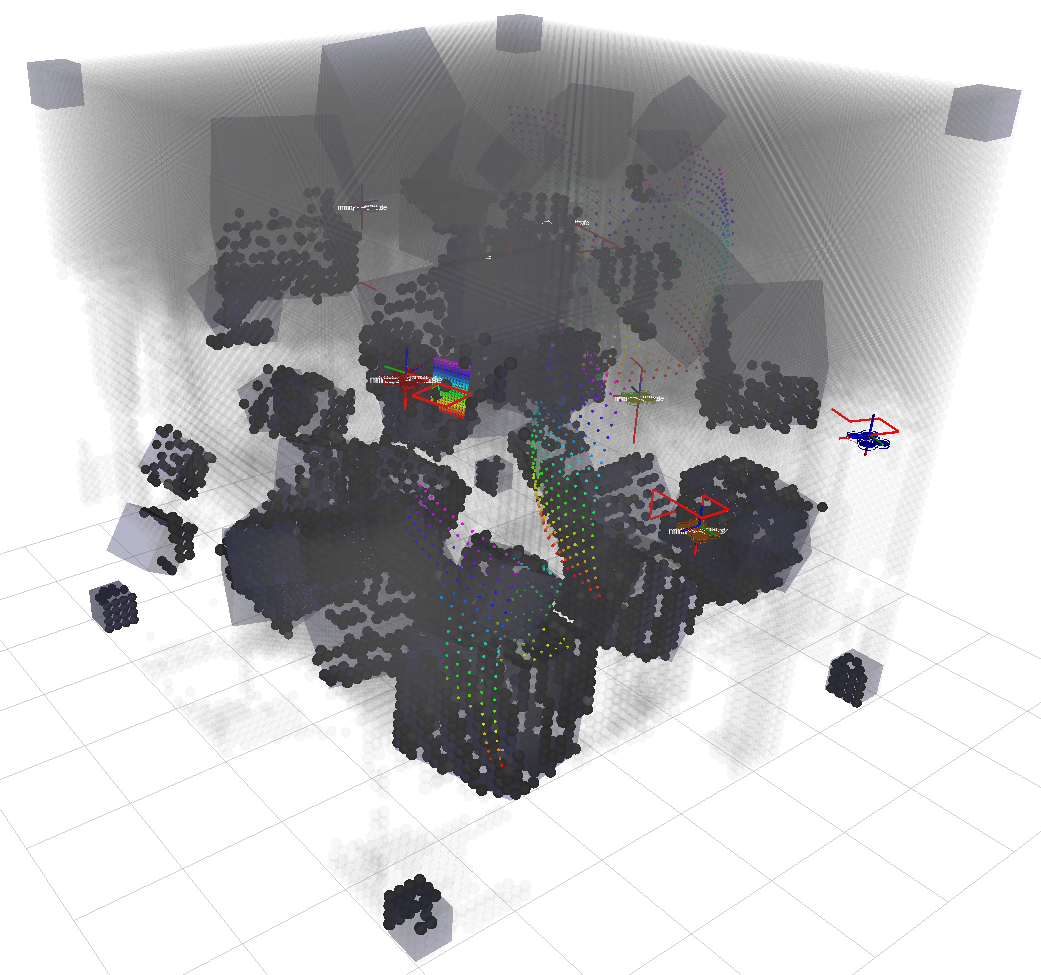}%
    } 
     & Boxes
     & $216\,\si{\cubic\metre}$ 
     & $199\,\si{\cubic\metre}$ 
     & \scriptsize
     Scattered boxes cause occlusions in a $6\,\si\metre$ cube.
     Robots start offset at bottom and move \emph{upward}.
    \\
    %
    %
    \\
    \raisebox{-0.5\height}{%
     \includegraphics[width=\figurewidth]{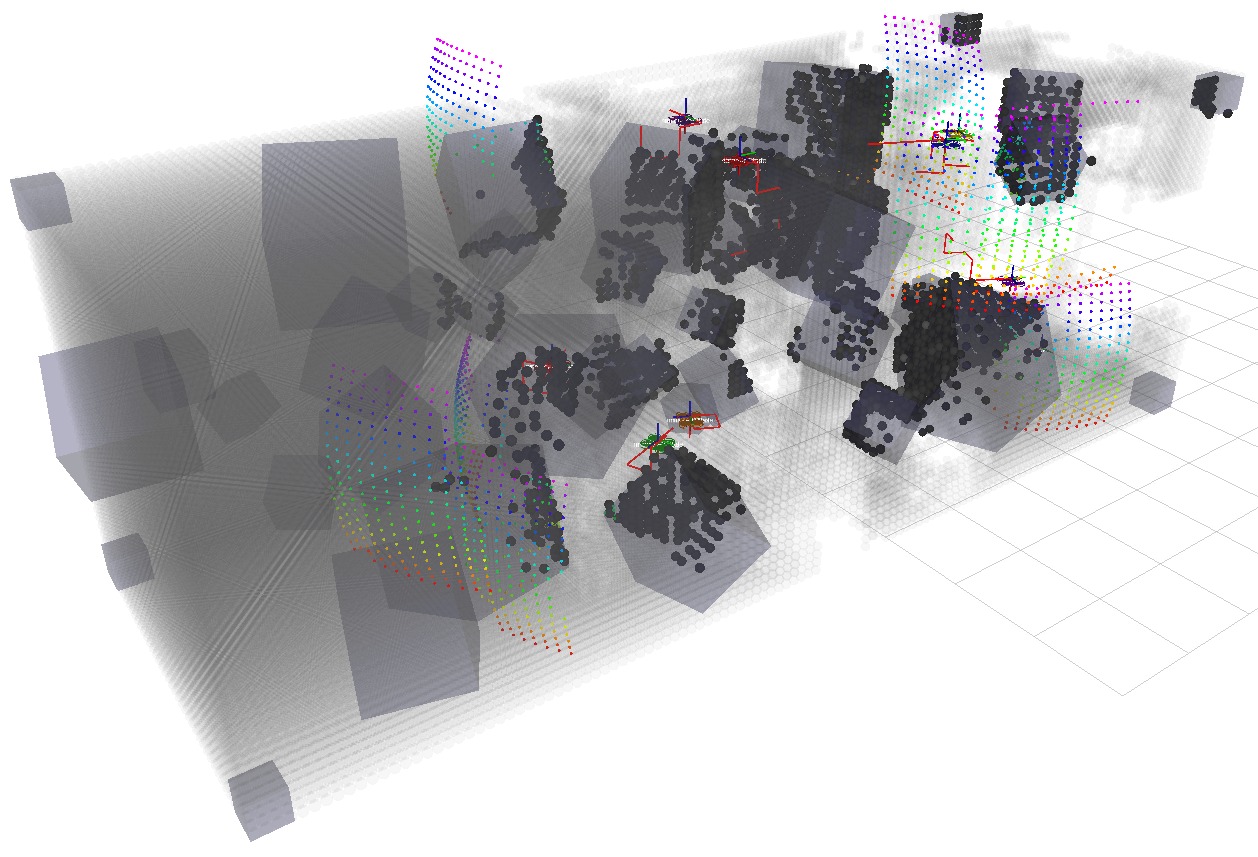}%
    } 
    & Hallway-Boxes
    & $217\,\si{\cubic\metre}$ 
    & $202\,\si{\cubic\metre}$ 
    & \scriptsize
    A rearrangement of the boxes environment into a $12\,\si{\metre}$ square
    prism.
    Robots start at one end and move toward the other.
    \\
    %
    %
    \raisebox{-0.5\height}{%
     \includegraphics[width=\figurewidth]{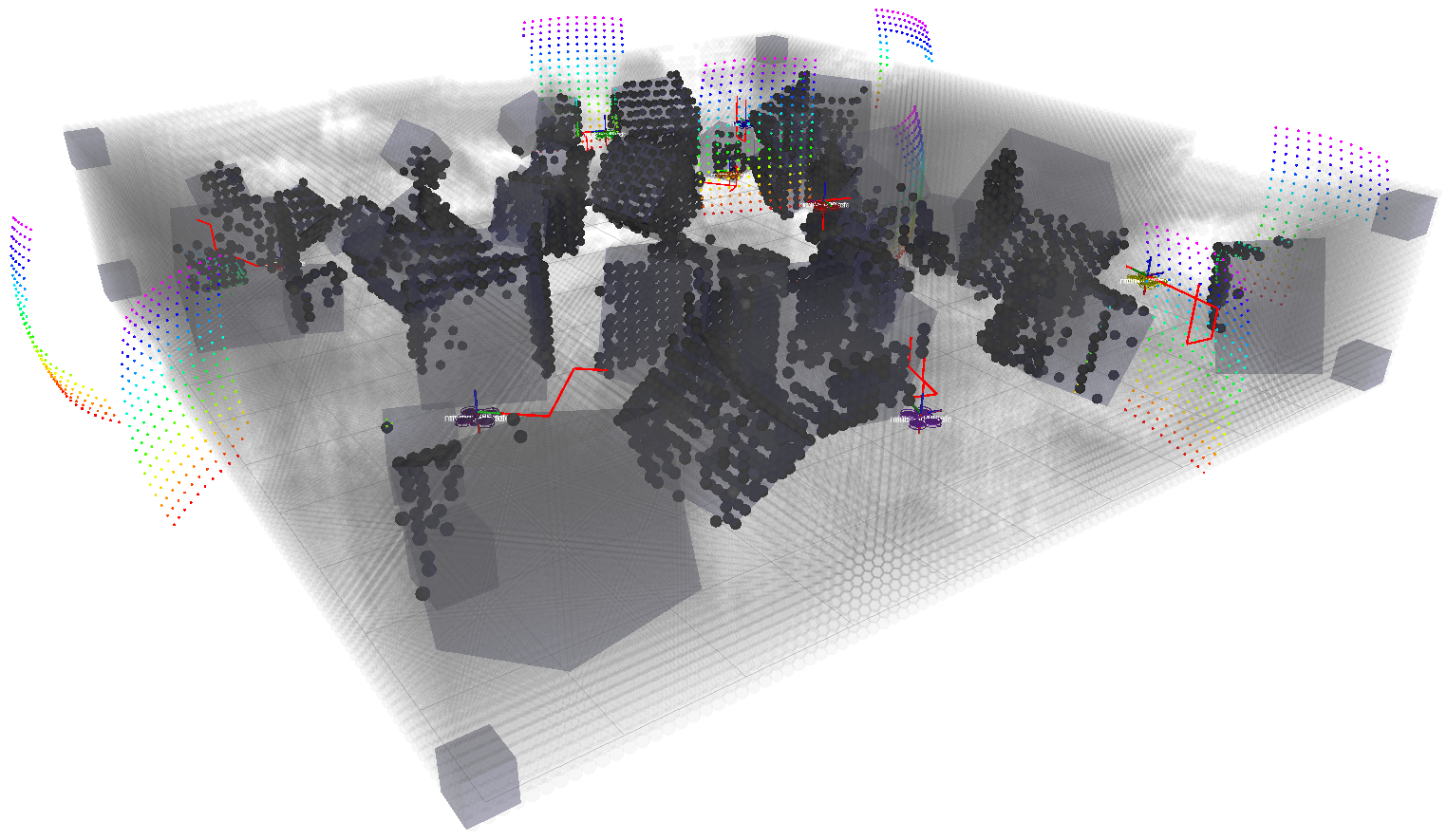}%
    } 
    & Plane-Boxes
    & $227\,\si{\cubic\metre}$ 
    & $212\,\si{\cubic\metre}$ 
    & \scriptsize
    Rearrangement of the boxes environment into $2\,\si{\metre}$ tall square
    planar configuration, and robots start in the center.
    This highlights performance in common, primarily two-dimensional
    environments.
    \\
    %
    %
    \raisebox{-0.5\height}{%
     \includegraphics[width=\figurewidth]{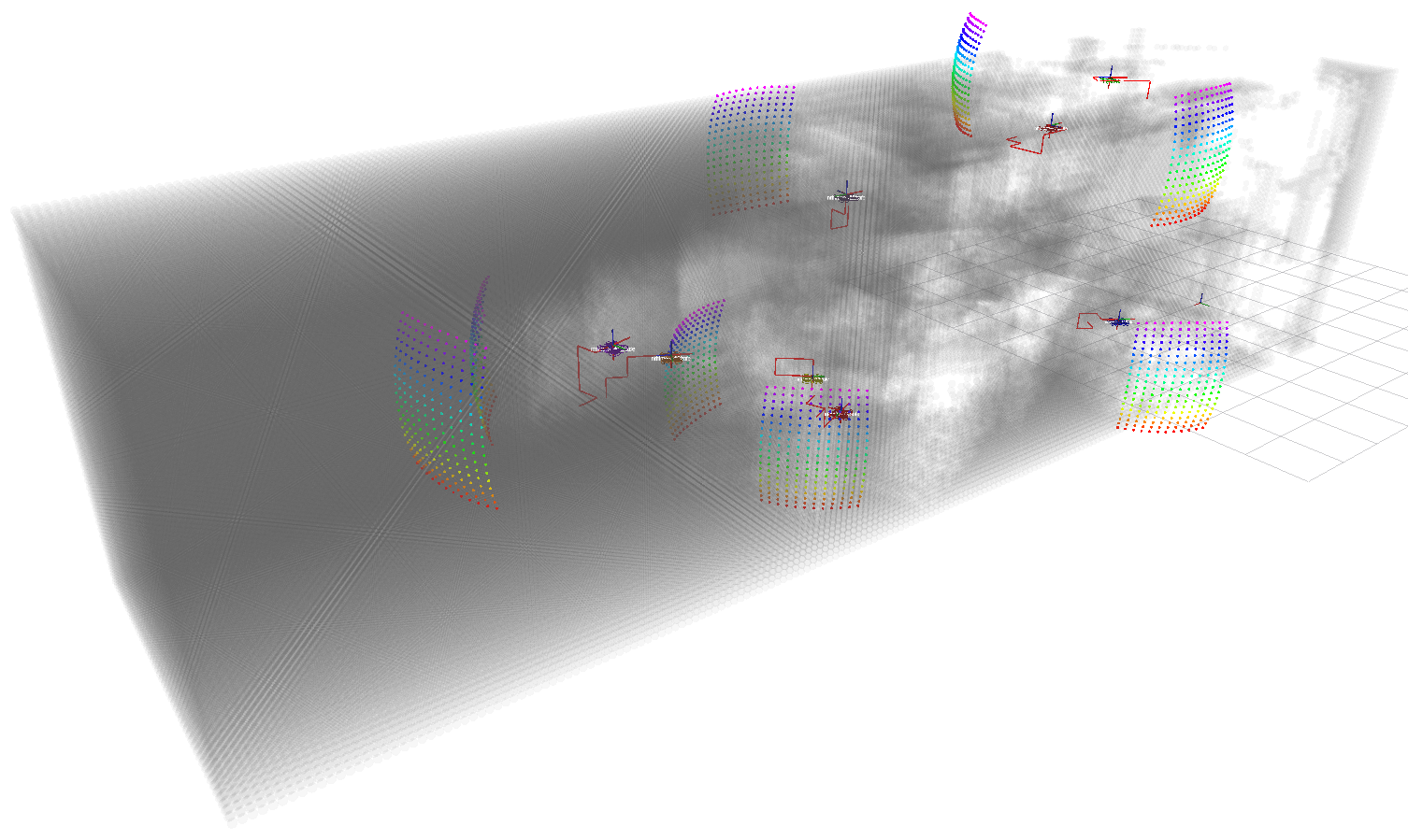}%
    } 
    & Empty
    & $500\,\si{\cubic\metre}$ 
    & $500\,\si{\cubic\metre}$ 
    & \scriptsize
    Robots start on one end of a $20\,\si{\metre}$ hallway-like square prism
    that is completely devoid of obstacles which highlight steady-state
    performance in open space.
    \\
    %
    %
    \raisebox{-0.5\height}{%
     \includegraphics[width=\figurewidth]{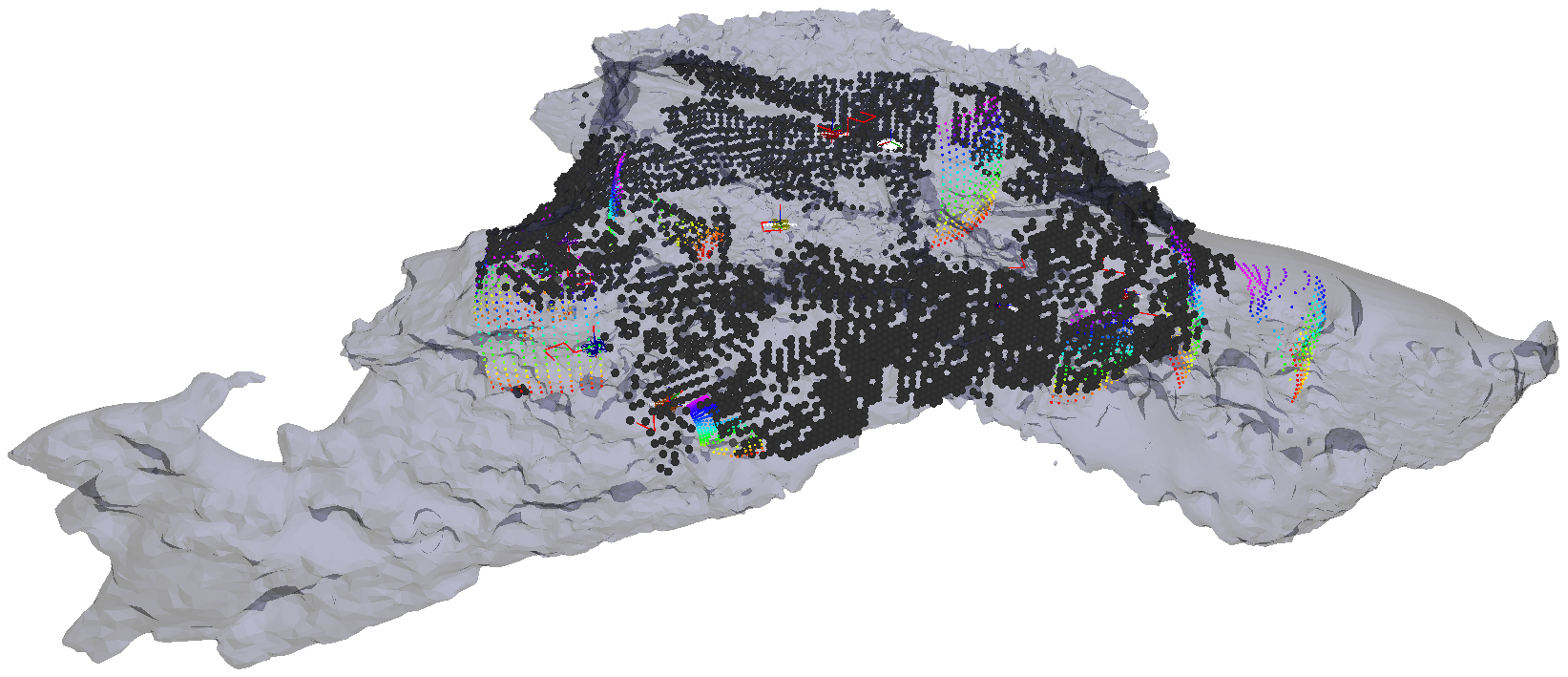}%
    } 
    & Skylight
    & N/A 
    & $220\,\si{\cubic\metre}$ 
    & \scriptsize
    Mesh based on survey data%
    ~\citep{pits_and_caves,jones2015,jones2016phd,tabib2016iros}
    from the Indian Tunnel
    skylight at Craters of the Moon National Park, scaled down to $\sim\!35\%$
    of actual size.
    Robots start at the top of the mouth (now $7\,\si{\metre}$ in diameter).
    \\
    %
    %
    \raisebox{-0.5\height}{%
     \includegraphics[width=0.9\figurewidth]{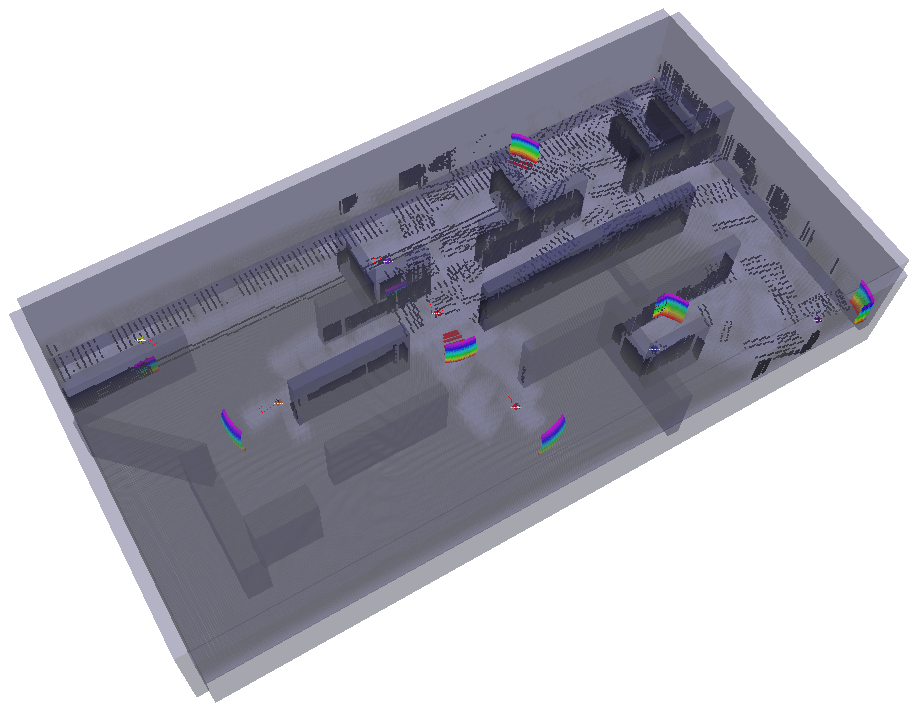}%
    } 
    & Office
    & $1300\,\si{\cubic\metre}$ 
    & $1180\,\si{\cubic\metre}$ 
    & \scriptsize
     A simulated
     $36\,\si{\metre}\times
     18\,\si{\metre}\times
     2\,\si{\metre}$
     office environment.
     This environment almost maze-like in design, and the height
     effectively restricts robots to planar motion.
     Robots start in the upper right corner.
    \\\bottomrule
  \end{tabularx}
\end{table}

\section{Methods for evaluation of results}

The following describes how we evaluate the performance of the submodular
maximization solvers (e.g. \rsp{}) and overall exploration performance
in terms of completion time.

\subsection{Online bounds on solution quality}
\label{sec:online_bounds_exploration}

While most of this thesis focuses on obtaining bounds on solution quality for
broad classes of problems, submodularity also produces certain online bounds
(Sec.~\ref{sec:online_bounds}, \citep{minoux1978})
which can provide tighter guarantees for individual
solutions~\citep{leskovec2007,golovin2011jair,krause2008}.
These bounds also apply to any feasible solution which makes them suitable for
comparing different kinds of planners.

Most works that we are aware of study these online bounds in the context of
cardinality-constrained problems.
Adapting these bounds to simple partition matroids produces some slight
differences compared to the cited works which we present below.
Specifically, maximization steps apply to blocks of the partition matroid
instead of the ground set.
Consider any---possibly incomplete---feasible solution $X\in\independence$
to an instance of Problem~\ref{prob:submodular_partition_matroid}
and an optimal solution $X^\opt$.
The following holds, respectively, by monotonicity, submodularity, and greedy
choice:
\begin{align}
  \setfun(X^\opt)
  &\leq \setfun(X,X^\opt)
  \leq \setfun(X) + \sum_{x^\opt \in X^\opt} \setfun(x^\opt | X)
  \nonumber\\
  &\leq \setfun(X) + \sum_{r \in \robots} \max_{x\in\block_r} \setfun(x | X).
  \label{eq:online_bound}
\end{align}
We apply two instances of the above bound.
For the first, $X$ is the full solution returned by the planner
(assigning actions to all robots) which we refer to simply as the \emph{online}
bound because it uses the current solution to bound the optimal solution.
Next, we call the case where $X=\emptyset$ is the empty set the \emph{oblivious}
bound\footnote{%
  Note that \eqref{eq:online_bound} provides an upper bound on $\setfun(X^\opt)$
  so we can characterize the suboptimality of other solutions by comparing to
  the right-hand-side.
} because this case can bound the optimal solution without planning.
This oblivious bound reduces to
$\setfun(X^\opt) \leq \sum_{r \in \robots} \max_{x\in\block_r} \setfun(x)$
which would produce an optimal solution if $\setfun$ were additive (modular).

\begin{figure}
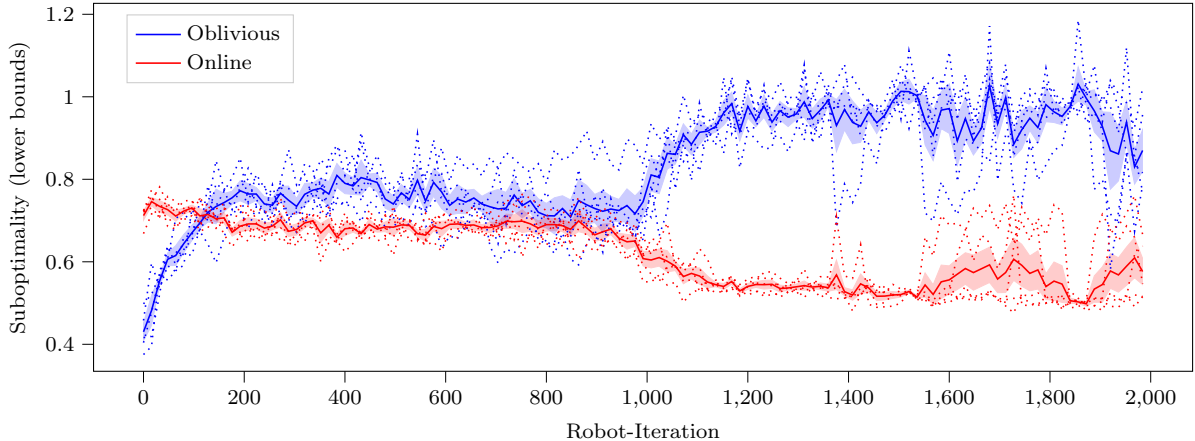

  \centering
  \setlength{\figurewidth}{\linewidth}
  \setlength{\figureheight}{0.4\linewidth}
  \scriptsize
  \inputfigure[online_and_oblivious_bounds]{%
  \datapath/test_standard_configuration_suboptimality_boxes_0b8bea2.tex}
  \caption[Illustration of online and oblivious solution bounds]{
    The above illustrates a representative example for the online and oblivious
    bounds on suboptimality (solution value over bound) with five trials for
    sequential planning and 16 robots, all starting near the \emph{same
    position}.
    Note that the online bound is tighter early when robots are close together,
    and the oblivious bound becomes tighter later, as robots spread out, and
    more so even later, after robots have observed most of the environment
    (environment coverage nears its maximum at about 1000 robot-iterations for
    these trials).
    Note that this bound is not exact as we approximate maximization steps in
    \eqref{eq:online_bound} with MCTS.
    Shaded regions show the standard error, and dotted lines show individual
    trials.
  }%
  \label{fig:online_bounds_example}
\end{figure}

Typically, the online case is tightest for solutions where robots
observe all nearby cells such as when robots are operating in close
proximity (illustrated in Fig.~\ref{fig:online_bounds_example}).\footnote{%
  Intuitively, the online bound might also be tight late when there is little
  left to explore.
  However, Fig.~\ref{fig:online_bounds_example} demonstrates that such behavior
  can be uncommon in practice, primarily (but not exclusively) due to the
  distance term in the objective~\eqref{eq:exploration_objective}.
}
On the other hand, the oblivious case is generally loose when robots are close
together with overlapping fields of view and tightest when robots are spread
out and observing disparate regions of the environment.
Regardless, observe that the tightest bound typically exceeds 70\% which is
significantly tighter than the a-priori bound of 1/2 for sequential planning.

Later, we will use these bounds to characterize solution quality across trials
with different planner configurations in lieu of comparison on common
subproblems.

\subsection{Exploration coverage rates, progress, and completion time}
\label{sec:completion_and_rates}

We evaluate task completion in terms of time
(simulation time-steps which we refer to as iterations)
to reach quotas for environment
coverage (see Sec.~\ref{sec:exploration_problem}).
We provide results for quotas at 90\% (completion the exploration task)
and 30\% (early progress in exploration) of the maximum coverage for each
environment (Table~\ref{tab:environments}).
The latter seeks to provide a measure that is more sensitive to variations in
suboptimality of receding-horizon planning and submodular maximization and
is less affected by challenges related to large-scale assignment and routing
(which we address less thoroughly, Sec.~\ref{sec:global_distance_reward}).
Note that all results for completion times provide statistics based on
completion times for trials individually.

To facilitate comparisons across environments and for different numbers of
robots, we also present results for completion time in terms of the coverage
rates per robot, per iteration.

Of course, evaluating fractions of the maximum exploration volume to determine
when the exploration task is complete would not be possible in practice for
truly unknown environments.
Instead, defining task completion for exploration as when the collection of
sampled informative views (see Sec.~\ref{sec:global_distance_reward}) becomes
empty---indicating that the robots are not aware of any more information-rich
regions of the environment---may work well in practice.

\section{Results}

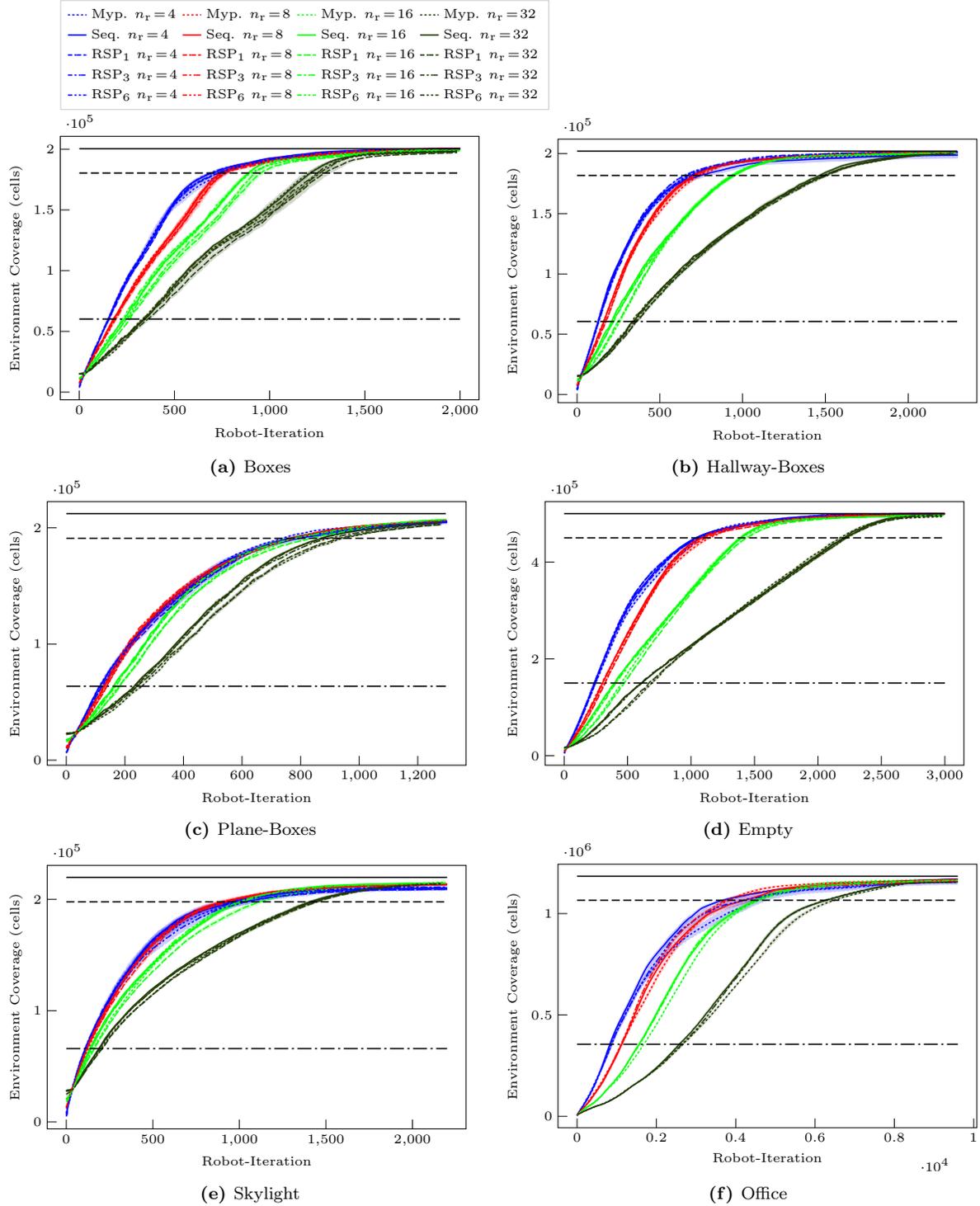
\begin{figure}
  \tiny
  \begin{subfigure}[b]{0.49\linewidth}
    \setlength{\figurewidth}{1.05\linewidth}
    \setlength{\figureheight}{0.7\figurewidth}
\begin{tikzpicture}

\definecolor{color0}{rgb}{0.12156862745098,0.227450980392157,0.00392156862745098}

\begin{axis}[
height=\figureheight,
legend cell align={left},
legend columns=4,
legend style={at={(0.00,1.10)}, anchor=south west, draw=white!80.0!black},
legend image post style={scale=0.5}, 
tick align=inside,
tick pos=left,
width=\figurewidth,
x grid style={lightgray!92.0261437908!black},
xlabel={Robot-Iteration},
xmin=-100, xmax=2100,
xtick style={color=black},
y grid style={lightgray!92.0261437908!black},
ylabel={Environment Coverage (cells)},
ymin=-6383.79138812876, ymax=210614.877920107,
ytick style={color=black}
]
\end{axis}

\end{tikzpicture}
    \vspace{-0.2cm}
    \caption{Boxes}
  \end{subfigure}
  \begin{subfigure}[b]{0.49\linewidth}
    \setlength{\figurewidth}{1.05\linewidth}
    \setlength{\figureheight}{0.7\figurewidth}
\begin{tikzpicture}

\definecolor{color0}{rgb}{0.12156862745098,0.227450980392157,0.00392156862745098}

\begin{axis}[
height=\figureheight,
tick align=inside,
tick pos=left,
width=\figurewidth,
x grid style={lightgray!92.0261437908!black},
xlabel={Robot-Iteration},
xmin=-115, xmax=2415,
xtick style={color=black},
y grid style={lightgray!92.0261437908!black},
ylabel={Environment Coverage (cells)},
ymin=-6779.41482546816, ymax=211905.210229784,
ytick style={color=black}
]
\end{axis}

\end{tikzpicture}
    \caption{Hallway-Boxes}
  \end{subfigure}
  \begin{subfigure}[b]{0.49\linewidth}
    \setlength{\figurewidth}{1.05\linewidth}
    \setlength{\figureheight}{0.7\figurewidth}
\begin{tikzpicture}

\definecolor{color0}{rgb}{0.12156862745098,0.227450980392157,0.00392156862745098}

\begin{axis}[
height=\figureheight,
tick align=inside,
tick pos=left,
width=\figurewidth,
x grid style={lightgray!92.0261437908!black},
xlabel={Robot-Iteration},
xmin=-65, xmax=1365,
xtick style={color=black},
y grid style={lightgray!92.0261437908!black},
ylabel={Environment Coverage (cells)},
ymin=-4325.7782908721, ymax=222509.703728137,
ytick style={color=black}
]
\end{axis}

\end{tikzpicture}
    \caption{Plane-Boxes}
  \end{subfigure}
  \begin{subfigure}[b]{0.49\linewidth}
    \setlength{\figurewidth}{1.05\linewidth}
    \setlength{\figureheight}{0.7\figurewidth}
\begin{tikzpicture}

\definecolor{color0}{rgb}{0.12156862745098,0.227450980392157,0.00392156862745098}

\begin{axis}[
height=\figureheight,
tick align=inside,
tick pos=left,
width=\figurewidth,
x grid style={lightgray!92.0261437908!black},
xlabel={Robot-Iteration},
xmin=-150, xmax=3150,
xtick style={color=black},
y grid style={lightgray!92.0261437908!black},
ylabel={Environment Coverage (cells)},
ymin=-19786.3786673849, ymax=524750.684698447,
ytick style={color=black}
]
\end{axis}

\end{tikzpicture}
    \caption{Empty}
  \end{subfigure}
  \begin{subfigure}[b]{0.49\linewidth}
    \setlength{\figurewidth}{1.05\linewidth}
    \setlength{\figureheight}{0.7\figurewidth}
\begin{tikzpicture}

\definecolor{color0}{rgb}{0.12156862745098,0.227450980392157,0.00392156862745098}

\begin{axis}[
height=\figureheight,
tick align=inside,
tick pos=left,
width=\figurewidth,
x grid style={lightgray!92.0261437908!black},
xlabel={Robot-Iteration},
xmin=-110, xmax=2310,
xtick style={color=black},
y grid style={lightgray!92.0261437908!black},
ylabel={Environment Coverage (cells)},
ymin=-6246.20904390939, ymax=230591.343287805,
ytick style={color=black}
]
\end{axis}

\end{tikzpicture}
    \caption{Skylight}
  \end{subfigure}
  \begin{subfigure}[b]{0.49\linewidth}
    \setlength{\figurewidth}{1.05\linewidth}
    \setlength{\figureheight}{0.7\figurewidth}
\begin{tikzpicture}

\definecolor{color0}{rgb}{0.12156862745098,0.227450980392157,0.00392156862745098}

\begin{axis}[
height=\figureheight,
tick align=inside,
tick pos=left,
width=\figurewidth,
x grid style={lightgray!92.0261437908!black},
xlabel={Robot-Iteration},
xmin=-480, xmax=10080,
xtick style={color=black},
y grid style={lightgray!92.0261437908!black},
ylabel={Environment Coverage (cells)},
ymin=-56349.5912806764, ymax=1243373.31387051,
ytick style={color=black}
]
\end{axis}

\end{tikzpicture}
    \caption{Office}
  \end{subfigure}
  \caption[Environment coverage results]{Environment coverage results for each
    environment. Shaded regions delineate the standard error.
    Black lines demarcate (from top to bottom) the maximum environment coverage,
    the completion threshold, and the early progress threshold.
    \emph{These plots highlight general performance trends.
      Other plots better highlight differences between planners.
    }
  }%
  \label{fig:environment_coverage}
\end{figure}

Figure~\ref{fig:environment_coverage} summarizes the exploration process from
the results for the simulations in terms of environment coverage
\eqref{eq:environment_coverage}
and highlights the maximum coverage values (Table~\ref{tab:environments})
and completion thresholds (Sec.~\ref{sec:completion_and_rates}).
Although, there is not always much variation across planner configurations,
these plots also illustrate consistency in coverage rates, graceful degradation
in performance, and reliably complete exploration.
The latter, reliable task completion, is evident in the asymptotic convergence
and decreasing variance at the end of each trial.
This is a product of the distance reward
(Sec.~\ref{sec:global_distance_reward}), and we note that \citet{corah2019ral}
provide additional results related to this feature.
The rest of this section will go into more detail on the former points regarding
time to completion.

\subsection{Comparison of early and final completion by environment}

\begin{figure}
  \tiny
  \begin{subfigure}[b]{0.49\linewidth}
    \setlength{\figurewidth}{1.0\linewidth}
    \setlength{\figureheight}{0.7\figurewidth}
\begin{tikzpicture}

\definecolor{color1}{rgb}{0,0.811764705882353,1}
\definecolor{color0}{rgb}{0.12156862745098,0.227450980392157,0.00392156862745098}

\begin{axis}[
height=\figureheight,
legend cell align={left},
legend style={at={(0.03,0.97)}, anchor=north west, draw=white!80.0!black},
legend image post style={scale=0.5}, 
tick align=outside,
tick pos=left,
width=\figurewidth,
x grid style={lightgray!92.0261437908!black},
xlabel={Num. Robots},
xmajorgrids,
xmin=2.6, xmax=33.4,
xtick style={color=black},
xtick={4,8,16,32},
y grid style={lightgray!92.0261437908!black},
ylabel={Coverage Rate (cells per iteration)},
ymajorgrids,
ymin=0, ymax=12000,
ytick style={color=black}
]
\path [draw=blue, fill=blue, opacity=0.2, line width=0.0pt]
(axis cs:4,1628.56832594905)
--(axis cs:8,2684.73449228558)
--(axis cs:16,4274.40430717587)
--(axis cs:32,5908.78674144032)
--(axis cs:32,5702.73531718065)
--(axis cs:16,4116.06734027767)
--(axis cs:8,2586.44689254277)
--(axis cs:4,1523.86546700417)
--cycle;
\path [draw=red, fill=red, opacity=0.2, line width=0.0pt]
(axis cs:4,1904.15194415664)
--(axis cs:8,3101.29898303008)
--(axis cs:16,4492.66282863667)
--(axis cs:32,5915.12838355264)
--(axis cs:32,5774.61801469451)
--(axis cs:16,4303.47287334948)
--(axis cs:8,2982.54125211602)
--(axis cs:4,1846.20946407503)
--cycle;
\path [draw=green, fill=green, opacity=0.2, line width=0.0pt]
(axis cs:4,2269.93688083467)
--(axis cs:8,4089.58766756175)
--(axis cs:16,6303.61915127255)
--(axis cs:32,8952.67234695781)
--(axis cs:32,8748.59818929961)
--(axis cs:16,6153.01007607515)
--(axis cs:8,3966.33915002507)
--(axis cs:4,2154.39777916322)
--cycle;
\path [draw=color0, fill=color0, opacity=0.2, line width=0.0pt]
(axis cs:4,2540.89730152441)
--(axis cs:8,4056.77493926824)
--(axis cs:16,6127.57935755988)
--(axis cs:32,7971.62837867086)
--(axis cs:32,7789.29811546052)
--(axis cs:16,5993.60397929049)
--(axis cs:8,3964.7303099926)
--(axis cs:4,2472.48572921293)
--cycle;
\path [draw=color1, fill=color1, opacity=0.2, line width=0.0pt]
(axis cs:4,2464.49074017957)
--(axis cs:8,4580.5844081605)
--(axis cs:16,7452.17994201518)
--(axis cs:32,11150.3438779731)
--(axis cs:32,10768.2550246409)
--(axis cs:16,7175.64930210709)
--(axis cs:8,4457.48377746408)
--(axis cs:4,2388.44952898698)
--cycle;
\path [draw=orange, fill=orange, opacity=0.2, line width=0.0pt]
(axis cs:4,1691.63132585636)
--(axis cs:8,2609.94968288816)
--(axis cs:16,3637.05560678646)
--(axis cs:32,4331.03896129429)
--(axis cs:32,4266.30377975079)
--(axis cs:16,3566.25400536949)
--(axis cs:8,2526.05781298134)
--(axis cs:4,1590.03399852351)
--cycle;
\addplot [semithick, blue]
table {%
4 1576.21689647661
8 2635.59069241418
16 4195.23582372677
32 5805.76102931049
};
\addlegendentry{Boxes}
\addplot [semithick, red]
table {%
4 1875.18070411584
8 3041.92011757305
16 4398.06785099307
32 5844.87319912357
};
\addlegendentry{Hallway-Boxes}
\addplot [semithick, green]
table {%
4 2212.16732999894
8 4027.96340879341
16 6228.31461367385
32 8850.63526812871
};
\addlegendentry{Plane-Boxes}
\addplot [semithick, color0]
table {%
4 2506.69151536867
8 4010.75262463042
16 6060.59166842519
32 7880.46324706569
};
\addlegendentry{Empty}
\addplot [semithick, color1]
table {%
4 2426.47013458327
8 4519.03409281229
16 7313.91462206114
32 10959.299451307
};
\addlegendentry{Skylight}
\addplot [semithick, orange]
table {%
4 1640.83266218994
8 2568.00374793475
16 3601.65480607798
32 4298.67137052254
};
\addlegendentry{Office}
\end{axis}

\end{tikzpicture}
\begin{tikzpicture}

\definecolor{color1}{rgb}{0,0.811764705882353,1}
\definecolor{color0}{rgb}{0.12156862745098,0.227450980392157,0.00392156862745098}

\begin{axis}[
height=\figureheight,
tick align=outside,
tick pos=left,
width=\figurewidth,
x grid style={lightgray!92.0261437908!black},
xlabel={Num. Robots},
xmin=2.6, xmax=33.4,
xtick style={color=black},
xtick={4,8,16,32},
xmajorgrids,
y grid style={lightgray!92.0261437908!black},
ylabel={Coverage Rate (cells per robot-iteration)},
ymin=0, ymax=650,
ymajorgrids,
ytick style={color=black}
]
\path [draw=blue, fill=blue, opacity=0.2, line width=0.0pt, dashed]
(axis cs:4,407.142081487262)
--(axis cs:8,335.591811535698)
--(axis cs:16,267.150269198492)
--(axis cs:32,184.64958567001)
--(axis cs:32,178.210478661895)
--(axis cs:16,257.254208767354)
--(axis cs:8,323.305861567847)
--(axis cs:4,380.966366751042)
--cycle;
\path [draw=red, fill=red, opacity=0.2, line width=0.0pt, dashed]
(axis cs:4,476.037986039159)
--(axis cs:8,387.66237287876)
--(axis cs:16,280.791426789792)
--(axis cs:32,184.84776198602)
--(axis cs:32,180.456812959203)
--(axis cs:16,268.967054584342)
--(axis cs:8,372.817656514502)
--(axis cs:4,461.552366018758)
--cycle;
\path [draw=green, fill=green, opacity=0.2, line width=0.0pt, dashed]
(axis cs:4,567.484220208668)
--(axis cs:8,511.198458445218)
--(axis cs:16,393.976196954534)
--(axis cs:32,279.771010842432)
--(axis cs:32,273.393693415613)
--(axis cs:16,384.563129754697)
--(axis cs:8,495.792393753133)
--(axis cs:4,538.599444790804)
--cycle;
\path [draw=color0, fill=color0, opacity=0.2, line width=0.0pt, dashed]
(axis cs:4,635.224325381101)
--(axis cs:8,507.09686740853)
--(axis cs:16,382.973709847493)
--(axis cs:32,249.113386833464)
--(axis cs:32,243.415566108141)
--(axis cs:16,374.600248705656)
--(axis cs:8,495.591288749075)
--(axis cs:4,618.121432303232)
--cycle;
\path [draw=color1, fill=color1, opacity=0.2, line width=0.0pt, dashed]
(axis cs:4,616.122685044893)
--(axis cs:8,572.573051020063)
--(axis cs:16,465.761246375949)
--(axis cs:32,348.44824618666)
--(axis cs:32,336.507969520028)
--(axis cs:16,448.478081381693)
--(axis cs:8,557.18547218301)
--(axis cs:4,597.112382246744)
--cycle;
\path [draw=orange, fill=orange, opacity=0.2, line width=0.0pt, dashed]
(axis cs:4,422.90783146409)
--(axis cs:8,326.24371036102)
--(axis cs:16,227.315975424154)
--(axis cs:32,135.344967540446)
--(axis cs:32,133.321993117212)
--(axis cs:16,222.890875335593)
--(axis cs:8,315.757226622667)
--(axis cs:4,397.508499630879)
--cycle;
\addplot [semithick, blue, dash pattern=on 3.7pt off 1.6pt]
table {%
4 394.054224119152
8 329.448836551772
16 262.202238982923
32 181.430032165953
};
\addplot [semithick, red, dash pattern=on 3.7pt off 1.6pt]
table {%
4 468.795176028959
8 380.240014696631
16 274.879240687067
32 182.652287472612
};
\addplot [semithick, green, dash pattern=on 3.7pt off 1.6pt]
table {%
4 553.041832499736
8 503.495426099176
16 389.269663354615
32 276.582352129022
};
\addplot [semithick, color0, dash pattern=on 3.7pt off 1.6pt]
table {%
4 626.672878842167
8 501.344078078803
16 378.786979276574
32 246.264476470803
};
\addplot [semithick, color1, dash pattern=on 3.7pt off 1.6pt]
table {%
4 606.617533645819
8 564.879261601537
16 457.119663878821
32 342.478107853344
};
\addplot [semithick, orange, dash pattern=on 3.7pt off 1.6pt]
table {%
4 410.208165547484
8 321.000468491844
16 225.103425379874
32 134.333480328829
};
\end{axis}

\end{tikzpicture}
    \caption{Early Progress Threshold}
    \label{subfig:early_rates}
  \end{subfigure}
  \begin{subfigure}[b]{0.49\linewidth}
    \setlength{\figurewidth}{1.0\linewidth}
    \setlength{\figureheight}{0.7\figurewidth}
\begin{tikzpicture}

\definecolor{color1}{rgb}{0,0.811764705882353,1}
\definecolor{color0}{rgb}{0.12156862745098,0.227450980392157,0.00392156862745098}

\begin{axis}[
height=\figureheight,
tick align=outside,
tick pos=left,
width=\figurewidth,
x grid style={lightgray!92.0261437908!black},
xlabel={Num. Robots},
xmajorgrids,
xmin=2.6, xmax=33.4,
xtick style={color=black},
xtick={4,8,16,32},
y grid style={lightgray!92.0261437908!black},
ylabel={Coverage Rate (cells per iteration)},
ymajorgrids,
ymin=0, ymax=12000,
ytick style={color=black}
]
\path [draw=blue, fill=blue, opacity=0.2, line width=0.0pt]
(axis cs:4,1030.68081844542)
--(axis cs:8,1951.81018283015)
--(axis cs:16,3326.97617896863)
--(axis cs:32,4718.68795190675)
--(axis cs:32,4581.3486321383)
--(axis cs:16,3238.65306649687)
--(axis cs:8,1885.28094723611)
--(axis cs:4,981.885051157037)
--cycle;
\path [draw=red, fill=red, opacity=0.2, line width=0.0pt]
(axis cs:4,1120.87627233832)
--(axis cs:8,2093.52126386907)
--(axis cs:16,3178.50109789958)
--(axis cs:32,3937.48257379047)
--(axis cs:32,3843.03899961445)
--(axis cs:16,3067.98115685735)
--(axis cs:8,2029.87919025477)
--(axis cs:4,1071.84949555257)
--cycle;
\path [draw=green, fill=green, opacity=0.2, line width=0.0pt]
(axis cs:4,1016.674883256)
--(axis cs:8,2026.99605197217)
--(axis cs:16,3999.02393666741)
--(axis cs:32,7046.84081974086)
--(axis cs:32,6953.53727222694)
--(axis cs:16,3888.34441634477)
--(axis cs:8,1950.81476159419)
--(axis cs:4,968.185515127211)
--cycle;
\path [draw=color0, fill=color0, opacity=0.2, line width=0.0pt]
(axis cs:4,1706.38003112173)
--(axis cs:8,3393.96997406593)
--(axis cs:16,5234.01726630126)
--(axis cs:32,6517.33440477127)
--(axis cs:32,6446.82738695951)
--(axis cs:16,5129.71389837089)
--(axis cs:8,3312.80950294136)
--(axis cs:4,1670.51785497861)
--cycle;
\path [draw=color1, fill=color1, opacity=0.2, line width=0.0pt]
(axis cs:4,819.424696872304)
--(axis cs:8,1782.09278314023)
--(axis cs:16,3221.18463510014)
--(axis cs:32,4522.34267302187)
--(axis cs:32,4394.75409917748)
--(axis cs:16,3115.15804802391)
--(axis cs:8,1696.10519455653)
--(axis cs:4,754.160279569368)
--cycle;
\path [draw=orange, fill=orange, opacity=0.2, line width=0.0pt]
(axis cs:4,1116.91243584369)
--(axis cs:8,2314.73758258225)
--(axis cs:16,3865.63216003465)
--(axis cs:32,5501.42135454299)
--(axis cs:32,5419.32102056875)
--(axis cs:16,3608.25015179828)
--(axis cs:8,2246.2675912235)
--(axis cs:4,878.699773160089)
--cycle;
\addplot [semithick, blue]
table {%
4 1006.28293480123
8 1918.54556503313
16 3282.81462273275
32 4650.01829202252
};
\addplot [semithick, red]
table {%
4 1096.36288394545
8 2061.70022706192
16 3123.24112737846
32 3890.26078670246
};
\addplot [semithick, green]
table {%
4 992.430199191606
8 1988.90540678318
16 3943.68417650609
32 7000.1890459839
};
\addplot [semithick, color0]
table {%
4 1688.44894305017
8 3353.38973850364
16 5181.86558233608
32 6482.08089586539
};
\addplot [semithick, color1]
table {%
4 786.792488220836
8 1739.09898884838
16 3168.17134156203
32 4458.54838609967
};
\addplot [semithick, orange]
table {%
4 997.80610450189
8 2280.50258690287
16 3736.94115591646
32 5460.37118755587
};
\end{axis}

\end{tikzpicture}
\begin{tikzpicture}

\definecolor{color1}{rgb}{0,0.811764705882353,1}
\definecolor{color0}{rgb}{0.12156862745098,0.227450980392157,0.00392156862745098}

\begin{axis}[
height=\figureheight,
tick align=outside,
tick pos=left,
width=\figurewidth,
x grid style={lightgray!92.0261437908!black},
xlabel={Num. Robots},
xmin=2.6, xmax=33.4,
xtick style={color=black},
xtick={4,8,16,32},
xmajorgrids,
y grid style={lightgray!92.0261437908!black},
ylabel={Coverage Rate (cells per robot-iteration)},
ymin=0, ymax=650,
ymajorgrids,
ytick style={color=black}
]
\path [draw=blue, fill=blue, opacity=0.2, line width=0.0pt, dashed]
(axis cs:4,257.670204611356)
--(axis cs:8,243.976272853769)
--(axis cs:16,207.936011185539)
--(axis cs:32,147.458998497086)
--(axis cs:32,143.167144754322)
--(axis cs:16,202.415816656055)
--(axis cs:8,235.660118404513)
--(axis cs:4,245.471262789259)
--cycle;
\path [draw=red, fill=red, opacity=0.2, line width=0.0pt, dashed]
(axis cs:4,280.219068084581)
--(axis cs:8,261.690157983634)
--(axis cs:16,198.656318618724)
--(axis cs:32,123.046330430952)
--(axis cs:32,120.094968737951)
--(axis cs:16,191.748822303584)
--(axis cs:8,253.734898781846)
--(axis cs:4,267.962373888143)
--cycle;
\path [draw=green, fill=green, opacity=0.2, line width=0.0pt, dashed]
(axis cs:4,254.168720814)
--(axis cs:8,253.374506496521)
--(axis cs:16,249.938996041713)
--(axis cs:32,220.213775616902)
--(axis cs:32,217.298039757092)
--(axis cs:16,243.021526021548)
--(axis cs:8,243.851845199274)
--(axis cs:4,242.046378781803)
--cycle;
\path [draw=color0, fill=color0, opacity=0.2, line width=0.0pt, dashed]
(axis cs:4,426.595007780432)
--(axis cs:8,424.246246758241)
--(axis cs:16,327.126079143829)
--(axis cs:32,203.666700149102)
--(axis cs:32,201.463355842485)
--(axis cs:16,320.607118648181)
--(axis cs:8,414.101187867669)
--(axis cs:4,417.629463744652)
--cycle;
\path [draw=color1, fill=color1, opacity=0.2, line width=0.0pt, dashed]
(axis cs:4,204.856174218076)
--(axis cs:8,222.761597892528)
--(axis cs:16,201.324039693759)
--(axis cs:32,141.323208531933)
--(axis cs:32,137.336065599296)
--(axis cs:16,194.697378001495)
--(axis cs:8,212.013149319567)
--(axis cs:4,188.540069892342)
--cycle;
\path [draw=orange, fill=orange, opacity=0.2, line width=0.0pt, dashed]
(axis cs:4,279.228108960923)
--(axis cs:8,289.342197822781)
--(axis cs:16,241.602010002165)
--(axis cs:32,171.919417329469)
--(axis cs:32,169.353781892773)
--(axis cs:16,225.515634487392)
--(axis cs:8,280.783448902937)
--(axis cs:4,219.674943290022)
--cycle;
\addplot [semithick, blue, dash pattern=on 3.7pt off 1.6pt]
table {%
4 251.570733700307
8 239.818195629141
16 205.175913920797
32 145.313071625704
};
\addplot [semithick, red, dash pattern=on 3.7pt off 1.6pt]
table {%
4 274.090720986362
8 257.71252838274
16 195.202570461154
32 121.570649584452
};
\addplot [semithick, green, dash pattern=on 3.7pt off 1.6pt]
table {%
4 248.107549797901
8 248.613175847897
16 246.480261031631
32 218.755907686997
};
\addplot [semithick, color0, dash pattern=on 3.7pt off 1.6pt]
table {%
4 422.112235762542
8 419.173717312955
16 323.866598896005
32 202.565027995793
};
\addplot [semithick, color1, dash pattern=on 3.7pt off 1.6pt]
table {%
4 196.698122055209
8 217.387373606048
16 198.010708847627
32 139.329637065615
};
\addplot [semithick, orange, dash pattern=on 3.7pt off 1.6pt]
table {%
4 249.451526125472
8 285.062823362859
16 233.558822244779
32 170.636599611121
};
\end{axis}

\end{tikzpicture}
    \caption{Complete Exploration}
    \label{subfig:complete_rates}
  \end{subfigure}
  \caption[Coverage rates per robot across environments]{%
    Coverage rates across environments up to
    (\subref{subfig:early_rates})
    the threshold for early progress
    and
    (\subref{subfig:complete_rates})
    up to completing the exploration task
    both \emph{per iteration} (top, solid)
    and \emph{per robot per iteration} (bottom, dashed).
    In general, the exploration process slows after the beginning of each trial,
    and contributions per-robot decrease gradually as more are added.
  }%
  \label{fig:coverage_rate_by_environment}
\end{figure}
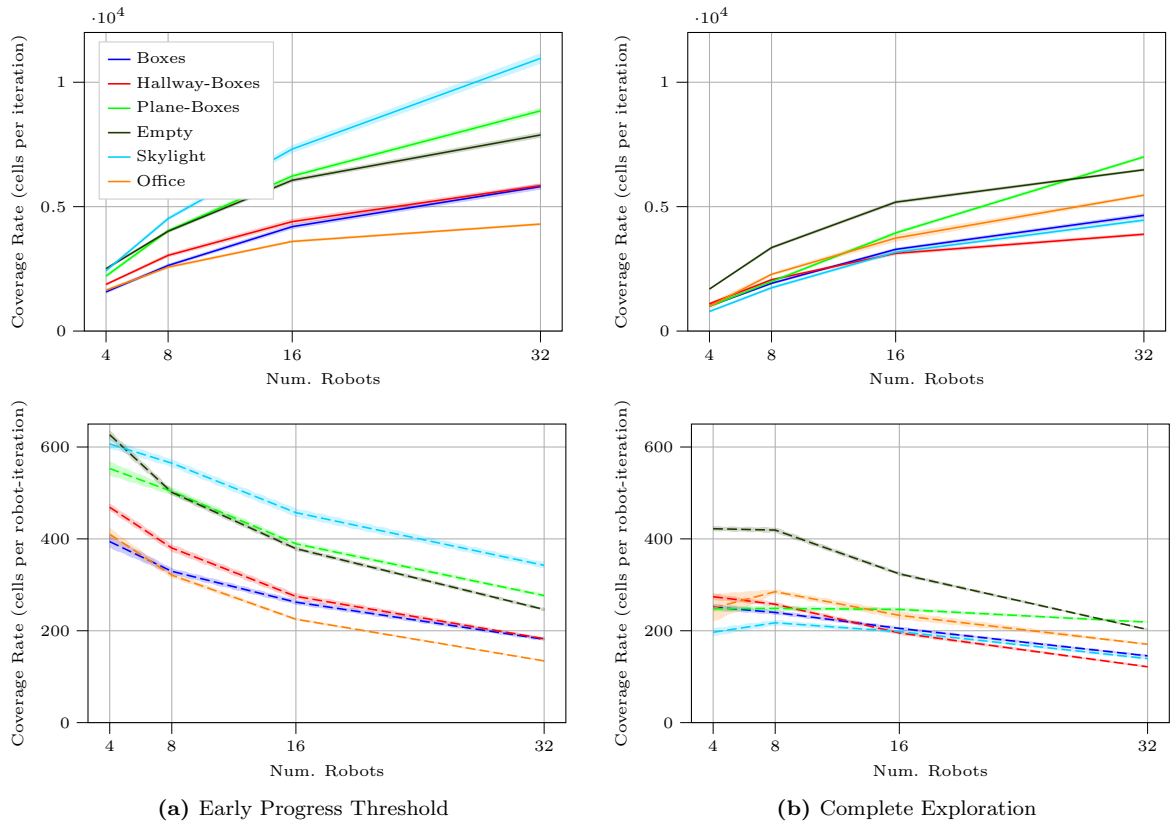

Figure~\ref{fig:coverage_rate_by_environment} illustrates how
coverage rates vary as robots reach the early progress and completion thresholds
for different environments and numbers of robots.\footnote{%
  This figure depicts results for \rsp[6] which performs reliably and avoids
  the case for Sequential in the Hallway-Boxes environment in which not all
  trials completed the exploration task.
}
All environments exhibit some similar trends as would be expected:
slowing exploration over time and with increasing numbers of robots.
Across all this data, coverage rates vary from
120 to
630 (cells per robot per iteration)
a factor of about $5\times$, and the Empty environment alone exhibits a
variation of $3\times$.
While not all of this slow-down is likely to be avoidable, there may be room to
improve performance significantly through the end of each trial via more
effective goal assignment and long-term planning.
On the other hand, coverage rates for completing the exploration task
are somewhat more consistent, excluding the Empty environment, and per robot
performance varies by about $2.4\times$.
Still, Plane-Boxes---where robots can spread out quickly to cover the
volume---is the only environment where new robots consistently maintain coverage
rates (and contribute to improving completion times).
On the other hand, robots slow down significantly after beginning exploration in
the Skylight environment, possibly because the robots can spread out quickly in
the central volume but slow down while covering the passages on either side.

Later, Sec.~\ref{sec:completion_by_planner}
(which also includes completion
times in Table~\ref{tab:exploration_performance})
will go into more detail on exploration times as a function of the method of
planning for multi-robot coordination.

\subsection{Planner suboptimality}

\begin{figure}
  \tiny
  \begin{subfigure}[b]{0.32\linewidth}
    \setlength{\figurewidth}{1.1\linewidth}
    \setlength{\figureheight}{0.9\figurewidth}
\begin{tikzpicture}

\begin{axis}[
height=\figureheight,
legend cell align={left},
legend columns=1,
legend style={draw=white!80.0!black},
legend image post style={scale=0.5}, 
tick align=inside,
tick pos=left,
width=\figurewidth,
x grid style={lightgray!92.0261437908!black},
xlabel={Num. Robots},
xmin=2.6, xmax=33.4,
xtick={4,8,16,32},
xmajorgrids,
xtick style={color=black},
y grid style={lightgray!92.0261437908!black},
ylabel={Suboptimality (lower bound)},
ymajorgrids,
ymin=0.658387038181694, ymax=0.981027569572757,
ytick style={color=black}
]
\path [draw=black, fill=black, opacity=0.2, line width=0.0pt]
(axis cs:4,0.965116017328073)
--(axis cs:8,0.896379488936513)
--(axis cs:16,0.807188122514159)
--(axis cs:32,0.721235218097543)
--(axis cs:32,0.716448979423886)
--(axis cs:16,0.79712554993359)
--(axis cs:8,0.888983192213121)
--(axis cs:4,0.958877032736742)
--cycle;
\path [draw=black, fill=black, opacity=0.2, line width=0.0pt, dash pattern=on 2pt off 1pt on 1pt off 0pt]
(axis cs:4,0.953988129804407)
--(axis cs:8,0.873517405085316)
--(axis cs:16,0.765830972566315)
--(axis cs:32,0.680367164951788)
--(axis cs:32,0.673052516881288)
--(axis cs:16,0.753484352792801)
--(axis cs:8,0.860517403297618)
--(axis cs:4,0.948977579349474)
--cycle;
\path [draw=black, fill=black, opacity=0.2, line width=0.0pt, dash pattern=on 2pt off 1pt on 1pt off 1pt on 1pt off 0pt]
(axis cs:4,0.966362090873163)
--(axis cs:8,0.893705246448072)
--(axis cs:16,0.791578582002038)
--(axis cs:32,0.710887418281532)
--(axis cs:32,0.707300760888764)
--(axis cs:16,0.784986343223645)
--(axis cs:8,0.885584688064257)
--(axis cs:4,0.95907908486403)
--cycle;
\path [draw=black, fill=black, opacity=0.2, line width=0.0pt, dash pattern=on 2pt off 1pt on 1pt off 1pt on 1pt off 1pt on 1pt off 0pt]
(axis cs:4,0.956885031013744)
--(axis cs:8,0.899650190421774)
--(axis cs:16,0.803384102939604)
--(axis cs:32,0.71653785062444)
--(axis cs:32,0.711659102755402)
--(axis cs:16,0.794107362470926)
--(axis cs:8,0.892603403139717)
--(axis cs:4,0.953678485574394)
--cycle;
\addplot [semithick, black]
table {%
4 0.961996525032407
8 0.892681340574817
16 0.802156836223875
32 0.718842098760715
};
\addlegendentry{Seq.}
\addplot [semithick, black, dash pattern=on 2pt off 1pt on 1pt off 0pt]
table {%
4 0.95148285457694
8 0.867017404191467
16 0.759657662679558
32 0.676709840916538
};
\addlegendentry{RSP$_1$}
\addplot [semithick, black, dash pattern=on 2pt off 1pt on 1pt off 1pt on 1pt off 0pt]
table {%
4 0.962720587868597
8 0.889644967256164
16 0.788282462612842
32 0.709094089585148
};
\addlegendentry{RSP$_3$}
\addplot [semithick, black, dash pattern=on 2pt off 1pt on 1pt off 1pt on 1pt off 1pt on 1pt off 0pt]
table {%
4 0.955281758294069
8 0.896126796780746
16 0.798745732705265
32 0.714098476689921
};
\addlegendentry{RSP$_6$}
\end{axis}

\end{tikzpicture}
    \caption{Boxes}
  \end{subfigure}
  \begin{subfigure}[b]{0.32\linewidth}
    \setlength{\figurewidth}{1.1\linewidth}
    \setlength{\figureheight}{0.9\figurewidth}
\begin{tikzpicture}

\begin{axis}[
height=\figureheight,
tick align=inside,
tick pos=left,
width=\figurewidth,
x grid style={lightgray!92.0261437908!black},
xlabel={Num. Robots},
xmajorgrids,
xmin=2.6, xmax=33.4,
xtick={4,8,16,32},
xtick style={color=black},
y grid style={lightgray!92.0261437908!black},
ylabel={Suboptimality (lower bound)},
ymajorgrids,
ymin=0.711726020625745, ymax=0.985500893261604,
ytick style={color=black}
]
\path [draw=black, fill=black, opacity=0.2, line width=0.0pt]
(axis cs:4,0.973056580869064)
--(axis cs:8,0.937612784104185)
--(axis cs:16,0.868929866053388)
--(axis cs:32,0.794605703820135)
--(axis cs:32,0.783155008757146)
--(axis cs:16,0.857586009001771)
--(axis cs:8,0.932424192673601)
--(axis cs:4,0.968062566198584)
--cycle;
\path [draw=black, fill=black, opacity=0.2, line width=0.0pt, dash pattern=on 2pt off 1pt on 1pt off 0pt]
(axis cs:4,0.96611232579288)
--(axis cs:8,0.909427040150689)
--(axis cs:16,0.825052525418944)
--(axis cs:32,0.735842722495936)
--(axis cs:32,0.724170333018284)
--(axis cs:16,0.81614416259534)
--(axis cs:8,0.900504726436803)
--(axis cs:4,0.957151754516616)
--cycle;
\path [draw=black, fill=black, opacity=0.2, line width=0.0pt, dash pattern=on 2pt off 1pt on 1pt off 1pt on 1pt off 0pt]
(axis cs:4,0.972924808120328)
--(axis cs:8,0.93037849979574)
--(axis cs:16,0.86213362806924)
--(axis cs:32,0.789618941579381)
--(axis cs:32,0.780068252263833)
--(axis cs:16,0.852447813973787)
--(axis cs:8,0.924063533123661)
--(axis cs:4,0.968175091746736)
--cycle;
\path [draw=black, fill=black, opacity=0.2, line width=0.0pt, dash pattern=on 2pt off 1pt on 1pt off 1pt on 1pt off 1pt on 1pt off 0pt]
(axis cs:4,0.971553613974181)
--(axis cs:8,0.934229092551815)
--(axis cs:16,0.867806284724248)
--(axis cs:32,0.797554722937503)
--(axis cs:32,0.792800583670256)
--(axis cs:16,0.859596181505249)
--(axis cs:8,0.92697046604468)
--(axis cs:4,0.966975969192826)
--cycle;
\addplot [semithick, black]
table {%
4 0.970559573533824
8 0.935018488388893
16 0.86325793752758
32 0.788880356288641
};
\addplot [semithick, black, dash pattern=on 2pt off 1pt on 1pt off 0pt]
table {%
4 0.961632040154748
8 0.904965883293746
16 0.820598344007142
32 0.73000652775711
};
\addplot [semithick, black, dash pattern=on 2pt off 1pt on 1pt off 1pt on 1pt off 0pt]
table {%
4 0.970549949933532
8 0.9272210164597
16 0.857290721021513
32 0.784843596921607
};
\addplot [semithick, black, dash pattern=on 2pt off 1pt on 1pt off 1pt on 1pt off 1pt on 1pt off 0pt]
table {%
4 0.969264791583504
8 0.930599779298248
16 0.863701233114749
32 0.795177653303879
};
\end{axis}

\end{tikzpicture}
    \caption{Hallway-Boxes}
  \end{subfigure}
  \begin{subfigure}[b]{0.32\linewidth}
    \setlength{\figurewidth}{1.1\linewidth}
    \setlength{\figureheight}{0.9\figurewidth}
\begin{tikzpicture}

\begin{axis}[
height=\figureheight,
tick align=inside,
tick pos=left,
width=\figurewidth,
x grid style={lightgray!92.0261437908!black},
xlabel={Num. Robots},
xmajorgrids,
xmin=2.6, xmax=33.4,
xtick={4,8,16,32},
xtick style={color=black},
y grid style={lightgray!92.0261437908!black},
ylabel={Suboptimality (lower bound)},
ymajorgrids,
ymin=0.727006781683169, ymax=0.998007210509476,
ytick style={color=black}
]
\path [draw=black, fill=black, opacity=0.2, line width=0.0pt]
(axis cs:4,0.985689009199189)
--(axis cs:8,0.95865769440926)
--(axis cs:16,0.901367414800235)
--(axis cs:32,0.798446560293545)
--(axis cs:32,0.788607291762268)
--(axis cs:16,0.898722483428229)
--(axis cs:8,0.953017872723794)
--(axis cs:4,0.981980224589827)
--cycle;
\path [draw=black, fill=black, opacity=0.2, line width=0.0pt, dash pattern=on 2pt off 1pt on 1pt off 0pt]
(axis cs:4,0.977865516086646)
--(axis cs:8,0.944937830645764)
--(axis cs:16,0.862403323377394)
--(axis cs:32,0.748903629279214)
--(axis cs:32,0.739324982993456)
--(axis cs:16,0.854609188214293)
--(axis cs:8,0.941125780246909)
--(axis cs:4,0.971692045717908)
--cycle;
\path [draw=black, fill=black, opacity=0.2, line width=0.0pt, dash pattern=on 2pt off 1pt on 1pt off 1pt on 1pt off 0pt]
(axis cs:4,0.983883503499852)
--(axis cs:8,0.957750679741461)
--(axis cs:16,0.897107667036728)
--(axis cs:32,0.786800945539797)
--(axis cs:32,0.77977416445793)
--(axis cs:16,0.887720886623516)
--(axis cs:8,0.953483884838492)
--(axis cs:4,0.981049282856595)
--cycle;
\path [draw=black, fill=black, opacity=0.2, line width=0.0pt, dash pattern=on 2pt off 1pt on 1pt off 1pt on 1pt off 1pt on 1pt off 0pt]
(axis cs:4,0.980440698370675)
--(axis cs:8,0.95829694912201)
--(axis cs:16,0.895794213965701)
--(axis cs:32,0.785304109354172)
--(axis cs:32,0.779232285519457)
--(axis cs:16,0.891097506441789)
--(axis cs:8,0.955640077682401)
--(axis cs:4,0.976232078671403)
--cycle;
\addplot [semithick, black]
table {%
4 0.983834616894508
8 0.955837783566527
16 0.900044949114232
32 0.793526926027906
};
\addplot [semithick, black, dash pattern=on 2pt off 1pt on 1pt off 0pt]
table {%
4 0.974778780902277
8 0.943031805446336
16 0.858506255795844
32 0.744114306136335
};
\addplot [semithick, black, dash pattern=on 2pt off 1pt on 1pt off 1pt on 1pt off 0pt]
table {%
4 0.982466393178223
8 0.955617282289976
16 0.892414276830122
32 0.783287554998863
};
\addplot [semithick, black, dash pattern=on 2pt off 1pt on 1pt off 1pt on 1pt off 1pt on 1pt off 0pt]
table {%
4 0.978336388521039
8 0.956968513402205
16 0.893445860203745
32 0.782268197436814
};
\end{axis}

\end{tikzpicture}
    \caption{Plane-Boxes}
  \end{subfigure}
  \begin{subfigure}[b]{0.32\linewidth}
    \setlength{\figurewidth}{1.1\linewidth}
    \setlength{\figureheight}{0.9\figurewidth}
\begin{tikzpicture}

\begin{axis}[
height=\figureheight,
tick align=inside,
tick pos=left,
width=\figurewidth,
x grid style={lightgray!92.0261437908!black},
xlabel={Num. Robots},
xmajorgrids,
xmin=2.6, xmax=33.4,
xtick={4,8,16,32},
xtick style={color=black},
y grid style={lightgray!92.0261437908!black},
ylabel={Suboptimality (lower bound)},
ymajorgrids,
ymin=0.692562618329813, ymax=0.999455380917714,
ytick style={color=black}
]
\path [draw=black, fill=black, opacity=0.2, line width=0.0pt]
(axis cs:4,0.985505709890991)
--(axis cs:8,0.951945491318529)
--(axis cs:16,0.87922065846736)
--(axis cs:32,0.800895602577976)
--(axis cs:32,0.788247770024664)
--(axis cs:16,0.871785444117147)
--(axis cs:8,0.946896644506391)
--(axis cs:4,0.983465038486744)
--cycle;
\path [draw=black, fill=black, opacity=0.2, line width=0.0pt, dash pattern=on 2pt off 1pt on 1pt off 0pt]
(axis cs:4,0.97105818144169)
--(axis cs:8,0.925816121304062)
--(axis cs:16,0.828404329560753)
--(axis cs:32,0.722652725561558)
--(axis cs:32,0.706512289356536)
--(axis cs:16,0.813188891503177)
--(axis cs:8,0.91792655533893)
--(axis cs:4,0.963665313116637)
--cycle;
\path [draw=black, fill=black, opacity=0.2, line width=0.0pt, dash pattern=on 2pt off 1pt on 1pt off 1pt on 1pt off 0pt]
(axis cs:4,0.977135088166194)
--(axis cs:8,0.950432468700303)
--(axis cs:16,0.886860830489869)
--(axis cs:32,0.81094573838937)
--(axis cs:32,0.796467303209899)
--(axis cs:16,0.879256225158327)
--(axis cs:8,0.945528631812611)
--(axis cs:4,0.972344772490698)
--cycle;
\path [draw=black, fill=black, opacity=0.2, line width=0.0pt, dash pattern=on 2pt off 1pt on 1pt off 1pt on 1pt off 1pt on 1pt off 0pt]
(axis cs:4,0.980764852112999)
--(axis cs:8,0.946827673858836)
--(axis cs:16,0.885570216982605)
--(axis cs:32,0.80033510898552)
--(axis cs:32,0.786229639679025)
--(axis cs:16,0.879026388502622)
--(axis cs:8,0.942154493935499)
--(axis cs:4,0.976898922780705)
--cycle;
\addplot [semithick, black]
table {%
4 0.984485374188867
8 0.94942106791246
16 0.875503051292254
32 0.79457168630132
};
\addplot [semithick, black, dash pattern=on 2pt off 1pt on 1pt off 0pt]
table {%
4 0.967361747279164
8 0.921871338321496
16 0.820796610531965
32 0.714582507459047
};
\addplot [semithick, black, dash pattern=on 2pt off 1pt on 1pt off 1pt on 1pt off 0pt]
table {%
4 0.974739930328446
8 0.947980550256457
16 0.883058527824098
32 0.803706520799635
};
\addplot [semithick, black, dash pattern=on 2pt off 1pt on 1pt off 1pt on 1pt off 1pt on 1pt off 0pt]
table {%
4 0.978831887446852
8 0.944491083897168
16 0.882298302742614
32 0.793282374332272
};
\end{axis}

\end{tikzpicture}
    \caption{Empty}
  \end{subfigure}
  \begin{subfigure}[b]{0.32\linewidth}
    \setlength{\figurewidth}{1.1\linewidth}
    \setlength{\figureheight}{0.9\figurewidth}
\begin{tikzpicture}

\begin{axis}[
height=\figureheight,
tick align=inside,
tick pos=left,
width=\figurewidth,
x grid style={lightgray!92.0261437908!black},
xlabel={Num. Robots},
xmajorgrids,
xmin=2.6, xmax=33.4,
xtick={4,8,16,32},
xtick style={color=black},
y grid style={lightgray!92.0261437908!black},
ylabel={Suboptimality (lower bound)},
ymajorgrids,
ymin=0.761482608230704, ymax=0.999886989203023,
ytick style={color=black}
]
\path [draw=black, fill=black, opacity=0.2, line width=0.0pt]
(axis cs:4,0.989050426431554)
--(axis cs:8,0.963642695230361)
--(axis cs:16,0.908591872466488)
--(axis cs:32,0.832681730457248)
--(axis cs:32,0.826728390200297)
--(axis cs:16,0.899680368339212)
--(axis cs:8,0.958744716434054)
--(axis cs:4,0.986704010310967)
--cycle;
\path [draw=black, fill=black, opacity=0.2, line width=0.0pt, dash pattern=on 2pt off 1pt on 1pt off 0pt]
(axis cs:4,0.986849812167453)
--(axis cs:8,0.95316283032738)
--(axis cs:16,0.876396278414805)
--(axis cs:32,0.783418124247361)
--(axis cs:32,0.772319171002173)
--(axis cs:16,0.865011950394842)
--(axis cs:8,0.946511701959897)
--(axis cs:4,0.984082638431111)
--cycle;
\path [draw=black, fill=black, opacity=0.2, line width=0.0pt, dash pattern=on 2pt off 1pt on 1pt off 1pt on 1pt off 0pt]
(axis cs:4,0.988741682368799)
--(axis cs:8,0.964232421702813)
--(axis cs:16,0.904385639129174)
--(axis cs:32,0.823449947254179)
--(axis cs:32,0.815733815700559)
--(axis cs:16,0.897201495402442)
--(axis cs:8,0.959069299106235)
--(axis cs:4,0.985939777795143)
--cycle;
\path [draw=black, fill=black, opacity=0.2, line width=0.0pt, dash pattern=on 2pt off 1pt on 1pt off 1pt on 1pt off 1pt on 1pt off 0pt]
(axis cs:4,0.988821458332432)
--(axis cs:8,0.964038578558087)
--(axis cs:16,0.908351874531167)
--(axis cs:32,0.827811391311039)
--(axis cs:32,0.820359776643949)
--(axis cs:16,0.904110433589344)
--(axis cs:8,0.96009649694951)
--(axis cs:4,0.985567868055238)
--cycle;
\addplot [semithick, black]
table {%
4 0.987877218371261
8 0.961193705832208
16 0.90413612040285
32 0.829705060328773
};
\addplot [semithick, black, dash pattern=on 2pt off 1pt on 1pt off 0pt]
table {%
4 0.985466225299282
8 0.949837266143639
16 0.870704114404823
32 0.777868647624767
};
\addplot [semithick, black, dash pattern=on 2pt off 1pt on 1pt off 1pt on 1pt off 0pt]
table {%
4 0.987340730081971
8 0.961650860404524
16 0.900793567265808
32 0.819591881477369
};
\addplot [semithick, black, dash pattern=on 2pt off 1pt on 1pt off 1pt on 1pt off 1pt on 1pt off 0pt]
table {%
4 0.987194663193835
8 0.962067537753798
16 0.906231154060255
32 0.824085583977494
};
\end{axis}

\end{tikzpicture}
    \caption{Skylight}
  \end{subfigure}
  \begin{subfigure}[b]{0.32\linewidth}
    \setlength{\figurewidth}{1.1\linewidth}
    \setlength{\figureheight}{0.9\figurewidth}
\begin{tikzpicture}

\begin{axis}[
height=\figureheight,
tick align=inside,
tick pos=left,
width=\figurewidth,
x grid style={lightgray!92.0261437908!black},
xlabel={Num. Robots},
xmajorgrids,
xmin=2.6, xmax=33.4,
xtick={4,8,16,32},
xtick style={color=black},
y grid style={lightgray!92.0261437908!black},
ylabel={Suboptimality (lower bound)},
ymajorgrids,
ymin=0.844703904258497, ymax=1.00040477604721,
ytick style={color=black}
]
\path [draw=black, fill=black, opacity=0.2, line width=0.0pt]
(axis cs:4,0.991739582588112)
--(axis cs:8,0.976005205461024)
--(axis cs:16,0.935121872854867)
--(axis cs:32,0.867507433009072)
--(axis cs:32,0.862327617942758)
--(axis cs:16,0.931128616926046)
--(axis cs:8,0.971739737261198)
--(axis cs:4,0.99081701279236)
--cycle;
\path [draw=black, fill=black, opacity=0.2, line width=0.0pt, dash pattern=on 2pt off 1pt on 1pt off 1pt on 1pt off 1pt on 1pt off 0pt]
(axis cs:4,0.993327463693177)
--(axis cs:8,0.975570604782392)
--(axis cs:16,0.930124613447741)
--(axis cs:32,0.856450672558353)
--(axis cs:32,0.851781216612529)
--(axis cs:16,0.924512089604038)
--(axis cs:8,0.972720360316847)
--(axis cs:4,0.99219533172419)
--cycle;
\addplot [semithick, black]
table {%
4 0.991278297690236
8 0.973872471361111
16 0.933125244890456
32 0.864917525475915
};
\addplot [semithick, black, dash pattern=on 2pt off 1pt on 1pt off 1pt on 1pt off 1pt on 1pt off 0pt]
table {%
4 0.992761397708684
8 0.97414548254962
16 0.927318351525889
32 0.854115944585441
};
\end{axis}

\end{tikzpicture}
    \vspace{-0.2cm}
    \caption{Office}
  \end{subfigure}
  \caption[Lower bounds on suboptimality for receding-horizon planning for
  exploration]{%
    Lower bounds on suboptimality for receding-horizon planning for exploration
    in each environment.
    Plots provide mean values and standard error
    (\emph{w.r.t.} the standard deviation of trial means)
    for data up to the completion time of each trial.
    These results demonstrate how \rsp{} planning performance approaches that of
    sequential planning with increasing numbers of planning rounds from
    $\numrounds=1$
    to
    $\numrounds=6$.
    We exclude the Myopic planner as the suboptimality bounds are not
    directly comparable to those of the other planners because the choice of
    parameters (Table~\ref{tab:exploration_performance})
    changes the objective values.
    Instead,
    \rsp[1]{} also plans myopically but has parameters that are comparable to
    other planner configurations, and
    Table~\ref{tab:suboptimality_exploration} includes data for all planners.
  }%
  \label{fig:exploration_suboptimality}
\end{figure}
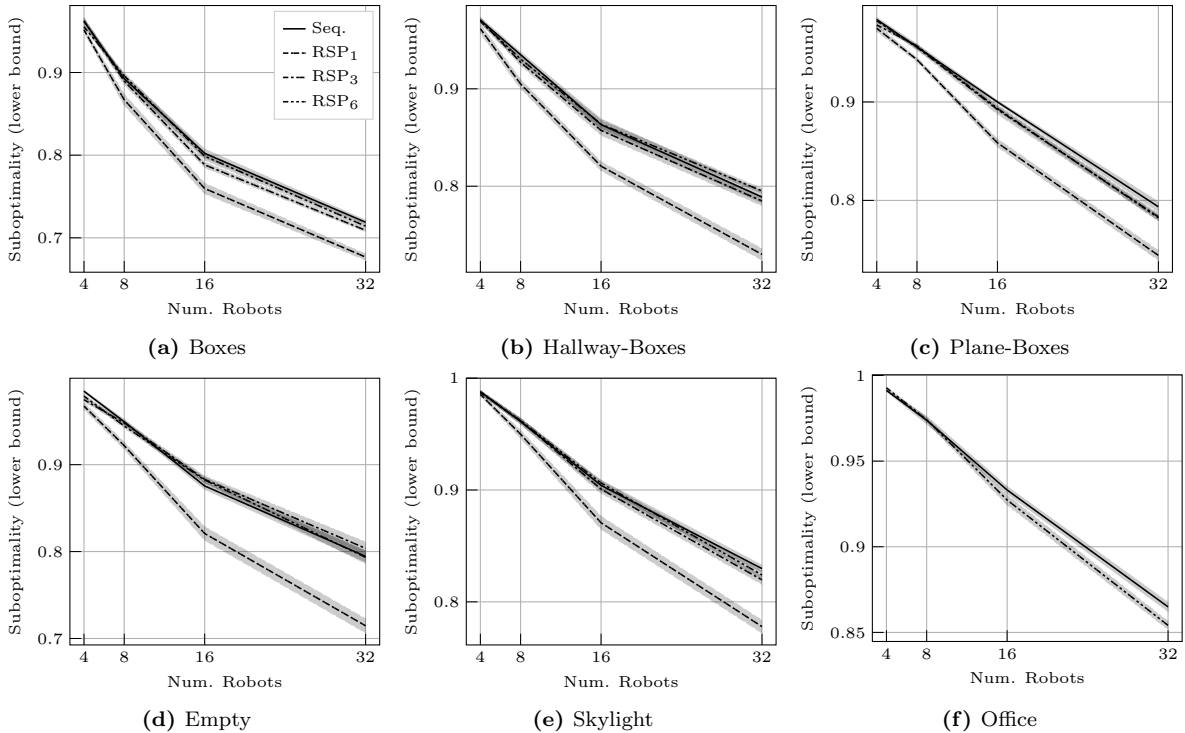

\begin{table}
  \caption[Lower bounds on suboptimality for receding-horizon planning]
  {%
    Lower bounds on suboptimality for receding-horizon planning for
    exploration in each environment.
    See Fig.~\ref{fig:exploration_suboptimality} for more detail.
  }\label{tab:suboptimality_exploration}
  \resizebox{0.49\linewidth}{!}{
    \tiny
    \setlength{\tabcolsep}{3pt}
    \pgfplotstabletypeset[
      every head row/.style={
        output empty row,
        before row={
          \textbf{Num. Robot}
          &
          \multicolumn{2}{c}{\textbf{Myopic}}
          &
          \multicolumn{2}{c}{\textbf{Sequential}}
          &
          \multicolumn{2}{c}{\rsp[1][\textbf]}
          &
          \multicolumn{2}{c}{\rsp[3][\textbf]}
          &
          \multicolumn{2}{c}{\rsp[6][\textbf]}
          \\
          &Avg.&Std.&Avg.&Std.&Avg.&Std.&Avg.&Std.&Avg.&Std.
          \\
        },
        after row=\belowtoprule
      },
      every row no 0/.style={before row={
          \multicolumn{2}{l}{\quad\textbf{Boxes}}\vspace{3pt}\\
      }},
      every row no 4/.style={before row={
          \multicolumn{2}{l}{\quad\textbf{Hallway-Boxes}}\vspace{3pt}\\
      }},
      every row no 8/.style={before row={
          \multicolumn{2}{l}{\quad\textbf{Plane-Boxes}}\vspace{3pt}\\
      }},
      every row no 12/.style={before row={
          \multicolumn{2}{l}{\quad\textbf{Empty}}\vspace{3pt}\\
      }},
      every row no 16/.style={before row={
          \multicolumn{2}{l}{\quad\textbf{Skylight}}\vspace{3pt}\\
      }},
      every row no 20/.style={before row={
          \multicolumn{2}{l}{\quad\textbf{Office}}\vspace{3pt}\\
      }},
      every odd column/.style={
        column type={r},
        fixed, precision=2, zerofill, 
      },
      every even column/.style={
        column type={l},
        fixed, precision=3, zerofill,
      },
      display columns/0/.style={
        column type={c},
        precision=0
      },
      col sep=comma,
      skip rows between index={12}{24},
      empty cells with={--} 
    ]{\datapath/suboptimality_bounds.csv}
  }
  \resizebox{0.49\linewidth}{!}{
    \tiny
    \setlength{\tabcolsep}{3pt}
    \pgfplotstabletypeset[
      every head row/.style={
        output empty row,
        before row={
          \textbf{Num. Robot}
          &
          \multicolumn{2}{c}{\textbf{Myopic}}
          &
          \multicolumn{2}{c}{\textbf{Sequential}}
          &
          \multicolumn{2}{c}{\rsp[1][\textbf]}
          &
          \multicolumn{2}{c}{\rsp[3][\textbf]}
          &
          \multicolumn{2}{c}{\rsp[6][\textbf]}
          \\
          &Avg.&Std.&Avg.&Std.&Avg.&Std.&Avg.&Std.&Avg.&Std.
          \\
        },
        after row=\belowtoprule
      },
      every row no 0/.style={before row={
          \multicolumn{2}{l}{\quad\textbf{Empty}}\vspace{3pt}\\
      }},
      every row no 4/.style={before row={
          \multicolumn{2}{l}{\quad\textbf{Skylight}}\vspace{3pt}\\
      }},
      every row no 8/.style={before row={
          \multicolumn{2}{l}{\quad\textbf{Office}}\vspace{3pt}\\
      }},
      every odd column/.style={
        column type={r},
        fixed, precision=2, zerofill, 
      },
      every even column/.style={
        column type={l},
        fixed, precision=3, zerofill,
      },
      display columns/0/.style={
        column type={c},
        precision=0
      },
      col sep=comma,
      skip rows between index={0}{12},
      empty cells with={--} 
    ]{\datapath/suboptimality_bounds.csv}
  }
\end{table}

Figure~\ref{fig:exploration_suboptimality} presents results on mean values of
the lower bound\footnote{
  Bounds are computed approximately via MCTS (for single-robot planning)
  and are only representative of suboptimality in multi-robot coordination.
}
on suboptimality
(the greater of the two bounds in Sec.~\ref{sec:online_bounds_exploration}),
and Table~\ref{tab:suboptimality_exploration} lists this data as well.
These plots clearly demonstrate how the suboptimality of \rsp{} planning
approaches that of sequential planning (Alg.~\ref{alg:local_greedy}) with
increasing numbers of planning rounds ($\numrounds$)
as the performance bounds for these planners would suggest\footnote{
  Strictly, the bounds for \rsp{} planning only establish convergence to the
  same worst case suboptimality as sequential planning (1/2), but we expect
  comparable suboptimality in practice.
}
(Theorem~\ref{theorem:adaptive_partitioning}).
Moreover, we see that actual suboptimality is consistently better than the worst
case bounds (even approaching one for fewer robots)
which is consistent with observations from related
works~\citep{leskovec2007,golovin2011jair,krause2008}.
Likewise, the performance gaps widen with increasing suboptimality and larger
numbers of robots
(with the difference reaching 8\% for in Empty).
The decrease in these bounds with increasing numbers of robots is representative
of robots operating in closer proximity and with overlapping observations over
the planning horizon (see \eqref{eq:receding_horizon_exploration}).

\subsection{Early progress and completion times by planner}
\label{sec:completion_by_planner}

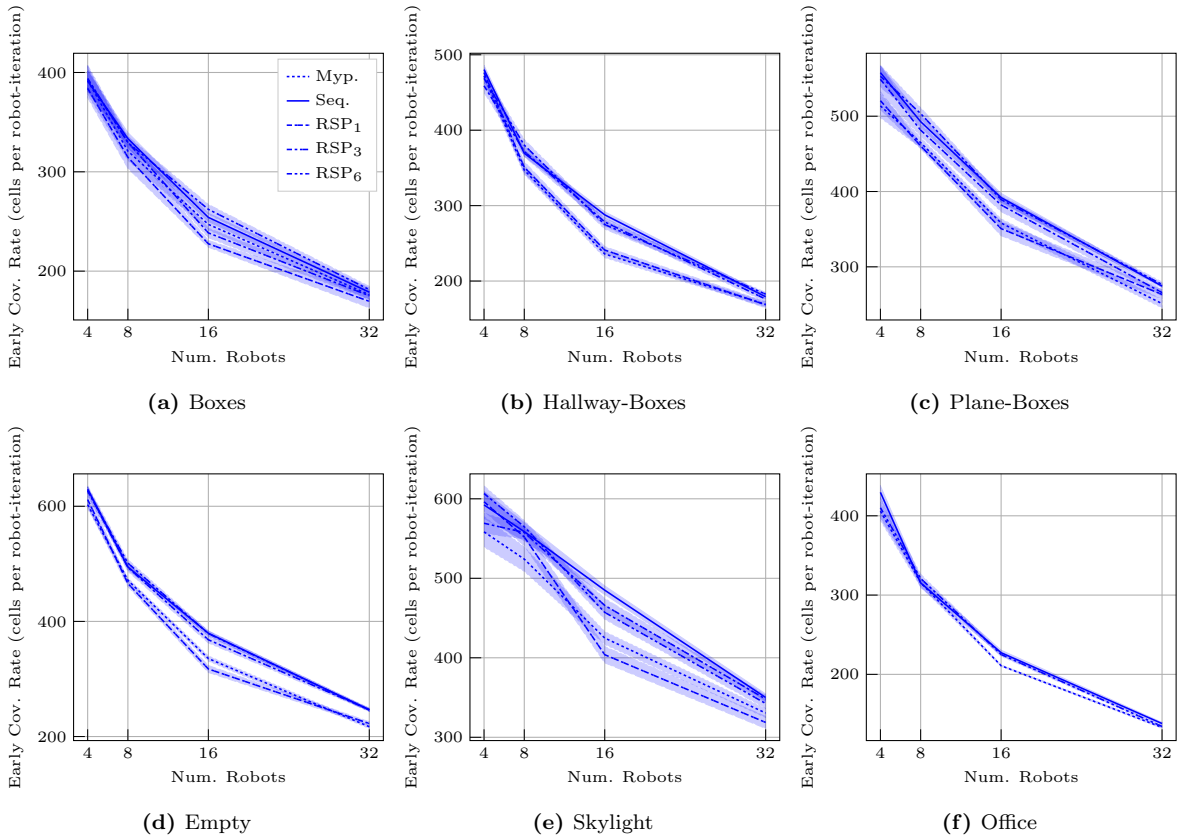
\begin{figure}
  \tiny
  \begin{subfigure}[b]{0.32\linewidth}
    \setlength{\figurewidth}{1.1\linewidth}
    \setlength{\figureheight}{0.9\figurewidth}
\begin{tikzpicture}

\begin{axis}[
height=\figureheight,
legend cell align={left},
legend columns=1,
legend style={draw=white!80.0!black},
legend image post style={scale=0.5}, 
tick align=inside,
tick pos=left,
width=\figurewidth,
x grid style={lightgray!92.0261437908!black},
xlabel={Num. Robots},
xmajorgrids,
xmin=2.6, xmax=33.4,
xtick={4,8,16,32},
xtick style={color=black},
y grid style={lightgray!92.0261437908!black},
ylabel={Early Cov. Rate (cells per robot-iteration)},
ymajorgrids,
ymin=150.835557787158, ymax=419.34715404441,
ytick style={color=black}
]
\path [draw=blue, fill=blue, opacity=0.2, line width=0.0pt, dash pattern=on 1pt off 1pt]
(axis cs:4,406.025303767122)
--(axis cs:8,330.003032151373)
--(axis cs:16,252.588672418127)
--(axis cs:32,180.989380729743)
--(axis cs:32,171.734850565152)
--(axis cs:16,241.527641167396)
--(axis cs:8,311.640864302991)
--(axis cs:4,377.879158444442)
--cycle;
\path [draw=blue, fill=blue, opacity=0.2, line width=0.0pt]
(axis cs:4,401.330979746092)
--(axis cs:8,339.467045739769)
--(axis cs:16,258.241899707259)
--(axis cs:32,182.870396710097)
--(axis cs:32,174.806220742626)
--(axis cs:16,249.964766381598)
--(axis cs:8,327.21677691174)
--(axis cs:4,381.402343255729)
--cycle;
\path [draw=blue, fill=blue, opacity=0.2, line width=0.0pt, dash pattern=on 2pt off 1pt on 1pt off 0pt]
(axis cs:4,393.301588961371)
--(axis cs:8,325.667980188408)
--(axis cs:16,230.923223717206)
--(axis cs:32,176.155426280188)
--(axis cs:32,163.040630344306)
--(axis cs:16,224.128718059205)
--(axis cs:8,303.559386902016)
--(axis cs:4,374.724565264765)
--cycle;
\path [draw=blue, fill=blue, opacity=0.2, line width=0.0pt, dash pattern=on 2pt off 1pt on 1pt off 1pt on 1pt off 0pt]
(axis cs:4,403.148760756815)
--(axis cs:8,334.666027343533)
--(axis cs:16,243.235443079407)
--(axis cs:32,178.984702704897)
--(axis cs:32,171.475760009638)
--(axis cs:16,233.72369863453)
--(axis cs:8,324.763392069369)
--(axis cs:4,383.200304928487)
--cycle;
\path [draw=blue, fill=blue, opacity=0.2, line width=0.0pt, dash pattern=on 2pt off 1pt on 1pt off 1pt on 1pt off 1pt on 1pt off 0pt]
(axis cs:4,407.142081487262)
--(axis cs:8,335.591811535698)
--(axis cs:16,267.150269198492)
--(axis cs:32,184.64958567001)
--(axis cs:32,178.210478661895)
--(axis cs:16,257.254208767354)
--(axis cs:8,323.305861567847)
--(axis cs:4,380.966366751042)
--cycle;
\addplot [semithick, blue, dash pattern=on 1pt off 1pt]
table {%
4 391.952231105782
8 320.821948227182
16 247.058156792761
32 176.362115647447
};
\addlegendentry{Myp.}
\addplot [semithick, blue]
table {%
4 391.366661500911
8 333.341911325754
16 254.103333044428
32 178.838308726362
};
\addlegendentry{Seq.}
\addplot [semithick, blue, dash pattern=on 2pt off 1pt on 1pt off 0pt]
table {%
4 384.013077113068
8 314.613683545212
16 227.525970888205
32 169.598028312247
};
\addlegendentry{RSP$_1$}
\addplot [semithick, blue, dash pattern=on 2pt off 1pt on 1pt off 1pt on 1pt off 0pt]
table {%
4 393.174532842651
8 329.714709706451
16 238.479570856968
32 175.230231357268
};
\addlegendentry{RSP$_3$}
\addplot [semithick, blue, dash pattern=on 2pt off 1pt on 1pt off 1pt on 1pt off 1pt on 1pt off 0pt]
table {%
4 394.054224119152
8 329.448836551772
16 262.202238982923
32 181.430032165953
};
\addlegendentry{RSP$_6$}
\end{axis}

\end{tikzpicture}
    \caption{Boxes}
  \end{subfigure}
  \begin{subfigure}[b]{0.32\linewidth}
    \setlength{\figurewidth}{1.1\linewidth}
    \setlength{\figureheight}{0.9\figurewidth}
\begin{tikzpicture}

\begin{axis}[
height=\figureheight,
tick align=inside,
tick pos=left,
width=\figurewidth,
x grid style={lightgray!92.0261437908!black},
xlabel={Num. Robots},
xmajorgrids,
xmin=2.6, xmax=33.4,
xtick={4,8,16,32},
xtick style={color=black},
y grid style={lightgray!92.0261437908!black},
ylabel={Early Cov. Rate (cells per robot-iteration)},
ymajorgrids,
ymin=148.387535457923, ymax=502.01431556101,
ytick style={color=black}
]
\path [draw=blue, fill=blue, opacity=0.2, line width=0.0pt, dash pattern=on 1pt off 1pt]
(axis cs:4,479.854833887333)
--(axis cs:8,352.792250782154)
--(axis cs:16,240.780540070001)
--(axis cs:32,172.004482661389)
--(axis cs:32,166.012359595595)
--(axis cs:16,230.947041700093)
--(axis cs:8,341.518173572474)
--(axis cs:4,464.001640419867)
--cycle;
\path [draw=blue, fill=blue, opacity=0.2, line width=0.0pt]
(axis cs:4,485.94037101087)
--(axis cs:8,373.313336999472)
--(axis cs:16,291.800464603777)
--(axis cs:32,181.989441286904)
--(axis cs:32,176.346338985668)
--(axis cs:16,283.752135661484)
--(axis cs:8,365.547098688045)
--(axis cs:4,474.844269455476)
--cycle;
\path [draw=blue, fill=blue, opacity=0.2, line width=0.0pt, dash pattern=on 2pt off 1pt on 1pt off 0pt]
(axis cs:4,483.060473938793)
--(axis cs:8,358.270115990422)
--(axis cs:16,245.603028099375)
--(axis cs:32,172.655142518545)
--(axis cs:32,164.461480008063)
--(axis cs:16,236.648050796479)
--(axis cs:8,343.520294213304)
--(axis cs:4,469.694152218055)
--cycle;
\path [draw=blue, fill=blue, opacity=0.2, line width=0.0pt, dash pattern=on 2pt off 1pt on 1pt off 1pt on 1pt off 0pt]
(axis cs:4,467.753242362337)
--(axis cs:8,378.6072392564)
--(axis cs:16,283.594448228504)
--(axis cs:32,179.535905060304)
--(axis cs:32,172.56297601265)
--(axis cs:16,273.125599489964)
--(axis cs:8,365.138893917255)
--(axis cs:4,450.093570750241)
--cycle;
\path [draw=blue, fill=blue, opacity=0.2, line width=0.0pt, dash pattern=on 2pt off 1pt on 1pt off 1pt on 1pt off 1pt on 1pt off 0pt]
(axis cs:4,476.037986039159)
--(axis cs:8,387.66237287876)
--(axis cs:16,280.791426789792)
--(axis cs:32,184.84776198602)
--(axis cs:32,180.456812959203)
--(axis cs:16,268.967054584342)
--(axis cs:8,372.817656514502)
--(axis cs:4,461.552366018758)
--cycle;
\addplot [semithick, blue, dash pattern=on 1pt off 1pt]
table {%
4 471.9282371536
8 347.155212177314
16 235.863790885047
32 169.008421128492
};
\addplot [semithick, blue]
table {%
4 480.392320233173
8 369.430217843758
16 287.77630013263
32 179.167890136286
};
\addplot [semithick, blue, dash pattern=on 2pt off 1pt on 1pt off 0pt]
table {%
4 476.377313078424
8 350.895205101863
16 241.125539447927
32 168.558311263304
};
\addplot [semithick, blue, dash pattern=on 2pt off 1pt on 1pt off 1pt on 1pt off 0pt]
table {%
4 458.923406556289
8 371.873066586827
16 278.360023859234
32 176.049440536477
};
\addplot [semithick, blue, dash pattern=on 2pt off 1pt on 1pt off 1pt on 1pt off 1pt on 1pt off 0pt]
table {%
4 468.795176028959
8 380.240014696631
16 274.879240687067
32 182.652287472612
};
\end{axis}

\end{tikzpicture}
    \caption{Hallway-Boxes}
  \end{subfigure}
  \begin{subfigure}[b]{0.32\linewidth}
    \setlength{\figurewidth}{1.1\linewidth}
    \setlength{\figureheight}{0.9\figurewidth}
\begin{tikzpicture}

\begin{axis}[
height=\figureheight,
tick align=inside,
tick pos=left,
width=\figurewidth,
x grid style={lightgray!92.0261437908!black},
xlabel={Num. Robots},
xmajorgrids,
xmin=2.6, xmax=33.4,
xtick={4,8,16,32},
xtick style={color=black},
y grid style={lightgray!92.0261437908!black},
ylabel={Early Cov. Rate (cells per robot-iteration)},
ymajorgrids,
ymin=229.430467127792, ymax=583.582017974424,
ytick style={color=black}
]
\path [draw=blue, fill=blue, opacity=0.2, line width=0.0pt, dash pattern=on 1pt off 1pt]
(axis cs:4,529.429859379003)
--(axis cs:8,471.670161895356)
--(axis cs:16,363.616323771144)
--(axis cs:32,256.759381584423)
--(axis cs:32,245.528264893548)
--(axis cs:16,352.724722570031)
--(axis cs:8,458.170409296666)
--(axis cs:4,498.004053140239)
--cycle;
\path [draw=blue, fill=blue, opacity=0.2, line width=0.0pt]
(axis cs:4,566.277462842558)
--(axis cs:8,500.050049689801)
--(axis cs:16,397.526406840742)
--(axis cs:32,276.504008710593)
--(axis cs:32,272.290370635175)
--(axis cs:16,386.31666111987)
--(axis cs:8,485.445974011807)
--(axis cs:4,549.363257896779)
--cycle;
\path [draw=blue, fill=blue, opacity=0.2, line width=0.0pt, dash pattern=on 2pt off 1pt on 1pt off 0pt]
(axis cs:4,532.360489896414)
--(axis cs:8,464.037723139003)
--(axis cs:16,360.011136022819)
--(axis cs:32,269.254502710256)
--(axis cs:32,257.433178377557)
--(axis cs:16,341.38003304451)
--(axis cs:8,457.028370727396)
--(axis cs:4,508.868120586736)
--cycle;
\path [draw=blue, fill=blue, opacity=0.2, line width=0.0pt, dash pattern=on 2pt off 1pt on 1pt off 1pt on 1pt off 0pt]
(axis cs:4,566.622464087138)
--(axis cs:8,489.336874195885)
--(axis cs:16,390.982767600363)
--(axis cs:32,270.044301059262)
--(axis cs:32,261.600743606292)
--(axis cs:16,373.373813575672)
--(axis cs:8,473.711393822827)
--(axis cs:4,531.367745258132)
--cycle;
\path [draw=blue, fill=blue, opacity=0.2, line width=0.0pt, dash pattern=on 2pt off 1pt on 1pt off 1pt on 1pt off 1pt on 1pt off 0pt]
(axis cs:4,567.484220208668)
--(axis cs:8,511.198458445218)
--(axis cs:16,393.976196954534)
--(axis cs:32,279.771010842432)
--(axis cs:32,273.393693415613)
--(axis cs:16,384.563129754697)
--(axis cs:8,495.792393753133)
--(axis cs:4,538.599444790804)
--cycle;
\addplot [semithick, blue, dash pattern=on 1pt off 1pt]
table {%
4 513.716956259621
8 464.920285596011
16 358.170523170588
32 251.143823238985
};
\addplot [semithick, blue]
table {%
4 557.820360369668
8 492.748011850804
16 391.921533980306
32 274.397189672884
};
\addplot [semithick, blue, dash pattern=on 2pt off 1pt on 1pt off 0pt]
table {%
4 520.614305241575
8 460.533046933199
16 350.695584533665
32 263.343840543907
};
\addplot [semithick, blue, dash pattern=on 2pt off 1pt on 1pt off 1pt on 1pt off 0pt]
table {%
4 548.995104672635
8 481.524134009356
16 382.178290588017
32 265.822522332777
};
\addplot [semithick, blue, dash pattern=on 2pt off 1pt on 1pt off 1pt on 1pt off 1pt on 1pt off 0pt]
table {%
4 553.041832499736
8 503.495426099176
16 389.269663354615
32 276.582352129022
};
\end{axis}

\end{tikzpicture}
    \caption{Plane-Boxes}
  \end{subfigure}
  \begin{subfigure}[b]{0.32\linewidth}
    \setlength{\figurewidth}{1.1\linewidth}
    \setlength{\figureheight}{0.9\figurewidth}
\begin{tikzpicture}

\begin{axis}[
height=\figureheight,
tick align=inside,
tick pos=left,
width=\figurewidth,
x grid style={lightgray!92.0261437908!black},
xlabel={Num. Robots},
xmajorgrids,
xmin=2.6, xmax=33.4,
xtick={4,8,16,32},
xtick style={color=black},
y grid style={lightgray!92.0261437908!black},
ylabel={Early Cov. Rate (cells per robot-iteration)},
ymajorgrids,
ymin=193.017133804998, ymax=656.281810694249,
ytick style={color=black}
]
\path [draw=blue, fill=blue, opacity=0.2, line width=0.0pt, dash pattern=on 1pt off 1pt]
(axis cs:4,609.429933983183)
--(axis cs:8,475.197764776998)
--(axis cs:16,339.942193307484)
--(axis cs:32,220.024234689349)
--(axis cs:32,214.074619118146)
--(axis cs:16,331.024460376872)
--(axis cs:8,467.875979834501)
--(axis cs:4,596.101976463404)
--cycle;
\path [draw=blue, fill=blue, opacity=0.2, line width=0.0pt]
(axis cs:4,632.874206702998)
--(axis cs:8,498.839426629749)
--(axis cs:16,383.454109022475)
--(axis cs:32,248.803523538587)
--(axis cs:32,244.184898502613)
--(axis cs:16,375.48294174561)
--(axis cs:8,490.227451393411)
--(axis cs:4,626.598507537944)
--cycle;
\path [draw=blue, fill=blue, opacity=0.2, line width=0.0pt, dash pattern=on 2pt off 1pt on 1pt off 0pt]
(axis cs:4,623.800564290163)
--(axis cs:8,470.930145511457)
--(axis cs:16,323.364713587916)
--(axis cs:32,225.747334023724)
--(axis cs:32,219.506545269876)
--(axis cs:16,311.374148345673)
--(axis cs:8,460.76799751688)
--(axis cs:4,599.565303446529)
--cycle;
\path [draw=blue, fill=blue, opacity=0.2, line width=0.0pt, dash pattern=on 2pt off 1pt on 1pt off 1pt on 1pt off 0pt]
(axis cs:4,633.344248537012)
--(axis cs:8,500.320673052249)
--(axis cs:16,373.557257187328)
--(axis cs:32,248.982989889695)
--(axis cs:32,244.022716612792)
--(axis cs:16,362.619539138906)
--(axis cs:8,487.045060046091)
--(axis cs:4,622.392833443075)
--cycle;
\path [draw=blue, fill=blue, opacity=0.2, line width=0.0pt, dash pattern=on 2pt off 1pt on 1pt off 1pt on 1pt off 1pt on 1pt off 0pt]
(axis cs:4,635.224325381101)
--(axis cs:8,507.09686740853)
--(axis cs:16,382.973709847493)
--(axis cs:32,249.113386833464)
--(axis cs:32,243.415566108141)
--(axis cs:16,374.600248705656)
--(axis cs:8,495.591288749075)
--(axis cs:4,618.121432303232)
--cycle;
\addplot [semithick, blue, dash pattern=on 1pt off 1pt]
table {%
4 602.765955223294
8 471.53687230575
16 335.483326842178
32 217.049426903747
};
\addplot [semithick, blue]
table {%
4 629.736357120471
8 494.53343901158
16 379.468525384043
32 246.4942110206
};
\addplot [semithick, blue, dash pattern=on 2pt off 1pt on 1pt off 0pt]
table {%
4 611.682933868346
8 465.849071514168
16 317.369430966795
32 222.6269396468
};
\addplot [semithick, blue, dash pattern=on 2pt off 1pt on 1pt off 1pt on 1pt off 0pt]
table {%
4 627.868540990043
8 493.68286654917
16 368.088398163117
32 246.502853251244
};
\addplot [semithick, blue, dash pattern=on 2pt off 1pt on 1pt off 1pt on 1pt off 1pt on 1pt off 0pt]
table {%
4 626.672878842167
8 501.344078078803
16 378.786979276574
32 246.264476470803
};
\end{axis}

\end{tikzpicture}
    \caption{Empty}
  \end{subfigure}
  \begin{subfigure}[b]{0.32\linewidth}
    \setlength{\figurewidth}{1.1\linewidth}
    \setlength{\figureheight}{0.9\figurewidth}
\begin{tikzpicture}

\begin{axis}[
height=\figureheight,
tick align=inside,
tick pos=left,
width=\figurewidth,
x grid style={lightgray!92.0261437908!black},
xlabel={Num. Robots},
xmajorgrids,
xmin=2.6, xmax=33.4,
xtick={4,8,16,32},
xtick style={color=black},
y grid style={lightgray!92.0261437908!black},
ylabel={Early Cov. Rate (cells per robot-iteration)},
ymajorgrids,
ymin=295.872511934403, ymax=631.372693288249,
ytick style={color=black}
]
\path [draw=blue, fill=blue, opacity=0.2, line width=0.0pt, dash pattern=on 1pt off 1pt]
(axis cs:4,577.495378376985)
--(axis cs:8,539.409112125976)
--(axis cs:16,432.904017032742)
--(axis cs:32,336.061557070944)
--(axis cs:32,324.99417723809)
--(axis cs:16,416.789885814235)
--(axis cs:8,508.853539237839)
--(axis cs:4,539.435909980671)
--cycle;
\path [draw=blue, fill=blue, opacity=0.2, line width=0.0pt]
(axis cs:4,609.191379608259)
--(axis cs:8,569.970644309579)
--(axis cs:16,490.382124551363)
--(axis cs:32,354.503557363196)
--(axis cs:32,345.547384393895)
--(axis cs:16,479.706329746651)
--(axis cs:8,547.438056043631)
--(axis cs:4,575.45287254872)
--cycle;
\path [draw=blue, fill=blue, opacity=0.2, line width=0.0pt, dash pattern=on 2pt off 1pt on 1pt off 0pt]
(axis cs:4,607.988052616907)
--(axis cs:8,562.864845821338)
--(axis cs:16,414.075896375286)
--(axis cs:32,326.253711155105)
--(axis cs:32,311.122520177759)
--(axis cs:16,393.664125942535)
--(axis cs:8,541.103106709522)
--(axis cs:4,584.556514204072)
--cycle;
\path [draw=blue, fill=blue, opacity=0.2, line width=0.0pt, dash pattern=on 2pt off 1pt on 1pt off 1pt on 1pt off 0pt]
(axis cs:4,581.991390934946)
--(axis cs:8,565.691082347848)
--(axis cs:16,474.540347393934)
--(axis cs:32,352.945706587659)
--(axis cs:32,343.218767369571)
--(axis cs:16,457.259963420323)
--(axis cs:8,549.611782096145)
--(axis cs:4,556.505135787821)
--cycle;
\path [draw=blue, fill=blue, opacity=0.2, line width=0.0pt, dash pattern=on 2pt off 1pt on 1pt off 1pt on 1pt off 1pt on 1pt off 0pt]
(axis cs:4,616.122685044893)
--(axis cs:8,572.573051020063)
--(axis cs:16,465.761246375949)
--(axis cs:32,348.44824618666)
--(axis cs:32,336.507969520028)
--(axis cs:16,448.478081381693)
--(axis cs:8,557.18547218301)
--(axis cs:4,597.112382246744)
--cycle;
\addplot [semithick, blue, dash pattern=on 1pt off 1pt]
table {%
4 558.465644178828
8 524.131325681907
16 424.846951423488
32 330.527867154517
};
\addplot [semithick, blue]
table {%
4 592.32212607849
8 558.704350176605
16 485.044227149007
32 350.025470878546
};
\addplot [semithick, blue, dash pattern=on 2pt off 1pt on 1pt off 0pt]
table {%
4 596.27228341049
8 551.98397626543
16 403.870011158911
32 318.688115666432
};
\addplot [semithick, blue, dash pattern=on 2pt off 1pt on 1pt off 1pt on 1pt off 0pt]
table {%
4 569.248263361384
8 557.651432221996
16 465.900155407129
32 348.082236978615
};
\addplot [semithick, blue, dash pattern=on 2pt off 1pt on 1pt off 1pt on 1pt off 1pt on 1pt off 0pt]
table {%
4 606.617533645819
8 564.879261601537
16 457.119663878821
32 342.478107853344
};
\end{axis}

\end{tikzpicture}
    \caption{Skylight}
  \end{subfigure}
  \begin{subfigure}[b]{0.32\linewidth}
    \setlength{\figurewidth}{1.1\linewidth}
    \setlength{\figureheight}{0.9\figurewidth}
\begin{tikzpicture}

\begin{axis}[
height=\figureheight,
tick align=inside,
tick pos=left,
width=\figurewidth,
x grid style={lightgray!92.0261437908!black},
xlabel={Num. Robots},
xmajorgrids,
xmin=2.6, xmax=33.4,
xtick={4,8,16,32},
xtick style={color=black},
y grid style={lightgray!92.0261437908!black},
ylabel={Early Cov. Rate (cells per robot-iteration)},
ymajorgrids,
ymin=116.471471734956, ymax=452.902624451101,
ytick style={color=black}
]
\path [draw=blue, fill=blue, opacity=0.2, line width=0.0pt, dash pattern=on 1pt off 1pt]
(axis cs:4,417.653844447662)
--(axis cs:8,320.497851553134)
--(axis cs:16,211.910142335176)
--(axis cs:32,134.445723705414)
--(axis cs:32,131.763796858417)
--(axis cs:16,209.931752930927)
--(axis cs:8,310.736785566461)
--(axis cs:4,394.230239399814)
--cycle;
\path [draw=blue, fill=blue, opacity=0.2, line width=0.0pt]
(axis cs:4,437.61029932764)
--(axis cs:8,320.938910893315)
--(axis cs:16,230.463443466398)
--(axis cs:32,139.390861869592)
--(axis cs:32,136.506190809485)
--(axis cs:16,224.123156182175)
--(axis cs:8,309.803461961324)
--(axis cs:4,421.543665037226)
--cycle;
\path [draw=blue, fill=blue, opacity=0.2, line width=0.0pt, dash pattern=on 2pt off 1pt on 1pt off 1pt on 1pt off 1pt on 1pt off 0pt]
(axis cs:4,422.90783146409)
--(axis cs:8,326.24371036102)
--(axis cs:16,227.315975424154)
--(axis cs:32,135.344967540446)
--(axis cs:32,133.321993117212)
--(axis cs:16,222.890875335593)
--(axis cs:8,315.757226622667)
--(axis cs:4,397.508499630879)
--cycle;
\addplot [semithick, blue, dash pattern=on 1pt off 1pt]
table {%
4 405.942041923738
8 315.617318559798
16 210.920947633052
32 133.104760281915
};
\addplot [semithick, blue]
table {%
4 429.576982182433
8 315.37118642732
16 227.293299824286
32 137.948526339539
};
\addplot [semithick, blue, dash pattern=on 2pt off 1pt on 1pt off 1pt on 1pt off 1pt on 1pt off 0pt]
table {%
4 410.208165547484
8 321.000468491844
16 225.103425379874
32 134.333480328829
};
\end{axis}

\end{tikzpicture}
    \caption{Office}
  \end{subfigure}
  \caption[Coverage rates per robot up to the early progress threshold]{%
    Coverage rates per robot up to the early progress threshold are generally
    similar across planners without many significant differences.
    Still, the differences that do exist consistently highlight deficiencies in
    the Myopic and \rsp[1] configurations
    (each plans myopically with different parameters).
    Shaded regions depict standard error.
  }%
  \label{fig:early_coverage_rate}
\end{figure}

\begin{table}
  \caption[
    Completion times and times for reaching the early progress threshold for
    each environment and planner.
  ]
  {%
    Completion times (in robot-iterations) and times for reaching the early
    progress threshold
    for each environment and planner (see Sec.~\ref{sec:completion_and_rates}).
  }\label{tab:completion_times}
  \begin{subfigure}[t]{0.49\linewidth}\resizebox{\linewidth}{!}{
      \tiny
      \setlength{\tabcolsep}{3pt}
      \pgfplotstabletypeset[
        every head row/.style={
          output empty row,
          before row={
            \textbf{Num. Robot}
            &
            \multicolumn{2}{c}{\textbf{Myopic}}
            &
            \multicolumn{2}{c}{\textbf{Sequential}}
            &
            \multicolumn{2}{c}{\rsp[1][\textbf]}
            &
            \multicolumn{2}{c}{\rsp[3][\textbf]}
            &
            \multicolumn{2}{c}{\rsp[6][\textbf]}
            \\
            &Avg.&Std.&Avg.&Std.&Avg.&Std.&Avg.&Std.&Avg.&Std.
            \\
          },
          after row=\belowtoprule
        },
        every row no 0/.style={before row={
            \multicolumn{2}{l}{\quad\textbf{Boxes}}\vspace{3pt}\\
        }},
        every row no 4/.style={before row={
            \multicolumn{2}{l}{\quad\textbf{Hallway-Boxes}}\vspace{3pt}\\
        }},
        every row no 8/.style={before row={
            \multicolumn{2}{l}{\quad\textbf{Plane-Boxes}}\vspace{3pt}\\
        }},
        every row no 12/.style={before row={
            \multicolumn{2}{l}{\quad\textbf{Empty}}\vspace{3pt}\\
        }},
        every row no 16/.style={before row={
            \multicolumn{2}{l}{\quad\textbf{Skylight}}\vspace{3pt}\\
        }},
        every row no 20/.style={before row={
            \multicolumn{2}{l}{\quad\textbf{Office}}\vspace{3pt}\\
        }},
        every odd column/.style={
          column type={r}
        },
        every even column/.style={
          column type={l}
        },
        display columns/0/.style={column type={c}},
        col sep=comma,
        precision=0,
        empty cells with={--} 
      ]{\datapath/early_progress_times.csv}
    }
    \caption{Early progress times}
  \end{subfigure}
  \begin{subfigure}[t]{0.49\linewidth}\resizebox{\linewidth}{!}{
      \tiny
      \setlength{\tabcolsep}{3pt}
      \pgfplotstabletypeset[
        every head row/.style={
          output empty row,
          before row={
            \textbf{Num. Robot}
            &
            \multicolumn{2}{c}{\textbf{Myopic}}
            &
            \multicolumn{2}{c}{\textbf{Sequential}}
            &
            \multicolumn{2}{c}{\rsp[1][\textbf]}
            &
            \multicolumn{2}{c}{\rsp[3][\textbf]}
            &
            \multicolumn{2}{c}{\rsp[6][\textbf]}
            \\
            &Avg.&Std.&Avg.&Std.&Avg.&Std.&Avg.&Std.&Avg.&Std.
            \\
          },
          after row=\belowtoprule
        },
        every row no 0/.style={before row={
            \multicolumn{2}{l}{\quad\textbf{Boxes}}\vspace{3pt}\\
        }},
        every row no 4/.style={before row={
            \multicolumn{2}{l}{\quad\textbf{Hallway-Boxes}}\vspace{3pt}\\
        }},
        every row no 8/.style={before row={
            \multicolumn{2}{l}{\quad\textbf{Plane-Boxes}}\vspace{3pt}\\
        }},
        every row no 12/.style={before row={
            \multicolumn{2}{l}{\quad\textbf{Empty}}\vspace{3pt}\\
        }},
        every row no 16/.style={before row={
            \multicolumn{2}{l}{\quad\textbf{Skylight}}\vspace{3pt}\\
        }},
        every row no 20/.style={before row={
            \multicolumn{2}{l}{\quad\textbf{Office}}\vspace{3pt}\\
        }},
        every odd column/.style={
          column type={r}
        },
        every even column/.style={
          column type={l}
        },
        display columns/0/.style={column type={c}},
        col sep=comma,
        precision=0,
        empty cells with={--} 
      ]{\datapath/completion_times.csv}
    }
    \caption{Completion times}
  \end{subfigure}
\end{table}

The picture for planning performance becomes more complex when we look at the
exploration process which is affected by factors such as the
design of the exploration objective \eqref{eq:exploration_objective},
(including the view and distance components)
and the use of a receding-horizon with fixed trajectories
\eqref{eq:receding_horizon_exploration}.
This is evident from Table~\ref{tab:completion_times} which lists statistics for
time to completion and time to reach the early progress threshold
(see Sec.~\ref{sec:completion_and_rates}).
Most planners for a given environment and number of robots perform comparably
both at the end of each trial and near the beginning.

The exception for complete trials is the Plane-Boxes environment where both the
Myopic and \rsp[1]{} planners perform worse for 32 robots (by about 6\%).
This is likely a product of how the small size and limited vertical mobility of
the Plane-Boxes environment produces more frequent and complex interactions
between robots.

Figure~\ref{fig:early_coverage_rate} compares performance across planners in
terms of the rate of increase in environment coverage
(via the data in Table~\ref{tab:exploration_performance}).
Although the variations in coverage rates are less significant than for
suboptimality, the trends that the data does represent are consistent with
expectations:
the Myopic and \rsp[1] planner configurations
(which each involve planning myopically, robots ignoring others' decisions)
perform worst whenever there are significant differences across planners
(e.g. by 14\% for 16 robots in the Hallway-Boxes environment).

The differences in these coverage rates also narrow somewhat going from
16 to 32 robots, unlike the trends for suboptimality
(Fig.~\ref{fig:exploration_suboptimality})
where the performance gaps widen.
The process for parameter selection
(Table~\ref{tab:exploration_performance})
which focused on completion times for 16 robots partially explains this
phenomenon.
However, this may suggest that the receding-horizon optimization problems
we formulate
become less representative of actual performance with increasing numbers and
crowding of robots.
One possible cause is that the regions robots are planning to observe toward the
end of their planning horizons are frequently being observed sooner by other robots.
This produces uncertainty in future actions which assuming fixed trajectories
does not account for but which
might be alleviated by selecting a smaller discount factor.

\section{Conclusion and future work}

The system design and results from this chapter significantly improved
on the earlier work in
Chapter~\ref{chapter:distributed_multi-robot_exploration}.
Adding a term for distance to informative regions to the objective of the
environment, ensured reliable task completion which would not be true for our
prior results (Fig.~\ref{fig:entropy_reduction}).
Likewise, employing a coverage-based objective also improved
performance
(by as much as 16\%, see Appendix~\ref{appendix:csqmi_and_coverage}).
Incidentally, more mundane changes also contributed.
Running experiments on a 16-core AMD Ryzen processor enabled us to dismiss
approximations such as down-sampling rays and to perform more effective parameter
tuning.

This chapter also overcame an important challenge that arose from this thesis by
resolving limitations on applying \rsp{} planning to exploration.
This came about by
demonstrating that mutual information on occupancy grids without noise is
3-increasing.
Moreover, this ensures that the suboptimality guarantees
for \rsp{} apply to receding-horizon planning for exploration:
constant-factor suboptimality, approaching 1/2,
for a fixed number of planning rounds ($\numrounds$) and any number of robots
subject to the extent of inter-robot interaction (via the pairwise weights).
To reiterate prior chapters, this provides an $O(\numrobot)$ speed-up compared
to sequential planning (Alg.~\ref{alg:local_greedy}) while approaching the same
suboptimality guarantee.
Additionally, while this chapter did not focus on distributed algorithms,
the next will present a distributed implementation of \rsp{}
and application to similar exploration tasks.

Still, realizing significant improvements in performance via submodular
maximization and \rsp{} planning proves challenging.
Our results narrow this gap by establishing significant
improvements in suboptimality on the receding-horizon subproblems
as well as improvements in coverage rates early in the exploration process.


\subsection{Improving exploration performance}

There are a few clear avenues for further reducing completion time
and sustaining improvements in coverage rates.

Regarding solution quality, the online lower bounds suggest that,
although solutions to the receding-horizon optimization subproblems
predominantly outperform worst case bounds (1/2), they may
still be far from optimal (as low as 70\%).
Iteratively improving solutions~\citep{vetta2002focs,atanasov2015icra}
or developing an applicable variant of the continuous greedy
algorithm~\citep{calinescu2011} could each improve suboptimality.
However, there is no guarantee that doing so would significantly improve
exploration performance---this is evident from the large gaps in
suboptimality for trials with 32 robots which do not translate into commensurate
improvements in completion times.


The exploration process also often slows after the beginning of a trial.
In some of these cases, better routing and goal
assignment~\citep{simmons2000aaai,mitchell2019icra} could alleviate this issue.
Such approaches could be implemented within our framework by swapping the
distance reward goal regions \eqref{eq:informative_space}
with distances to specific goals.



\chapter{Implementation of Distributed, Receding-Horizon Planning for
Exploration}
\setdatapath{./fig/communication}
\label{chapter:distributed_communication}

While the last chapter focused on task performance in exploration, this chapter
focuses on the implementation of a distributed, receding-horizon planner.
The planner is synchronous and runs in real time in an anytime fashion.
Moreover, the implementation is simple
and does not depend on information about other robots, aside from the
sensor data that those robots use to construct maps of the environment.

A number of works propose distributed or decentralized algorithms
for online%
\footnote{
  As opposed to offline (equivalently non-adaptive) problems
  where a plan is computed in entirety before execution time~\citep{singh2009}.
}
multi-robot informative path planning
problems~\citep{%
atanasov2015icra,schlotfeldt2018ral,best2019ijrr,sukkar2019icra,regev2016iros}
much like those that arise in this thesis.
Fewer present results for planning in real
time in simulation or with real
robots~\citep{charrow2015wafr,schlotfeldt2018ral,sukkar2019icra}
as we do in Chapter~\ref{chapter:distributed_multi-robot_exploration}
and typically with no more than three robots.
Furthermore, actual implementations may still be
centralized, again in line with our earlier work~\citep{schlotfeldt2018ral}.
On the other hand,
\citet{sukkar2019icra} provide results for two robots in an agricultural active
perception task with a distributed implementation of Dec-MCTS.
With this chapter, we seek to provide an implementation of a distributed sensor
planning system that is suitable for large numbers of robots.

Auction algorithms can also solve submodular maximization
problems~\citep{williams2017icra} and are perhaps more mature as
\citet{ponda2012}
provides results for a task assignment problem with six robots with wireless
communication for both aerial and ground robots and as many as six robots at
once.
Those results demonstrate a version of
the Consensus-Based Bundle Algorithm (CBBA)~\citep{choi2009tro}.

At a high level, agents in auction algorithms like CBBA compute marginal gains
(bids) and iteratively communicate sets of assignments to neighbors, seeking to
come to consensus on the maximization steps of Alg.~\ref{alg:greedy_submodular}.
We provide a comparison to analogous auction algorithms in terms of solution
quality and messaging costs.
While auctions converge quickly given full access to objective information, our
results indicate that constraints on information access
(so that agents cannot accurately compute rewards for others' actions)
can slow convergence
or harm solution quality for sensing and coverage problems.
This is unlike the setting that \citet{choi2009tro} study which is more amenable
to constraints on information access.

\section{Background}

Let us begin with a brief background on distributed algorithms and communication
networks.

\subsection{Distributed models of computation: synchronous and asynchronous
algorithms}

This chapter describes a partially synchronous implementation of \rrsp{}
based on message timing and with distributed memory~\citep{lynch1996}.
For contrast, the directed graph structure of \rsp{} also lends itself to
asynchronous implementation.
For asynchronous models, computation times, message arrival times, and order
all may vary.
An asynchronous version of \rsp{} could be implemented by providing access to
neighbors in the directed graph structure of the planner and making each robot
wait to start planning until receiving decisions from all in-neighbors.

\subsection{Communication typologies}

Given a network of agents, there are many ways by which those agents may
communicate.
One the most common is \emph{unicast} which we also refer to as
\emph{point-to-point} communication.
For unicast communication, individual agents send messages to other individuals.
On the other extreme, \emph{broadcast} refers to sending messages from one agent
to all others.
The actual implementation we present effectively implements broadcasts via ROS.
Communication neighborhoods for \rrsp{} are also naturally amenable to multicast
communication as the robots send identical messages to each of their neighbors.

\subsection{Existing networking and communication systems}

This chapter presents a distributed algorithm for implementation on teams of
mobile robots.
However, operation in subterranean and urban environments can limit
communication between robots~\citep{murphy2016}.
Designers may then enable communication between robots that are not immediately
connected by establishing a mesh network over the team of robots.

One complication with this approach is that connectivity between robots can
change as they move about the environment.
Networks that address this challenge by allowing for links to change and
reconstructing routes online are called mobile ad hoc networks (MANETs).
One example of a MANET protocol that may be appropriate for use alongside the
algorithm we propose is Better Approach to Mobile Adhoc Networking
(BATMAN)~\citep{batman}.
However, we note that support for multicast communication appears incomplete,
and implementations of \rsp{} may be limited to unicast communication, depending
on the underlying mesh network.

\section{Distributed Planning for Exploration}

This section describes a distributed implementation of \rsp{} with a focus on
application to exploration problems much like those in prior chapters.
Toward this end, consider a team of robots $\robots=\{1,\ldots,\numrobot\}$
exploring some environment.
The robots plan in a distributed manner and collectively solve receding-horizon
sensing problems (Prob.~\ref{prob:submodular_partition_matroid}) via methods for
submodular maximization while simultaneously maintaining distributed
representations of the environment.
The following describes that distributed planner and its operation.

\subsection{Distributed computation model and assumptions}
\label{sec:communication_computation_model}

The distributed algorithm in this chapter is based on a partially synchronous,
distributed memory model.
The implementation itself is timing-based~\citep{lynch1996}, and
robots solve receding-horizon submodular maximization problems
(Prob.~\ref{prob:submodular_partition_matroid}) in \emph{epochs} with fixed
start and end times.
For the purpose of presentation, we assume that robots transmit messages
reliably and instantaneously, potentially on some mesh network.
Likewise, we assume robots have access to synchronized clocks;
although this work does not address clock-synchronization, the time scales in
this work (seconds) are much slower than the accuracy (milliseconds)
that common methods such as the Network Time Protocol (NTP) can
provide~\citep{mills1991}.

Regarding information access, robots have access to local models of the
environment as in Sec.~\ref{sec:background_model_of_computation} and are only
able to accurately approximate marginal gains for their own actions such as due
to latency in updates from distant robots.

\subsection{Maintaining the environment model}

Robots have access to some local approximation of the environment $\data_i$
e.g. an occupancy map.
Maintaining models of the environment involves some sort of distributed
perception system.
If robots simply share sensor data, the number of messages sent is quadratic
with the number of robots.
Compressed representations~\citep{corah2019ral} can alleviate this burden
somewhat as well as modifying the system to avoid the need for a complete
models.

\subsection{Communication and neighborhoods}

On the lines of the definition of the \rrsp{} planner in
Sec.~\ref{section:range}, we assume robots communicate with a local neighborhood
$\communicationneighbors_i$ for each robot $i\in\robots$.
This neighborhood may be based on metric distance or a number of hops in a
communication graph.

Then, aside from maintaining the local environment model,
the only access to (or knowledge of) the multi-robot team and mesh network
that we assume is the ability to communicate with these
neighbors $\communicationneighbors_i$.
In our analysis, we assume point-to-point messaging on the mesh network, with
messages possibly traveling for multiple hops.
However, the design is amenable to other modalities such as multicast.

\subsection{Long term goals}
\label{sec:long_term}

The exploration system in this chapter also includes a view distance reward
like the one Sec.~\ref{sec:global_distance_reward} describes.
To avoid growth in computation time, we compute the distance field only on a
sub-map around the robot as in our prior work~\citep{corah2019ral}.
More generally, a distributed road mapping strategy would be
appropriate and in line with recent works on robotic
exploration~\citep{wang2019ral_exploration,witting2018iros}.

\subsection[Implementation of a synchronous, distributed, receding-horizon
\rsp{} planner]
{Implementation of a synchronous, distributed, receding-horizon
\rsp[][\textbf]{} planner}

Algorithm~\ref{alg:distributed_synchronous} describes the implementation of
\rsp{}.
Robots solve instances of receding-horizon planning problems over the course of
each epoch, once every $\distributedplanningduration$ seconds, and planning for
individual robots runs for
$\planningduration=\frac{\distributedplanningduration}{\numrounds}$
seconds
(less some time for slack in practice),
recalling that $\numrounds$ refers to the number of sequential planning rounds
for the \rsp{} planner.

At the beginning of each epoch, robots sample their assignments to planning
rounds, uniformly at random
(lines~\ref{line:epoch_start}--\ref{line:random_assignment}).
Each robot $i \in \robots$ then waits for the beginning of its respective round
(line~\ref{line:round_start}) and, meanwhile, listens for planned actions
from robots in its neighborhood $\communicationneighbors_i$ and receives
$X^\mathrm{d}_{\inneighbor_i}$ (line~\ref{line:received_messages}).
The robot then plans sensing actions (line~\ref{line:plan_anytime})
over a receding horizon in an anytime fashion---in our case via MCTS---given
those received actions, its state $\state_i$, and its local representation of
the environment $\data_i$.
Finally, robots send their planned actions to their neighbors and execute their
plans (lines~\ref{line:send_action}--\ref{line:execute_action}).

The following sub-sections expound on the implementation and behavior of this
distributed algorithm.

\begin{algorithm}[t]
  \caption[Synchronous, distributed implementation of \rrsp{}]{%
    Synchronous, distributed implementation of Range-limited Randomized
    Sequential Partitions (\rrsp{}) from the perspective of robot $i$
  }%
  \label{alg:distributed_synchronous}
  \begin{minipage}{\linewidth}
    \begin{algorithmic}[1]
      \State $\epoch \gets \text{number of the current planning epoch}$
      \State $\numrounds \gets \text{number of planning rounds}$
      \State $\distributedplanningduration \gets
        \text{duration for execution of the distributed planer}$
      \State $\planningduration \gets \distributedplanningduration / \numrounds$
        \hfill(planning time per robot)
      \State $\communicationneighbors_i \gets
        \text{Robots (neighbors) within communication range}$
      \State $\data_i \gets
        \text{belief (e.g. map) available to robot } i$
      \State $\state_i \gets \text{state (e.g. position) of robot } i$
      \vspace{1em}
      \State Wait for the start of the current planning epoch at time
      $\epoch\distributedplanningduration$
      \label{line:epoch_start}
      \State $d_i \sim \{0,\ldots,\numrounds\!-\!1\}$
        \hfill(randomly select one of $\numrounds$ planning rounds)
        \label{line:random_assignment}
      \label{line:sampling}
      \vspace{1em}
      \State At time
      $\epoch\distributedplanningduration +
      d \planningduration$
      \label{line:round_start}
      \State \method{Receive}: $X^\mathrm{d}_{\inneighbor_i}$
        \hfill(listen for action selections from nearby robots
        $\inneighbor_i \subseteq \communicationneighbors_i$)%
        \label{line:received_messages}
      \State $x^\mathrm{d}_i \gets
      \method{PlanAnytime}(\data_i,
                           \state_i,
                           X^\mathrm{d}_{\inneighbor_i},
                           \planningduration)$
      \hfill(plan given available time)
      \label{line:plan_anytime}
      \State \method{Send}: $x^\mathrm{d}_i$ to $\communicationneighbors_i$
      \hfill(send plan with epoch $\epoch$ for neighbors in later rounds)
      \label{line:send_action}
      \State \method{Execute}: $x^\mathrm{d}_i$
      \label{line:execute_action}
    \end{algorithmic}
  \end{minipage}
\end{algorithm}

\subsubsection{Message content}
\label{sec:message_content}

The messages that Alg.~\ref{alg:distributed_synchronous} uses to send and receive
decisions consist of:
\begin{enumerate}
  \item The unique id of the sender
  \item The number of the current planning epoch
  \item A representation of the finite-horizon plan
    (e.g. the initial state and time and a sequence of action indices)
\end{enumerate}
A typical decision message consists of about 130 bytes.

\subsubsection{Minimalist design}

The implementation of Alg.~\ref{alg:distributed_synchronous} differs from other
distributed algorithms in this thesis
(aside from being the only actual distributed implementation)
because it avoids depending on objects related to the planning process.
Instead, the implementation focuses on handling messages and scheduling
planning times.
As such, the planning step (\method{PlanAnytime}) is simply a callback with
messages and planning time as inputs and outputs.

We chose this design to simplify the process of augmenting existing single-robot
sensor planning systems with distributed reasoning.
The main limitation of this approach is that the user becomes responsible for
tracking any statistics related to planning such as objective value or
solution quality.

\subsubsection{Robustness to communication failure}

Additionally,
the \emph{realization of the directed graph structure for the planning process}
(as described in Sec.~\ref{sec:dag_planning})
\emph{is implicit}
given the set of decisions each robot receives.
When a robot begins planning for itself, it simply uses the messages that are
available at that scheduled planning time.
Messages that arrive late or fail to arrive
(such as due to a crashed robot or loss of communication in a subterranean
environment)
do not prevent or delay the planning process.
Instead, failures in messaging contribute to increasing suboptimality
according to the realization of the directed planner graph
(as in Theorem~\ref{theorem:graphical_bound}
or Theorem~\ref{theorem:sum_submodular}).
This non-blocking behavior is valuable, considering that a naive implementation
of a sequential planner that waits to receive messages from previous robots
will not produce a complete plan on a disconnected network or if there are
communication failures.

\subsubsection{Toward preventing collisions between robots}

The planner that we present in this chapter is also not aware of possible
collisions between robots unlike the planner in
Chapter~\ref{chapter:distributed_multi-robot_exploration}.
Moreover, the approach to preventing inter-robot collisions in
Chapter~\ref{chapter:distributed_multi-robot_exploration} depends on sequential
reasoning to determine which robots execute planned trajectories and which
execute fallbacks.
However, a planner could be designed that allows robots to update their
decisions in parallel because:
\begin{itemize}
  \item Any robot can commit to a new plan if doing so avoids collisions with
    other robots' old and new plans.
  \item Local and global maxima (with respect collision-neighbors)\footnote{%
      A \emph{collision-neighbor} is a robot that is near enough that some
      pair of trajectories over the finite horizon could be in collision.
    }
    given any ordering
    can always commit to new plans, using the maxima to break symmetry.
\end{itemize}
Such an approach would preserve most of the guarantees from
Chapter~\ref{chapter:distributed_multi-robot_exploration} (e.g. liveness) with
only local rules.

Alternatively, other collision-avoidance tools such as barrier
methods~\citep{wang2017tro}
(which have also been specialized for aerial
robots~\citep{luo2019,wang2017icra})
can prevent collisions by augmenting planners that
are not necessarily collision-aware.
Barriers methods benefit from being minimally invasive (changing robots' plans
only when necessary).
At the same time, such changes may harm sensing performance along a robot's
trajectory so further study on the topic of inter-robot collisions would be
beneficial.

\subsection{Characterizing timing and synchronization}

Because messages contain the number of the planning epoch, robots are able to
determine whether to accept messages (if they belong to the current epoch)
or reject them (because they did not arrive in time for a previous epoch).
This also provides a mechanism for evaluating whether timing mechanisms are
working.
Due to random assignment to planning rounds (line~\ref{line:random_assignment})
and because robots should receive messages from prior rounds on time, the
nominal message acceptance rate is\footnote{
  This calculation includes unnecessary messages sent during the last round.
  Omitting those messages would require adaptation of this formula.
}
\begin{align}
  \E\left[\frac{n_\mathrm{accept}}{n_\mathrm{accept}+n_\mathrm{reject}}\right]
  &= \frac{1}{2}\left(1-\frac{1}{\numrounds}\right).
  \label{eq:acceptance_rate}
\end{align}
Tracking whether the empirical acceptance rate matches expectations then enables
designers to determine whether the planning system is respecting timing
constraints and whether latencies in messaging are preventing messages from
being received in time.

Note that this is distinct from rejecting messages that are outside the
communication range.
Hypothetically, a robot might also receive and ignore messages from a distant
robot but at later times due to multi-hop communication.
Such messages would then have to be excluded from these statistics on acceptance
rates (based on timing).

\section{Asymptotic behavior for messaging}

To begin, the number of sequential message transmissions---which we refer to as
the communication span---for Alg.~\ref{alg:distributed_synchronous} is constant
($\numrounds-1$ or $\numrounds$ if including messages which are sent during the
last round and so not used).
The total number of messages sent is in $O(\numrobot\numneighbors)$
where $\numneighbors = \max_{i\in\robots} |\communicationneighbors_i|$
is the largest number of communication neighbors over all robots.
The total number of message hops in a mesh network
depends on the length of the longest shortest-path $\neighbordistance$
from any robot $i$ to a communication
neighbor $j \in \communicationneighbors_i$
and comes to $O(\numrobot\numneighbors\neighbordistance)$.
The total communication volume, considering the number of decisions being sent,
is the same.
The numbers of messages, message hops, and communication volume are then constant
per-robot if $\numneighbors$ and $\neighbordistance$ are bounded.

In general, distributed perception is more expensive.
In the worst case, robots may share all sensor data (or summaries thereof) and
exchange $O(\numrobot^2)$ messages per unit time.
Then, if the diameter of the mesh network
(the length of the longest shortest-path between any two robots)
is
$\networkdiameter$,
the number of message hops and communication volume come to
$O(\networkdiameter\numrobot^2)$

Mechanisms such as compressed representations~\citep{corah2017icra}
can reduce these communication costs.
Further improvement may be possible by maintaining spatially local models, but
some sort of global modeling is necessary in the framework we present such as to
compute distances to long-term goals or views (Sec.~\ref{sec:long_term}).

\section{Results}

The results for this chapter include two studies.
The first set of results
(Sec.~\ref{sec:communication_results})
characterize communication costs in relation to other possible distributed
solvers in a setting based on the coverage experiments in
Chapter~\ref{chapter:scalable_multi-agent_coverage}.
This supports the second set of results
(Sec.~\ref{sec:distributed_results})
which demonstrate the synchronous, distributed implementation of
\rsp{} (Alg.~\ref{alg:distributed_synchronous})
via a simulation of exploration with dynamic aerial robots
(as in Chapter~\ref{chapter:distributed_multi-robot_exploration})
which runs in real time but in a setting where communication costs are trivial.

\subsection{Communication networks and messaging study}
\label{sec:communication_results}

This study seeks to characterize communication costs for distributed submodular
maximization on a mesh network in comparison to other relevant approaches.
The submodular maximization problems are based on the coverage problems in
Sec.~\ref{sec:area_coverage_results}.
The objective of these problems is to maximize area coverage over a unit square.
Actions cover circles with radius $\sensorrange$
and are distributed around agent centers within a radius of $\agentrange$.
The radii are set so as to normalize the sum of the areas of all sensing actions
for a given number of agents, as described in Table~\ref{tab:radii}.
The simulations vary the number of agents from 10 to 100 in increments of 10,
and we provide results for 50 trials for each configuration.

Agents communicate decisions on an undirected communication graph
with edges between any pair of agents within a radius of
$\communicationrange=3\agentrange$.
Due to the choice of sensor and agent radii, this ensures that there is an edge
between any two agents that may have non-zero redundancy.
Unlike the experiments in Chapter~\ref{chapter:scalable_multi-agent_coverage},
we require agents to form a connected communication graph
(to avoid complications with comparisons to other solvers).
To generate such agent positions, the position of the first agent center is
sampled randomly and the rest by sampling a random agent and a position within
range of that agent and inside the unit square.

The solver configurations compare \rrsp{} to sequential planning
(Alg.~\ref{alg:local_greedy}) and two auction algorithms
(which converge to results equivalent to Alg.~\ref{alg:greedy_submodular}).
The results for \rrsp{} encompass planning with
different numbers of rounds $\numrounds \in \{4, 8, 16\}$.
To reduce suboptimality and simplify comparisons, the communication range for
\rrsp{} is set to one step (or $\communicationrange$).
As such, the cost of ignoring decisions outside the communication range
\eqref{eq:range_cost} is zero,
and agents only communicate with their immediate neighbors during each round
of the planning process.
For these results, we assume that agents have access to their
communication-neighbors' planning rounds\footnote{
  This could be implemented by sharing random seeds used to generate the
  assignments to planning rounds.
}
$d_j$ for $j\in\communicationneighbors_i$
and only send messages to agents within range and \emph{in later rounds}
(whereas Alg.~\ref{alg:distributed_synchronous} includes communication with
\emph{all}
agents within range).
For the sequential planner, agents plan in the order they were generated.
Messages travel along shortest paths, and we take advantage of the structure of
the assignment process to consolidate decisions into individual messages sent
between adjacent agents in the planning sequence.
The auction algorithms are adapted from CBBA by \citet{choi2009tro}
in order to be applicable to general submodular objectives.
These two auction algorithms---detailed in
Appendix~\ref{appendix:auction_algorithms}---differ in terms of information
access:
the \emph{global auction} algorithm
(Alg.~\ref{alg:global_auction})
requires full access to the objective $\setfun$
and the ability to re-evaluate marginal gains for other agents' actions;
and for the \emph{local auction} algorithm
(Alg.~\ref{alg:local_auction})
agents only compute values of their own actions,
and they communicate assignments in lists along with the corresponding marginal
gains.
While the global auction algorithm will converge faster, this comes at the cost
of a strict requirement that all agents have complete and consistent access to
the objective\footnote{%
  Issues related to inconsistencies in the objective can prevent convergence
  which is an important topic for related works on auction algorithms%
  ~\citep{johnson2017jais}.
}
\emph{which violates the assumptions on information access}
(Sec.~\ref{sec:communication_computation_model}).
The local auction converges more slowly but respects constraints on information
access for local models.
These auction algorithms represent two extremes.
Specialized implementations might interpolate between the two such as by
tracking whether assignments change within a given distance of each agent,
somewhat like the specialized update rules for CBBA.
In the spirit of online planning, we also provide results for early stopping
after a given numbers of steps (matching the numbers of rounds for \rrsp{})
as well as for planning to convergence.\footnote{%
  We state that an auction algorithm has converged once all agents have a
  complete set of matching assignments.
}

\begin{figure}
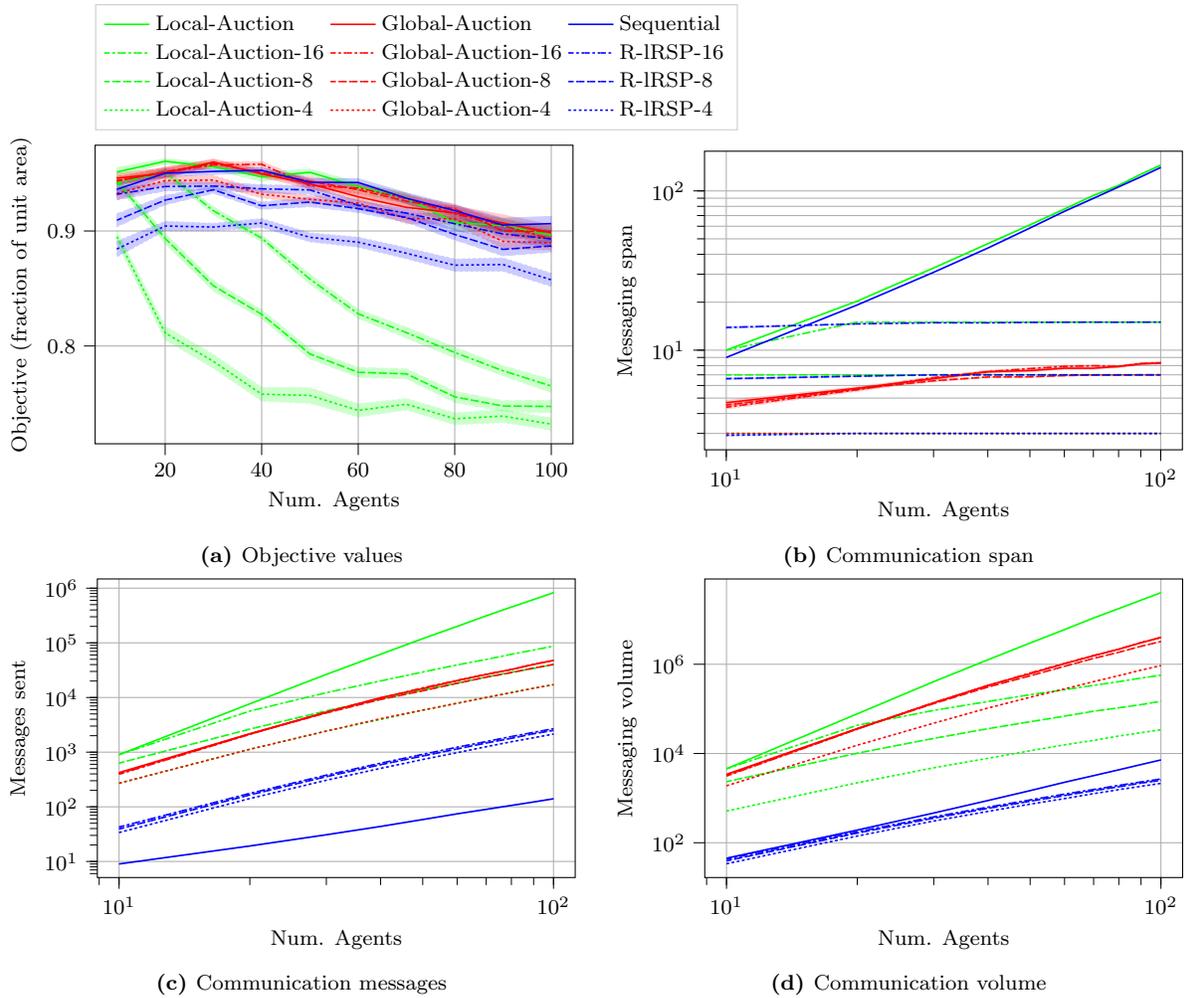

  \begin{subfigure}[b]{0.49\linewidth}
    \setlength{\figurewidth}{\linewidth}
    \setlength{\figureheight}{0.7\figurewidth}
    \scriptsize
    \inputfigure[communication_objective]{\datapath/messaging/objective.tex}
    \vspace{-0.2cm}
    \caption{Objective values}
    \label{subfig:communication_objective}
  \end{subfigure}
  \begin{subfigure}[b]{0.49\linewidth}
    \setlength{\figurewidth}{\linewidth}
    \setlength{\figureheight}{0.7\figurewidth}
    \scriptsize
    \inputfigure[communication_span]{\datapath/messaging/span.tex}
    \caption{Communication span}
    \label{subfig:communication_span}
  \end{subfigure}
  \begin{subfigure}[t]{0.49\linewidth}
    \setlength{\figurewidth}{\linewidth}
    \setlength{\figureheight}{0.7\figurewidth}
    \scriptsize
    \inputfigure[communication_messages]{\datapath/messaging/messages.tex}
    \caption{Communication messages}
    \label{subfig:communication_messages}
  \end{subfigure}
  \begin{subfigure}[t]{0.49\linewidth}
    \setlength{\figurewidth}{\linewidth}
    \setlength{\figureheight}{0.7\figurewidth}
    \scriptsize
    \inputfigure[communication_volume]{\datapath/messaging/volume.tex}
    \caption{Communication volume}
    \label{subfig:communication_volume}
  \end{subfigure}
  \caption[Objective values and messaging statistics for various distributed
  submodular maximization algorithms]
  {%
    Objective values and messaging statistics for various distributed submodular
    maximization algorithms.
    Shaded regions (although small) show standard error.
    (\subref{subfig:communication_objective})
    Objective values (out of one) remain consistent
    with increasing numbers of agents, except for the local auction solver
    which degrades significantly when querying solutions before convergence.
    \rsp{} requires eight or more communication rounds to obtain
    performance comparable to the sequential solver or global-information
    auction.
    (\subref{subfig:communication_span})
    When considering the span (number of sequential message hops),
    only \rsp{} obtains constant values, but the global-information auction
    solvers still retain a small spans, peaking at slightly over nine.
    (\subref{subfig:communication_messages})
    Numbers of messages (including hops) and
    (\subref{subfig:communication_messages})
    total message volume (by numbers of decisions and hops traveled)
    all grow super-linearly as the smallest communication radius (0.48)
    covers nearly half the length of the environment.
    While sequential planning benefits from being able to consolidate messages,
    \rsp{} obtains the smallest communication volumes and consistently
    outperforms all auction planners by about at least an order of magnitude on
    both metrics.
  }%
  \label{fig:communication_results}
\end{figure}

Figure~\ref{fig:communication_results} plots objective values and communication
costs for varying numbers of agents.
The numbers of messages include each hop on the communication graph.
However, only the sequential algorithm sends messages that travel for multiple
hops.
The communication volume then weights messages and hops by numbers of decisions
being sent
(ignoring the marginal gains that the local auction algorithm also sends).
The span is the number of sequential message transmissions or hops so that
\rrsp[4] has a span of 3, and the span for auction algorithms with early
stopping is at most as such.

While these results reflect a large variation in the number of agents, the
smallest communication radius is still nearly half the length of the unit
square.
This may explain the slow growth in spans (Fig.~\ref{subfig:communication_span})
for the global auction algorithm.
Likewise, the large communication radii cause messaging costs to increase
quickly compared to asymptotic rates for \rrsp{} and both auction
algorithms (due to increasing numbers of neighbors)
and more slowly for sequential planning
(as paths between sequential agents are short).
Overall, \rrsp{} maintains constant span, small communication
volume,\footnote{%
  The auction algorithms produce large communication volumes and numbers of
  messages
  because all agents plan and communicate during each communication round
  and because agents communicate sets of decisions rather than individual
  decisions like \rsp{}.
}
and consistent performance in terms of objective values and requires roughly
eight rounds to approach objective values for the global auction solver.
Interestingly, the convergence times for global auctions grow slowly, but this
may be a product of small network diameters\footnote{%
  The diameter is the length of the longest shortest-path between any two
  agents.
}
in our results.
The global auction solver may then be appropriate for sensor planning in similar
settings if designers can accept greater communication costs and strict
constraints on information access.
On the other hand, the local auction solver
better reflects our assumptions on information access,
but early stopping significantly harms objective values.%
\footnote{
  The distributed exploration system exhibited in
  Sec.~\ref{sec:distributed_results} would also violate the
  requirements of the global auction solver, simply because the robots query the
  maps at different times so that the models for planning are not entirely
  consistent with each other.
}
These results also exclusively address communication;
even when either auction algorithm converges more quickly,
the robot whose action is assigned last
completes maximization steps for each prior assignment
and incurs a computational cost equivalent to sequential planning for the
entire team (Alg.~\ref{alg:local_greedy}).
Conversely, for \rrsp{} each robot completes only a single maximization step.

\subsection{Anytime distributed planning for exploration}
\label{sec:distributed_results}

\begin{figure}
  \begin{subfigure}[b]{0.24\linewidth}
    \includegraphics[width=\linewidth,trim={50 100 200 100},clip]{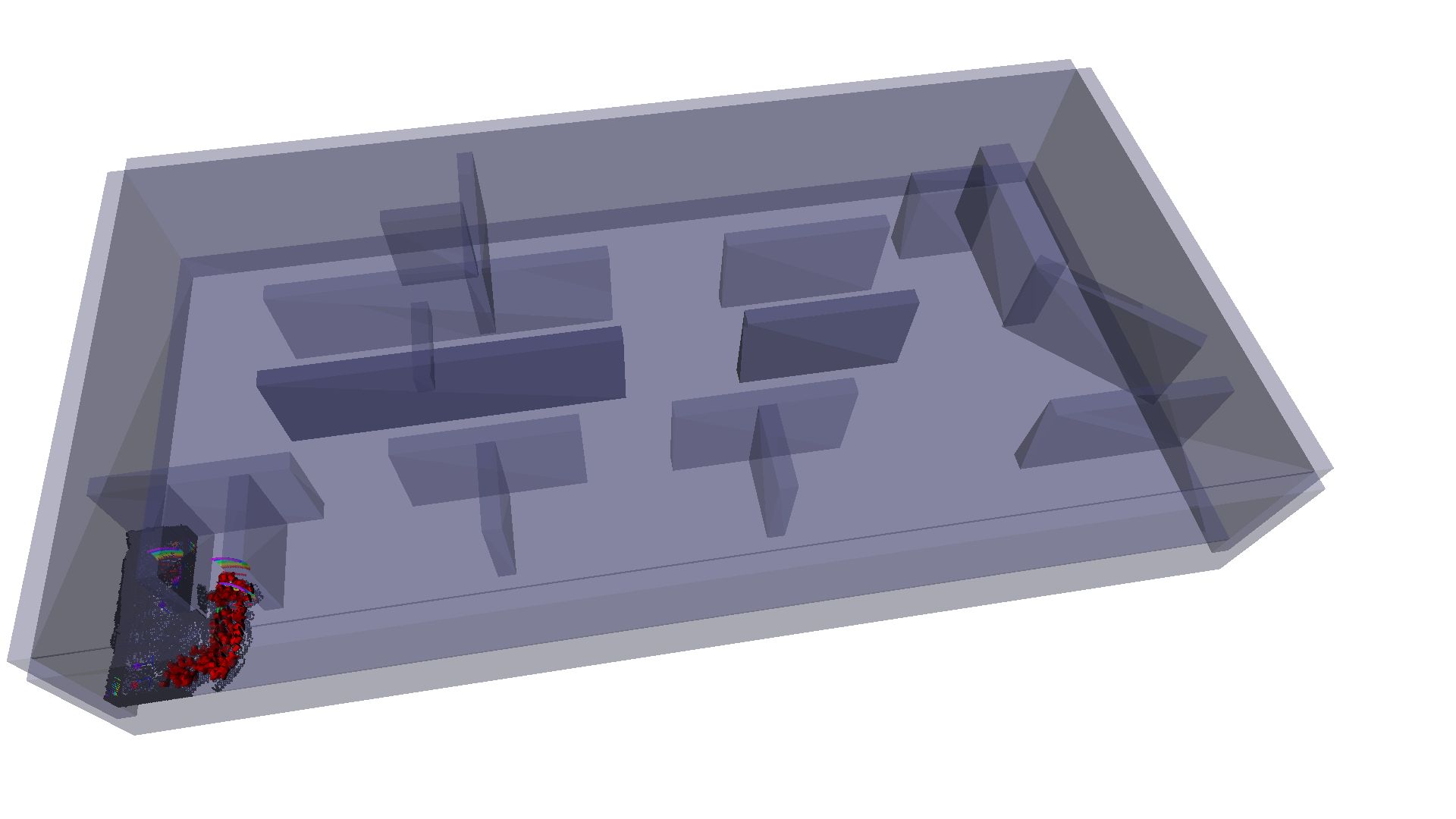}
    \caption{27 seconds}
  \end{subfigure}
  \begin{subfigure}[b]{0.24\linewidth}
    \includegraphics[width=\linewidth,trim={50 100 200 100},clip]{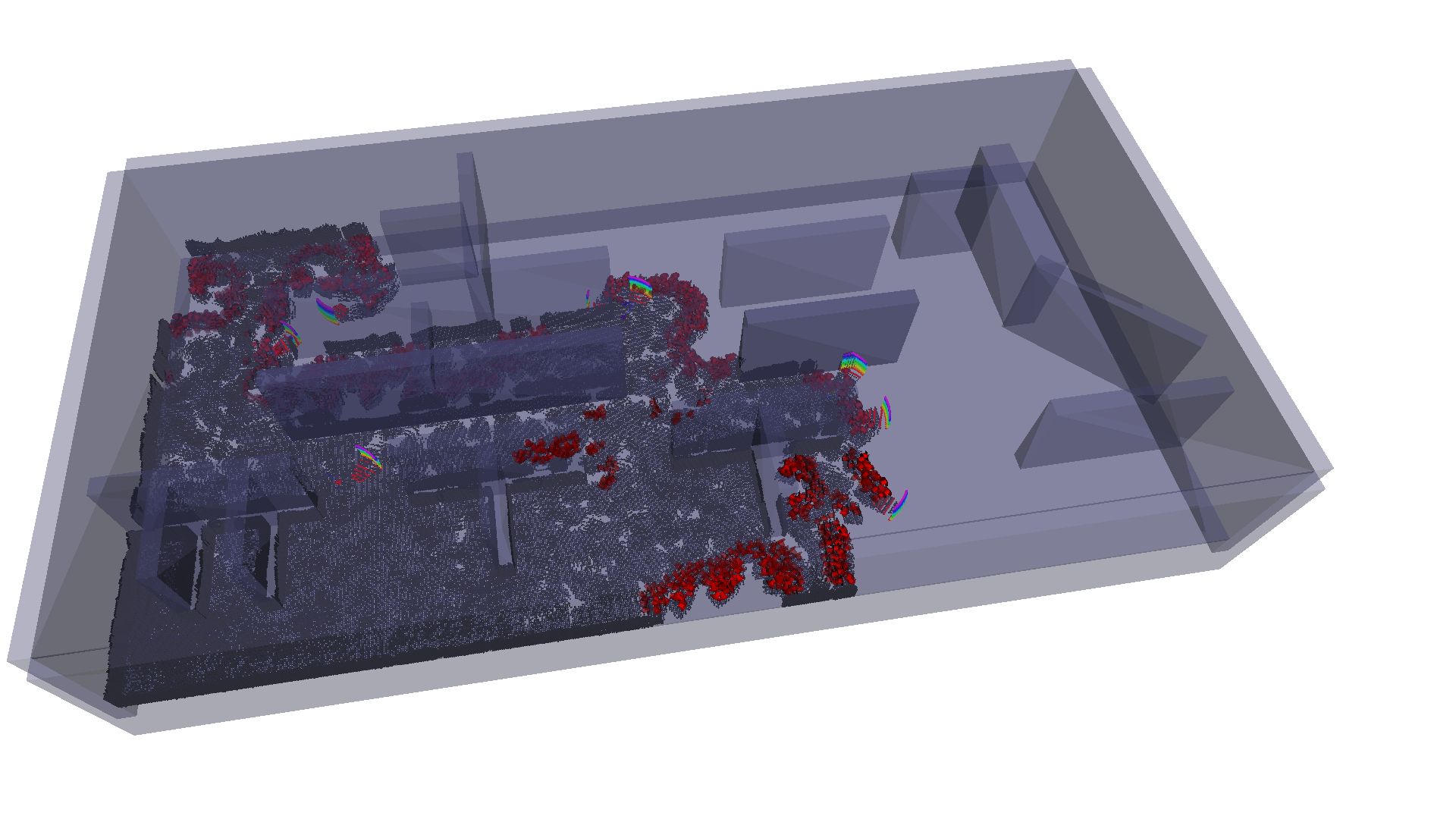}
    \caption{418 seconds}
  \end{subfigure}
  \begin{subfigure}[b]{0.24\linewidth}
    \includegraphics[width=\linewidth,trim={50 100 200 100},clip]{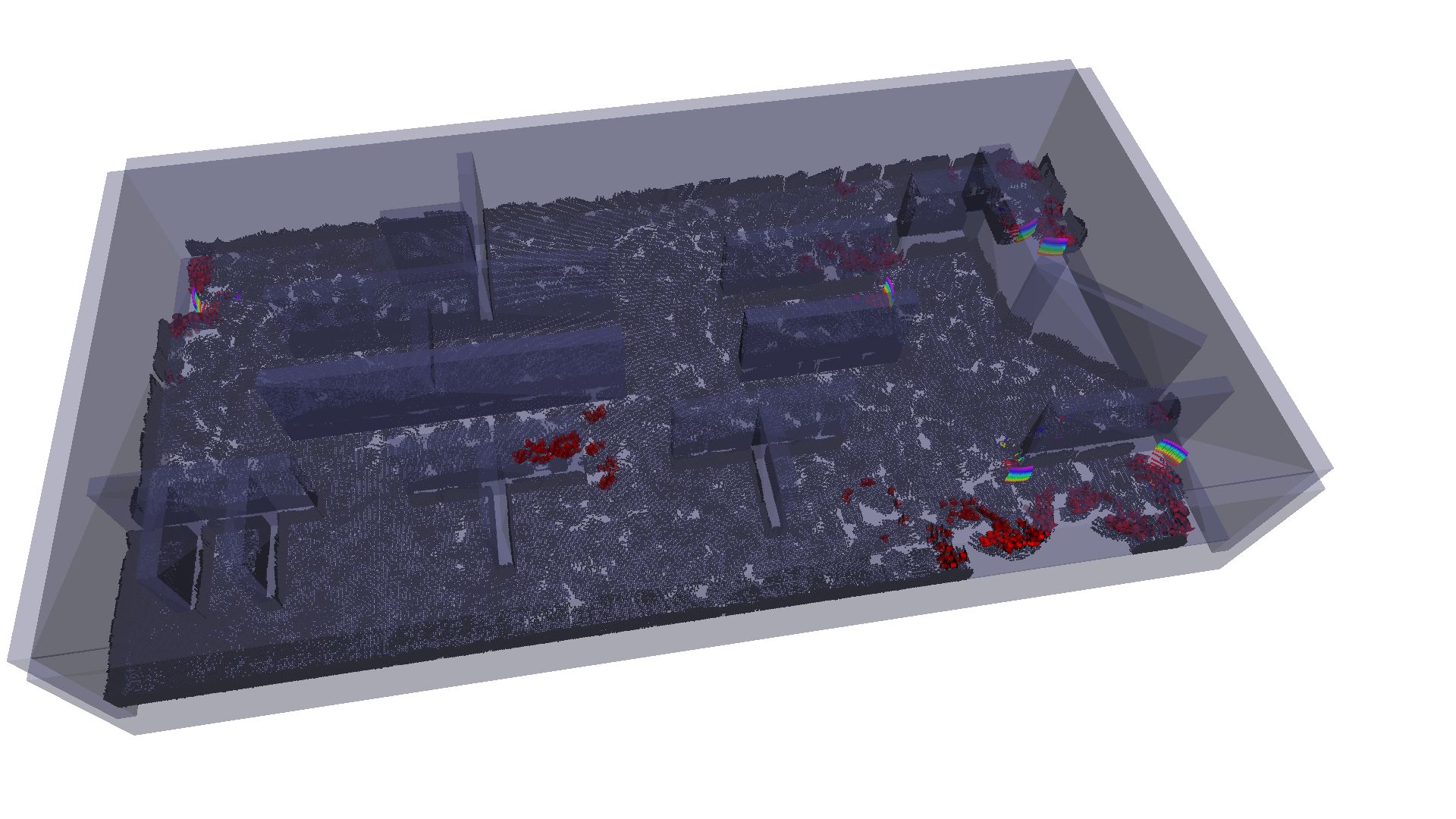}
    \caption{809 seconds}
  \end{subfigure}
  \begin{subfigure}[b]{0.24\linewidth}
    \includegraphics[width=\linewidth,trim={50 100 200 100},clip]{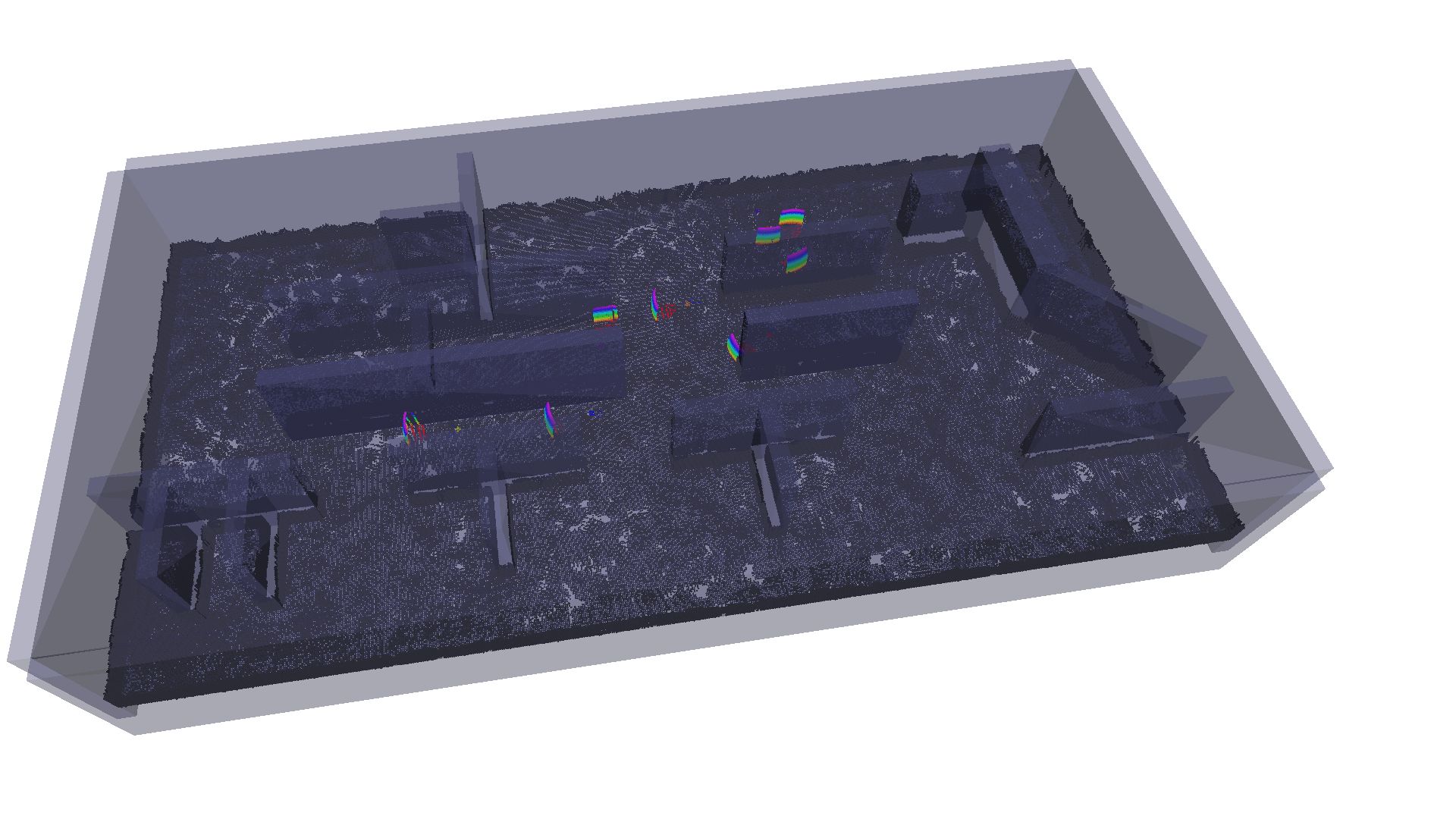}
    \caption{1200 seconds}
  \end{subfigure}
  \caption[Visualization of exploration of an office environment with
  distributed planning in real time]{%
    The above images visualize exploration of an office environment
    with eight robots and \rsp{} planning with $\numrounds=3$.
    Time stamps are approximate.
    A video of the exploration process can be found at:
    \url{https://youtu.be/MMr9NxT_J8c}
  }%
  \label{fig:distributed_visualization}
\end{figure}

\begin{figure}
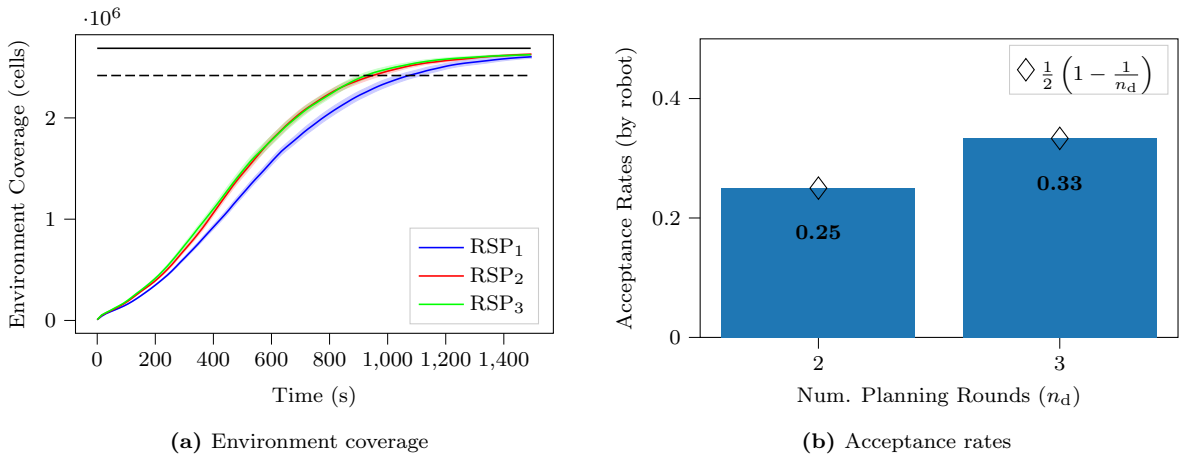
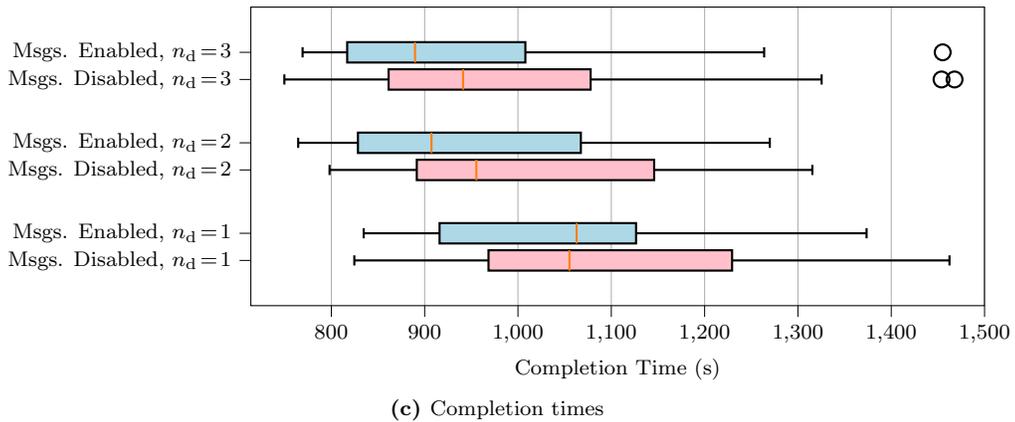

  \begin{subfigure}[t]{0.49\linewidth}
    \setlength{\figurewidth}{\linewidth}
    \setlength{\figureheight}{0.7\figurewidth}
    \scriptsize
    \inputfigure[distributed_coverage]{\datapath/environment_coverage.tex}
    \caption{Environment coverage}
    \label{subfig:distributed_environment_coverage}
  \end{subfigure}
  \begin{subfigure}[t]{0.49\linewidth}
    \setlength{\figurewidth}{\linewidth}
    \setlength{\figureheight}{0.7\figurewidth}
    \scriptsize
    \inputfigure[acceptance_rates]{\datapath/acceptance_rates.tex}
    \caption{Acceptance rates}%
    \label{subfig:acceptance_rates}
  \end{subfigure}
  \setlength{\figureheight}{0.343\linewidth}
  \vskip 0.3cm
  \begin{subfigure}{0.7\linewidth}
    \setlength{\figurewidth}{\linewidth}
    \scriptsize
    \hspace{-1cm}
    \inputfigure[distributed_completion]{\datapath/messaging_completion.tex}
    \vspace{-0.4cm}
    \caption{Completion times}%
    \label{subfig:distributed_completion_times}
  \end{subfigure}
  \caption[Results for exploration with the distributed \rsp{} implementation]{%
    Results for exploration with the distributed \rsp{} implementation.
    (\subref{subfig:distributed_environment_coverage})
    Mean environment coverage and standard error (shaded regions) for different
    numbers of robots.
    Black lines demarcate the maximum environment coverage and a 90\% threshold
    for task completion.
    (\subref{subfig:acceptance_rates})
    Message acceptance rates closely match predictions \eqref{eq:acceptance_rate}
    which indicates that the synchronous execution of anytime planning rounds
    is functioning properly.
    (\subref{subfig:distributed_completion_times})
    Planning with two or three rounds ($\numrounds$) improves median completion
    times by about 5\%
    compared to the same planners after disabling communication of control
    actions.
    Note that the cases for $\numrounds=1$ effectively represent the same
    configuration.
  }%
  \label{fig:distributed_planning_results}
\end{figure}

Having, addressed the question of communication costs, let us now evaluate
the distributed planner in the context of simulated multi-robot exploration
(with Fig.~\ref{fig:distributed_visualization} visualizing the exploration
process for these experiments).
For these results, the robot and sensor models are identical to the distributed
model in Sec.~\ref{sec:collision_experiments}.
Likewise, individual robots plan with Monte-Carlo tree
search~\citep{chaslot2010,browne2012} with the same motion primitive library.
Aside from the distributed implementation, the primary differences in the
exploration system, compared to
Chapter~\ref{chapter:distributed_multi-robot_exploration},
are that planners do not check for inter-robot collisions,
inclusion of a view distance term (see Secs.~\ref{sec:long_term}
and~\ref{sec:global_distance_reward}), and
use of the optimistic coverage objective (Sec.~\ref{sec:optimistic_coverage}).
The distributed, receding-horizon planner
(Alg.~\ref{alg:distributed_synchronous})
runs in real time in an anytime
fashion with an epoch duration of 3 seconds
and a horizon of 4 seconds.
Because the epoch duration is constant, robots planning with greater
numbers of rounds ($\numrounds$) have decreasing amounts of time to plan for
themselves.
The simulation results include 30 trials with 8 robots and
planning with distributed \rsp{} for 1, 2, and 3 planning rounds.
The simulations
in the Office environment pictured in Table~\ref{tab:environments}.
The mesh is scaled to 166\% of the size in
Chapter~\ref{chapter:time_sensitive_sensing} so that the longer dimension is now
$60\si\metre$, and the exploration volume\footnote{%
  The discrepancy in exploration volume is a product of differences in handling
  height.
  Here, robots have a height limit of $0.6\si\metre$ without a specific bounding
  box while Chapter~\ref{chapter:time_sensitive_sensing} provided a $2\si\metre$
  tall bounding box.
}
is $2687\si{\cubic\metre}$.

The implementation of the distributed planner takes advantage of
ROS~\citep{quigley2009icra} for messaging, and planners for each robot run in
separate processes (nodes).
The simulations themselves run on a single desktop computer with a 16-core
Ryzen 2950X processor.
While the processor provides ample capacity to run computation in parallel,
communication costs are not well-represented.
Providing projected communication costs can ameliorate this issue somewhat.
Regarding messaging span and latencies, the largest span
(for $\numrounds\!=\!3$)
produces a span of two.
Even significant latency, say $0.1\si\second$, produces a time cost of
$0.2\si\second$ which is small compared to the total planning time
(three seconds).
The implementation of the distributed planner does not include
range limits on communication so that robots effectively communicate via
broadcast.
However, there is no barrier to including range limits, particularly at larger
scales.
Recalling the discussion of message content in Sec.~\ref{sec:message_content},
the total cost of a communication round
(assuming point-to-point communication between all robots)
for eight robots is
$7.1\si{\kibi\byte}$
or, conservatively, at most
$7.1\si{\kibi\byte\per\second}$.
The simulated robots also maintain maps by assimilating camera data from the
entire team.
Still, all planning runs on spatially local sub-maps, and our approach is
compatible with methods for distributed mapping that can significantly reduce
communication costs such as via Gaussian mixture
models~\citep{corah2019ral,tabib2019fsr,tabib2020}.
Extrapolating from results for Gaussian mixture mapping
\citep[Fig.~7]{corah2019ral},\footnote{
  In these results~\citep{corah2019ral}, three robots produce about
  $0.02\si{\mebi\byte\per\second}$ of data or $6.8\si{\kibi\byte\per\second}$
  per robot.
}
maintaining distributed maps for eight robots,
again by point-to-point communication,
would require a total of about $400\si{\kibi\byte\per\second}$
of communication volume.

Figure~\ref{fig:distributed_planning_results} plots results for environment
coverage, message acceptance, and task completion.
The synchronous planner functions as designed
(Fig.~\ref{subfig:acceptance_rates}) with message acceptance rates matching
predictions \eqref{eq:acceptance_rate}.
This is consistent with the requirement that robots have access to decisions
from prior rounds for \rsp{} planning and subsequently to apply
suboptimality guarantees for \rsp{}
(Theorem~\ref{theorem:adaptive_partitioning}).
In our implementation, robots query their maps when they begin planning, at the
beginning of the assigned round rather than the beginning of the epoch.
Because of this the effective latency can vary, the average decreasing with
increasing $\numrounds$.
To address this, we ran a second set of experiments with
\emph{communication for \rsp{} disabled} which isolates the impact of
distributed planning.
This narrows the gap in task completion time to about 5\%.
Still, the scope of these simulation results is narrow as the intent is to
demonstrate functionality of the distributed planner rather than to characterize
task performance (which is the aim of
Chapter~\ref{chapter:time_sensitive_sensing}).

\section{Conclusion}

This chapter has demonstrated a distributed implementation of \rsp{} for
receding-horizon sensor planning in the context of multi-robot exploration.
The design is simple, requires minimal knowledge of the network structure, and
is naturally robust to communication failures.
Doing so realizes one of the main goals of this thesis,
making sensor planning for large numbers of robots tractable by reducing the
growth in planning time for sequential planners
(or conversely the increasing division of fixed planning time)
to a constant number of rounds with fixed duration.
As the exploration experiments focused on computation in a setting where
communication is relatively trivial, we also included a numerical study that
inspects communication costs.
This study demonstrated that \rrsp{} can provide a significant reduction in
communication compared to likely competitors, auction algorithms.
Moreover, our approach can also provide good solution quality with only locally
consistent models while auctions struggle to do so under time constraints.
Again, our results demonstrate distributed, receding-horizon planning in
real time in simulation with eight robots, significantly more than for
comparable works, and we hope that this work can serve to enable further
development of sensing systems with large numbers of robots.

\chapter{Conclusions and Future Work}
\setdatapath{./fig/conclusion}
\label{chapter:conclusion}

This thesis has developed and advanced methods for receding-horizon sensor
planning for teams of robots.
Specifically, receding-horizon planning for problems such as exploration
involves reasoning about likely observations at different camera views,
reasoning about overlaps and redundancy between views, and collectively
optimizing trajectories and views for teams of robots.

Existing works are able to address some of the challenges related to solving
these problems via greedy algorithms for maximizing submodular
objectives---submodularity being a common property amongst many sensing
objectives~\citep{fisher1978,singh2009}.
However, these algorithms are not particularly amenable to planning in real time
for large numbers of robots, particularly in distributed settings:
numbers of computation and communication rounds both grow linearly with the
number of robots.

Seeking to reduce computation time for submodular maximization problems
leads to results indicating that constant-factor computation time
(more specifically, adaptivity, Sec.~\ref{sec:quantify_parallel})
and solution quality cannot co-exist in
general~\citep{gharesifard2017,grimsman2018tcns,balkanski2018stoc}.
This led us to develop Randomized Sequential Partitions (\rsp{}) and methods for
analysis of pairwise redundancy for multi-robot sensing problems.
In Chapter~\ref{chapter:scalable_multi-agent_coverage}, we identified
3-increasing functions---which include general coverage objectives---as the
class of functions where redundancy between robots' actions decreases
monotonically given others' decisions.
Later, we applied this result for 3-increasing function to a case of mutual
information for exploration in Chapter~\ref{chapter:time_sensitive_sensing} via
interpretation of the mutual information objective as expected coverage.
Further, in Chapter~\ref{chapter:target_tracking}, developing similar methods
for redundancy analysis for sums of submodular functions enabled application of
\rsp{} for target tracking problems, demonstrating consistent task performance
for as many as 96 robots in simulation results.
Finally, we described a distributed implementation of \rsp{} in
Chapter~\ref{chapter:distributed_communication} and provided simulation results
for exploration in real time with eight robots.

\section{Future work}

This thesis work opens up numerous directions for future work which
can be divided roughly between submodular function maximization
and active sensing for robotics.

\subsection{Submodular function maximization}

This thesis also did not thoroughly characterize classes of 3-increasing
objectives or more general classes that satisfy bounds on redundancy such as
sums of submodular functions.
For example, we did not identify any meaningful objectives that are 3-increasing
but do not satisfy alternating monotonicity conditions as coverage objectives
do.

Additionally, although we focused on greedy algorithms, the continuous greedy
algorithm~\citep{calinescu2011} could improve solution quality.
While applying the continuous greedy algorithm to sensing problems, particular
informative path planning
problems with large spaces of actions, may prove challenging, there is interest
in applying such methods in robotics~\citep{robey2019}.
Our methods for redundancy analysis could possibly be used to develop versions
of such algorithms that converge with small numbers of integration steps.

Continuous analogues of submodular functions have also been a recent topic of
interest~\citep{bian2017,xie2019,robey2019}.
Reformulating our results for 3-increasing functions in continuous domains could
produce interesting results.
However, the nature of possible impacts in this area are unclear.

Future works may also view this text as providing existence proofs that
characterize certain sensing and submodular maximization problems.
For example, a corollary to our results is that the adaptive complexity
(Sec.~\ref{sec:low_adaptivity_algorithms})
of a wide variety of receding-horizon sensing problems is $O(1)$.
Likewise, our results could be interpreted as providing upper bounds on
the amount of information access that is necessary to achieve a given bound on
solution quality;
doing so could inform design of more complex optimization systems or machine
learning methods.

\subsection{Active sensing}

To begin, this thesis studied two sensing problems: exploration and target
tracking.
However, \rsp{} methods are applicable to a broad variety of sensing problems.
An interesting property of the problems we studied is that equilibrium
conditions generally perform well without explicit coordination
(i.e. with myopic planning):
so long as robots communicate observations to each other, they tend to
distribute themselves evenly across the environment, automatically.
Other sensing problems, one being target coverage, may not exhibit such
equilibrium behavior and may benefit more significantly from coordinated
sensor planning.

Online adaptation of the planner structure was discussed in
Chapter~\ref{chapter:scalable_multi-agent_coverage} but not revived afterward.
Moreover, selecting the number of rounds produces a potentially challenging
tradeoff between planning time and solution quality.
Development of  impactful methods for adaptation of \rsp{} planning,
particularly for planning in real time as in
Chapter~\ref{chapter:distributed_communication} is attractive.
Ideal numbers of planning rounds may also vary spatially, and this may pose a
challenging problem, particularly for planning in real time.

While Chapter~\ref{chapter:target_tracking} applied results for sums of
submodular functions to target tracking, the methods can also apply more
generally to multi-objective sensing problems.
The analysis for redundancy could capture changes in dependency
between robots through the course of a complex task.
For example, robots searching for the source of a gas leak in a building may
experience large amount of redundancy that decreases after they localize the
leak and focusing, instead, on a mapping task with more local interactions.

The applications in this thesis also focused on receding-horizon planning
\emph{with fixed trajectories}.
However, the problems we solve could be addressed more directly as variants of
POMDPs.
\citet{satsangi2018auro} recursively apply greedy algorithms to obtain
guarantees for certain submodular sensing processes.
Similar results could be obtained for partition matroids in multi-robot
problems, and the contributions of this thesis could be applied as well,
possibly to develop efficient, distributed solvers for active perception
processes.

In Chapter~\ref{chapter:time_sensitive_sensing}, we provided some incidental
contributions toward objective design for exploration, and in kind with
\citet{henderson2020icra}, we identified some limitations of prior works
on mutual information objectives.
In one such direction, the interpretation of mutual information without noise as
expected coverage could produce more effective approximations of joint mutual
information (see Appendix~\ref{appendix:csqmi_and_coverage}).
The comparison to CSQMI in same appendix also strongly suggests that improving
approximations for joint mutual information could be fruitful in terms of task
performance.
Such approximations should also be compared to recent work on
approximating conditional mutual information~\citep{henderson2020icra}.

As a whole, the methods for planning in this thesis are limited by the focus on
receding-horizon planning.
Further, realizing consistent improvements in task performance via \rsp{} may
require better and more complete planning methodologies.
Chapter~\ref{chapter:time_sensitive_sensing} cited spatially
global routing and assignment---such as via distributed auctions---as likely
directions for improving performance.
Likewise, development of methods for planning with distributed road maps would
be an important step toward improving our methods for computing view distance
rewards.

\appendix

\chapter{Additional Technical Discussion}
\setdatapath{./fig/technical_discussion/}
\label{chapter:technical_discussion}

\section{Expected coverage and mutual information for exploration with
independent cells are not necessarily adaptive submodular}
\label{appendix:not_adaptive_submodular}

Although this thesis does not take advantage of adaptive submodularity,
the work was developed in the wake of a number of works that apply adaptive
submodularity to active sensing problems in
robotics~\citep{choudhury2018ijrr,javdani2013icra,chen2015aaai}.
For this reason, considering whether and how adaptive submodularity might apply
to problems such as exploration is important to this thesis.

In simple terms, adaptive submodularity~\cite{golovin2011jair} seeks to apply
the concept of submodularity (and subsequent suboptimality guarantees)
to adaptive settings---wherein agents obtain
realizations of actions or observations
(e.g. whether a measurement succeeded or the realization of a actual camera
view)
at each step of a greedy decision process
(as in Alg.~\ref{alg:greedy_submodular}).
In our work, the receding-horizon planning sub-problems which we formulate
(Sec.~\ref{sec:receding_horizon_planning})
are
\emph{non-adaptive}, and the theory we develop primarily applies to the
sub-problem solutions.
However, \emph{the decision process by which the robots replan and explore is
adaptive}, and we do not provide guarantees for this process as a whole.

Methods based on adaptive submodularity provide the prospect of general
suboptimality guarantees for the entire exploration process.
Moreover, \citet{choudhury2018ijrr} study an exploration problem with an
expected coverage objective equivalent to \eqref{eq:expected_coverage}
and cite routing constraints (e.g. path planning) as the primary theoretic
limitation for applying adaptive submodularity to exploration
(versus existing results for cardinality constraints~\citep{golovin2011jair}).
However, we note that they only \emph{assume} that the objective is
(or is approximately)
adaptive submodular.
Although this is true for some special cases~\citep[Sec.~7.1]{golovin2011jair},
such exploration objectives are not adaptive submodular in general
which we will prove by counterexample after introducing adaptive submodularity
more formally.

Consider a function
$\setfun : 2^\ground \times \mathcal{O}^\ground$
where $\mathcal{O}$ is a set of observations or outcomes.
Here, $X \subseteq \ground$ encodes the set of selections
and $\Phi \in \mathcal{O}^\ground$
encodes the outcomes for each observation.
Further, agents receive an observation whenever they execute an action,
and there is some probability distribution $\Phi$ over the outcomes for each
action in $\ground$.
Then, following \citet{golovin2011jair}, we can define the expected marginal
gain as
\begin{align}
  \Delta(x|\psi) &= \E\left[
    \setfun(\mathrm{dom}(\psi)\cup\{x\},\Phi)
    -
    \setfun(\mathrm{dom}(\psi),\Phi)
    \mid
    \Phi \sim \psi
  \right]
\end{align}
where $x\in\ground$, $\psi$ is a partial realization that consists of
action-outcome pairs,
$\mathrm{dom}(\psi)$ is the associated set of actions in $\ground$,
and $\Phi \sim \psi$ indicates that the full realization
$\Phi$ is drawn conditional on the partial realization $\psi$.
A function $\setfun$ is adaptive submodular
\emph{with respect to the distribution over outcomes} $\mathcal{O}^\ground$
if for all
$\psi$, $\psi' \subseteq \psi$, and $x\in\ground$, then
\begin{align}
  \Delta(x|\psi) \geq \Delta(x|\psi').
  \label{eq:adaptive_submodularity}
\end{align}
In other words, the expected marginal gains for all actions must decrease
\emph{regardless of the outcome of a selection}.

Now, let us move on to the counterexamples for mutual information\footnote{%
  The fact that mutual information is not adaptive submodular in general is
  well-known~\citep{golovin2010nips}.
  We prove that the same holds more narrowly for exploration and as part of
  the proof for expected coverage.
}
and expected coverage.\footnote{%
  Note that this does not preclude some cases being exactly or approximately
  adaptive submodular as in the work of \citet{choudhury2018ijrr}.
}
Note that we now use more traditional notation for mutual information since
that already includes mechanisms for encoding outcomes for observations and
expected marginal gains which are equivalent to the ones \citet{golovin2011jair}
describe.

\begin{figure}
  \def\svgwidth{0.7\linewidth}
  \input{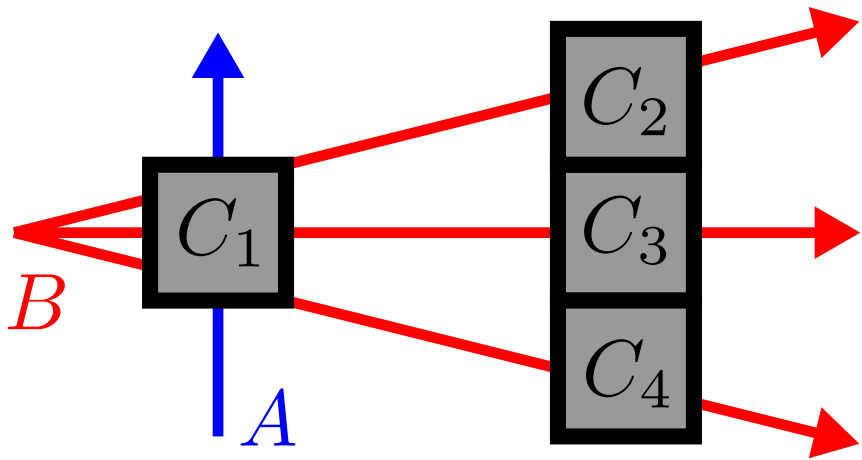}
  \caption[Adaptive submodularity counter-example]{
    The above illustrates a counter-example which demonstrates that mutual
    information for ranging observations with independent cells is not
    necessarily adaptive submodular.
    The target variable, the environment, consists of four cells
    $\environment=[C_1, \ldots, C_4]$
    which are each independent and \emph{free} (0) or \emph{occupied} (1) with
    probability 0.5 (each cell with 1 bit of entropy).
    Two available observations $\textcolor{red}{B}$ and $\textcolor{blue}{A}$
    provide distances to the nearest occupied cell along a given ray, as
    illustrated.
  }
  \label{fig:adaptive_submodular_counter_example}
\end{figure}

\begin{theorem}[Noiseless mutual information for depth sensors with independent
  cells is not necessarily adaptive submodular]
  \label{theorem:mutual_information_adaptive_submodular}
  Consider an environment $\environment$ which consists of Bernoulli cells that
  are each free or occupied with some probability and possible observations
  $\ground$ from a depth sensor which provide ranges to the nearest occupied
  cell along collections of rays according to descriptions in
  Chapter~\ref{chapter:time_sensitive_sensing} and
  Sec.~\ref{sec:mutual_information_exploration}.

  The mutual information $\I(\environment; X)$ for $X \subseteq \ground$
  is not adaptive submodular.
\end{theorem}
\begin{proof}
  Consider the scenario
  that
  Fig.~\ref{fig:adaptive_submodular_counter_example}
  illustrates
  and the resulting distribution over environments $\environment$
  and ground set $\ground = \{A, B\}$.

  The mutual information between $B$ and $E$ will violate adaptive
  submodularity.
  Now, note that $B$ either observes the value of only one cell
  (if $C_1$ is occupied)
  or else all four.
  As such, the mutual information between the two is initially
  \begin{align}
    \MI(B;E) = 1 + 0.5 \cdot 3 = 2.5
  \end{align}
  given the expected information gain
  \eqref{eq:mapping_information_independent}
  and the entropies of the cells.
  However, the mutual information for $B$ increases if we observe $A$ and
  determine that $C_1$ is free
  \begin{align}
    \MI(B;E|C_1=0) = 3.
  \end{align}
  This increase in mutual information violates adaptive submodularity
  \eqref{eq:adaptive_submodularity}
  which completes the proof.
\end{proof}

\begin{corollary}[Expected coverage for exploration is not necessarily adaptive
  submodular]
  \label{corollary:expected_coverage_adaptive_submodular}
  Expected coverage as described in Sec.~\ref{sec:expected_coverage_exploration}
  is not necessarily adaptive submodular.
\end{corollary}

\begin{proof}
  This follows from Theorem~\ref{theorem:mutual_information_adaptive_submodular}
  because noiseless mutual information with independent cells is
  an expected coverage objective
  (Theorem~\ref{theorem:mapping_information_is_coverage}).
\end{proof}

Still, there is still room to apply adaptive submodularity or similar
properties to exploration policies.
In fact, theory for objectives based on the size of the hypothesis space
still
applies~\citep{golovin2011jair,javdani2013icra,javdani2014aistats,chen2015aaai}.
Further, \citet{gupta2017} provide results for a routing problem in a similar
setting.

However, reducing the size of the hypothesis space has its limitations.
Consider the reduction in the hypothesis space $h$
and the following hypothetical bound for a greedy maximization process:
\begin{align}
  h^\mathrm{g} &\geq \alpha h^\star
\end{align}
For a uniform prior over $n$ hypotheses, the relationship between a reduction
in the hypothesis space $h$ and the information gain $\MI$ is
\begin{align}
  \MI = \log_2\left(\frac{n}{n-h}\right) = \log_2(n) - \log_2(n - h),
\end{align}
and
\begin{align}
  h = n - 2^{\log_2(n) - \MI}.
\end{align}
Substituting into the bound
\begin{align}
  \MI^\mathrm{g}
  &\geq
  \log_2\left(
    \frac{n}{
      n - \alpha (h^\star)
    }
  \right)
\end{align}
Consider if $h^\star=n$ (such as for mapping an entire environment) so that
\begin{align}
  \MI^\mathrm{g}
  &\geq
  \log_2\left(
    \frac{1}{
      1 - \alpha
    }
  \right)
\end{align}
which works out to \emph{one bit} for $\alpha=1/2$ and \emph{1.44 bits} for
$\alpha = 1 - 1/e$.

In general, we can expect bounds on the reduction of the hypothesis space to be
most useful for small hypothesis spaces (such as to distinguish between a small
number of likely environments predicted by a learner
or localizing objects~\citep{javdani2013icra}).
However, such bounds would be less impactful for the exploration problems that
we study which exhibit large information gain and exponentially
hypothesis spaces.

\section{Analysis for scaling target tracking to large numbers of robots}
\label{appendix:scaling_analysis}

The analysis in this section establishes sufficient conditions for the cost of
distributed planning $\distcost$~\eqref{eq:distributed_planning_cost} for each
robot to be constant (in expectation) for planners with a fixed number of
sequential steps, independent of the number of robots.
Afterward, we discuss how this analysis relates to the design and analysis of
target tracking systems.

Consider a distribution of robots and targets on $\real^n$ with at most $\alpha$
robots and $\beta$ targets on average per unit volume.
Then assume that the channel capacities~\eqref{eq:target_capacities} between
each robot $i\in\robots$ and target $j\in\targets$ satisfy a non-increasing
upper bound
$\phi: \real_{\geq 0} \rightarrow \real_{\geq 0}$
(possibly in expectation) so that
$C_{i,j} \leq \phi(||\robotposition_i - \targetposition_j||_2)$
where $\robotposition_i$ and $\targetposition_j$ are the robot position and
target \emph{mean} position in $\real^n$.
Now, taking the expectation of the total weight
for targets distributed on a $n$-ball centered around one
robot and for robots on another ball centered on each target, each ball with
radius $R$,
produces an upper bound on the expectation for robots and
targets within a radius of $R/2$ of the given robot.
Given the bound on interaction between robots in terms of interactions between
robots and targets~\eqref{eq:weight_by_component}, and designating the
zero-centered ball with radius $R$ as $B$ this expectation has the form:
\begin{align}
\E\bigg[\sum_{j\in \robots\setminus\{i\}} \weights(i,j)\bigg]
  &=
  \int_B\int_B
  \alpha\beta
  \min(\phi(||\vec{x}||_2), \phi(||\vec{y}||_2))
  \dd{\vec{x}}\dd{\vec{y}}.
  \intertext{By integrating over the surface of each ball
  (each an $(n-1)$-sphere with surface area $S_{n-1}$)}
  &=
  \alpha\beta
  \int_0^{R}\int_0^{R}
  {S_{n-1}}^2
  {r_1}^{n-1}{r_2}^{n-1}
  \min(\phi(r_1), \phi(r_2))
  \dd{r_1}\dd{r_2}.
  \intertext{Given that $\phi$ is non-increasing, separating the minimum
  produces:}
  &=
  \alpha\beta
  {S_{n-1}}^2
  \left(
    \int_0^{R}
    \int_0^{r_2}
    {r_1}^{n-1}{r_2}^{n-1}
    \phi(r_2)
    \dd{r_1} \dd{r_2}
  \right.
    \nonumber\\
  &\hspace{15ex}
  \left.
    +
    \int_0^{R}
    \int_{r_2}^{R}
    {r_1}^{n-1}{r_2}^{n-1}
    \phi(r_1)
    \dd{r_1} \dd{r_2}
  \right),
  \\
  \intertext{and by swapping the bounds of the second integral, combining, and
  evaluating the inner integral, we get:}
  &=
  \alpha\beta
  {S_{n-1}}^2
  \left(
    \int_0^{R}
    \int_0^{r_2}
    {r_1}^{n-1}{r_2}^{n-1}
    \phi(r_2) \dd{r_1} \dd{r_2}
  \right.
    \nonumber\\
  &\hspace{15ex}
  \left.
    +
    \int_0^{R}
    \int_0^{r_1}
    {r_1}^{n-1}{r_2}^{n-1}
    \phi(r_1) \dd{r_2} \dd{r_1}
  \right)
  \\
  &=
  2
  \alpha\beta
  {S_{n-1}}^2
  \int_0^{R}
  \int_0^{r_1}
  {r_1}^{n-1}{r_2}^{n-1}
  \phi(r_1) \dd{r_2} \dd{r_1}
  \\
  &=
  \frac{%
    2
    \alpha\beta
    {S_{n-1}}^2
  }{n}
  \int_0^{R}
  {r_1}^{2n-1} \phi(r_1)
  \dd{r_1}.
  \label{eq:target_tracking_scaling}
\end{align}
The above integral~\eqref{eq:target_tracking_scaling} converges in the limit
if $\phi \in O(1/x^{2n+\epsilon})$, or most relevantly, on a plane this
condition comes to $\phi \in O(1/x^{4+\epsilon})$.\footnote{%
  This requirement on interactions between
  \emph{robots and targets} is stricter than the equivalent one
  \emph{between robots} (see Theorem~\ref{theorem:scalability}) as
  interactions between robots must decrease as $O(1/x^{n+\epsilon})$.
}
Given that sensor models purely with sensor noise proportional to distance do
not fit this constraint, designers may wish to consider the effects of
constraints on the total sensor range or the number of observations one robot
can obtain at once.

\subsection{Scaling and sensor models}

The sensitivity to how quickly interactions between robots and targets fall off
motivates attention to sensor design and modeling lest planners perform poorly
for large numbers or require additional computation time.
For example, additive Gaussian noise with the standard deviation proportional to
distance (as is common in range sensing models~\citep{charrow2014ijrr}) is
insufficient which is evident from the channel capacity of a Gaussian
channel~\citep[Chap.~9]{cover2012}.

At the same time, robots in realizable systems cannot obtain and process
observations of unbounded numbers of targets, and features such as a maximum
sensor range can model these limits.
Still, the observation of whether a target is within range provides some
information.
Here, a narrow tails assumption on the probability distributions for the
targets---such as that the \emph{tails approach zero exponentially}---ensures
that $\phi$ decreases sufficiently quickly.
As a result of this sensitivity to the tails,
the scaling behavior~\eqref{eq:target_tracking_scaling} may be difficult to
characterize a-priori, even when known to be bounded.

\section{An argument in favor of coverage over entropy reduction for evaluating
exploration performance}
\label{appendix:entropy_vs_coverage}

\begin{figure}
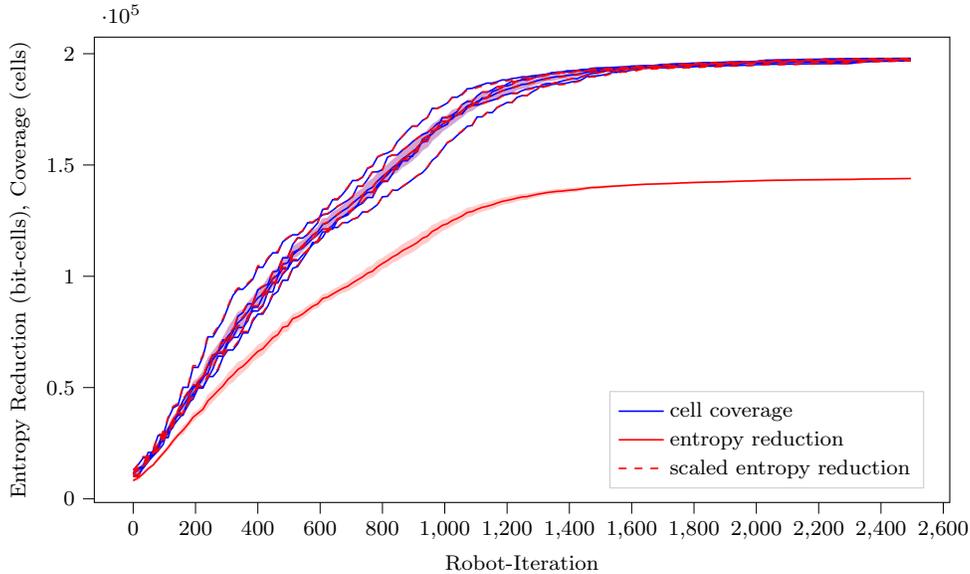

  \centering
  \setlength{\figurewidth}{0.8\linewidth}
  \setlength{\figureheight}{0.6\figurewidth}
  \scriptsize
  \inputfigure[compare_entropy_coverage]{\datapath/compare_entropy_coverage.tex}
  \caption[Comparison of environment coverage and entropy reduction]{%
    Compare entropy reduction and environment coverage for five
    exploration trials with 16 robots in the \emph{boxes} environment
    (jagged lines are individual trials).
    Notice that these coincide exactly when scaled to match.
  }%
  \label{fig:compare_entropy_coverage}
\end{figure}

Chapter~\ref{chapter:time_sensitive_sensing} applies coverage-based performance
measures while, previously
in Chapter~\ref{chapter:distributed_multi-robot_exploration}, we used entropy
reduction to evaluate exploration performance.
However, the systems for exploration that we study in this thesis are subject
to little or no noise and are tuned so that cell occupancy values converge
quickly.
Anecdotally, we have observed significant changes in maximum entropy
reduction for exploration trials in the same environment as systems and
parameters have changed over time.
Moreover, entropy results for our systems closely match environment coverage
when scaled by a constant factor (Fig.~\ref{fig:compare_entropy_coverage})
which likely reflects design decisions such as limits on occupancy
certainty (which are common in practice~\citep[(4)]{hornung2013})
rather than any feature of performance.
Instead, a coverage-based metric provides a consistent measure of the
amount of volume that robots have observed (given cell volume) and is less
affected by design parameters such as limits on occupancy uncertainty.

\section{Evaluating the performance of an approximate mutual information
objective}
\label{appendix:csqmi_and_coverage}

\begin{figure}
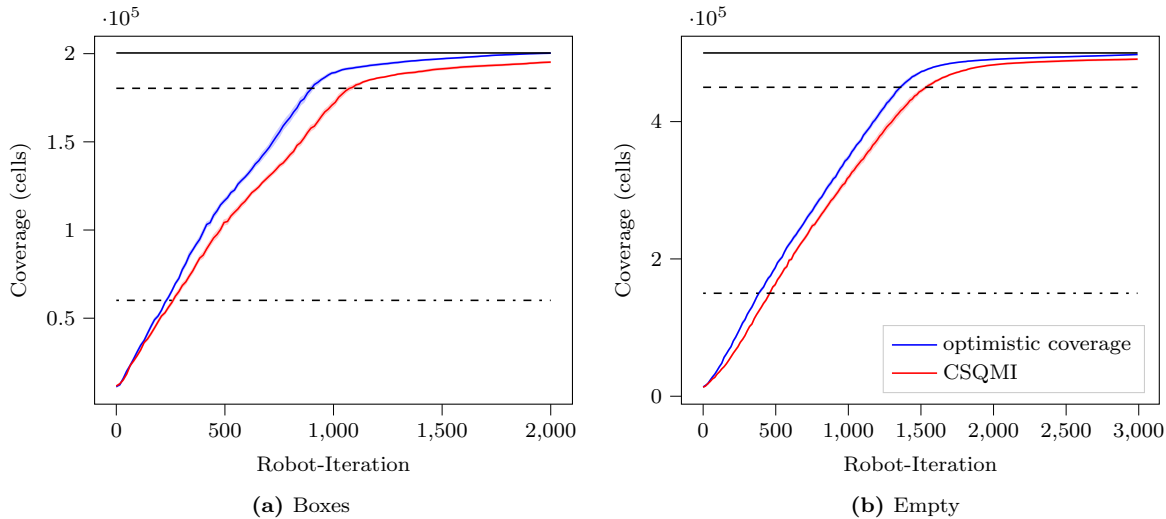

  \setlength{\figureheight}{0.4\linewidth}
  \scriptsize
  \begin{subfigure}{0.49\linewidth}
    \setlength{\figurewidth}{\linewidth}
    \input{\datapath/csqmi_and_coverage/test_csqmi_coverage_coverage_boxes_afdb01c.tex}
    \caption{Boxes}
  \end{subfigure}
  \begin{subfigure}{0.49\linewidth}
    \setlength{\figurewidth}{\linewidth}
    \input{\datapath/csqmi_and_coverage/test_csqmi_coverage_coverage_empty_afdb01c.tex}
    \caption{Empty}
  \end{subfigure}
  \caption[Comparison of exploration with CSQMI and optimistic coverage]{%
    The above plots compare exploration with optimistic coverage and CSQMI
    objectives  in terms of environment coverage in the Boxes and Empty
    environments
    (for ten trials with sequential planning and 16 robots).
    Black lines demarcate (from top to bottom) the maximum environment coverage,
    the completion threshold, and the early progress threshold.
  }%
  \label{fig:csqmi_and_coverage}
\end{figure}

Initially, we chose the optimistic coverage objective in
Chapter~\ref{chapter:time_sensitive_sensing}
primarily because exact evaluation is tractable and because optimistic coverage
is
3-increasing.\footnote{
  We made this decision before identifying the connection between mutual
  information and expected coverage.
}
As such, we wish to clarify how this decision affects performance, particularly
in comparison to the CSQMI objective applied in
Chapter~\ref{chapter:distributed_multi-robot_exploration}.
Additionally, we are not aware of any other works that compare ray-based
mutual information~\citep{charrow2015icra,julian2014}
to coverage-like approaches which are also
common~\citep{delmerico2018auro,simmons2000aaai,bircher2018receding}.
On the other hand, \citet{zhang2016aaai} found that Shannon mutual information
objectives behaves similarly as CSQMI.

Figure~\ref{fig:csqmi_and_coverage} presents results for exploration in the
Boxes and Empty environments where optimistic coverage provides a 16\%
and 11\% improvement in average completion time, respectively, compared to
exploration with a CSQMI objective with an occupancy prior of 0.125.\footnote{
  Chapter~\ref{chapter:distributed_multi-robot_exploration} uses the same value
  for the prior.
}
Note that some improvement in performance is expected for the Empty
environment because the optimistic coverage objective provides oracle values.

These results are somewhat limited.
First, we do not vary the parameters of the CSQMI objective such as the prior
and parameters for the independence test.
In particular, the results suggest that one way to improve performance may be
to select a lower occupancy prior (compared to 0.125).
Additionally, we did not put the exploration system with the CSQMI objective
through the same parameter tuning process as we did for optimistic coverage.
Still, the average completion times for CSQMI in the Boxes and Empty
environments
(1067 and 1526 robot-iterations, respectively)
exceed the average for any configuration of the optimistic coverage objective
that we tested.

That said, these results indicate several ways by which exploration
performance could be improved for systems based on mutual information
objectives:

To begin, selecting priors with low occupancy probabilities may improve
performance
(in agreement with \citet{henderson2020icra})
given that optimistic coverage is equivalent to mutual information with a prior
approaching zero.
Toward this end, normalizing the mutual information
(such as by dividing by the entropy of unobserved cells)
would mitigate the issue of the mutual information trending toward zero for
small priors.
The motivation for such normalization is self-evident from the form of
\eqref{eq:mapping_information_independent}, and we note that
the optimistic coverage objective effectively realizes this normalization in the
limit.

\pagebreak[4]
Additionally, approximation of
joint mutual information\footnote{
  Arguably, accurate evaluation of joint mutual information is much more
  important to this work than others on exploration
  since we study the collective contributions of teams of robots.
}
via the method of \citet{charrow2015icra}) may significantly affect
accuracy for small occupancy priors.\footnote{
  One case where the approximation of the joint by \citet{charrow2015icra}
  would perform poorly is when many rays traverse a single, previously
  unobserved cell but collectively observe many unobserved cells.
  However, the contributions of only one or few rays would be counted.
}
Notably, recent work by \citet{henderson2020icra} provides accurate values for
individual camera views by treating the view as a continuous integral
as well as new bounds for multiple views,
and this approach could significantly improve approximations for multiple views
and robots.
For either case, optimistic coverage, having the advantage of being exact for a
certain regime, could provide a useful point of comparison for evaluating the
impact of approximation.

\subsection{Upper and lower bounds on mutual information}

As stated, approximations of mutual information for multiple rays and camera
views frequently apply upper and lower bounds on joint mutual
information~\citep{charrow2015icra,henderson2019thesis}.
The interpretation of noiseless mutual information as expected coverage
likewise leads to useful bounds on mutual information.
Specifically, the expectation for the mutual information in
\eqref{eq:mapping_information_independent}
commutes and can be computed cell-wise.
The expectations for individual cells can then be bounded
via upper and lower bounds on probabilities that an observation along a given
ray will infer the occupancy value for that cell.\footnote{%
  Such probabilities are trivial to compute~\citep{charrow2015icra,julian2014}.
}
The maximum probability that any of several rays will observe a cell produces
a lower bound on the mutual information while summing
probabilities of observation (and capping at one) produces an upper bound.
Notably, both of these bounds become tight as the prior probability of occupancy
approaches zero because
the probability of observing all cells in range approaches one.

\section[Comparison of suboptimality for \rsp{} and \dgreedy{}]{%
  Comparison of suboptimality for \rsp[][\textbf]{} and \dgreedy[][\textbf]{}}
\label{appendix:dsga_and_rsp}

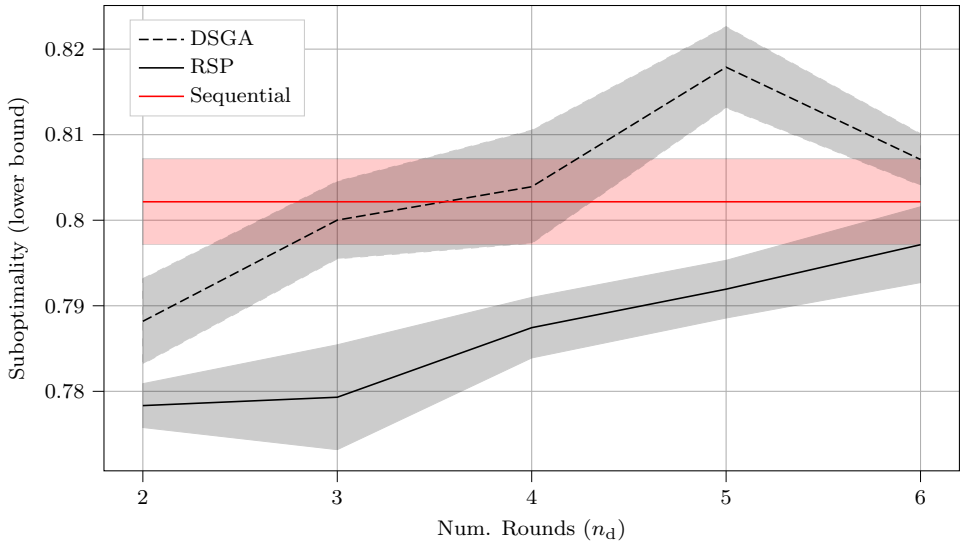
\begin{figure}
  \scriptsize
  \setlength{\figurewidth}{0.8\linewidth}
  \setlength{\figureheight}{0.6\figurewidth}
\begin{tikzpicture}

\begin{axis}[
height=\figureheight,
legend cell align={left},
legend columns=1,
legend style={at={(0.03,0.97)}, anchor=north west, draw=white!80.0!black},
tick align=outside,
tick pos=left,
width=\figurewidth,
x grid style={lightgray!92.0261437908!black},
xlabel={Num. Rounds (\(\displaystyle n_\mathrm{d}\))},
xmajorgrids,
xmin=1.8, xmax=6.2,
xtick style={color=black},
xtick={2,3,4,5,6},
y grid style={lightgray!92.0261437908!black},
ylabel={Suboptimality (lower bound)},
ymajorgrids,
ymin=0.770707172391841, ymax=0.825097440095319,
ytick style={color=black}
]
\path [draw=black, fill=red, opacity=0.2, line width=0.0pt]
(axis cs:2,0.807188122514159)
--(axis cs:6,0.807188122514159)
--(axis cs:6,0.79712554993359)
--(axis cs:2,0.79712554993359)
--cycle;
\path [draw=black, fill=black, opacity=0.2, line width=0.0pt, dashed]
(axis cs:2,0.793171318086755)
--(axis cs:3,0.804537699595286)
--(axis cs:4,0.81052359810951)
--(axis cs:5,0.822625155199706)
--(axis cs:6,0.810110309055341)
--(axis cs:6,0.804085596362334)
--(axis cs:5,0.81314927711493)
--(axis cs:4,0.797304478058364)
--(axis cs:3,0.79549005505364)
--(axis cs:2,0.783203445556816)
--cycle;
\path [draw=black, fill=black, opacity=0.2, line width=0.0pt]
(axis cs:2,0.780893719500921)
--(axis cs:3,0.785439292667904)
--(axis cs:4,0.790977161443961)
--(axis cs:5,0.795314030587651)
--(axis cs:6,0.801569527503976)
--(axis cs:6,0.792707173832404)
--(axis cs:5,0.788569182909218)
--(axis cs:4,0.783901777317209)
--(axis cs:3,0.773179457287454)
--(axis cs:2,0.77576960287052)
--cycle;
\addplot [semithick, black, dash pattern=on 3.7pt off 1.6pt]
table {%
2 0.788187381821785
3 0.800013877324463
4 0.803914038083937
5 0.817887216157318
6 0.807097952708837
};
\addlegendentry{DSGA}
\addplot [semithick, black]
table {%
2 0.778331661185721
3 0.779309374977679
4 0.787439469380585
5 0.791941606748434
6 0.79713835066819
};
\addlegendentry{RSP}
\addplot [semithick, red]
table {%
2 0.802156836223875
6 0.802156836223875
};
\addlegendentry{Sequential}
\end{axis}

\end{tikzpicture}
  \caption[Comparison of lower bounds on suboptimality for planning for
  exploration with \rsp{} and \dgreedy{}]{%
    Comparison of lower bounds on suboptimality for planning for exploration with
    \rsp{} and \dgreedy{} for ten simulation trials in the Boxes
    environment.
    The red line is based suboptimality results for sequential planning in the
    same configuration from Table~\ref{tab:suboptimality_exploration}.
  }%
  \label{fig:dsga_and_rsp}
\end{figure}

The proposed \rsp{} planners overcome a number of limitations of \dgreedy{}
(Chapter~\ref{chapter:distributed_multi-robot_exploration}):
\dgreedy{} does not provide offline suboptimality guarantees,
distributed implementations would involve broadcasts and reductions over all
robots during each round,
and the assignments themselves are sequential.
We provide the following comparison as a point of reference as \dgreedy{} may
be impractical in practice.

Figure~\ref{fig:dsga_and_rsp} compares lower bounds on suboptimality
(Sec.~\ref{sec:online_bounds_exploration})
for \dgreedy{} and \rsp{} given different numbers of planning
rounds ($\numrounds$).
On average, \dgreedy{} outperforms \rsp{} planning in these results.
Even though \rsp{} provides stricter performance guarantees, this result is not
surprising.
While \rsp{} partitions robots randomly,
\dgreedy{} effectively partitions robots in the midst of the planning process.
This could enable \dgreedy{} to better take advantage of problem structure
or random improvements in planner results across rounds
(as \dgreedy{} selects a subset of results from \emph{all} robots during each
round).

\section{Description of auction methods for distributed submodular maximization}
\label{appendix:auction_algorithms}

Chapter~\ref{chapter:distributed_communication} compares \rsp{} planning to
auction methods much like those which are common for task
assignment~\citep{choi2009tro}.
Such auction algorithms converge to greedy solutions
(Alg.~\ref{alg:greedy_submodular})
and can even be extended to complex constraints~\citep{williams2017icra}.
However,
the auction algorithms that \citet{choi2009tro}
describe (e.g. CBBA) are specific to the assignment problems they study and do
not immediately apply for general submodular objectives.
Specifically, they take advantage of how the value of an assignment to one robot
is independent of assignments to others
to avoid recomputing marginal gains (bids) when updating assignments.

The following algorithms
(Algs.~\ref{alg:global_auction}~and~\ref{alg:local_auction})
are similar and are suitable for the kinds of assignment
problems we study in this thesis (e.g. \eqref{eq:receding_horizon_sensing})
The two differ based on assumptions on how agents evaluate objective values.
The first requires full information---the ability to evaluate marginal gains for
any agents' actions---in order for agents to reorder existing assignments.
This likely produces behavior that is comparable to CBBA which can often avoid
re-evaluating marginal gains.
For simplicity we assume that both run synchronously on an undirected
communication graph where each agent $i$ has neighbors $\neighbor_i$.
Both algorithms eventually converge to the maximum value\footnote{%
  Like \citet{choi2009tro} we assume that there is a deterministic mechanism to
  break symmetry in comparisons (e.g. via agent and action indices).
}
across all agents
at each step of the assignment process and so eventually obtain solutions
equivalent to the general greedy algorithm (Alg.~\ref{alg:greedy_submodular}).
Although we do not seek to prove this point, the arguments would be similar to
those by \citet{choi2009tro}.

Algorithm~\ref{alg:global_auction} requires consistent, global access to the
objective $\setfun$.
This algorithm consists of
exchanging assignments with neighbors
and simply
executing that general greedy algorithm
(Alg.~\ref{alg:greedy_submodular})
on a reduced ground set to update the agent's assignments,
much like the approach of \citet{luo2016iros}.
The reduced ground set consists of the agent's
own block of control actions along with its and its neighbors current
assignments.
The output then preserves maxima at each assignment position and includes some
best-guess at the assignment to that agent ($i$).
This greedy process can enable agents to collectively identify
multiple elements of the final solution during a single synchronous
communication round and sometimes converges in fewer synchronous communication
rounds than the number of agents.
However, this approach \emph{strictly} requires global access to the objective
$\setfun$ because the agents evaluate the value of others' assignments during
the greedy decision process and to ensure that the outputs are consistent and
deterministic.
As such, this algorithm does not represent a practical solution to the problems
we study as asymmetry in information access is common in distributed
settings~\citep{choi2009tro}.\footnote{%
  For example, robots in Chapter~\ref{chapter:target_tracking} use
  different local approximations for target tracking, and
  for the distributed exploration implementation in
  Chapter~\ref{chapter:distributed_communication}, even though robots construct
  maps with sensor data from all robots, the maps are not synchronized
  and generally differ.
}

\begin{algorithm}
  \caption[Auction algorithm with global information]{
    Auction algorithm with global information.
    One synchronous iteration from the perspective of agent $i$.
  }%
  \label{alg:global_auction}
  \begin{minipage}{\linewidth}
    \begin{algorithmic}[1]
      \State $X_i \gets \text{Current set of assignments for agent $i$}$
      \State $\block_i \gets \text{Set of actions available to $i$}$
      \State $\neighbor_i \gets \text{Set of communication neighbors}$
      \vspace{1em}
      \State \method{Send}: $X_i$ to $\neighbor_i$
      \State \method{Receive}: $X_j$ from $j$ for $j \in \neighbor_i$
      \vspace{1em}
      \State $G \gets \bigcup_{j \in \neighbor_i \cup i} X_j \cup \block_i$
      \State $X_i \gets \method{Greedy}(G)$
      \hfill(Executing Alg.~\ref{alg:greedy_submodular}
      for objective $\setfun$ and partition matroid $\independence$)
    \end{algorithmic}
  \end{minipage}
\end{algorithm}

Algorithm~\ref{alg:local_auction}, which performs auctions with local
information, is somewhat more complex
because agents are not allowed to evaluate marginal gains for others'
assignments so that these values have to be communicated along with the
decisions and so that updates require more care.
The main body (lines~\ref{line:local_send}--\ref{line:local_update_call})
of this algorithm is similar to the last in that agents exchange assignments
with neighbors before updating the local solution.
Except, this time, values and assignments are exchanged in lists in assignment
order.
The update itself is based on the first position where the assignments differ
and consists of three cases:
(line~\ref{line:local_dominates})
the current assignments dominate (and already include an assignment to $i$);
(line~\ref{line:other_dominates_includes_local})
the other's assignments dominate and include some assignment to $i$
(in block $\block_i$ of the partition matroid),
obtained via the following procedure;
or finally
(line~\ref{line:other_dominates_excluding_local})
agent $i$ must update the other's assignments with an assignment to itself.
Seeking to emulate Alg.~\ref{alg:greedy_submodular},
agent $i$ checks whether its best marginal gain
beats the value of the previous winning assignment $V_{\mathrm{o},m}$
at each point $m$ in the decision process.
Once identifying a winning assignment position
(or implicitly beating a null assignment)
the agent concatenates its assignment and value
(using ``$\cdot$'' to represent list concatenation)
onto the sub-lists of prior assignments and values
(line~\ref{line:local_auction_assignment})
and returns this result.
The agent thereby discards assignments at later positions in the sequence
as inserting the assignment for $i$ invalidates any following marginal gains.
This behavior also implies that, unlike Alg.~\ref{alg:global_auction}, this
auction algorithm requires at least one synchronous round per each
element of the final solution (i.e. one per robot) as the incremental
construction of the solution in line~\ref{line:local_auction_assignment} ensures
that the agents also collectively produce the full solution sequentially.

Both of these algorithms can be implemented more efficiently from a
computational perspective.
But, given that the discussion in
Chapter~\ref{chapter:distributed_communication} focuses on communication costs,
efficient computation will not be necessary for this text.
Still, even efficient implementations of these auction algorithms
will be more computation intensive than \rsp{}
(where agents together complete a single pass over each block $\block_i$ of
the ground set).
The auction algorithms both execute a complete pass over the ground set in the
first synchronous round alone and may require many times more computation to
converge.
Likewise, auction algorithms incur greater communication costs because all
agents exchange assignments during each round (rather than a subset)
and because agents exchange sets of assignments rather than individual
assignments as in \rsp{}.

\begin{algorithm}
  \caption[Auction algorithm with local information]{
    Auction algorithm with local information.
    One synchronous iteration from the perspective of agent $i$.
  }%
  \label{alg:local_auction}
  \begin{minipage}{\linewidth}
    \begin{algorithmic}[1]
      \State $X_i \gets \text{\emph{List} of assignments for agent $i$
      (in assignment order)}$
      \State $V_i \gets \text{List of marginal gains (values)
                              for assignments $X_i$}$
      \State $\block_i \gets \text{Set of actions available to $i$}$
      \State $\neighbor_i \gets \text{Set of communication neighbors}$
      \vspace{1em}
      \State \method{Send}: $X_i,~V_i$ to $\neighbor_i$
      \label{line:local_send}
      \State \method{Receive}: $X_j,~V_j$ from $j$ for $j \in \neighbor_i$
      \vspace{1em}
      \For{$j \in \neighbor_i$}
        \State $X_i,~V_i \gets \method{UpdateAssignments}(X_i,V_i,X_j,V_j)$
        \label{line:local_update_call}
      \EndFor
      \vspace{1em}
      \CommentLine{Update local assignment in light of one from an other agent}
      \Procedure{UpdateAssignments}{$X_\mathrm{\ell}, V_\mathrm{\ell},
                                    X_\mathrm{o}, V_\mathrm{o}$}
        \State $n_\mathrm{diff}
          \gets \min_{n} V_{\mathrm{\ell},n} \neq V_{\mathrm{o},n}$
        \If{$V_{\mathrm{\ell},n} \geq V_{\mathrm{o},n}$}
        \label{line:local_dominates}
        \hfill(Agent $i$'s assignments dominate. Return.)
          \State \Return $X_\mathrm{\ell},~V_\mathrm{\ell}$
        \ElsIf{$\block_i \cap X_\mathrm{o} \neq \emptyset$}
        \label{line:other_dominates_includes_local}
        \hfill(Other's assignments reflect $i$'s choices)
          \State \Return $X_\mathrm{o},~V_\mathrm{o}$
        \Else
        \label{line:other_dominates_excluding_local}
          \For{$m \in \{n,\ldots,|X_\mathrm{o}|\}$}
            \hfill(Search for a viable assignment position)
            \State $x \gets \argmax_{x' \in \block_i}
            \setfun(x' | X_{\mathrm{o},1:m-1})$
            \State $v \gets \setfun(x | X_{\mathrm{o},1:m-1})$
            \If{$v > V_{\mathrm{o},m}$}
            \hfill($i$'s bid wins at position $m$)
              \State \Return $X_{\mathrm{o},1:m-1} \cdot x,%
                ~V_{\mathrm{o},1:m-1} \cdot v$
              \hfill(values $V_\mathrm{o}$ valid up to $m-1$)
              \label{line:local_auction_assignment}
            \EndIf
          \EndFor
        \EndIf
      \EndProcedure
    \end{algorithmic}
  \end{minipage}
\end{algorithm}

\chapter{Assorted Proofs}
\setdatapath{./fig/assorted_proofs}
\label{chapter:proofs}

\section{Proof for representations of derivatives of set functions}
\label{appendix:derivative_representation}

Equation~\eqref{eq:recursive_derivative} defines the $n^\mathrm{th}$ derivative
of a set function as
\begin{align}
  \setfun(Y_1;\ldots;Y_n|X) &=
  \setfun(Y_1;\ldots;Y_{n-1}|X,Y_n) - \setfun(Y_1;\ldots;Y_{n-1}|X).
  \label{eq:recursive_derivative_appendix}
\end{align}
This is equivalent to the definition by \citet{foldes2005},
\begin{align}
  \setfun(Y_1;\ldots;Y_n|X) &=
  \sum_{Y\subseteq \{Y_1,\ldots,Y_n\}} (-1)^{n-|Y|} \setfun(X \cup \hat Y),
  \quad \hat Y = \bigcup_{Y' \in Y} Y'.
  \label{eq:nth_derivative}
\end{align}
with the exceptions that we define derivatives with respect to sets for
convenience
(where the extension follows from closure properties~\citet{foldes2005})
and omit set differences by requiring disjoint sets.
\begin{proof}
  The proof follows by expanding the right hand side of
  \eqref{eq:recursive_derivative_appendix} and combining the resulting
  expressions.
  Define the union of a collection of sets as $\phi(X)=\bigcup_{X' \in X} X'$.
  Then
  \begin{align*}
    &\setfun(Y_1;\ldots;Y_{n-1}|X,Y_n) - \setfun(Y_1;\ldots;Y_{n-1}|X)
    \\
    &=
    \sum_{Y\subseteq \{Y_1,\ldots,Y_{n-1}\}} (-1)^{n-1-|Y|}
    \setfun(X \cup Y_n \cup \phi(Y))
    -
    \sum_{Y\subseteq \{Y_1,\ldots,Y_{n-1}\}}
      (-1)^{n-1-|Y|} \setfun(X \cup \phi(Y))
    \\
    &=
    \sum_{Y\subseteq \{Y_1,\ldots,Y_{n-1}\}} (-1)^{n-(|Y|+1)}
    \setfun(X \cup Y_n \cup \phi(Y))
    +
    \sum_{Y\subseteq \{Y_1,\ldots,Y_{n-1}\}}
      (-1)^{n-|Y|} \setfun(X \cup \phi(Y))
    \\
    &=
    \sum_{Y\subseteq \{Y_1,\ldots,Y_{n}\}} (-1)^{n-|Y|} \setfun(X \cup \phi(Y))
    = \setfun(Y_1;\ldots;Y_n|X).
  \end{align*}
\end{proof}

\section{Proof of Lemma~\ref{lemma:chain_rule}, chain rule for derivatives of
set functions}
\label{appendix:chain_rule}

\begin{proof}
  The proof follows by expanding the derivative \eqref{eq:recursive_derivative}, forming a
  telescoping sum, and rewriting the summands as individual derivatives:
  \begin{align}
    \setfun(Y_1;\ldots;Y_n|X)
    &=
    \setfun(Y_1;\ldots;Y_{n-1}|Y_n,X)
    - \setfun(Y_1;\ldots;Y_{n-1}|X) \nonumber\\
    &=
    \sum_{i=1}^{|Y_n|}
    \big(
      \setfun(Y_1;\ldots;Y_{n-1}|Y_{n,1:i},X)
      - \setfun(Y_1;\ldots;Y_{n-1}|Y_{n,1:i-1},X)
    \big)
    \nonumber\\
    &=
    \sum_{i=1}^{|Y_n|}
    \setfun(Y_1;\ldots;Y_{n-1};y_{n,i}|Y_{n,1:i-1},X).
  \end{align}
\end{proof}

\section[Proof of Theorem~\ref{theorem:distributed_suboptimality}, post-hoc
suboptimality of \dgreedy{}]{%
Proof of Theorem~\ref{theorem:distributed_suboptimality}, post-hoc
suboptimality of \dgreedy[][\textbf]{}}
\label{appendix:distributed_proof}
\begin{proof}
  The proof of the suboptimality bound relating \dgreedy{}
  to \greedy{}
  incorporates suboptimality of the single-robot planner
  and is similar to~\citep{singh2009} or~\citep{atanasov2015icra}.
  We obtain the following
  by monotonicity and by rearranging the resulting telescoping sum
  \begin{align}
    \MI(M;Y^\star)
    &\leq \MI(M;Y^\star) +
    \sum_{i=1}^{\numrobot}
    \MI(M;Y^\mathrm{d}_{i}|Y^\mathrm{d}_{1:i-1}, Y^\star_{i+1:\numrobot})
    \nonumber\\
    &= \MI(M;Y^\mathrm{d}) +
    \sum_{i=1}^{\numrobot}
    \MI(M;Y^\star_{i}|Y^\mathrm{d}_{1:i-1}, Y^\star_{i+1:\numrobot}).
    \label{eq:proof_start}
  \end{align}
  By submodularity
  \begin{align*}
    \MI(M;Y^\star_{i}|Y^\mathrm{d}_{1:i-1}, Y^\star_{i+1:\numrobot}) \leq
      \MI(M;Y^\star_{i} | Y^\mathrm{d}_{1:i-1}).
  \end{align*}
  Without loss of generality, assume that agent indices correspond to the
  selection order and rewrite in terms of the planning rounds
  such that
  $
  \MI(M;Y^\star_{i} | Y^\mathrm{d}_{1:i-1})
  =
  \MI(M;Y^\star_{D_{j,k}}|Y^\mathrm{d}_{D_{j,1:k-1}\cup F_{j-1}})
  $
  and note that although $Y^\star$ is formally a set, the mapping from
  elements to robots can be obtained by the intersections
  $Y^\star\cap\mathcal{Y}_i$ given that the sets $\mathcal{Y}_i$ are disjoint.
  Then, by submodularity,
  \begin{align*}
    \MI(M;Y^\star_{i} | Y^\mathrm{d}_{1:i-1})
    \leq \MI(M;Y^\star_{D_{j,k}}|Y^\mathrm{d}_{F_{j-1}}).
  \end{align*}
  By \eqref{eq:approximation} and the greedy maximization step in
  Alg.~\ref{alg:distributed_greedy}
  \begin{align}
    \MI(M;Y^\star_{D_{i,j}}|Y^\mathrm{d}_{F_{i-1}})
    \leq
    \eta \MI(M;Y^\mathrm{d}_{D_{i,j}}|Y^\mathrm{d}_{F_{i-1}}).
    \label{eq:greedy_choice_auro}
  \end{align}
  Substitute \eqref{eq:greedy_choice_auro} and preceding inequalities
  into \eqref{eq:proof_start} to obtain
  \begin{align}
    \begin{multlined}
      \MI(M;Y^\star)
      \leq
      \MI(M;Y^\mathrm{d}) +
      \eta\sum_{i=1}^{\numrounds}
      \sum_{j=1}^{|D_i|}
      \MI(M;Y^\mathrm{d}_{D_{i,j}}|Y^\mathrm{d}_{F_{i-1}}).
    \end{multlined}
    \label{eq:substitute_greedy_choice}
  \end{align}
  The distributed objective can be rewritten as a sum so that
  \linebreak
  $
  \MI(M;Y^\mathrm{d})=
  \sum_{i=1}^{\numrounds}
  \sum_{j=1}^{|D_i|}
  \MI(M;Y^\mathrm{d}_{D_{i,j}}|Y^\mathrm{d}_{D_{i,1:j-1}\cup F_{i-1}})
  $
  and substituted into \eqref{eq:substitute_greedy_choice}
  to obtain
  \begin{align}
    \begin{multlined}
      \MI(M;Y^\star)
      \leq
      (1+\eta)\MI(M;Y^\mathrm{d}) +
      \eta\sum_{i=1}^{\numrounds}
      \sum_{j=1}^{|D_i|}
      \left(
        \MI(M;Y^\mathrm{d}_{D_{i,j}}|Y^\mathrm{d}_{F_{i-1}})
        -\MI(M;Y^\mathrm{d}_{D_{i,j}}|Y^\mathrm{d}_{D_{i,1:j-1}\cup F_{i-1}})
      \right)
    \end{multlined}
    \label{eq:planner_difference}
  \end{align}
  which expresses the suboptimality
  in terms of decreases in the conditional mutual information
  from when the plans were first obtained from the planner
  (Alg.~\ref{alg:distributed_greedy}, line~\ref{line:updated_objective})
  to when they were assigned,
  (Alg.~\ref{alg:distributed_greedy}, line~\ref{line:reduction})
  potentially after other assignments in the same round.
  By rewriting mutual information in terms of entropies we can rearrange to
  obtain the following
  \begin{align*}
    \begin{multlined}
      \MI(M;Y_1) - \MI(M;Y_1|Y_2)
      = \MI(Y_1;Y_2) - \MI(Y_1;Y_2|M).
    \end{multlined}
  \end{align*}
  \emergencystretch 3em
  If $Y_1$ and $Y_2$ are conditionally independent given $M$, then the mutual
  information,
  $\MI(Y_1;Y_2|M)=0$ and so
  $
  \MI(M;Y_1) - \MI(M;Y_1|Y_2)
  =
  \MI(Y_1;Y_2)
  $.
  By substitution into \eqref{eq:planner_difference} we can obtain the slightly
  more concise and final expression for the suboptimality in terms of the mutual
  information between observations
  \begin{align}
    \begin{multlined}
      \MI(M;Y^\star)
      \leq
      (1+\eta)\MI(M;Y^\mathrm{d}) +
      \eta\sum_{i=1}^{\numrounds}
      \sum_{j=1}^{|D_i|}
      \MI(Y^\mathrm{d}_{D_{i,j}};Y^\mathrm{d}_{D_{i,1:j-1}}|Y^\mathrm{d}_{F_{i-1}}).
    \end{multlined}
    \label{eq:distributed_final}
  \end{align}
\end{proof}

\section[Proof of Theorem~\ref{theorem:worst-case}, worst case suboptimality of
\dgreedy]{%
  Proof of Theorem~\ref{theorem:worst-case}, worst case suboptimality of
\dgreedy[][\textbf]}
\label{appendix:worst_case_proof}
\begin{proof}
  Equation~\eqref{eq:worst_case} follows from Theorem~1
  by~\citet{grimsman2018tcns} which proves a $k+1$ bound where $k$ is the clique
  cover number of a directed acyclic graph associated with the planner structure.
  In this directed graph, each robot represents a vertex,
  and the graph has a directed edge $(a,b)$ between robots $a,b\in\robots$
  if $b$ maximizes its objective~\eqref{eq:single_robot_objective} conditional
  on the sequence of observations selected by $a$.
  Here, a clique, which is a complete subgraph, is analogous to a set of robots
  that plan sequentially given the choices by all prior robots in the clique.
  For Alg.~\ref{alg:distributed_greedy}, any set of robots $A \subseteq \robots$
  with at most one robot from each planning round
  ($|A\cap D_i| \leq 1$ for $i \in \{1,\ldots,\numrounds\}$) forms a clique in the
  associated directed graph.
  A clique cover of size
  $\lceil \numrobot/\numrounds \rceil=\max_{i\in\{1,\ldots,\numrounds\}} |D_i|$
  can be obtained by selecting cliques with a single robot (as available)
  from each planning round ($D_1,\ldots,D_{\numrounds}$) without replacement.
  Then, \eqref{eq:worst_case} follows by substitution of
  \eqref{eq:approximation} to obtain a factor of $\eta$.
\end{proof}

\section{Proof of Lemma~\ref{lemma:approximate_suboptimality}, suboptimality
of general assignments}
\label{appendix:approximate_suboptimality}

\begin{proof}
  This result follows a similar approach as the other proofs related to
  sequential submodular maximization that arise throughout this thesis with
  slight deviation to assist in book-keeping:
  \begin{align}
    \setfun(X^\opt) &\leq \setfun(X^\mathrm{d}, X^\opt) \label{eq:line1}\\
               &=    \setfun(X^\mathrm{d}) + \sum_{i=1}^{\numrobots}
                     \setfun(x^\star_i | X^\star_{1:i-1}, X^\mathrm{d})
                     \label{eq:line2}\\
               &\leq \setfun(X^\mathrm{d}) + \sum_{i=1}^{\numrobots}
                     \setfun(x^\star_i | X^\mathrm{d}_{1:i-1})
                     \label{eq:line3}\\
               &\leq \setfun(X^\mathrm{d}) + \sum_{i=1}^{\numrobots}
                     \left(
                       \setfun(x^\mathrm{d}_i | X^\mathrm{d}_{1:i-1})
                       + \cost_i(\setfun, X^\mathrm{d}_{1:i-1})
                     \right)
                     \label{eq:line4}
                     \\
               &=    2\setfun(X^\mathrm{d})
                     + \sum_{i=1}^{\numrobots}
                     \cost_i(\setfun, X^\mathrm{d}_{1:i-1}).
                     \label{eq:line5}
  \end{align}
  Above,
  \eqref{eq:line1} follows from monotonicity;
  \eqref{eq:line2} expands a telescoping series;
  \eqref{eq:line3} follows from submodularity;
  \eqref{eq:line4} upper bounds the incremental values optimal robot decisions
  with the incremental value of each actual decision and its suboptimality
  \eqref{eq:single_robot_suboptimality}; and
  \eqref{eq:line5} collapses the telescoping series.
\end{proof}

\section{Proof of Theorem~\ref{theorem:sum_submodular}, suboptimality of
distributed planning for target tracking}
\label{appendix:proof_of_sum_submodular}

\begin{proof}
  Theorem~\ref{theorem:sum_submodular} consists of two parts,
  \eqref{eq:costs_bound} and \eqref{eq:weights_bound}.
  We prove each in turn.
  Since the costs in both equations involve sums over robots, both proofs
  analyze costs with respect to some robot $i\in\robots$.

  \subsubsection*{Proof of Theorem~\ref{theorem:sum_submodular}, part 1
  \eqref{eq:costs_bound}}
  According to the standard greedy algorithm (Alg.~\ref{alg:local_greedy}),
  robot $i$ would plan conditional on decisions by robots $\{1,\ldots,i-1\}$.
  However, in Alg.~\ref{alg:distributed_target_tracking} that robot instead
  plans conditional on a subset of these robots
  $\neighbor_i\subseteq\{1,\ldots,i-1\}$ and ignores
  $\ignore_i=\{1,\ldots,i-1\}\setminus\neighbor_i$.
  We can then write the cost of a suboptimal decision from
  \eqref{eq:approximate_suboptimality} in terms of $\distcost_i$ as
  \begin{align}
    \begin{split}
      \cost_i(\setfun, X^\mathrm{d}_{1:i-1})
      &\leq
      \cost_i(\setfun, X^\mathrm{d}_{\neighbor_i})
      + \setfun(x_i^\mathrm{d}|X^\mathrm{d}_{\neighbor_i})
      - \setfun(x_i^\mathrm{d}|X^\mathrm{d}_{1:i-1})
      \\
      &=
      \cost_i(\setfun, X^\mathrm{d}_{\neighbor_i}) + \distcost_i,
    \end{split}
    \label{eq:substitute_distributed_cost}
  \end{align}
  where the first step follows by substituting the cost
  model \eqref{eq:single_robot_suboptimality} and
  given that
  \linebreak[4]
  $\max_{x\in\block_i} \setfun(x|X^\mathrm{d}_{1:i-1})
  \leq
  \max_{x\in\block_i} \setfun(x|X^\mathrm{d}_{\neighbor_i})$
  due to submodularity, and the second follows from the definition of the cost
  of distributed planning \eqref{eq:distributed_planning_cost}.
  To incorporate the cost of suboptimal planning $\plannercost_i$, observe that
  \begin{align}
    \begin{split}
      \cost_i(\setfun, X^\mathrm{d}_{\neighbor_i})
      &=
      \cost_i(\setfun, X^\mathrm{d}_{\neighbor_i})
      + \plannercost_i - \plannercost_i
      \\
      &=
      \plannercost_i
      + \cost_i(\setfun, X^\mathrm{d}_{\neighbor_i})
      - \cost_i(\approxsetfun_i, X^\mathrm{d}_{\neighbor_i})
    \end{split}
    \label{eq:substitute_planner_cost}
  \end{align}
  which follows from the definition of the planning cost
  \eqref{eq:planner_cost}.
  The cost of approximation of the objective $\objectivecost_i$
  upper bounds the difference of the last two terms in
  \eqref{eq:substitute_planner_cost}:
  \begin{align}
    \begin{split}
      \cost_i(\setfun, X^\mathrm{d}_{\neighbor_i})
      - \cost_i(\approxsetfun_i, X^\mathrm{d}_{\neighbor_i})
      &=
      \approxsetfun_i(x_i^\mathrm{d} | X^\mathrm{d}_{\neighbor_i})
      -
      \setfun(x_i^\mathrm{d} | X^\mathrm{d}_{\neighbor_i})
      \\&\quad
      +
      \max_{x \in \block_i}
      \setfun(x | X^\mathrm{d}_{\neighbor_i})
      -
      \max_{x \in \block_i}
      \approxsetfun_i(x | X^\mathrm{d}_{\neighbor_i})
      \\
      &\leq
      \approxsetfun_i(x_i^\mathrm{d} | X^\mathrm{d}_{\neighbor_i})
      -
      \setfun(x_i^\mathrm{d} | X^\mathrm{d}_{\neighbor_i})
      \\&\quad
      +
      \setfun(\hat x | X^\mathrm{d}_{\neighbor_i})
      -
      \approxsetfun_i(\hat x | X^\mathrm{d}_{\neighbor_i}),
      \quad
      \text{for }
      \hat x \in \argmax \setfun(\hat x | X^\mathrm{d}_{\neighbor_i})
      \\
      &\leq \objectivecost_i
      \label{eq:substitute_objective_cost}
    \end{split}
  \end{align}
  by expanding and rearranging the costs \eqref{eq:single_robot_suboptimality}
  on the left-hand-side, using an upper bound to match the arguments of the last
  two terms, and by using the definition of the objective cost
  \eqref{eq:objective_cost} to bound the two differences.

  Then, the expression for the costs in \eqref{eq:costs_bound}
  \begin{align}
    \cost_i(\setfun, X^\mathrm{d}_{1:i-1})
    &\leq
    \objectivecost_i
    + \plannercost_i
    + \distcost_i
    \label{eq:general_and_specific_costs}
  \end{align}
  follows by substituting the prior three equations into each other:
  \eqref{eq:substitute_planner_cost} into
  \eqref{eq:substitute_distributed_cost} and
  \eqref{eq:substitute_objective_cost} into the result.
  Finally, substituting this inequality \eqref{eq:general_and_specific_costs}
  into \eqref{eq:approximate_suboptimality} from
  Lemma~\ref{lemma:approximate_suboptimality} on the suboptimality of general
  assignments yields the expression for \eqref{eq:costs_bound} which completes
  the first part of this proof.

  \subsubsection*{Proof of Theorem~\ref{theorem:sum_submodular}, part 2
  \eqref{eq:weights_bound}}
  The second part of Theorem~\ref{theorem:sum_submodular} \eqref{eq:weights_bound}
  follows by referring to definition of $\distcost_i$ in
  \eqref{eq:distributed_planning_cost}, applying the chain rule
  \eqref{eq:chain_rule}, and substituting the bound on second derivatives
  \eqref{eq:derivative_bound} and the definitions of the weights
  \eqref{eq:weights} and \eqref{eq:weight_by_component} in turn:
  \begin{align}
    \distcost_i
    &=
    - \setfun(x_i^\mathrm{d};X^\mathrm{d}_{\ignore_i}|X^\mathrm{d}_{\neighbor_i})
    \\
    &=
    - \sum_{j \in \ignore_i}
    \setfun\big(
      x_i^\mathrm{d};x^\mathrm{d}_{j}|
      X^\mathrm{d}_{\neighbor_i}, X^\mathrm{d}_{\ignore_i\cap\{1:j-1\}}
    \big)
    \\
    &\leq
    \sum_{j \in \ignore_i}
    \weights(i,j)
    \leq
    \sum_{j \in \ignore_i}
    \approxweights(i,j).
  \end{align}
  Then,~\eqref{eq:weights_bound} follows by summing over $\robots$.
  This completes this second and last part of the proof of
  Theorem~\ref{theorem:sum_submodular}.
\end{proof}

\section{Proof of Theorem~\ref{theorem:mapping_information_is_coverage},
noiseless mutual information with independent cells is 3-increasing}
\label{appendix:proof_that_mapping_information_is_coverage}

The following proof takes advantage of cell independence liberally to write
mutual information in terms of the expected entropy of the cells that the robots
will observe.

\begin{proof}
  We can write the mutual information \eqref{eq:mutual_information_entropy}
  between
  the environment $\environment$
  and
  future observations $\observations(X)$
  in terms of entropies:
  \begin{align}
    \MI(\environment; \observations(X))
  &=
  \H(\environment) - \H(\environment | \observations(X)).
  \intertext{The conditional entropy can then be rewritten
    in terms of the expected entropy
    \eqref{eq:conditional_entropy_expectation}
    given the direct observations of
    cell occupancy \eqref{eq:exploration_observation}
    associated with a hypothetical instantiation of the environment
    $\environment'$
    while
    abbreviating observed cells as $C'=\covercells(X,\environment')$:
  }
  &=
  \H(\environment)
  -
  \E_{\environment' \sim \environmentsguess}
  \left[
    \H(\environment |
    \environment_{C'}\! =\! \environment'_{C'}
    )
  \right],
  \intertext{and that conditional entropy is simply the entropy of the cells
    that have not yet been observed
    $D' = \cells \setminus C'$
    due to independence:
  }
  &=
  \H(\environment)
  -
  \E_{\environment' \sim \environmentsguess}
  \left[
    \H(\environment_{D'})
  \right].
  \\
  \intertext{Then, bringing the entropy of $\environment$ into the expectation
    does not change its value,
    and separating the independent observed and unobserved cells
    simplifies the expression:
  }
  &=
  \E_{\environment' \sim \environmentsguess}
  \left[
    \H(\environment_{C'})
    +
    \H(\environment_{D'})
    -
    \H(\environment_{D'})
  \right].
  \\
  &=
  \E_{\environment' \sim \environmentsguess}
  \left[
    \H(\environment_{C'})
  \right].
\end{align}

Finally, the joint entropy of the cells the robot will observe
is the sum of their individual entropies
\begin{align}
  \MI(\environment; \observations(X)) &=
  \E_{\environment' \sim \environmentsguess}
  \left[
    \sum\nolimits_{i \in \covercells(X, \environment')} \H(\cell_i)
  \right].
\end{align}
This expresses a weighted expected coverage objective
\eqref{eq:expected_coverage} where the weight $\cellweight(i)$ of each cell $i
\in \cells$ is equal to its entropy $\H(\cell_i)$.
Observing that weighted expected coverage is 3-increasing
(Sec.~\ref{sec:expected_coverage_monotonicity})
completes the proof.
\end{proof}



\backmatter

\renewcommand{\bibsection}{\chapter{\bibname}}

\bibliographystyle{plainnat}
\bibliography{bibliography}

\end{document}